\documentclass[journal]{IEEEtran}
\pdfoutput=1
\usepackage[T1]{fontenc}
\usepackage[utf8]{inputenc}
\usepackage{newunicodechar} 
\usepackage[english]{babel}
\usepackage{amsmath}
\usepackage{todonotes}
\usepackage{ifpdf}
\ifpdf
	\usepackage{multirow}
\else
\fi

\usepackage{cite}

\ifCLASSINFOpdf

\usepackage{graphicx}
\usepackage{tikz}
\usetikzlibrary{arrows, decorations.markings}
\usepackage{pgfplots}
\usepackage{pgfplotstable}
\usepgfplotslibrary{colorbrewer}
\usetikzlibrary{pgfplots.groupplots}
\usepgfplotslibrary{statistics}
\pgfplotsset{compat=1.3}
\else
\fi

\usepackage{algorithm}
\usepackage{algpseudocode}

\makeatletter
\def\BState{\State\hskip-\ALG@thistlm}
\makeatother

\renewcommand{\algorithmicrequire}{\textbf{Input:}}
\renewcommand{\algorithmicensure}{\textbf{Output:}}

\algnewcommand
\algorithmicforeach{\textbf{foreach}}
\algdef{S}[FOR]{ForEach}[1]{\algorithmicforeach\ #1\ \algorithmicdo}
\algnewcommand{\LineComment}[1]{\State \(\triangleright\)
	~#1}

\newcommand{\removelatexerror}{\let\@latex@error\@gobble}

\usepackage{array}

\ifCLASSOPTIONcompsoc
\usepackage[caption=false,font=normalsize,labelfont=sf,textfont=sf]{subfig}
\else
\usepackage[caption=false,font=footnotesize]{subfig}
\fi

\usepackage{etoolbox}

\makeatletter 
\pretocmd\@bibitem{\color{black}\csname keycolor#1\endcsname}{}{\fail}
\newcommand\citecolor[1]{\@namedef{keycolor#1}{\color{blue}}}
\makeatother

\usepackage{xcolor}
\newcommand{\edit}[1]{{\color{blue}{#1}}}
\newcommand{\editt}[1]{{\color{red}{#1}}}

\newcommand{\fix}[1]{{\color{red}#1}}

\pgfplotstableread{
	sigma	corr	si
	0.01	1.0000	0.0833
	0.02	1.0000	0.0833
	0.03	1.0000	0.0833
	0.04	1.0000	0.0833
	0.05	1.0000	0.0833
	0.06	1.0000	0.0833
	0.07	1.0000	0.0833
	0.08	1.0000	0.0833
	0.09	1.0000	0.0833
	0.10	1.0000	0.0833
	0.11	1.0000	0.0833
	0.12	1.0000	0.0833
	0.13	1.0000	0.0833
	0.14	1.0000	0.0833
	0.15	1.0000	0.0833
	0.16	1.0000	0.0833
	0.17	1.0000	0.0833
	0.18	1.0000	0.0833
	0.19	1.0000	0.0833
	0.20	1.0000	0.0833
	0.21	1.0000	0.0833
	0.22	1.0000	0.0833
	0.23	1.0000	0.0833
	0.24	1.0000	0.0832
	0.25	1.0000	0.0831
	0.26	1.0000	0.0829
	0.27	1.0000	0.0826
	0.28	1.0000	0.0821
	0.29	1.0000	0.0813
	0.30	1.0000	0.0802
	0.31	1.0000	0.0786
	0.32	0.9999	0.0763
	0.33	0.9999	0.0732
	0.34	0.9998	0.0689
	0.35	0.9997	0.0631
	0.36	0.9995	0.0556
	0.37	0.9993	0.0458
	0.38	0.9990	0.0335
	0.39	0.9986	0.0185
	0.40	0.9980	0.0007
	0.41	0.9973	0.0197
	0.42	0.9965	0.0425
	0.43	0.9954	0.0674
	0.44	0.9942	0.0938
	0.45	0.9927	0.1215
	0.46	0.9910	0.1497
	0.47	0.9890	0.1781
	0.48	0.9868	0.2062
	0.49	0.9843	0.2337
	0.50	0.9815	0.2604
	0.51	0.9785	0.2860
	0.52	0.9752	0.3104
	0.53	0.9716	0.3335
	0.54	0.9678	0.3553
	0.55	0.9638	0.3758
	0.56	0.9595	0.3951
	0.57	0.9550	0.4131
	0.58	0.9504	0.4299
	0.59	0.9455	0.4457
	0.60	0.9405	0.4603
	0.61	0.9353	0.4741
	0.62	0.9301	0.4869
	0.63	0.9247	0.4988
	0.64	0.9192	0.5100
	0.65	0.9136	0.5205
	0.66	0.9080	0.5303
	0.67	0.9024	0.5395
	0.68	0.8967	0.5482
	0.69	0.8910	0.5563
	0.70	0.8853	0.5639
	0.71	0.8797	0.5711
	0.72	0.8740	0.5779
	0.73	0.8684	0.5843
	0.74	0.8628	0.5904
	0.75	0.8572	0.5961
	0.76	0.8517	0.6015
	0.77	0.8463	0.6067
	0.78	0.8409	0.6115
	0.79	0.8356	0.6162
	0.80	0.8303	0.6206
	0.81	0.8252	0.6248
	0.82	0.8201	0.6288
	0.83	0.8150	0.6327
	0.84	0.8101	0.6363
	0.85	0.8052	0.6399
	0.86	0.8004	0.6432
	0.87	0.7956	0.6464
	0.88	0.7910	0.6495
	0.89	0.7864	0.6525
	0.90	0.7819	0.6553
	0.91	0.7775	0.6581
	0.92	0.7731	0.6607
	0.93	0.7688	0.6633
	0.94	0.7646	0.6657
	0.95	0.7605	0.6681
	0.96	0.7564	0.6704
	0.97	0.7524	0.6726
	0.98	0.7484	0.6747
	0.99	0.7446	0.6768
	1.00	0.7407	0.6788
	1.01	0.7370	0.6807
	1.02	0.7333	0.6826
	1.03	0.7297	0.6844
	1.04	0.7261	0.6861
	1.05	0.7226	0.6879
	1.06	0.7191	0.6895
	1.07	0.7157	0.6911
	1.08	0.7123	0.6927
	1.09	0.7090	0.6942
	1.10	0.7057	0.6957
	1.11	0.7025	0.6972
	1.12	0.6994	0.6986
	1.13	0.6962	0.7000
	1.14	0.6932	0.7013
	1.15	0.6901	0.7026
	1.16	0.6872	0.7039
	1.17	0.6842	0.7051
	1.18	0.6813	0.7063
	1.19	0.6785	0.7075
	1.20	0.6756	0.7087
	1.21	0.6729	0.7098
	1.22	0.6701	0.7109
	1.23	0.6674	0.7120
	1.24	0.6647	0.7131
	1.25	0.6621	0.7141
	1.26	0.6595	0.7151
	1.27	0.6569	0.7161
	1.28	0.6544	0.7171
	1.29	0.6518	0.7180
	1.30	0.6494	0.7190
	1.31	0.6469	0.7199
	1.32	0.6445	0.7208
	1.33	0.6421	0.7217
	1.34	0.6398	0.7225
	1.35	0.6374	0.7234
	1.36	0.6351	0.7242
	1.37	0.6328	0.7250
	1.38	0.6306	0.7258
	1.39	0.6284	0.7266
	1.40	0.6262	0.7274
	1.41	0.6240	0.7281
	1.42	0.6218	0.7289
	1.43	0.6197	0.7296
	1.44	0.6176	0.7303
	1.45	0.6155	0.7310
	1.46	0.6134	0.7317
	1.47	0.6114	0.7324
	1.48	0.6094	0.7331
	1.49	0.6074	0.7337
	1.50	0.6054	0.7344
	1.51	0.6035	0.7350
	1.52	0.6015	0.7357
	1.53	0.5996	0.7363
	1.54	0.5977	0.7369
	1.55	0.5958	0.7375
	1.56	0.5940	0.7381
	1.57	0.5921	0.7387
	1.58	0.5903	0.7393
	1.59	0.5885	0.7398
	1.60	0.5867	0.7404
	1.61	0.5849	0.7409
	1.62	0.5831	0.7415
	1.63	0.5814	0.7420
	1.64	0.5797	0.7425
	1.65	0.5780	0.7431
	1.66	0.5763	0.7436
	1.67	0.5746	0.7441
	1.68	0.5729	0.7446
	1.69	0.5713	0.7451
	1.70	0.5696	0.7456
	1.71	0.5680	0.7460
	1.72	0.5664	0.7465
	1.73	0.5648	0.7470
	1.74	0.5632	0.7474
	1.75	0.5617	0.7479
	1.76	0.5601	0.7483
	1.77	0.5586	0.7488
	1.78	0.5570	0.7492
	1.79	0.5555	0.7497
	1.80	0.5540	0.7501
	1.81	0.5525	0.7505
	1.82	0.5510	0.7509
	1.83	0.5496	0.7513
	1.84	0.5481	0.7517
	1.85	0.5467	0.7521
	1.86	0.5452	0.7525
	1.87	0.5438	0.7529
	1.88	0.5424	0.7533
	1.89	0.5410	0.7537
	1.90	0.5396	0.7541
	1.91	0.5382	0.7545
	1.92	0.5369	0.7548
	1.93	0.5355	0.7552
	1.94	0.5342	0.7555
	1.95	0.5328	0.7559
	1.96	0.5315	0.7563
	1.97	0.5302	0.7566
	1.98	0.5289	0.7570
	1.99	0.5276	0.7573
	2.00	0.5263	0.7576
	2.01	0.5250	0.7580
	2.02	0.5237	0.7583
	2.03	0.5225	0.7586
	2.04	0.5212	0.7590
	2.05	0.5200	0.7593
	2.06	0.5188	0.7596
	2.07	0.5175	0.7599
	2.08	0.5163	0.7602
	2.09	0.5151	0.7605
	2.10	0.5139	0.7608
	2.11	0.5127	0.7611
	2.12	0.5115	0.7614
	2.13	0.5104	0.7617
	2.14	0.5092	0.7620
	2.15	0.5080	0.7623
	2.16	0.5069	0.7626
	2.17	0.5057	0.7629
	2.18	0.5046	0.7632
	2.19	0.5035	0.7634
	2.20	0.5024	0.7637
	2.21	0.5013	0.7640
	2.22	0.5002	0.7643
	2.23	0.4991	0.7645
	2.24	0.4980	0.7648
	2.25	0.4969	0.7651
	2.26	0.4958	0.7653
	2.27	0.4947	0.7656
	2.28	0.4937	0.7658
	2.29	0.4926	0.7661
	2.30	0.4916	0.7663
	2.31	0.4905	0.7666
	2.32	0.4895	0.7668
	2.33	0.4885	0.7671
	2.34	0.4875	0.7673
	2.35	0.4864	0.7676
	2.36	0.4854	0.7678
	2.37	0.4844	0.7680
	2.38	0.4834	0.7683
	2.39	0.4824	0.7685
	2.40	0.4815	0.7687
	2.41	0.4805	0.7690
	2.42	0.4795	0.7692
	2.43	0.4786	0.7694
	2.44	0.4776	0.7696
	2.45	0.4766	0.7698
	2.46	0.4757	0.7701
	2.47	0.4748	0.7703
	2.48	0.4738	0.7705
	2.49	0.4729	0.7707
	2.50	0.4720	0.7709
	2.51	0.4710	0.7711
	2.52	0.4701	0.7713
	2.53	0.4692	0.7715
	2.54	0.4683	0.7717
	2.55	0.4674	0.7719
	2.56	0.4665	0.7721
	2.57	0.4657	0.7723
	2.58	0.4648	0.7725
	2.59	0.4639	0.7727
	2.60	0.4630	0.7729
	2.61	0.4622	0.7731
	2.62	0.4613	0.7733
	2.63	0.4604	0.7735
	2.64	0.4596	0.7737
	2.65	0.4587	0.7739
	2.66	0.4579	0.7741
	2.67	0.4571	0.7743
	2.68	0.4562	0.7744
	2.69	0.4554	0.7746
	2.70	0.4546	0.7748
	2.71	0.4538	0.7750
	2.72	0.4530	0.7752
	2.73	0.4521	0.7753
	2.74	0.4513	0.7755
	2.75	0.4505	0.7757
	2.76	0.4497	0.7759
	2.77	0.4490	0.7760
	2.78	0.4482	0.7762
	2.79	0.4474	0.7764
	2.80	0.4466	0.7765
	2.81	0.4458	0.7767
	2.82	0.4451	0.7769
	2.83	0.4443	0.7770
	2.84	0.4435	0.7772
	2.85	0.4428	0.7774
	2.86	0.4420	0.7775
	2.87	0.4413	0.7777
	2.88	0.4405	0.7778
	2.89	0.4398	0.7780
	2.90	0.4390	0.7781
	2.91	0.4383	0.7783
	2.92	0.4376	0.7785
	2.93	0.4368	0.7786
	2.94	0.4361	0.7788
	2.95	0.4354	0.7789
	2.96	0.4347	0.7791
	2.97	0.4340	0.7792
	2.98	0.4333	0.7794
	2.99	0.4326	0.7795
	3.00	0.4319	0.7797
	3.01	0.4312	0.7798
	3.02	0.4305	0.7799
	3.03	0.4298	0.7801
	3.04	0.4291	0.7802
	3.05	0.4284	0.7804
	3.06	0.4277	0.7805
	3.07	0.4270	0.7806
	3.08	0.4264	0.7808
	3.09	0.4257	0.7809
	3.10	0.4250	0.7811
	3.11	0.4243	0.7812
	3.12	0.4237	0.7813
	3.13	0.4230	0.7815
	3.14	0.4224	0.7816
	3.15	0.4217	0.7817
	3.16	0.4211	0.7819
	3.17	0.4204	0.7820
	3.18	0.4198	0.7821
	3.19	0.4191	0.7823
	3.20	0.4185	0.7824
	3.21	0.4178	0.7825
	3.22	0.4172	0.7826
	3.23	0.4166	0.7828
	3.24	0.4159	0.7829
	3.25	0.4153	0.7830
	3.26	0.4147	0.7831
	3.27	0.4141	0.7833
	3.28	0.4135	0.7834
	3.29	0.4128	0.7835
	3.30	0.4122	0.7836
	3.31	0.4116	0.7837
	3.32	0.4110	0.7839
	3.33	0.4104	0.7840
	3.34	0.4098	0.7841
	3.35	0.4092	0.7842
	3.36	0.4086	0.7843
	3.37	0.4080	0.7845
	3.38	0.4074	0.7846
	3.39	0.4068	0.7847
	3.40	0.4062	0.7848
	3.41	0.4057	0.7849
	3.42	0.4051	0.7850
	3.43	0.4045	0.7851
	3.44	0.4039	0.7852
	3.45	0.4033	0.7854
	3.46	0.4028	0.7855
	3.47	0.4022	0.7856
	3.48	0.4016	0.7857
	3.49	0.4010	0.7858
	3.50	0.4005	0.7859
	3.51	0.3999	0.7860
	3.52	0.3994	0.7861
	3.53	0.3988	0.7862
	3.54	0.3982	0.7863
	3.55	0.3977	0.7864
	3.56	0.3971	0.7865
	3.57	0.3966	0.7866
	3.58	0.3960	0.7868
	3.59	0.3955	0.7869
	3.60	0.3949	0.7870
	3.61	0.3944	0.7871
	3.62	0.3939	0.7872
	3.63	0.3933	0.7873
	3.64	0.3928	0.7874
	3.65	0.3922	0.7875
	3.66	0.3917	0.7876
	3.67	0.3912	0.7877
	3.68	0.3906	0.7878
	3.69	0.3901	0.7879
	3.70	0.3896	0.7880
	3.71	0.3891	0.7881
	3.72	0.3885	0.7882
	3.73	0.3880	0.7882
	3.74	0.3875	0.7883
	3.75	0.3870	0.7884
	3.76	0.3865	0.7885
	3.77	0.3860	0.7886
	3.78	0.3854	0.7887
	3.79	0.3849	0.7888
	3.80	0.3844	0.7889
	3.81	0.3839	0.7890
	3.82	0.3834	0.7891
	3.83	0.3829	0.7892
	3.84	0.3824	0.7893
	3.85	0.3819	0.7894
	3.86	0.3814	0.7895
	3.87	0.3809	0.7895
	3.88	0.3804	0.7896
	3.89	0.3799	0.7897
	3.90	0.3794	0.7898
	3.91	0.3789	0.7899
	3.92	0.3785	0.7900
	3.93	0.3780	0.7901
	3.94	0.3775	0.7902
	3.95	0.3770	0.7902
	3.96	0.3765	0.7903
	3.97	0.3760	0.7904
	3.98	0.3756	0.7905
	3.99	0.3751	0.7906
	4.00	0.3746	0.7907
	4.01	0.3741	0.7908
	4.02	0.3736	0.7908
	4.03	0.3732	0.7909
	4.04	0.3727	0.7910
	4.05	0.3722	0.7911
	4.06	0.3718	0.7912
	4.07	0.3713	0.7913
	4.08	0.3708	0.7913
	4.09	0.3704	0.7914
}\tableCorrSiA

\pgfplotstableread{
	sigma	corr	si
	0.01	1.0000	0.0833
	0.02	1.0000	0.0833
	0.03	1.0000	0.0833
	0.04	1.0000	0.0833
	0.05	1.0000	0.0833
	0.06	1.0000	0.0833
	0.07	1.0000	0.0833
	0.08	1.0000	0.0833
	0.09	1.0000	0.0833
	0.10	1.0000	0.0833
	0.11	1.0000	0.0833
	0.12	1.0000	0.0833
	0.13	1.0000	0.0833
	0.14	1.0000	0.0833
	0.15	1.0000	0.0833
	0.16	1.0000	0.0833
	0.17	1.0000	0.0833
	0.18	1.0000	0.0833
	0.19	1.0000	0.0833
	0.20	1.0000	0.0833
	0.21	1.0000	0.0833
	0.22	1.0000	0.0833
	0.23	1.0000	0.0833
	0.24	1.0000	0.0833
	0.25	1.0000	0.0832
	0.26	1.0000	0.0832
	0.27	1.0000	0.0831
	0.28	1.0000	0.0829
	0.29	1.0000	0.0826
	0.30	1.0000	0.0823
	0.31	1.0000	0.0818
	0.32	0.9999	0.0811
	0.33	0.9999	0.0803
	0.34	0.9998	0.0792
	0.35	0.9997	0.0778
	0.36	0.9995	0.0760
	0.37	0.9993	0.0738
	0.38	0.9990	0.0712
	0.39	0.9986	0.0681
	0.40	0.9980	0.0644
	0.41	0.9974	0.0601
	0.42	0.9965	0.0551
	0.43	0.9955	0.0495
	0.44	0.9942	0.0434
	0.45	0.9928	0.0368
	0.46	0.9911	0.0297
	0.47	0.9892	0.0222
	0.48	0.9870	0.0143
	0.49	0.9845	0.0062
	0.50	0.9818	0.0022
	0.51	0.9788	0.0108
	0.52	0.9756	0.0195
	0.53	0.9721	0.0283
	0.54	0.9684	0.0371
	0.55	0.9644	0.0460
	0.56	0.9602	0.0547
	0.57	0.9558	0.0634
	0.58	0.9512	0.0720
	0.59	0.9465	0.0805
	0.60	0.9415	0.0888
	0.61	0.9365	0.0969
	0.62	0.9313	0.1049
	0.63	0.9260	0.1126
	0.64	0.9206	0.1202
	0.65	0.9152	0.1276
	0.66	0.9097	0.1348
	0.67	0.9041	0.1417
	0.68	0.8986	0.1485
	0.69	0.8930	0.1550
	0.70	0.8874	0.1614
	0.71	0.8818	0.1676
	0.72	0.8762	0.1735
	0.73	0.8706	0.1793
	0.74	0.8651	0.1849
	0.75	0.8596	0.1903
	0.76	0.8542	0.1956
	0.77	0.8488	0.2007
	0.78	0.8435	0.2056
	0.79	0.8382	0.2104
	0.80	0.8330	0.2150
	0.81	0.8279	0.2195
	0.82	0.8228	0.2238
	0.83	0.8178	0.2281
	0.84	0.8129	0.2322
	0.85	0.8080	0.2361
	0.86	0.8032	0.2400
	0.87	0.7985	0.2437
	0.88	0.7938	0.2474
	0.89	0.7892	0.2509
	0.90	0.7847	0.2544
	0.91	0.7803	0.2577
	0.92	0.7759	0.2610
	0.93	0.7716	0.2642
	0.94	0.7674	0.2673
	0.95	0.7632	0.2703
	0.96	0.7591	0.2732
	0.97	0.7551	0.2761
	0.98	0.7511	0.2789
	0.99	0.7472	0.2816
	1.00	0.7433	0.2842
	1.01	0.7395	0.2868
	1.02	0.7358	0.2894
	1.03	0.7321	0.2919
	1.04	0.7285	0.2943
	1.05	0.7249	0.2967
	1.06	0.7214	0.2990
	1.07	0.7179	0.3012
	1.08	0.7145	0.3035
	1.09	0.7111	0.3056
	1.10	0.7078	0.3078
	1.11	0.7046	0.3098
	1.12	0.7013	0.3119
	1.13	0.6982	0.3139
	1.14	0.6950	0.3158
	1.15	0.6920	0.3178
	1.16	0.6889	0.3197
	1.17	0.6859	0.3215
	1.18	0.6829	0.3233
	1.19	0.6800	0.3251
	1.20	0.6771	0.3268
	1.21	0.6743	0.3286
	1.22	0.6715	0.3302
	1.23	0.6687	0.3319
	1.24	0.6660	0.3335
	1.25	0.6633	0.3351
	1.26	0.6606	0.3367
	1.27	0.6580	0.3382
	1.28	0.6554	0.3398
	1.29	0.6529	0.3413
	1.30	0.6503	0.3427
	1.31	0.6478	0.3442
	1.32	0.6454	0.3456
	1.33	0.6429	0.3470
	1.34	0.6405	0.3484
	1.35	0.6381	0.3497
	1.36	0.6358	0.3511
	1.37	0.6334	0.3524
	1.38	0.6311	0.3537
	1.39	0.6289	0.3550
	1.40	0.6266	0.3562
	1.41	0.6244	0.3574
	1.42	0.6222	0.3587
	1.43	0.6200	0.3599
	1.44	0.6179	0.3611
	1.45	0.6158	0.3622
	1.46	0.6137	0.3634
	1.47	0.6116	0.3645
	1.48	0.6095	0.3656
	1.49	0.6075	0.3667
	1.50	0.6055	0.3678
	1.51	0.6035	0.3689
	1.52	0.6015	0.3700
	1.53	0.5996	0.3710
	1.54	0.5977	0.3721
	1.55	0.5957	0.3731
	1.56	0.5939	0.3741
	1.57	0.5920	0.3751
	1.58	0.5901	0.3761
	1.59	0.5883	0.3770
	1.60	0.5865	0.3780
	1.61	0.5847	0.3789
	1.62	0.5829	0.3799
	1.63	0.5811	0.3808
	1.64	0.5794	0.3817
	1.65	0.5777	0.3826
	1.66	0.5760	0.3835
	1.67	0.5743	0.3844
	1.68	0.5726	0.3853
	1.69	0.5709	0.3861
	1.70	0.5693	0.3870
	1.71	0.5676	0.3878
	1.72	0.5660	0.3887
	1.73	0.5644	0.3895
	1.74	0.5628	0.3903
	1.75	0.5612	0.3911
	1.76	0.5597	0.3919
	1.77	0.5581	0.3927
	1.78	0.5566	0.3935
	1.79	0.5551	0.3942
	1.80	0.5535	0.3950
	1.81	0.5520	0.3957
	1.82	0.5506	0.3965
	1.83	0.5491	0.3972
	1.84	0.5476	0.3980
	1.85	0.5462	0.3987
	1.86	0.5448	0.3994
	1.87	0.5433	0.4001
	1.88	0.5419	0.4008
	1.89	0.5405	0.4015
	1.90	0.5391	0.4022
	1.91	0.5378	0.4029
	1.92	0.5364	0.4035
	1.93	0.5350	0.4042
	1.94	0.5337	0.4049
	1.95	0.5324	0.4055
	1.96	0.5310	0.4062
	1.97	0.5297	0.4068
	1.98	0.5284	0.4075
	1.99	0.5271	0.4081
	2.00	0.5258	0.4087
	2.01	0.5246	0.4093
	2.02	0.5233	0.4099
	2.03	0.5220	0.4105
	2.04	0.5208	0.4111
	2.05	0.5196	0.4117
	2.06	0.5183	0.4123
	2.07	0.5171	0.4129
	2.08	0.5159	0.4135
	2.09	0.5147	0.4141
	2.10	0.5135	0.4146
	2.11	0.5123	0.4152
	2.12	0.5112	0.4158
	2.13	0.5100	0.4163
	2.14	0.5088	0.4169
	2.15	0.5077	0.4174
	2.16	0.5066	0.4180
	2.17	0.5054	0.4185
	2.18	0.5043	0.4190
	2.19	0.5032	0.4196
	2.20	0.5021	0.4201
	2.21	0.5010	0.4206
	2.22	0.4999	0.4211
	2.23	0.4988	0.4216
	2.24	0.4977	0.4221
	2.25	0.4967	0.4226
	2.26	0.4956	0.4231
	2.27	0.4945	0.4236
	2.28	0.4935	0.4241
	2.29	0.4924	0.4246
	2.30	0.4914	0.4251
	2.31	0.4904	0.4255
	2.32	0.4894	0.4260
	2.33	0.4883	0.4265
	2.34	0.4873	0.4270
	2.35	0.4863	0.4274
	2.36	0.4853	0.4279
	2.37	0.4844	0.4283
	2.38	0.4834	0.4288
	2.39	0.4824	0.4292
	2.40	0.4814	0.4297
	2.41	0.4805	0.4301
	2.42	0.4795	0.4306
	2.43	0.4786	0.4310
	2.44	0.4776	0.4314
	2.45	0.4767	0.4319
	2.46	0.4757	0.4323
	2.47	0.4748	0.4327
	2.48	0.4739	0.4331
	2.49	0.4730	0.4335
	2.50	0.4721	0.4340
	2.51	0.4712	0.4344
	2.52	0.4703	0.4348
	2.53	0.4694	0.4352
	2.54	0.4685	0.4356
	2.55	0.4676	0.4360
	2.56	0.4667	0.4364
	2.57	0.4659	0.4368
	2.58	0.4650	0.4372
	2.59	0.4641	0.4375
	2.60	0.4633	0.4379
	2.61	0.4624	0.4383
	2.62	0.4616	0.4387
	2.63	0.4608	0.4391
	2.64	0.4599	0.4394
	2.65	0.4591	0.4398
	2.66	0.4583	0.4402
	2.67	0.4574	0.4406
	2.68	0.4566	0.4409
	2.69	0.4558	0.4413
	2.70	0.4550	0.4416
	2.71	0.4542	0.4420
	2.72	0.4534	0.4424
	2.73	0.4526	0.4427
	2.74	0.4518	0.4431
	2.75	0.4511	0.4434
	2.76	0.4503	0.4437
	2.77	0.4495	0.4441
	2.78	0.4487	0.4444
	2.79	0.4480	0.4448
	2.80	0.4472	0.4451
	2.81	0.4465	0.4454
	2.82	0.4457	0.4458
	2.83	0.4450	0.4461
	2.84	0.4442	0.4464
	2.85	0.4435	0.4468
	2.86	0.4427	0.4471
	2.87	0.4420	0.4474
	2.88	0.4413	0.4477
	2.89	0.4406	0.4480
	2.90	0.4398	0.4484
	2.91	0.4391	0.4487
	2.92	0.4384	0.4490
	2.93	0.4377	0.4493
	2.94	0.4370	0.4496
	2.95	0.4363	0.4499
	2.96	0.4356	0.4502
	2.97	0.4349	0.4505
	2.98	0.4342	0.4508
	2.99	0.4335	0.4511
	3.00	0.4329	0.4514
	3.01	0.4322	0.4517
	3.02	0.4315	0.4520
	3.03	0.4308	0.4523
	3.04	0.4302	0.4526
	3.05	0.4295	0.4529
	3.06	0.4288	0.4532
	3.07	0.4282	0.4535
	3.08	0.4275	0.4537
	3.09	0.4269	0.4540
	3.10	0.4262	0.4543
	3.11	0.4256	0.4546
	3.12	0.4250	0.4549
	3.13	0.4243	0.4551
	3.14	0.4237	0.4554
	3.15	0.4231	0.4557
	3.16	0.4224	0.4560
	3.17	0.4218	0.4562
	3.18	0.4212	0.4565
	3.19	0.4206	0.4568
	3.20	0.4200	0.4570
	3.21	0.4193	0.4573
	3.22	0.4187	0.4576
	3.23	0.4181	0.4578
	3.24	0.4175	0.4581
	3.25	0.4169	0.4583
	3.26	0.4163	0.4586
	3.27	0.4157	0.4588
	3.28	0.4151	0.4591
	3.29	0.4146	0.4594
	3.30	0.4140	0.4596
	3.31	0.4134	0.4599
	3.32	0.4128	0.4601
	3.33	0.4122	0.4603
	3.34	0.4117	0.4606
	3.35	0.4111	0.4608
	3.36	0.4105	0.4611
	3.37	0.4100	0.4613
	3.38	0.4094	0.4616
	3.39	0.4088	0.4618
	3.40	0.4083	0.4620
	3.41	0.4077	0.4623
	3.42	0.4072	0.4625
	3.43	0.4066	0.4628
	3.44	0.4061	0.4630
	3.45	0.4055	0.4632
	3.46	0.4050	0.4635
	3.47	0.4044	0.4637
	3.48	0.4039	0.4639
	3.49	0.4034	0.4641
	3.50	0.4028	0.4644
	3.51	0.4023	0.4646
	3.52	0.4018	0.4648
	3.53	0.4013	0.4650
	3.54	0.4007	0.4653
	3.55	0.4002	0.4655
	3.56	0.3997	0.4657
	3.57	0.3992	0.4659
	3.58	0.3987	0.4661
	3.59	0.3982	0.4664
	3.60	0.3976	0.4666
	3.61	0.3971	0.4668
	3.62	0.3966	0.4670
	3.63	0.3961	0.4672
	3.64	0.3956	0.4674
	3.65	0.3951	0.4676
	3.66	0.3946	0.4678
	3.67	0.3941	0.4681
	3.68	0.3937	0.4683
	3.69	0.3932	0.4685
	3.70	0.3927	0.4687
	3.71	0.3922	0.4689
	3.72	0.3917	0.4691
	3.73	0.3912	0.4693
	3.74	0.3908	0.4695
	3.75	0.3903	0.4697
	3.76	0.3898	0.4699
	3.77	0.3893	0.4701
	3.78	0.3889	0.4703
	3.79	0.3884	0.4705
	3.80	0.3879	0.4707
	3.81	0.3874	0.4709
	3.82	0.3870	0.4711
	3.83	0.3865	0.4713
	3.84	0.3861	0.4715
	3.85	0.3856	0.4717
	3.86	0.3851	0.4718
	3.87	0.3847	0.4720
	3.88	0.3842	0.4722
	3.89	0.3838	0.4724
	3.90	0.3833	0.4726
	3.91	0.3829	0.4728
	3.92	0.3824	0.4730
	3.93	0.3820	0.4732
	3.94	0.3816	0.4734
	3.95	0.3811	0.4735
	3.96	0.3807	0.4737
	3.97	0.3802	0.4739
	3.98	0.3798	0.4741
	3.99	0.3794	0.4743
	4.00	0.3789	0.4744
	4.01	0.3785	0.4746
	4.02	0.3781	0.4748
	4.03	0.3777	0.4750
	4.04	0.3772	0.4752
	4.05	0.3768	0.4753
	4.06	0.3764	0.4755
	4.07	0.3760	0.4757
	4.08	0.3755	0.4759
	4.09	0.3751	0.4760
}\tableCorrSiB

\pgfplotstableread{
	sigma	corr	si
	0.01	1.0000	0.0833
	0.02	1.0000	0.0833
	0.03	1.0000	0.0833
	0.04	1.0000	0.0833
	0.05	1.0000	0.0833
	0.06	1.0000	0.0833
	0.07	1.0000	0.0833
	0.08	1.0000	0.0833
	0.09	1.0000	0.0833
	0.10	1.0000	0.0833
	0.11	1.0000	0.0833
	0.12	1.0000	0.0833
	0.13	1.0000	0.0833
	0.14	1.0000	0.0833
	0.15	1.0000	0.0833
	0.16	1.0000	0.0833
	0.17	1.0000	0.0833
	0.18	1.0000	0.0833
	0.19	1.0000	0.0833
	0.20	1.0000	0.0833
	0.21	1.0000	0.0833
	0.22	1.0000	0.0833
	0.23	1.0000	0.0833
	0.24	1.0000	0.0833
	0.25	1.0000	0.0833
	0.26	1.0000	0.0832
	0.27	1.0000	0.0831
	0.28	1.0000	0.0830
	0.29	1.0000	0.0827
	0.30	1.0000	0.0825
	0.31	1.0000	0.0821
	0.32	0.9999	0.0815
	0.33	0.9999	0.0808
	0.34	0.9998	0.0800
	0.35	0.9997	0.0789
	0.36	0.9996	0.0775
	0.37	0.9993	0.0759
	0.38	0.9990	0.0739
	0.39	0.9986	0.0715
	0.40	0.9981	0.0687
	0.41	0.9975	0.0655
	0.42	0.9966	0.0619
	0.43	0.9956	0.0577
	0.44	0.9944	0.0532
	0.45	0.9930	0.0483
	0.46	0.9914	0.0431
	0.47	0.9896	0.0376
	0.48	0.9874	0.0318
	0.49	0.9851	0.0258
	0.50	0.9825	0.0197
	0.51	0.9796	0.0134
	0.52	0.9765	0.0071
	0.53	0.9731	0.0006
	0.54	0.9696	0.0058
	0.55	0.9657	0.0123
	0.56	0.9617	0.0187
	0.57	0.9575	0.0251
	0.58	0.9531	0.0315
	0.59	0.9485	0.0377
	0.60	0.9438	0.0439
	0.61	0.9389	0.0499
	0.62	0.9339	0.0558
	0.63	0.9288	0.0617
	0.64	0.9236	0.0673
	0.65	0.9184	0.0729
	0.66	0.9131	0.0783
	0.67	0.9077	0.0836
	0.68	0.9023	0.0887
	0.69	0.8969	0.0938
	0.70	0.8915	0.0986
	0.71	0.8861	0.1034
	0.72	0.8807	0.1080
	0.73	0.8753	0.1126
	0.74	0.8699	0.1170
	0.75	0.8646	0.1213
	0.76	0.8593	0.1254
	0.77	0.8540	0.1295
	0.78	0.8488	0.1335
	0.79	0.8437	0.1373
	0.80	0.8386	0.1411
	0.81	0.8335	0.1448
	0.82	0.8285	0.1483
	0.83	0.8236	0.1518
	0.84	0.8187	0.1552
	0.85	0.8139	0.1586
	0.86	0.8092	0.1618
	0.87	0.8045	0.1650
	0.88	0.7999	0.1681
	0.89	0.7954	0.1711
	0.90	0.7909	0.1741
	0.91	0.7864	0.1770
	0.92	0.7821	0.1798
	0.93	0.7778	0.1826
	0.94	0.7735	0.1853
	0.95	0.7694	0.1880
	0.96	0.7652	0.1906
	0.97	0.7612	0.1931
	0.98	0.7571	0.1956
	0.99	0.7532	0.1981
	1.00	0.7493	0.2005
	1.01	0.7454	0.2028
	1.02	0.7416	0.2051
	1.03	0.7379	0.2074
	1.04	0.7342	0.2096
	1.05	0.7306	0.2118
	1.06	0.7270	0.2139
	1.07	0.7234	0.2160
	1.08	0.7199	0.2181
	1.09	0.7165	0.2201
	1.10	0.7131	0.2220
	1.11	0.7097	0.2240
	1.12	0.7064	0.2259
	1.13	0.7031	0.2278
	1.14	0.6999	0.2296
	1.15	0.6967	0.2314
	1.16	0.6936	0.2332
	1.17	0.6905	0.2350
	1.18	0.6874	0.2367
	1.19	0.6843	0.2384
	1.20	0.6814	0.2401
	1.21	0.6784	0.2417
	1.22	0.6755	0.2433
	1.23	0.6726	0.2449
	1.24	0.6697	0.2465
	1.25	0.6669	0.2480
	1.26	0.6641	0.2495
	1.27	0.6614	0.2510
	1.28	0.6586	0.2525
	1.29	0.6560	0.2539
	1.30	0.6533	0.2553
	1.31	0.6507	0.2567
	1.32	0.6481	0.2581
	1.33	0.6455	0.2595
	1.34	0.6430	0.2608
	1.35	0.6405	0.2621
	1.36	0.6380	0.2634
	1.37	0.6355	0.2647
	1.38	0.6331	0.2660
	1.39	0.6307	0.2672
	1.40	0.6283	0.2685
	1.41	0.6260	0.2697
	1.42	0.6236	0.2709
	1.43	0.6213	0.2721
	1.44	0.6191	0.2732
	1.45	0.6168	0.2744
	1.46	0.6146	0.2755
	1.47	0.6124	0.2766
	1.48	0.6102	0.2777
	1.49	0.6080	0.2788
	1.50	0.6059	0.2799
	1.51	0.6038	0.2810
	1.52	0.6017	0.2820
	1.53	0.5996	0.2830
	1.54	0.5976	0.2841
	1.55	0.5955	0.2851
	1.56	0.5935	0.2861
	1.57	0.5915	0.2871
	1.58	0.5896	0.2880
	1.59	0.5876	0.2890
	1.60	0.5857	0.2899
	1.61	0.5838	0.2909
	1.62	0.5819	0.2918
	1.63	0.5800	0.2927
	1.64	0.5782	0.2936
	1.65	0.5763	0.2945
	1.66	0.5745	0.2954
	1.67	0.5727	0.2963
	1.68	0.5709	0.2971
	1.69	0.5691	0.2980
	1.70	0.5674	0.2988
	1.71	0.5656	0.2997
	1.72	0.5639	0.3005
	1.73	0.5622	0.3013
	1.74	0.5605	0.3021
	1.75	0.5588	0.3029
	1.76	0.5572	0.3037
	1.77	0.5555	0.3045
	1.78	0.5539	0.3053
	1.79	0.5523	0.3060
	1.80	0.5507	0.3068
	1.81	0.5491	0.3076
	1.82	0.5475	0.3083
	1.83	0.5460	0.3090
	1.84	0.5444	0.3098
	1.85	0.5429	0.3105
	1.86	0.5414	0.3112
	1.87	0.5399	0.3119
	1.88	0.5384	0.3126
	1.89	0.5369	0.3133
	1.90	0.5354	0.3140
	1.91	0.5340	0.3147
	1.92	0.5325	0.3153
	1.93	0.5311	0.3160
	1.94	0.5297	0.3167
	1.95	0.5283	0.3173
	1.96	0.5269	0.3180
	1.97	0.5255	0.3186
	1.98	0.5241	0.3192
	1.99	0.5228	0.3199
	2.00	0.5214	0.3205
	2.01	0.5201	0.3211
	2.02	0.5188	0.3217
	2.03	0.5174	0.3223
	2.04	0.5161	0.3229
	2.05	0.5148	0.3235
	2.06	0.5136	0.3241
	2.07	0.5123	0.3247
	2.08	0.5110	0.3253
	2.09	0.5098	0.3258
	2.10	0.5085	0.3264
	2.11	0.5073	0.3270
	2.12	0.5061	0.3275
	2.13	0.5049	0.3281
	2.14	0.5036	0.3286
	2.15	0.5025	0.3292
	2.16	0.5013	0.3297
	2.17	0.5001	0.3302
	2.18	0.4989	0.3308
	2.19	0.4978	0.3313
	2.20	0.4966	0.3318
	2.21	0.4955	0.3323
	2.22	0.4943	0.3329
	2.23	0.4932	0.3334
	2.24	0.4921	0.3339
	2.25	0.4910	0.3344
	2.26	0.4899	0.3349
	2.27	0.4888	0.3354
	2.28	0.4877	0.3358
	2.29	0.4867	0.3363
	2.30	0.4856	0.3368
	2.31	0.4845	0.3373
	2.32	0.4835	0.3378
	2.33	0.4824	0.3382
	2.34	0.4814	0.3387
	2.35	0.4804	0.3392
	2.36	0.4794	0.3396
	2.37	0.4783	0.3401
	2.38	0.4773	0.3405
	2.39	0.4763	0.3410
	2.40	0.4753	0.3414
	2.41	0.4744	0.3419
	2.42	0.4734	0.3423
	2.43	0.4724	0.3428
	2.44	0.4714	0.3432
	2.45	0.4705	0.3436
	2.46	0.4695	0.3440
	2.47	0.4686	0.3445
	2.48	0.4677	0.3449
	2.49	0.4667	0.3453
	2.50	0.4658	0.3457
	2.51	0.4649	0.3461
	2.52	0.4640	0.3465
	2.53	0.4631	0.3469
	2.54	0.4622	0.3474
	2.55	0.4613	0.3478
	2.56	0.4604	0.3481
	2.57	0.4595	0.3485
	2.58	0.4586	0.3489
	2.59	0.4578	0.3493
	2.60	0.4569	0.3497
	2.61	0.4560	0.3501
	2.62	0.4552	0.3505
	2.63	0.4543	0.3509
	2.64	0.4535	0.3512
	2.65	0.4527	0.3516
	2.66	0.4518	0.3520
	2.67	0.4510	0.3524
	2.68	0.4502	0.3527
	2.69	0.4494	0.3531
	2.70	0.4485	0.3535
	2.71	0.4477	0.3538
	2.72	0.4469	0.3542
	2.73	0.4461	0.3545
	2.74	0.4453	0.3549
	2.75	0.4446	0.3552
	2.76	0.4438	0.3556
	2.77	0.4430	0.3559
	2.78	0.4422	0.3563
	2.79	0.4415	0.3566
	2.80	0.4407	0.3570
	2.81	0.4399	0.3573
	2.82	0.4392	0.3576
	2.83	0.4384	0.3580
	2.84	0.4377	0.3583
	2.85	0.4369	0.3586
	2.86	0.4362	0.3590
	2.87	0.4355	0.3593
	2.88	0.4348	0.3596
	2.89	0.4340	0.3599
	2.90	0.4333	0.3603
	2.91	0.4326	0.3606
	2.92	0.4319	0.3609
	2.93	0.4312	0.3612
	2.94	0.4305	0.3615
	2.95	0.4298	0.3618
	2.96	0.4291	0.3621
	2.97	0.4284	0.3624
	2.98	0.4277	0.3628
	2.99	0.4270	0.3631
	3.00	0.4263	0.3634
	3.01	0.4256	0.3637
	3.02	0.4250	0.3640
	3.03	0.4243	0.3643
	3.04	0.4236	0.3646
	3.05	0.4230	0.3649
	3.06	0.4223	0.3651
	3.07	0.4217	0.3654
	3.08	0.4210	0.3657
	3.09	0.4204	0.3660
	3.10	0.4197	0.3663
	3.11	0.4191	0.3666
	3.12	0.4184	0.3669
	3.13	0.4178	0.3672
	3.14	0.4172	0.3674
	3.15	0.4165	0.3677
	3.16	0.4159	0.3680
	3.17	0.4153	0.3683
	3.18	0.4147	0.3685
	3.19	0.4140	0.3688
	3.20	0.4134	0.3691
	3.21	0.4128	0.3694
	3.22	0.4122	0.3696
	3.23	0.4116	0.3699
	3.24	0.4110	0.3702
	3.25	0.4104	0.3704
	3.26	0.4098	0.3707
	3.27	0.4092	0.3710
	3.28	0.4086	0.3712
	3.29	0.4080	0.3715
	3.30	0.4074	0.3717
	3.31	0.4068	0.3720
	3.32	0.4063	0.3722
	3.33	0.4057	0.3725
	3.34	0.4051	0.3728
	3.35	0.4045	0.3730
	3.36	0.4040	0.3733
	3.37	0.4034	0.3735
	3.38	0.4028	0.3738
	3.39	0.4023	0.3740
	3.40	0.4017	0.3743
	3.41	0.4012	0.3745
	3.42	0.4006	0.3747
	3.43	0.4000	0.3750
	3.44	0.3995	0.3752
	3.45	0.3990	0.3755
	3.46	0.3984	0.3757
	3.47	0.3979	0.3759
	3.48	0.3973	0.3762
	3.49	0.3968	0.3764
	3.50	0.3962	0.3767
	3.51	0.3957	0.3769
	3.52	0.3952	0.3771
	3.53	0.3946	0.3774
	3.54	0.3941	0.3776
	3.55	0.3936	0.3778
	3.56	0.3931	0.3780
	3.57	0.3925	0.3783
	3.58	0.3920	0.3785
	3.59	0.3915	0.3787
	3.60	0.3910	0.3789
	3.61	0.3905	0.3792
	3.62	0.3900	0.3794
	3.63	0.3895	0.3796
	3.64	0.3890	0.3798
	3.65	0.3885	0.3801
	3.66	0.3879	0.3803
	3.67	0.3874	0.3805
	3.68	0.3869	0.3807
	3.69	0.3865	0.3809
	3.70	0.3860	0.3811
	3.71	0.3855	0.3814
	3.72	0.3850	0.3816
	3.73	0.3845	0.3818
	3.74	0.3840	0.3820
	3.75	0.3835	0.3822
	3.76	0.3830	0.3824
	3.77	0.3825	0.3826
	3.78	0.3821	0.3828
	3.79	0.3816	0.3830
	3.80	0.3811	0.3832
	3.81	0.3806	0.3834
	3.82	0.3802	0.3836
	3.83	0.3797	0.3838
	3.84	0.3792	0.3840
	3.85	0.3787	0.3842
	3.86	0.3783	0.3844
	3.87	0.3778	0.3846
	3.88	0.3773	0.3849
	3.89	0.3769	0.3851
	3.90	0.3764	0.3852
	3.91	0.3760	0.3854
	3.92	0.3755	0.3856
	3.93	0.3751	0.3858
	3.94	0.3746	0.3860
	3.95	0.3741	0.3862
	3.96	0.3737	0.3864
	3.97	0.3732	0.3866
	3.98	0.3728	0.3868
	3.99	0.3724	0.3870
	4.00	0.3719	0.3872
	4.01	0.3715	0.3874
	4.02	0.3710	0.3876
	4.03	0.3706	0.3877
	4.04	0.3701	0.3879
	4.05	0.3697	0.3881
	4.06	0.3693	0.3883
	4.07	0.3688	0.3885
	4.08	0.3684	0.3887
	4.09	0.3680	0.3888
}\tableCorrSiC

\pgfplotstableread{
	sigma	corr	si
	0.01	1.0000	0.0833
	0.02	1.0000	0.0833
	0.03	1.0000	0.0833
	0.04	1.0000	0.0833
	0.05	1.0000	0.0833
	0.06	1.0000	0.0833
	0.07	1.0000	0.0833
	0.08	1.0000	0.0833
	0.09	1.0000	0.0833
	0.10	1.0000	0.0833
	0.11	1.0000	0.0833
	0.12	1.0000	0.0833
	0.13	1.0000	0.0833
	0.14	1.0000	0.0833
	0.15	1.0000	0.0833
	0.16	1.0000	0.0833
	0.17	1.0000	0.0833
	0.18	1.0000	0.0833
	0.19	1.0000	0.0833
	0.20	1.0000	0.0833
	0.21	1.0000	0.0833
	0.22	1.0000	0.0833
	0.23	1.0000	0.0833
	0.24	1.0000	0.0833
	0.25	1.0000	0.0832
	0.26	1.0000	0.0831
	0.27	1.0000	0.0829
	0.28	1.0000	0.0827
	0.29	1.0000	0.0823
	0.30	1.0000	0.0818
	0.31	1.0000	0.0811
	0.32	0.9999	0.0801
	0.33	0.9999	0.0788
	0.34	0.9998	0.0771
	0.35	0.9997	0.0749
	0.36	0.9995	0.0722
	0.37	0.9993	0.0687
	0.38	0.9990	0.0644
	0.39	0.9986	0.0591
	0.40	0.9980	0.0529
	0.41	0.9973	0.0456
	0.42	0.9964	0.0372
	0.43	0.9954	0.0278
	0.44	0.9941	0.0175
	0.45	0.9926	0.0064
	0.46	0.9909	0.0056
	0.47	0.9889	0.0182
	0.48	0.9867	0.0314
	0.49	0.9842	0.0450
	0.50	0.9814	0.0590
	0.51	0.9784	0.0732
	0.52	0.9751	0.0875
	0.53	0.9716	0.1018
	0.54	0.9678	0.1161
	0.55	0.9638	0.1301
	0.56	0.9595	0.1439
	0.57	0.9551	0.1573
	0.58	0.9505	0.1704
	0.59	0.9457	0.1830
	0.60	0.9407	0.1953
	0.61	0.9356	0.2072
	0.62	0.9304	0.2186
	0.63	0.9251	0.2295
	0.64	0.9197	0.2401
	0.65	0.9143	0.2502
	0.66	0.9088	0.2598
	0.67	0.9032	0.2691
	0.68	0.8977	0.2780
	0.69	0.8921	0.2865
	0.70	0.8865	0.2946
	0.71	0.8810	0.3024
	0.72	0.8755	0.3098
	0.73	0.8700	0.3170
	0.74	0.8645	0.3238
	0.75	0.8591	0.3304
	0.76	0.8538	0.3366
	0.77	0.8485	0.3427
	0.78	0.8432	0.3485
	0.79	0.8381	0.3540
	0.80	0.8330	0.3594
	0.81	0.8280	0.3645
	0.82	0.8230	0.3695
	0.83	0.8181	0.3742
	0.84	0.8133	0.3788
	0.85	0.8086	0.3833
	0.86	0.8040	0.3875
	0.87	0.7994	0.3916
	0.88	0.7949	0.3956
	0.89	0.7905	0.3995
	0.90	0.7861	0.4032
	0.91	0.7818	0.4068
	0.92	0.7776	0.4103
	0.93	0.7735	0.4137
	0.94	0.7694	0.4170
	0.95	0.7654	0.4202
	0.96	0.7615	0.4232
	0.97	0.7576	0.4262
	0.98	0.7538	0.4292
	0.99	0.7501	0.4320
	1.00	0.7464	0.4347
	1.01	0.7427	0.4374
	1.02	0.7392	0.4400
	1.03	0.7357	0.4426
	1.04	0.7322	0.4450
	1.05	0.7288	0.4475
	1.06	0.7254	0.4498
	1.07	0.7221	0.4521
	1.08	0.7189	0.4543
	1.09	0.7157	0.4565
	1.10	0.7125	0.4587
	1.11	0.7094	0.4608
	1.12	0.7063	0.4628
	1.13	0.7033	0.4648
	1.14	0.7003	0.4668
	1.15	0.6974	0.4687
	1.16	0.6945	0.4705
	1.17	0.6916	0.4724
	1.18	0.6888	0.4742
	1.19	0.6860	0.4759
	1.20	0.6832	0.4776
	1.21	0.6805	0.4793
	1.22	0.6778	0.4810
	1.23	0.6751	0.4826
	1.24	0.6725	0.4842
	1.25	0.6699	0.4857
	1.26	0.6674	0.4873
	1.27	0.6648	0.4888
	1.28	0.6623	0.4903
	1.29	0.6599	0.4917
	1.30	0.6574	0.4931
	1.31	0.6550	0.4945
	1.32	0.6526	0.4959
	1.33	0.6503	0.4973
	1.34	0.6479	0.4986
	1.35	0.6456	0.4999
	1.36	0.6433	0.5012
	1.37	0.6411	0.5024
	1.38	0.6388	0.5037
	1.39	0.6366	0.5049
	1.40	0.6344	0.5061
	1.41	0.6323	0.5073
	1.42	0.6301	0.5085
	1.43	0.6280	0.5096
	1.44	0.6259	0.5107
	1.45	0.6238	0.5119
	1.46	0.6218	0.5130
	1.47	0.6197	0.5140
	1.48	0.6177	0.5151
	1.49	0.6157	0.5162
	1.50	0.6137	0.5172
	1.51	0.6118	0.5182
	1.52	0.6098	0.5192
	1.53	0.6079	0.5202
	1.54	0.6060	0.5212
	1.55	0.6041	0.5221
	1.56	0.6022	0.5231
	1.57	0.6004	0.5240
	1.58	0.5985	0.5250
	1.59	0.5967	0.5259
	1.60	0.5949	0.5268
	1.61	0.5931	0.5277
	1.62	0.5914	0.5285
	1.63	0.5896	0.5294
	1.64	0.5878	0.5303
	1.65	0.5861	0.5311
	1.66	0.5844	0.5319
	1.67	0.5827	0.5328
	1.68	0.5810	0.5336
	1.69	0.5793	0.5344
	1.70	0.5777	0.5352
	1.71	0.5760	0.5359
	1.72	0.5744	0.5367
	1.73	0.5728	0.5375
	1.74	0.5712	0.5382
	1.75	0.5696	0.5390
	1.76	0.5680	0.5397
	1.77	0.5665	0.5404
	1.78	0.5649	0.5412
	1.79	0.5634	0.5419
	1.80	0.5618	0.5426
	1.81	0.5603	0.5433
	1.82	0.5588	0.5440
	1.83	0.5573	0.5446
	1.84	0.5558	0.5453
	1.85	0.5544	0.5460
	1.86	0.5529	0.5466
	1.87	0.5514	0.5473
	1.88	0.5500	0.5479
	1.89	0.5486	0.5486
	1.90	0.5472	0.5492
	1.91	0.5458	0.5498
	1.92	0.5444	0.5504
	1.93	0.5430	0.5510
	1.94	0.5416	0.5516
	1.95	0.5402	0.5522
	1.96	0.5389	0.5528
	1.97	0.5375	0.5534
	1.98	0.5362	0.5540
	1.99	0.5349	0.5546
	2.00	0.5335	0.5551
	2.01	0.5322	0.5557
	2.02	0.5309	0.5562
	2.03	0.5297	0.5568
	2.04	0.5284	0.5573
	2.05	0.5271	0.5579
	2.06	0.5258	0.5584
	2.07	0.5246	0.5589
	2.08	0.5233	0.5594
	2.09	0.5221	0.5600
	2.10	0.5209	0.5605
	2.11	0.5197	0.5610
	2.12	0.5184	0.5615
	2.13	0.5172	0.5620
	2.14	0.5160	0.5625
	2.15	0.5149	0.5630
	2.16	0.5137	0.5634
	2.17	0.5125	0.5639
	2.18	0.5114	0.5644
	2.19	0.5102	0.5649
	2.20	0.5091	0.5653
	2.21	0.5079	0.5658
	2.22	0.5068	0.5663
	2.23	0.5057	0.5667
	2.24	0.5045	0.5672
	2.25	0.5034	0.5676
	2.26	0.5023	0.5680
	2.27	0.5012	0.5685
	2.28	0.5002	0.5689
	2.29	0.4991	0.5693
	2.30	0.4980	0.5698
	2.31	0.4969	0.5702
	2.32	0.4959	0.5706
	2.33	0.4948	0.5710
	2.34	0.4938	0.5714
	2.35	0.4927	0.5718
	2.36	0.4917	0.5722
	2.37	0.4907	0.5726
	2.38	0.4897	0.5730
	2.39	0.4887	0.5734
	2.40	0.4876	0.5738
	2.41	0.4866	0.5742
	2.42	0.4857	0.5746
	2.43	0.4847	0.5750
	2.44	0.4837	0.5753
	2.45	0.4827	0.5757
	2.46	0.4817	0.5761
	2.47	0.4808	0.5765
	2.48	0.4798	0.5768
	2.49	0.4789	0.5772
	2.50	0.4779	0.5775
	2.51	0.4770	0.5779
	2.52	0.4761	0.5783
	2.53	0.4751	0.5786
	2.54	0.4742	0.5790
	2.55	0.4733	0.5793
	2.56	0.4724	0.5796
	2.57	0.4715	0.5800
	2.58	0.4706	0.5803
	2.59	0.4697	0.5806
	2.60	0.4688	0.5810
	2.61	0.4679	0.5813
	2.62	0.4671	0.5816
	2.63	0.4662	0.5820
	2.64	0.4653	0.5823
	2.65	0.4645	0.5826
	2.66	0.4636	0.5829
	2.67	0.4628	0.5832
	2.68	0.4619	0.5835
	2.69	0.4611	0.5839
	2.70	0.4602	0.5842
	2.71	0.4594	0.5845
	2.72	0.4586	0.5848
	2.73	0.4578	0.5851
	2.74	0.4569	0.5854
	2.75	0.4561	0.5857
	2.76	0.4553	0.5860
	2.77	0.4545	0.5863
	2.78	0.4537	0.5865
	2.79	0.4529	0.5868
	2.80	0.4521	0.5871
	2.81	0.4514	0.5874
	2.82	0.4506	0.5877
	2.83	0.4498	0.5880
	2.84	0.4490	0.5882
	2.85	0.4483	0.5885
	2.86	0.4475	0.5888
	2.87	0.4468	0.5891
	2.88	0.4460	0.5893
	2.89	0.4453	0.5896
	2.90	0.4445	0.5899
	2.91	0.4438	0.5901
	2.92	0.4430	0.5904
	2.93	0.4423	0.5907
	2.94	0.4416	0.5909
	2.95	0.4409	0.5912
	2.96	0.4401	0.5914
	2.97	0.4394	0.5917
	2.98	0.4387	0.5919
	2.99	0.4380	0.5922
	3.00	0.4373	0.5924
	3.01	0.4366	0.5927
	3.02	0.4359	0.5929
	3.03	0.4352	0.5932
	3.04	0.4345	0.5934
	3.05	0.4338	0.5937
	3.06	0.4331	0.5939
	3.07	0.4325	0.5941
	3.08	0.4318	0.5944
	3.09	0.4311	0.5946
	3.10	0.4304	0.5949
	3.11	0.4298	0.5951
	3.12	0.4291	0.5953
	3.13	0.4285	0.5956
	3.14	0.4278	0.5958
	3.15	0.4272	0.5960
	3.16	0.4265	0.5962
	3.17	0.4259	0.5965
	3.18	0.4252	0.5967
	3.19	0.4246	0.5969
	3.20	0.4240	0.5971
	3.21	0.4233	0.5973
	3.22	0.4227	0.5976
	3.23	0.4221	0.5978
	3.24	0.4215	0.5980
	3.25	0.4208	0.5982
	3.26	0.4202	0.5984
	3.27	0.4196	0.5986
	3.28	0.4190	0.5988
	3.29	0.4184	0.5990
	3.30	0.4178	0.5993
	3.31	0.4172	0.5995
	3.32	0.4166	0.5997
	3.33	0.4160	0.5999
	3.34	0.4154	0.6001
	3.35	0.4148	0.6003
	3.36	0.4142	0.6005
	3.37	0.4136	0.6007
	3.38	0.4131	0.6009
	3.39	0.4125	0.6011
	3.40	0.4119	0.6013
	3.41	0.4113	0.6015
	3.42	0.4108	0.6017
	3.43	0.4102	0.6019
	3.44	0.4096	0.6020
	3.45	0.4091	0.6022
	3.46	0.4085	0.6024
	3.47	0.4079	0.6026
	3.48	0.4074	0.6028
	3.49	0.4068	0.6030
	3.50	0.4063	0.6032
	3.51	0.4057	0.6034
	3.52	0.4052	0.6035
	3.53	0.4047	0.6037
	3.54	0.4041	0.6039
	3.55	0.4036	0.6041
	3.56	0.4030	0.6043
	3.57	0.4025	0.6045
	3.58	0.4020	0.6046
	3.59	0.4014	0.6048
	3.60	0.4009	0.6050
	3.61	0.4004	0.6052
	3.62	0.3999	0.6053
	3.63	0.3994	0.6055
	3.64	0.3988	0.6057
	3.65	0.3983	0.6059
	3.66	0.3978	0.6060
	3.67	0.3973	0.6062
	3.68	0.3968	0.6064
	3.69	0.3963	0.6065
	3.70	0.3958	0.6067
	3.71	0.3953	0.6069
	3.72	0.3948	0.6070
	3.73	0.3943	0.6072
	3.74	0.3938	0.6074
	3.75	0.3933	0.6075
	3.76	0.3928	0.6077
	3.77	0.3923	0.6079
	3.78	0.3918	0.6080
	3.79	0.3913	0.6082
	3.80	0.3908	0.6083
	3.81	0.3904	0.6085
	3.82	0.3899	0.6087
	3.83	0.3894	0.6088
	3.84	0.3889	0.6090
	3.85	0.3885	0.6091
	3.86	0.3880	0.6093
	3.87	0.3875	0.6094
	3.88	0.3870	0.6096
	3.89	0.3866	0.6098
	3.90	0.3861	0.6099
	3.91	0.3856	0.6101
	3.92	0.3852	0.6102
	3.93	0.3847	0.6104
	3.94	0.3843	0.6105
	3.95	0.3838	0.6107
	3.96	0.3834	0.6108
	3.97	0.3829	0.6110
	3.98	0.3825	0.6111
	3.99	0.3820	0.6113
	4.00	0.3816	0.6114
	4.01	0.3811	0.6115
	4.02	0.3807	0.6117
	4.03	0.3802	0.6118
	4.04	0.3798	0.6120
	4.05	0.3793	0.6121
	4.06	0.3789	0.6123
	4.07	0.3785	0.6124
	4.08	0.3780	0.6125
	4.09	0.3776	0.6127
}\tableCorrSiD

\pgfplotstableread{
	sigma	corr	si
	0.01	1.0000	0.0833
	0.02	1.0000	0.0833
	0.03	1.0000	0.0833
	0.04	1.0000	0.0833
	0.05	1.0000	0.0833
	0.06	1.0000	0.0833
	0.07	1.0000	0.0833
	0.08	1.0000	0.0833
	0.09	1.0000	0.0833
	0.10	1.0000	0.0833
	0.11	1.0000	0.0833
	0.12	1.0000	0.0833
	0.13	1.0000	0.0833
	0.14	1.0000	0.0833
	0.15	1.0000	0.0833
	0.16	1.0000	0.0833
	0.17	1.0000	0.0833
	0.18	1.0000	0.0833
	0.19	1.0000	0.0833
	0.20	1.0000	0.0833
	0.21	1.0000	0.0833
	0.22	1.0000	0.0833
	0.23	1.0000	0.0833
	0.24	1.0000	0.0833
	0.25	1.0000	0.0832
	0.26	1.0000	0.0832
	0.27	1.0000	0.0830
	0.28	1.0000	0.0829
	0.29	1.0000	0.0826
	0.30	1.0000	0.0822
	0.31	1.0000	0.0817
	0.32	0.9999	0.0810
	0.33	0.9999	0.0801
	0.34	0.9998	0.0790
	0.35	0.9997	0.0775
	0.36	0.9995	0.0756
	0.37	0.9993	0.0733
	0.38	0.9990	0.0706
	0.39	0.9986	0.0672
	0.40	0.9980	0.0633
	0.41	0.9973	0.0587
	0.42	0.9965	0.0534
	0.43	0.9954	0.0474
	0.44	0.9942	0.0409
	0.45	0.9927	0.0338
	0.46	0.9910	0.0263
	0.47	0.9891	0.0183
	0.48	0.9869	0.0099
	0.49	0.9844	0.0012
	0.50	0.9817	0.0078
	0.51	0.9787	0.0170
	0.52	0.9754	0.0263
	0.53	0.9719	0.0357
	0.54	0.9682	0.0451
	0.55	0.9642	0.0545
	0.56	0.9600	0.0638
	0.57	0.9556	0.0731
	0.58	0.9510	0.0822
	0.59	0.9462	0.0912
	0.60	0.9413	0.1000
	0.61	0.9362	0.1087
	0.62	0.9310	0.1171
	0.63	0.9257	0.1253
	0.64	0.9204	0.1333
	0.65	0.9149	0.1411
	0.66	0.9094	0.1486
	0.67	0.9039	0.1560
	0.68	0.8984	0.1631
	0.69	0.8928	0.1700
	0.70	0.8872	0.1766
	0.71	0.8817	0.1831
	0.72	0.8761	0.1893
	0.73	0.8706	0.1953
	0.74	0.8651	0.2012
	0.75	0.8597	0.2068
	0.76	0.8543	0.2122
	0.77	0.8490	0.2175
	0.78	0.8438	0.2226
	0.79	0.8385	0.2276
	0.80	0.8334	0.2323
	0.81	0.8283	0.2370
	0.82	0.8234	0.2414
	0.83	0.8184	0.2458
	0.84	0.8136	0.2500
	0.85	0.8088	0.2541
	0.86	0.8041	0.2580
	0.87	0.7994	0.2619
	0.88	0.7949	0.2656
	0.89	0.7904	0.2692
	0.90	0.7860	0.2727
	0.91	0.7816	0.2762
	0.92	0.7774	0.2795
	0.93	0.7731	0.2827
	0.94	0.7690	0.2859
	0.95	0.7649	0.2889
	0.96	0.7609	0.2919
	0.97	0.7570	0.2948
	0.98	0.7531	0.2977
	0.99	0.7493	0.3004
	1.00	0.7455	0.3032
	1.01	0.7418	0.3058
	1.02	0.7381	0.3084
	1.03	0.7345	0.3109
	1.04	0.7310	0.3133
	1.05	0.7275	0.3158
	1.06	0.7241	0.3181
	1.07	0.7207	0.3204
	1.08	0.7173	0.3227
	1.09	0.7141	0.3249
	1.10	0.7108	0.3270
	1.11	0.7076	0.3291
	1.12	0.7045	0.3312
	1.13	0.7013	0.3333
	1.14	0.6983	0.3352
	1.15	0.6952	0.3372
	1.16	0.6923	0.3391
	1.17	0.6893	0.3410
	1.18	0.6864	0.3428
	1.19	0.6835	0.3446
	1.20	0.6807	0.3464
	1.21	0.6779	0.3482
	1.22	0.6751	0.3499
	1.23	0.6724	0.3516
	1.24	0.6697	0.3532
	1.25	0.6670	0.3548
	1.26	0.6644	0.3564
	1.27	0.6618	0.3580
	1.28	0.6592	0.3596
	1.29	0.6567	0.3611
	1.30	0.6541	0.3626
	1.31	0.6517	0.3641
	1.32	0.6492	0.3655
	1.33	0.6468	0.3669
	1.34	0.6444	0.3683
	1.35	0.6420	0.3697
	1.36	0.6396	0.3711
	1.37	0.6373	0.3724
	1.38	0.6350	0.3738
	1.39	0.6327	0.3751
	1.40	0.6305	0.3763
	1.41	0.6282	0.3776
	1.42	0.6260	0.3788
	1.43	0.6238	0.3801
	1.44	0.6217	0.3813
	1.45	0.6195	0.3825
	1.46	0.6174	0.3837
	1.47	0.6153	0.3848
	1.48	0.6132	0.3860
	1.49	0.6112	0.3871
	1.50	0.6091	0.3882
	1.51	0.6071	0.3893
	1.52	0.6051	0.3904
	1.53	0.6031	0.3915
	1.54	0.6012	0.3925
	1.55	0.5992	0.3936
	1.56	0.5973	0.3946
	1.57	0.5954	0.3956
	1.58	0.5935	0.3966
	1.59	0.5917	0.3976
	1.60	0.5898	0.3986
	1.61	0.5880	0.3996
	1.62	0.5861	0.4005
	1.63	0.5843	0.4015
	1.64	0.5826	0.4024
	1.65	0.5808	0.4033
	1.66	0.5790	0.4042
	1.67	0.5773	0.4051
	1.68	0.5756	0.4060
	1.69	0.5739	0.4069
	1.70	0.5722	0.4078
	1.71	0.5705	0.4086
	1.72	0.5688	0.4095
	1.73	0.5672	0.4103
	1.74	0.5655	0.4111
	1.75	0.5639	0.4120
	1.76	0.5623	0.4128
	1.77	0.5607	0.4136
	1.78	0.5591	0.4144
	1.79	0.5575	0.4152
	1.80	0.5560	0.4159
	1.81	0.5544	0.4167
	1.82	0.5529	0.4175
	1.83	0.5514	0.4182
	1.84	0.5499	0.4190
	1.85	0.5484	0.4197
	1.86	0.5469	0.4204
	1.87	0.5454	0.4211
	1.88	0.5440	0.4219
	1.89	0.5425	0.4226
	1.90	0.5411	0.4233
	1.91	0.5397	0.4240
	1.92	0.5383	0.4247
	1.93	0.5369	0.4253
	1.94	0.5355	0.4260
	1.95	0.5341	0.4267
	1.96	0.5327	0.4273
	1.97	0.5314	0.4280
	1.98	0.5300	0.4286
	1.99	0.5287	0.4293
	2.00	0.5274	0.4299
	2.01	0.5261	0.4305
	2.02	0.5248	0.4311
	2.03	0.5235	0.4318
	2.04	0.5222	0.4324
	2.05	0.5209	0.4330
	2.06	0.5196	0.4336
	2.07	0.5184	0.4342
	2.08	0.5171	0.4348
	2.09	0.5159	0.4353
	2.10	0.5147	0.4359
	2.11	0.5134	0.4365
	2.12	0.5122	0.4370
	2.13	0.5110	0.4376
	2.14	0.5098	0.4382
	2.15	0.5087	0.4387
	2.16	0.5075	0.4393
	2.17	0.5063	0.4398
	2.18	0.5052	0.4403
	2.19	0.5040	0.4409
	2.20	0.5029	0.4414
	2.21	0.5017	0.4419
	2.22	0.5006	0.4424
	2.23	0.4995	0.4430
	2.24	0.4984	0.4435
	2.25	0.4973	0.4440
	2.26	0.4962	0.4445
	2.27	0.4951	0.4450
	2.28	0.4940	0.4455
	2.29	0.4930	0.4460
	2.30	0.4919	0.4464
	2.31	0.4908	0.4469
	2.32	0.4898	0.4474
	2.33	0.4887	0.4479
	2.34	0.4877	0.4483
	2.35	0.4867	0.4488
	2.36	0.4857	0.4493
	2.37	0.4847	0.4497
	2.38	0.4837	0.4502
	2.39	0.4827	0.4506
	2.40	0.4817	0.4511
	2.41	0.4807	0.4515
	2.42	0.4797	0.4520
	2.43	0.4787	0.4524
	2.44	0.4778	0.4528
	2.45	0.4768	0.4533
	2.46	0.4758	0.4537
	2.47	0.4749	0.4541
	2.48	0.4740	0.4545
	2.49	0.4730	0.4549
	2.50	0.4721	0.4554
	2.51	0.4712	0.4558
	2.52	0.4703	0.4562
	2.53	0.4694	0.4566
	2.54	0.4685	0.4570
	2.55	0.4676	0.4574
	2.56	0.4667	0.4578
	2.57	0.4658	0.4582
	2.58	0.4649	0.4586
	2.59	0.4640	0.4589
	2.60	0.4632	0.4593
	2.61	0.4623	0.4597
	2.62	0.4614	0.4601
	2.63	0.4606	0.4605
	2.64	0.4597	0.4608
	2.65	0.4589	0.4612
	2.66	0.4581	0.4616
	2.67	0.4572	0.4619
	2.68	0.4564	0.4623
	2.69	0.4556	0.4627
	2.70	0.4548	0.4630
	2.71	0.4539	0.4634
	2.72	0.4531	0.4637
	2.73	0.4523	0.4641
	2.74	0.4515	0.4644
	2.75	0.4507	0.4648
	2.76	0.4500	0.4651
	2.77	0.4492	0.4655
	2.78	0.4484	0.4658
	2.79	0.4476	0.4661
	2.80	0.4469	0.4665
	2.81	0.4461	0.4668
	2.82	0.4453	0.4671
	2.83	0.4446	0.4675
	2.84	0.4438	0.4678
	2.85	0.4431	0.4681
	2.86	0.4423	0.4684
	2.87	0.4416	0.4688
	2.88	0.4409	0.4691
	2.89	0.4401	0.4694
	2.90	0.4394	0.4697
	2.91	0.4387	0.4700
	2.92	0.4380	0.4703
	2.93	0.4373	0.4706
	2.94	0.4366	0.4709
	2.95	0.4359	0.4712
	2.96	0.4352	0.4715
	2.97	0.4345	0.4718
	2.98	0.4338	0.4721
	2.99	0.4331	0.4724
	3.00	0.4324	0.4727
	3.01	0.4317	0.4730
	3.02	0.4310	0.4733
	3.03	0.4304	0.4736
	3.04	0.4297	0.4739
	3.05	0.4290	0.4742
	3.06	0.4284	0.4745
	3.07	0.4277	0.4747
	3.08	0.4270	0.4750
	3.09	0.4264	0.4753
	3.10	0.4257	0.4756
	3.11	0.4251	0.4758
	3.12	0.4245	0.4761
	3.13	0.4238	0.4764
	3.14	0.4232	0.4767
	3.15	0.4225	0.4769
	3.16	0.4219	0.4772
	3.17	0.4213	0.4775
	3.18	0.4207	0.4777
	3.19	0.4200	0.4780
	3.20	0.4194	0.4783
	3.21	0.4188	0.4785
	3.22	0.4182	0.4788
	3.23	0.4176	0.4790
	3.24	0.4170	0.4793
	3.25	0.4164	0.4795
	3.26	0.4158	0.4798
	3.27	0.4152	0.4800
	3.28	0.4146	0.4803
	3.29	0.4140	0.4805
	3.30	0.4134	0.4808
	3.31	0.4128	0.4810
	3.32	0.4123	0.4813
	3.33	0.4117	0.4815
	3.34	0.4111	0.4818
	3.35	0.4105	0.4820
	3.36	0.4100	0.4822
	3.37	0.4094	0.4825
	3.38	0.4088	0.4827
	3.39	0.4083	0.4830
	3.40	0.4077	0.4832
	3.41	0.4072	0.4834
	3.42	0.4066	0.4837
	3.43	0.4060	0.4839
	3.44	0.4055	0.4841
	3.45	0.4049	0.4843
	3.46	0.4044	0.4846
	3.47	0.4039	0.4848
	3.48	0.4033	0.4850
	3.49	0.4028	0.4852
	3.50	0.4022	0.4855
	3.51	0.4017	0.4857
	3.52	0.4012	0.4859
	3.53	0.4007	0.4861
	3.54	0.4001	0.4863
	3.55	0.3996	0.4866
	3.56	0.3991	0.4868
	3.57	0.3986	0.4870
	3.58	0.3980	0.4872
	3.59	0.3975	0.4874
	3.60	0.3970	0.4876
	3.61	0.3965	0.4878
	3.62	0.3960	0.4880
	3.63	0.3955	0.4883
	3.64	0.3950	0.4885
	3.65	0.3945	0.4887
	3.66	0.3940	0.4889
	3.67	0.3935	0.4891
	3.68	0.3930	0.4893
	3.69	0.3925	0.4895
	3.70	0.3920	0.4897
	3.71	0.3915	0.4899
	3.72	0.3910	0.4901
	3.73	0.3905	0.4903
	3.74	0.3900	0.4905
	3.75	0.3896	0.4907
	3.76	0.3891	0.4909
	3.77	0.3886	0.4911
	3.78	0.3881	0.4913
	3.79	0.3876	0.4914
	3.80	0.3872	0.4916
	3.81	0.3867	0.4918
	3.82	0.3862	0.4920
	3.83	0.3858	0.4922
	3.84	0.3853	0.4924
	3.85	0.3848	0.4926
	3.86	0.3844	0.4928
	3.87	0.3839	0.4930
	3.88	0.3834	0.4931
	3.89	0.3830	0.4933
	3.90	0.3825	0.4935
	3.91	0.3821	0.4937
	3.92	0.3816	0.4939
	3.93	0.3812	0.4941
	3.94	0.3807	0.4942
	3.95	0.3803	0.4944
	3.96	0.3798	0.4946
	3.97	0.3794	0.4948
	3.98	0.3789	0.4950
	3.99	0.3785	0.4951
	4.00	0.3780	0.4953
	4.01	0.3776	0.4955
	4.02	0.3772	0.4957
	4.03	0.3767	0.4958
	4.04	0.3763	0.4960
	4.05	0.3759	0.4962
	4.06	0.3754	0.4963
	4.07	0.3750	0.4965
	4.08	0.3746	0.4967
	4.09	0.3741	0.4969
}\tableCorrSiE

\pgfplotstableread{
	sigma	corr	si
	0.01	1.0000	0.0833
	0.02	1.0000	0.0833
	0.03	1.0000	0.0833
	0.04	1.0000	0.0833
	0.05	1.0000	0.0833
	0.06	1.0000	0.0833
	0.07	1.0000	0.0833
	0.08	1.0000	0.0833
	0.09	1.0000	0.0833
	0.10	1.0000	0.0833
	0.11	1.0000	0.0833
	0.12	1.0000	0.0833
	0.13	1.0000	0.0833
	0.14	1.0000	0.0833
	0.15	1.0000	0.0833
	0.16	1.0000	0.0833
	0.17	1.0000	0.0833
	0.18	1.0000	0.0833
	0.19	1.0000	0.0833
	0.20	1.0000	0.0833
	0.21	1.0000	0.0833
	0.22	1.0000	0.0833
	0.23	1.0000	0.0833
	0.24	1.0000	0.0833
	0.25	1.0000	0.0832
	0.26	1.0000	0.0832
	0.27	1.0000	0.0831
	0.28	1.0000	0.0829
	0.29	1.0000	0.0827
	0.30	1.0000	0.0823
	0.31	1.0000	0.0819
	0.32	0.9999	0.0812
	0.33	0.9999	0.0804
	0.34	0.9998	0.0794
	0.35	0.9997	0.0781
	0.36	0.9996	0.0765
	0.37	0.9993	0.0745
	0.38	0.9990	0.0721
	0.39	0.9986	0.0692
	0.40	0.9981	0.0658
	0.41	0.9975	0.0618
	0.42	0.9966	0.0572
	0.43	0.9956	0.0521
	0.44	0.9945	0.0465
	0.45	0.9931	0.0404
	0.46	0.9914	0.0340
	0.47	0.9896	0.0271
	0.48	0.9875	0.0199
	0.49	0.9851	0.0125
	0.50	0.9825	0.0048
	0.51	0.9797	0.0031
	0.52	0.9766	0.0110
	0.53	0.9733	0.0191
	0.54	0.9697	0.0272
	0.55	0.9660	0.0353
	0.56	0.9620	0.0434
	0.57	0.9578	0.0514
	0.58	0.9535	0.0593
	0.59	0.9489	0.0671
	0.60	0.9443	0.0747
	0.61	0.9395	0.0823
	0.62	0.9346	0.0896
	0.63	0.9296	0.0968
	0.64	0.9245	0.1039
	0.65	0.9193	0.1107
	0.66	0.9141	0.1174
	0.67	0.9089	0.1239
	0.68	0.9036	0.1302
	0.69	0.8983	0.1363
	0.70	0.8930	0.1423
	0.71	0.8877	0.1481
	0.72	0.8824	0.1537
	0.73	0.8772	0.1592
	0.74	0.8720	0.1645
	0.75	0.8668	0.1697
	0.76	0.8616	0.1747
	0.77	0.8565	0.1796
	0.78	0.8514	0.1843
	0.79	0.8464	0.1889
	0.80	0.8415	0.1934
	0.81	0.8366	0.1977
	0.82	0.8318	0.2019
	0.83	0.8270	0.2060
	0.84	0.8223	0.2100
	0.85	0.8176	0.2139
	0.86	0.8130	0.2177
	0.87	0.8085	0.2214
	0.88	0.8040	0.2250
	0.89	0.7996	0.2285
	0.90	0.7953	0.2319
	0.91	0.7910	0.2352
	0.92	0.7868	0.2384
	0.93	0.7826	0.2416
	0.94	0.7785	0.2447
	0.95	0.7745	0.2477
	0.96	0.7705	0.2506
	0.97	0.7666	0.2535
	0.98	0.7627	0.2563
	0.99	0.7589	0.2591
	1.00	0.7551	0.2618
	1.01	0.7514	0.2644
	1.02	0.7477	0.2670
	1.03	0.7441	0.2695
	1.04	0.7405	0.2720
	1.05	0.7370	0.2744
	1.06	0.7335	0.2768
	1.07	0.7301	0.2791
	1.08	0.7267	0.2814
	1.09	0.7233	0.2836
	1.10	0.7200	0.2858
	1.11	0.7168	0.2879
	1.12	0.7136	0.2900
	1.13	0.7104	0.2921
	1.14	0.7072	0.2941
	1.15	0.7041	0.2961
	1.16	0.7010	0.2981
	1.17	0.6980	0.3000
	1.18	0.6950	0.3019
	1.19	0.6920	0.3038
	1.20	0.6891	0.3056
	1.21	0.6862	0.3074
	1.22	0.6833	0.3091
	1.23	0.6805	0.3109
	1.24	0.6777	0.3126
	1.25	0.6749	0.3142
	1.26	0.6722	0.3159
	1.27	0.6695	0.3175
	1.28	0.6668	0.3191
	1.29	0.6641	0.3207
	1.30	0.6615	0.3222
	1.31	0.6589	0.3237
	1.32	0.6563	0.3252
	1.33	0.6538	0.3267
	1.34	0.6512	0.3282
	1.35	0.6487	0.3296
	1.36	0.6463	0.3310
	1.37	0.6438	0.3324
	1.38	0.6414	0.3338
	1.39	0.6390	0.3351
	1.40	0.6366	0.3365
	1.41	0.6342	0.3378
	1.42	0.6319	0.3391
	1.43	0.6296	0.3403
	1.44	0.6273	0.3416
	1.45	0.6250	0.3428
	1.46	0.6228	0.3441
	1.47	0.6206	0.3453
	1.48	0.6184	0.3464
	1.49	0.6162	0.3476
	1.50	0.6140	0.3488
	1.51	0.6119	0.3499
	1.52	0.6097	0.3510
	1.53	0.6076	0.3522
	1.54	0.6055	0.3533
	1.55	0.6035	0.3543
	1.56	0.6014	0.3554
	1.57	0.5994	0.3565
	1.58	0.5974	0.3575
	1.59	0.5954	0.3585
	1.60	0.5934	0.3596
	1.61	0.5914	0.3606
	1.62	0.5895	0.3616
	1.63	0.5876	0.3625
	1.64	0.5857	0.3635
	1.65	0.5838	0.3645
	1.66	0.5819	0.3654
	1.67	0.5800	0.3663
	1.68	0.5782	0.3673
	1.69	0.5763	0.3682
	1.70	0.5745	0.3691
	1.71	0.5727	0.3700
	1.72	0.5709	0.3709
	1.73	0.5691	0.3717
	1.74	0.5674	0.3726
	1.75	0.5656	0.3734
	1.76	0.5639	0.3743
	1.77	0.5622	0.3751
	1.78	0.5605	0.3759
	1.79	0.5588	0.3768
	1.80	0.5571	0.3776
	1.81	0.5555	0.3784
	1.82	0.5538	0.3791
	1.83	0.5522	0.3799
	1.84	0.5506	0.3807
	1.85	0.5489	0.3814
	1.86	0.5473	0.3822
	1.87	0.5458	0.3829
	1.88	0.5442	0.3837
	1.89	0.5426	0.3844
	1.90	0.5411	0.3851
	1.91	0.5395	0.3859
	1.92	0.5380	0.3866
	1.93	0.5365	0.3873
	1.94	0.5350	0.3880
	1.95	0.5335	0.3886
	1.96	0.5320	0.3893
	1.97	0.5306	0.3900
	1.98	0.5291	0.3907
	1.99	0.5277	0.3913
	2.00	0.5262	0.3920
	2.01	0.5248	0.3926
	2.02	0.5234	0.3933
	2.03	0.5220	0.3939
	2.04	0.5206	0.3945
	2.05	0.5192	0.3951
	2.06	0.5178	0.3957
	2.07	0.5165	0.3964
	2.08	0.5151	0.3970
	2.09	0.5138	0.3976
	2.10	0.5124	0.3981
	2.11	0.5111	0.3987
	2.12	0.5098	0.3993
	2.13	0.5085	0.3999
	2.14	0.5072	0.4005
	2.15	0.5059	0.4010
	2.16	0.5046	0.4016
	2.17	0.5034	0.4021
	2.18	0.5021	0.4027
	2.19	0.5009	0.4032
	2.20	0.4996	0.4038
	2.21	0.4984	0.4043
	2.22	0.4972	0.4048
	2.23	0.4959	0.4054
	2.24	0.4947	0.4059
	2.25	0.4935	0.4064
	2.26	0.4924	0.4069
	2.27	0.4912	0.4074
	2.28	0.4900	0.4079
	2.29	0.4888	0.4084
	2.30	0.4877	0.4089
	2.31	0.4865	0.4094
	2.32	0.4854	0.4099
	2.33	0.4842	0.4104
	2.34	0.4831	0.4108
	2.35	0.4820	0.4113
	2.36	0.4809	0.4118
	2.37	0.4798	0.4123
	2.38	0.4787	0.4127
	2.39	0.4776	0.4132
	2.40	0.4765	0.4136
	2.41	0.4754	0.4141
	2.42	0.4743	0.4145
	2.43	0.4733	0.4150
	2.44	0.4722	0.4154
	2.45	0.4712	0.4158
	2.46	0.4701	0.4163
	2.47	0.4691	0.4167
	2.48	0.4681	0.4171
	2.49	0.4671	0.4176
	2.50	0.4660	0.4180
	2.51	0.4650	0.4184
	2.52	0.4640	0.4188
	2.53	0.4630	0.4192
	2.54	0.4620	0.4196
	2.55	0.4611	0.4200
	2.56	0.4601	0.4204
	2.57	0.4591	0.4208
	2.58	0.4581	0.4212
	2.59	0.4572	0.4216
	2.60	0.4562	0.4220
	2.61	0.4553	0.4224
	2.62	0.4543	0.4227
	2.63	0.4534	0.4231
	2.64	0.4525	0.4235
	2.65	0.4515	0.4239
	2.66	0.4506	0.4243
	2.67	0.4497	0.4246
	2.68	0.4488	0.4250
	2.69	0.4479	0.4254
	2.70	0.4470	0.4257
	2.71	0.4461	0.4261
	2.72	0.4452	0.4264
	2.73	0.4443	0.4268
	2.74	0.4435	0.4271
	2.75	0.4426	0.4275
	2.76	0.4417	0.4278
	2.77	0.4409	0.4282
	2.78	0.4400	0.4285
	2.79	0.4392	0.4288
	2.80	0.4383	0.4292
	2.81	0.4375	0.4295
	2.82	0.4367	0.4298
	2.83	0.4358	0.4302
	2.84	0.4350	0.4305
	2.85	0.4342	0.4308
	2.86	0.4334	0.4311
	2.87	0.4326	0.4315
	2.88	0.4317	0.4318
	2.89	0.4309	0.4321
	2.90	0.4301	0.4324
	2.91	0.4294	0.4327
	2.92	0.4286	0.4330
	2.93	0.4278	0.4333
	2.94	0.4270	0.4336
	2.95	0.4262	0.4339
	2.96	0.4255	0.4343
	2.97	0.4247	0.4346
	2.98	0.4239	0.4348
	2.99	0.4232	0.4351
	3.00	0.4224	0.4354
	3.01	0.4217	0.4357
	3.02	0.4209	0.4360
	3.03	0.4202	0.4363
	3.04	0.4195	0.4366
	3.05	0.4187	0.4369
	3.06	0.4180	0.4372
	3.07	0.4173	0.4374
	3.08	0.4166	0.4377
	3.09	0.4158	0.4380
	3.10	0.4151	0.4383
	3.11	0.4144	0.4385
	3.12	0.4137	0.4388
	3.13	0.4130	0.4391
	3.14	0.4123	0.4394
	3.15	0.4116	0.4396
	3.16	0.4109	0.4399
	3.17	0.4102	0.4402
	3.18	0.4096	0.4404
	3.19	0.4089	0.4407
	3.20	0.4082	0.4409
	3.21	0.4075	0.4412
	3.22	0.4069	0.4415
	3.23	0.4062	0.4417
	3.24	0.4055	0.4420
	3.25	0.4049	0.4422
	3.26	0.4042	0.4425
	3.27	0.4036	0.4427
	3.28	0.4029	0.4430
	3.29	0.4023	0.4432
	3.30	0.4017	0.4435
	3.31	0.4010	0.4437
	3.32	0.4004	0.4439
	3.33	0.3998	0.4442
	3.34	0.3991	0.4444
	3.35	0.3985	0.4447
	3.36	0.3979	0.4449
	3.37	0.3973	0.4451
	3.38	0.3967	0.4454
	3.39	0.3960	0.4456
	3.40	0.3954	0.4458
	3.41	0.3948	0.4461
	3.42	0.3942	0.4463
	3.43	0.3936	0.4465
	3.44	0.3930	0.4468
	3.45	0.3924	0.4470
	3.46	0.3919	0.4472
	3.47	0.3913	0.4474
	3.48	0.3907	0.4477
	3.49	0.3901	0.4479
	3.50	0.3895	0.4481
	3.51	0.3890	0.4483
	3.52	0.3884	0.4485
	3.53	0.3878	0.4487
	3.54	0.3872	0.4490
	3.55	0.3867	0.4492
	3.56	0.3861	0.4494
	3.57	0.3856	0.4496
	3.58	0.3850	0.4498
	3.59	0.3844	0.4500
	3.60	0.3839	0.4502
	3.61	0.3834	0.4504
	3.62	0.3828	0.4506
	3.63	0.3823	0.4508
	3.64	0.3817	0.4511
	3.65	0.3812	0.4513
	3.66	0.3806	0.4515
	3.67	0.3801	0.4517
	3.68	0.3796	0.4519
	3.69	0.3791	0.4521
	3.70	0.3785	0.4523
	3.71	0.3780	0.4525
	3.72	0.3775	0.4526
	3.73	0.3770	0.4528
	3.74	0.3765	0.4530
	3.75	0.3759	0.4532
	3.76	0.3754	0.4534
	3.77	0.3749	0.4536
	3.78	0.3744	0.4538
	3.79	0.3739	0.4540
	3.80	0.3734	0.4542
	3.81	0.3729	0.4544
	3.82	0.3724	0.4546
	3.83	0.3719	0.4547
	3.84	0.3714	0.4549
	3.85	0.3710	0.4551
	3.86	0.3705	0.4553
	3.87	0.3700	0.4555
	3.88	0.3695	0.4557
	3.89	0.3690	0.4558
	3.90	0.3685	0.4560
	3.91	0.3681	0.4562
	3.92	0.3676	0.4564
	3.93	0.3671	0.4566
	3.94	0.3666	0.4567
	3.95	0.3662	0.4569
	3.96	0.3657	0.4571
	3.97	0.3652	0.4573
	3.98	0.3648	0.4574
	3.99	0.3643	0.4576
	4.00	0.3639	0.4578
	4.01	0.3634	0.4580
	4.02	0.3630	0.4581
	4.03	0.3625	0.4583
	4.04	0.3621	0.4585
	4.05	0.3616	0.4586
	4.06	0.3612	0.4588
	4.07	0.3607	0.4590
	4.08	0.3603	0.4591
	4.09	0.3598	0.4593
}\tableCorrSiF

\pgfplotstableread{
	sigma	corr	si
	0.01	1.0000	0.0833
	0.02	1.0000	0.0833
	0.03	1.0000	0.0833
	0.04	1.0000	0.0833
	0.05	1.0000	0.0833
	0.06	1.0000	0.0833
	0.07	1.0000	0.0833
	0.08	1.0000	0.0833
	0.09	1.0000	0.0833
	0.10	1.0000	0.0833
	0.11	1.0000	0.0833
	0.12	1.0000	0.0833
	0.13	1.0000	0.0833
	0.14	1.0000	0.0833
	0.15	1.0000	0.0833
	0.16	1.0000	0.0833
	0.17	1.0000	0.0833
	0.18	1.0000	0.0833
	0.19	1.0000	0.0833
	0.20	1.0000	0.0833
	0.21	1.0000	0.0833
	0.22	1.0000	0.0833
	0.23	1.0000	0.0833
	0.24	1.0000	0.0833
	0.25	1.0000	0.0832
	0.26	1.0000	0.0832
	0.27	1.0000	0.0831
	0.28	1.0000	0.0829
	0.29	1.0000	0.0826
	0.30	1.0000	0.0823
	0.31	1.0000	0.0818
	0.32	0.9999	0.0811
	0.33	0.9999	0.0803
	0.34	0.9998	0.0792
	0.35	0.9997	0.0778
	0.36	0.9996	0.0760
	0.37	0.9993	0.0739
	0.38	0.9990	0.0713
	0.39	0.9986	0.0681
	0.40	0.9981	0.0644
	0.41	0.9974	0.0601
	0.42	0.9966	0.0552
	0.43	0.9956	0.0497
	0.44	0.9943	0.0436
	0.45	0.9929	0.0370
	0.46	0.9912	0.0299
	0.47	0.9893	0.0225
	0.48	0.9872	0.0147
	0.49	0.9848	0.0065
	0.50	0.9821	0.0018
	0.51	0.9791	0.0104
	0.52	0.9759	0.0191
	0.53	0.9725	0.0278
	0.54	0.9688	0.0367
	0.55	0.9649	0.0455
	0.56	0.9607	0.0542
	0.57	0.9564	0.0629
	0.58	0.9518	0.0715
	0.59	0.9471	0.0800
	0.60	0.9422	0.0883
	0.61	0.9372	0.0964
	0.62	0.9321	0.1044
	0.63	0.9268	0.1121
	0.64	0.9215	0.1197
	0.65	0.9161	0.1271
	0.66	0.9106	0.1343
	0.67	0.9051	0.1413
	0.68	0.8995	0.1480
	0.69	0.8940	0.1546
	0.70	0.8884	0.1610
	0.71	0.8828	0.1672
	0.72	0.8773	0.1732
	0.73	0.8718	0.1790
	0.74	0.8663	0.1846
	0.75	0.8608	0.1901
	0.76	0.8554	0.1953
	0.77	0.8501	0.2004
	0.78	0.8448	0.2054
	0.79	0.8395	0.2102
	0.80	0.8344	0.2148
	0.81	0.8293	0.2194
	0.82	0.8242	0.2237
	0.83	0.8192	0.2280
	0.84	0.8143	0.2321
	0.85	0.8095	0.2361
	0.86	0.8047	0.2400
	0.87	0.8000	0.2438
	0.88	0.7954	0.2474
	0.89	0.7908	0.2510
	0.90	0.7863	0.2545
	0.91	0.7819	0.2579
	0.92	0.7776	0.2612
	0.93	0.7733	0.2644
	0.94	0.7690	0.2675
	0.95	0.7649	0.2705
	0.96	0.7608	0.2735
	0.97	0.7568	0.2764
	0.98	0.7528	0.2792
	0.99	0.7489	0.2820
	1.00	0.7450	0.2847
	1.01	0.7412	0.2874
	1.02	0.7375	0.2899
	1.03	0.7338	0.2925
	1.04	0.7301	0.2949
	1.05	0.7266	0.2973
	1.06	0.7230	0.2997
	1.07	0.7195	0.3020
	1.08	0.7161	0.3043
	1.09	0.7127	0.3065
	1.10	0.7094	0.3087
	1.11	0.7061	0.3108
	1.12	0.7028	0.3129
	1.13	0.6996	0.3149
	1.14	0.6964	0.3169
	1.15	0.6933	0.3189
	1.16	0.6902	0.3209
	1.17	0.6872	0.3228
	1.18	0.6841	0.3246
	1.19	0.6812	0.3264
	1.20	0.6782	0.3282
	1.21	0.6753	0.3300
	1.22	0.6724	0.3318
	1.23	0.6696	0.3335
	1.24	0.6668	0.3351
	1.25	0.6640	0.3368
	1.26	0.6613	0.3384
	1.27	0.6586	0.3400
	1.28	0.6559	0.3416
	1.29	0.6532	0.3431
	1.30	0.6506	0.3446
	1.31	0.6480	0.3461
	1.32	0.6454	0.3476
	1.33	0.6429	0.3491
	1.34	0.6404	0.3505
	1.35	0.6379	0.3519
	1.36	0.6354	0.3533
	1.37	0.6330	0.3546
	1.38	0.6306	0.3560
	1.39	0.6282	0.3573
	1.40	0.6258	0.3586
	1.41	0.6235	0.3599
	1.42	0.6211	0.3612
	1.43	0.6188	0.3624
	1.44	0.6166	0.3637
	1.45	0.6143	0.3649
	1.46	0.6121	0.3661
	1.47	0.6098	0.3673
	1.48	0.6077	0.3684
	1.49	0.6055	0.3696
	1.50	0.6033	0.3707
	1.51	0.6012	0.3718
	1.52	0.5991	0.3729
	1.53	0.5970	0.3740
	1.54	0.5949	0.3751
	1.55	0.5928	0.3762
	1.56	0.5908	0.3772
	1.57	0.5888	0.3783
	1.58	0.5867	0.3793
	1.59	0.5847	0.3803
	1.60	0.5828	0.3813
	1.61	0.5808	0.3823
	1.62	0.5789	0.3832
	1.63	0.5769	0.3842
	1.64	0.5750	0.3852
	1.65	0.5731	0.3861
	1.66	0.5713	0.3870
	1.67	0.5694	0.3880
	1.68	0.5675	0.3889
	1.69	0.5657	0.3898
	1.70	0.5639	0.3906
	1.71	0.5621	0.3915
	1.72	0.5603	0.3924
	1.73	0.5585	0.3932
	1.74	0.5567	0.3941
	1.75	0.5550	0.3949
	1.76	0.5533	0.3957
	1.77	0.5515	0.3966
	1.78	0.5498	0.3974
	1.79	0.5481	0.3982
	1.80	0.5464	0.3990
	1.81	0.5448	0.3997
	1.82	0.5431	0.4005
	1.83	0.5415	0.4013
	1.84	0.5398	0.4020
	1.85	0.5382	0.4028
	1.86	0.5366	0.4035
	1.87	0.5350	0.4043
	1.88	0.5334	0.4050
	1.89	0.5319	0.4057
	1.90	0.5303	0.4064
	1.91	0.5287	0.4071
	1.92	0.5272	0.4078
	1.93	0.5257	0.4085
	1.94	0.5242	0.4092
	1.95	0.5227	0.4099
	1.96	0.5212	0.4105
	1.97	0.5197	0.4112
	1.98	0.5182	0.4118
	1.99	0.5167	0.4125
	2.00	0.5153	0.4131
	2.01	0.5139	0.4138
	2.02	0.5124	0.4144
	2.03	0.5110	0.4150
	2.04	0.5096	0.4156
	2.05	0.5082	0.4162
	2.06	0.5068	0.4168
	2.07	0.5054	0.4174
	2.08	0.5041	0.4180
	2.09	0.5027	0.4186
	2.10	0.5014	0.4192
	2.11	0.5000	0.4198
	2.12	0.4987	0.4203
	2.13	0.4974	0.4209
	2.14	0.4961	0.4215
	2.15	0.4948	0.4220
	2.16	0.4935	0.4226
	2.17	0.4922	0.4231
	2.18	0.4909	0.4237
	2.19	0.4896	0.4242
	2.20	0.4884	0.4247
	2.21	0.4871	0.4253
	2.22	0.4859	0.4258
	2.23	0.4847	0.4263
	2.24	0.4834	0.4268
	2.25	0.4822	0.4273
	2.26	0.4810	0.4278
	2.27	0.4798	0.4283
	2.28	0.4786	0.4288
	2.29	0.4775	0.4293
	2.30	0.4763	0.4298
	2.31	0.4751	0.4302
	2.32	0.4740	0.4307
	2.33	0.4728	0.4312
	2.34	0.4717	0.4317
	2.35	0.4706	0.4321
	2.36	0.4694	0.4326
	2.37	0.4683	0.4330
	2.38	0.4672	0.4335
	2.39	0.4661	0.4339
	2.40	0.4650	0.4344
	2.41	0.4639	0.4348
	2.42	0.4628	0.4353
	2.43	0.4618	0.4357
	2.44	0.4607	0.4361
	2.45	0.4597	0.4366
	2.46	0.4586	0.4370
	2.47	0.4576	0.4374
	2.48	0.4565	0.4378
	2.49	0.4555	0.4382
	2.50	0.4545	0.4386
	2.51	0.4535	0.4390
	2.52	0.4525	0.4394
	2.53	0.4515	0.4398
	2.54	0.4505	0.4402
	2.55	0.4495	0.4406
	2.56	0.4485	0.4410
	2.57	0.4475	0.4414
	2.58	0.4466	0.4418
	2.59	0.4456	0.4422
	2.60	0.4447	0.4425
	2.61	0.4437	0.4429
	2.62	0.4428	0.4433
	2.63	0.4418	0.4437
	2.64	0.4409	0.4440
	2.65	0.4400	0.4444
	2.66	0.4391	0.4448
	2.67	0.4382	0.4451
	2.68	0.4372	0.4455
	2.69	0.4363	0.4458
	2.70	0.4355	0.4462
	2.71	0.4346	0.4465
	2.72	0.4337	0.4469
	2.73	0.4328	0.4472
	2.74	0.4319	0.4475
	2.75	0.4311	0.4479
	2.76	0.4302	0.4482
	2.77	0.4294	0.4485
	2.78	0.4285	0.4489
	2.79	0.4277	0.4492
	2.80	0.4269	0.4495
	2.81	0.4260	0.4499
	2.82	0.4252	0.4502
	2.83	0.4244	0.4505
	2.84	0.4236	0.4508
	2.85	0.4228	0.4511
	2.86	0.4219	0.4514
	2.87	0.4212	0.4517
	2.88	0.4204	0.4521
	2.89	0.4196	0.4524
	2.90	0.4188	0.4527
	2.91	0.4180	0.4530
	2.92	0.4172	0.4533
	2.93	0.4165	0.4536
	2.94	0.4157	0.4539
	2.95	0.4149	0.4542
	2.96	0.4142	0.4544
	2.97	0.4134	0.4547
	2.98	0.4127	0.4550
	2.99	0.4119	0.4553
	3.00	0.4112	0.4556
	3.01	0.4105	0.4559
	3.02	0.4098	0.4562
	3.03	0.4090	0.4564
	3.04	0.4083	0.4567
	3.05	0.4076	0.4570
	3.06	0.4069	0.4573
	3.07	0.4062	0.4575
	3.08	0.4055	0.4578
	3.09	0.4048	0.4581
	3.10	0.4041	0.4583
	3.11	0.4034	0.4586
	3.12	0.4027	0.4589
	3.13	0.4020	0.4591
	3.14	0.4014	0.4594
	3.15	0.4007	0.4596
	3.16	0.4000	0.4599
	3.17	0.3994	0.4602
	3.18	0.3987	0.4604
	3.19	0.3981	0.4607
	3.20	0.3974	0.4609
	3.21	0.3968	0.4612
	3.22	0.3961	0.4614
	3.23	0.3955	0.4617
	3.24	0.3948	0.4619
	3.25	0.3942	0.4621
	3.26	0.3936	0.4624
	3.27	0.3929	0.4626
	3.28	0.3923	0.4629
	3.29	0.3917	0.4631
	3.30	0.3911	0.4633
	3.31	0.3905	0.4636
	3.32	0.3899	0.4638
	3.33	0.3893	0.4640
	3.34	0.3887	0.4643
	3.35	0.3881	0.4645
	3.36	0.3875	0.4647
	3.37	0.3869	0.4649
	3.38	0.3863	0.4652
	3.39	0.3857	0.4654
	3.40	0.3851	0.4656
	3.41	0.3845	0.4658
	3.42	0.3840	0.4661
	3.43	0.3834	0.4663
	3.44	0.3828	0.4665
	3.45	0.3823	0.4667
	3.46	0.3817	0.4669
	3.47	0.3811	0.4672
	3.48	0.3806	0.4674
	3.49	0.3800	0.4676
	3.50	0.3795	0.4678
	3.51	0.3789	0.4680
	3.52	0.3784	0.4682
	3.53	0.3778	0.4684
	3.54	0.3773	0.4686
	3.55	0.3768	0.4688
	3.56	0.3762	0.4690
	3.57	0.3757	0.4692
	3.58	0.3752	0.4694
	3.59	0.3746	0.4696
	3.60	0.3741	0.4698
	3.61	0.3736	0.4700
	3.62	0.3731	0.4702
	3.63	0.3726	0.4704
	3.64	0.3721	0.4706
	3.65	0.3716	0.4708
	3.66	0.3710	0.4710
	3.67	0.3705	0.4712
	3.68	0.3700	0.4714
	3.69	0.3695	0.4716
	3.70	0.3690	0.4718
	3.71	0.3685	0.4720
	3.72	0.3681	0.4722
	3.73	0.3676	0.4724
	3.74	0.3671	0.4726
	3.75	0.3666	0.4727
	3.76	0.3661	0.4729
	3.77	0.3656	0.4731
	3.78	0.3651	0.4733
	3.79	0.3647	0.4735
	3.80	0.3642	0.4737
	3.81	0.3637	0.4738
	3.82	0.3633	0.4740
	3.83	0.3628	0.4742
	3.84	0.3623	0.4744
	3.85	0.3619	0.4746
	3.86	0.3614	0.4747
	3.87	0.3609	0.4749
	3.88	0.3605	0.4751
	3.89	0.3600	0.4753
	3.90	0.3596	0.4754
	3.91	0.3591	0.4756
	3.92	0.3587	0.4758
	3.93	0.3582	0.4760
	3.94	0.3578	0.4761
	3.95	0.3574	0.4763
	3.96	0.3569	0.4765
	3.97	0.3565	0.4766
	3.98	0.3560	0.4768
	3.99	0.3556	0.4770
	4.00	0.3552	0.4771
	4.01	0.3547	0.4773
	4.02	0.3543	0.4775
	4.03	0.3539	0.4776
	4.04	0.3535	0.4778
	4.05	0.3530	0.4780
	4.06	0.3526	0.4781
	4.07	0.3522	0.4783
	4.08	0.3518	0.4785
	4.09	0.3514	0.4786
}\tableCorrSiG

\pgfplotstableread{
	sigma	corr	si
	0.01	1.0000	0.0833
	0.02	1.0000	0.0833
	0.03	1.0000	0.0833
	0.04	1.0000	0.0833
	0.05	1.0000	0.0833
	0.06	1.0000	0.0833
	0.07	1.0000	0.0833
	0.08	1.0000	0.0833
	0.09	1.0000	0.0833
	0.10	1.0000	0.0833
	0.11	1.0000	0.0833
	0.12	1.0000	0.0833
	0.13	1.0000	0.0833
	0.14	1.0000	0.0833
	0.15	1.0000	0.0833
	0.16	1.0000	0.0833
	0.17	1.0000	0.0833
	0.18	1.0000	0.0833
	0.19	1.0000	0.0833
	0.20	1.0000	0.0833
	0.21	1.0000	0.0833
	0.22	1.0000	0.0833
	0.23	1.0000	0.0833
	0.24	1.0000	0.0833
	0.25	1.0000	0.0832
	0.26	1.0000	0.0832
	0.27	1.0000	0.0830
	0.28	1.0000	0.0828
	0.29	1.0000	0.0826
	0.30	1.0000	0.0822
	0.31	1.0000	0.0816
	0.32	0.9999	0.0809
	0.33	0.9999	0.0800
	0.34	0.9998	0.0788
	0.35	0.9997	0.0773
	0.36	0.9995	0.0753
	0.37	0.9993	0.0729
	0.38	0.9990	0.0700
	0.39	0.9986	0.0666
	0.40	0.9980	0.0624
	0.41	0.9973	0.0576
	0.42	0.9965	0.0520
	0.43	0.9954	0.0458
	0.44	0.9942	0.0389
	0.45	0.9927	0.0315
	0.46	0.9910	0.0236
	0.47	0.9891	0.0152
	0.48	0.9869	0.0064
	0.49	0.9844	0.0028
	0.50	0.9817	0.0122
	0.51	0.9787	0.0218
	0.52	0.9755	0.0315
	0.53	0.9720	0.0414
	0.54	0.9683	0.0513
	0.55	0.9644	0.0611
	0.56	0.9602	0.0709
	0.57	0.9559	0.0806
	0.58	0.9513	0.0902
	0.59	0.9466	0.0996
	0.60	0.9418	0.1088
	0.61	0.9368	0.1178
	0.62	0.9317	0.1266
	0.63	0.9265	0.1352
	0.64	0.9213	0.1436
	0.65	0.9159	0.1517
	0.66	0.9106	0.1596
	0.67	0.9052	0.1673
	0.68	0.8997	0.1747
	0.69	0.8943	0.1819
	0.70	0.8888	0.1888
	0.71	0.8834	0.1955
	0.72	0.8780	0.2020
	0.73	0.8727	0.2082
	0.74	0.8673	0.2143
	0.75	0.8620	0.2201
	0.76	0.8568	0.2257
	0.77	0.8516	0.2312
	0.78	0.8465	0.2365
	0.79	0.8414	0.2415
	0.80	0.8364	0.2465
	0.81	0.8315	0.2512
	0.82	0.8267	0.2559
	0.83	0.8219	0.2603
	0.84	0.8172	0.2647
	0.85	0.8125	0.2688
	0.86	0.8079	0.2729
	0.87	0.8034	0.2769
	0.88	0.7990	0.2807
	0.89	0.7947	0.2844
	0.90	0.7904	0.2880
	0.91	0.7861	0.2915
	0.92	0.7820	0.2950
	0.93	0.7779	0.2983
	0.94	0.7739	0.3015
	0.95	0.7699	0.3047
	0.96	0.7660	0.3077
	0.97	0.7622	0.3107
	0.98	0.7585	0.3136
	0.99	0.7547	0.3165
	1.00	0.7511	0.3192
	1.01	0.7475	0.3220
	1.02	0.7440	0.3246
	1.03	0.7405	0.3272
	1.04	0.7370	0.3297
	1.05	0.7337	0.3322
	1.06	0.7303	0.3346
	1.07	0.7270	0.3369
	1.08	0.7238	0.3392
	1.09	0.7206	0.3415
	1.10	0.7175	0.3437
	1.11	0.7144	0.3459
	1.12	0.7113	0.3480
	1.13	0.7083	0.3501
	1.14	0.7053	0.3521
	1.15	0.7024	0.3541
	1.16	0.6995	0.3561
	1.17	0.6966	0.3580
	1.18	0.6938	0.3599
	1.19	0.6910	0.3617
	1.20	0.6883	0.3635
	1.21	0.6856	0.3653
	1.22	0.6829	0.3671
	1.23	0.6802	0.3688
	1.24	0.6776	0.3705
	1.25	0.6750	0.3722
	1.26	0.6725	0.3738
	1.27	0.6700	0.3754
	1.28	0.6675	0.3770
	1.29	0.6650	0.3785
	1.30	0.6625	0.3801
	1.31	0.6601	0.3816
	1.32	0.6577	0.3831
	1.33	0.6554	0.3845
	1.34	0.6531	0.3859
	1.35	0.6507	0.3874
	1.36	0.6485	0.3887
	1.37	0.6462	0.3901
	1.38	0.6440	0.3915
	1.39	0.6418	0.3928
	1.40	0.6396	0.3941
	1.41	0.6374	0.3954
	1.42	0.6353	0.3967
	1.43	0.6331	0.3979
	1.44	0.6310	0.3992
	1.45	0.6289	0.4004
	1.46	0.6269	0.4016
	1.47	0.6248	0.4028
	1.48	0.6228	0.4039
	1.49	0.6208	0.4051
	1.50	0.6188	0.4062
	1.51	0.6169	0.4074
	1.52	0.6149	0.4085
	1.53	0.6130	0.4096
	1.54	0.6111	0.4106
	1.55	0.6092	0.4117
	1.56	0.6073	0.4128
	1.57	0.6055	0.4138
	1.58	0.6036	0.4148
	1.59	0.6018	0.4158
	1.60	0.6000	0.4168
	1.61	0.5982	0.4178
	1.62	0.5964	0.4188
	1.63	0.5947	0.4198
	1.64	0.5929	0.4207
	1.65	0.5912	0.4217
	1.66	0.5895	0.4226
	1.67	0.5878	0.4235
	1.68	0.5861	0.4244
	1.69	0.5844	0.4253
	1.70	0.5827	0.4262
	1.71	0.5811	0.4271
	1.72	0.5795	0.4280
	1.73	0.5778	0.4288
	1.74	0.5762	0.4297
	1.75	0.5746	0.4305
	1.76	0.5731	0.4313
	1.77	0.5715	0.4321
	1.78	0.5699	0.4330
	1.79	0.5684	0.4338
	1.80	0.5669	0.4345
	1.81	0.5654	0.4353
	1.82	0.5639	0.4361
	1.83	0.5624	0.4369
	1.84	0.5609	0.4376
	1.85	0.5594	0.4384
	1.86	0.5580	0.4391
	1.87	0.5565	0.4399
	1.88	0.5551	0.4406
	1.89	0.5537	0.4413
	1.90	0.5522	0.4420
	1.91	0.5508	0.4427
	1.92	0.5494	0.4434
	1.93	0.5481	0.4441
	1.94	0.5467	0.4448
	1.95	0.5453	0.4455
	1.96	0.5440	0.4462
	1.97	0.5426	0.4468
	1.98	0.5413	0.4475
	1.99	0.5400	0.4481
	2.00	0.5387	0.4488
	2.01	0.5374	0.4494
	2.02	0.5361	0.4501
	2.03	0.5348	0.4507
	2.04	0.5335	0.4513
	2.05	0.5323	0.4519
	2.06	0.5310	0.4525
	2.07	0.5298	0.4531
	2.08	0.5285	0.4537
	2.09	0.5273	0.4543
	2.10	0.5261	0.4549
	2.11	0.5249	0.4555
	2.12	0.5237	0.4561
	2.13	0.5225	0.4567
	2.14	0.5213	0.4572
	2.15	0.5201	0.4578
	2.16	0.5189	0.4583
	2.17	0.5178	0.4589
	2.18	0.5166	0.4594
	2.19	0.5155	0.4600
	2.20	0.5143	0.4605
	2.21	0.5132	0.4610
	2.22	0.5121	0.4616
	2.23	0.5110	0.4621
	2.24	0.5099	0.4626
	2.25	0.5088	0.4631
	2.26	0.5077	0.4636
	2.27	0.5066	0.4641
	2.28	0.5055	0.4646
	2.29	0.5045	0.4651
	2.30	0.5034	0.4656
	2.31	0.5023	0.4661
	2.32	0.5013	0.4666
	2.33	0.5002	0.4671
	2.34	0.4992	0.4676
	2.35	0.4982	0.4680
	2.36	0.4972	0.4685
	2.37	0.4962	0.4690
	2.38	0.4951	0.4694
	2.39	0.4941	0.4699
	2.40	0.4931	0.4704
	2.41	0.4922	0.4708
	2.42	0.4912	0.4713
	2.43	0.4902	0.4717
	2.44	0.4892	0.4721
	2.45	0.4883	0.4726
	2.46	0.4873	0.4730
	2.47	0.4864	0.4734
	2.48	0.4854	0.4739
	2.49	0.4845	0.4743
	2.50	0.4835	0.4747
	2.51	0.4826	0.4751
	2.52	0.4817	0.4755
	2.53	0.4808	0.4760
	2.54	0.4799	0.4764
	2.55	0.4790	0.4768
	2.56	0.4781	0.4772
	2.57	0.4772	0.4776
	2.58	0.4763	0.4780
	2.59	0.4754	0.4784
	2.60	0.4745	0.4787
	2.61	0.4736	0.4791
	2.62	0.4728	0.4795
	2.63	0.4719	0.4799
	2.64	0.4710	0.4803
	2.65	0.4702	0.4807
	2.66	0.4693	0.4810
	2.67	0.4685	0.4814
	2.68	0.4677	0.4818
	2.69	0.4668	0.4821
	2.70	0.4660	0.4825
	2.71	0.4652	0.4829
	2.72	0.4644	0.4832
	2.73	0.4635	0.4836
	2.74	0.4627	0.4839
	2.75	0.4619	0.4843
	2.76	0.4611	0.4846
	2.77	0.4603	0.4850
	2.78	0.4596	0.4853
	2.79	0.4588	0.4857
	2.80	0.4580	0.4860
	2.81	0.4572	0.4863
	2.82	0.4564	0.4867
	2.83	0.4557	0.4870
	2.84	0.4549	0.4873
	2.85	0.4541	0.4877
	2.86	0.4534	0.4880
	2.87	0.4526	0.4883
	2.88	0.4519	0.4886
	2.89	0.4511	0.4890
	2.90	0.4504	0.4893
	2.91	0.4497	0.4896
	2.92	0.4489	0.4899
	2.93	0.4482	0.4902
	2.94	0.4475	0.4905
	2.95	0.4468	0.4908
	2.96	0.4460	0.4911
	2.97	0.4453	0.4914
	2.98	0.4446	0.4918
	2.99	0.4439	0.4921
	3.00	0.4432	0.4924
	3.01	0.4425	0.4926
	3.02	0.4418	0.4929
	3.03	0.4411	0.4932
	3.04	0.4405	0.4935
	3.05	0.4398	0.4938
	3.06	0.4391	0.4941
	3.07	0.4384	0.4944
	3.08	0.4377	0.4947
	3.09	0.4371	0.4950
	3.10	0.4364	0.4952
	3.11	0.4357	0.4955
	3.12	0.4351	0.4958
	3.13	0.4344	0.4961
	3.14	0.4338	0.4964
	3.15	0.4331	0.4966
	3.16	0.4325	0.4969
	3.17	0.4318	0.4972
	3.18	0.4312	0.4974
	3.19	0.4306	0.4977
	3.20	0.4299	0.4980
	3.21	0.4293	0.4982
	3.22	0.4287	0.4985
	3.23	0.4280	0.4988
	3.24	0.4274	0.4990
	3.25	0.4268	0.4993
	3.26	0.4262	0.4995
	3.27	0.4256	0.4998
	3.28	0.4250	0.5000
	3.29	0.4243	0.5003
	3.30	0.4237	0.5005
	3.31	0.4231	0.5008
	3.32	0.4225	0.5010
	3.33	0.4219	0.5013
	3.34	0.4214	0.5015
	3.35	0.4208	0.5018
	3.36	0.4202	0.5020
	3.37	0.4196	0.5023
	3.38	0.4190	0.5025
	3.39	0.4184	0.5027
	3.40	0.4178	0.5030
	3.41	0.4173	0.5032
	3.42	0.4167	0.5034
	3.43	0.4161	0.5037
	3.44	0.4156	0.5039
	3.45	0.4150	0.5041
	3.46	0.4144	0.5044
	3.47	0.4139	0.5046
	3.48	0.4133	0.5048
	3.49	0.4128	0.5051
	3.50	0.4122	0.5053
	3.51	0.4116	0.5055
	3.52	0.4111	0.5057
	3.53	0.4105	0.5060
	3.54	0.4100	0.5062
	3.55	0.4095	0.5064
	3.56	0.4089	0.5066
	3.57	0.4084	0.5068
	3.58	0.4078	0.5071
	3.59	0.4073	0.5073
	3.60	0.4068	0.5075
	3.61	0.4063	0.5077
	3.62	0.4057	0.5079
	3.63	0.4052	0.5081
	3.64	0.4047	0.5083
	3.65	0.4042	0.5086
	3.66	0.4036	0.5088
	3.67	0.4031	0.5090
	3.68	0.4026	0.5092
	3.69	0.4021	0.5094
	3.70	0.4016	0.5096
	3.71	0.4011	0.5098
	3.72	0.4006	0.5100
	3.73	0.4001	0.5102
	3.74	0.3996	0.5104
	3.75	0.3991	0.5106
	3.76	0.3986	0.5108
	3.77	0.3981	0.5110
	3.78	0.3976	0.5112
	3.79	0.3971	0.5114
	3.80	0.3966	0.5116
	3.81	0.3961	0.5118
	3.82	0.3956	0.5120
	3.83	0.3951	0.5122
	3.84	0.3946	0.5124
	3.85	0.3941	0.5125
	3.86	0.3937	0.5127
	3.87	0.3932	0.5129
	3.88	0.3927	0.5131
	3.89	0.3922	0.5133
	3.90	0.3918	0.5135
	3.91	0.3913	0.5137
	3.92	0.3908	0.5139
	3.93	0.3903	0.5140
	3.94	0.3899	0.5142
	3.95	0.3894	0.5144
	3.96	0.3889	0.5146
	3.97	0.3885	0.5148
	3.98	0.3880	0.5150
	3.99	0.3876	0.5151
	4.00	0.3871	0.5153
	4.01	0.3866	0.5155
	4.02	0.3862	0.5157
	4.03	0.3857	0.5159
	4.04	0.3853	0.5160
	4.05	0.3848	0.5162
	4.06	0.3844	0.5164
	4.07	0.3839	0.5166
	4.08	0.3835	0.5167
	4.09	0.3831	0.5169
}\tableCorrSiH

\pgfplotstableread{
	sigma	corr	si
	0.01	1.0000	0.0833
	0.02	1.0000	0.0833
	0.03	1.0000	0.0833
	0.04	1.0000	0.0833
	0.05	1.0000	0.0833
	0.06	1.0000	0.0833
	0.07	1.0000	0.0833
	0.08	1.0000	0.0833
	0.09	1.0000	0.0833
	0.10	1.0000	0.0833
	0.11	1.0000	0.0833
	0.12	1.0000	0.0833
	0.13	1.0000	0.0833
	0.14	1.0000	0.0833
	0.15	1.0000	0.0833
	0.16	1.0000	0.0833
	0.17	1.0000	0.0833
	0.18	1.0000	0.0833
	0.19	1.0000	0.0833
	0.20	1.0000	0.0833
	0.21	1.0000	0.0833
	0.22	1.0000	0.0833
	0.23	1.0000	0.0833
	0.24	1.0000	0.0833
	0.25	1.0000	0.0832
	0.26	1.0000	0.0832
	0.27	1.0000	0.0830
	0.28	1.0000	0.0828
	0.29	1.0000	0.0826
	0.30	1.0000	0.0822
	0.31	1.0000	0.0816
	0.32	0.9999	0.0809
	0.33	0.9999	0.0800
	0.34	0.9998	0.0787
	0.35	0.9997	0.0772
	0.36	0.9996	0.0752
	0.37	0.9993	0.0728
	0.38	0.9990	0.0698
	0.39	0.9986	0.0663
	0.40	0.9981	0.0621
	0.41	0.9974	0.0572
	0.42	0.9966	0.0515
	0.43	0.9955	0.0452
	0.44	0.9943	0.0382
	0.45	0.9929	0.0307
	0.46	0.9913	0.0226
	0.47	0.9894	0.0141
	0.48	0.9872	0.0051
	0.49	0.9849	0.0042
	0.50	0.9822	0.0137
	0.51	0.9794	0.0235
	0.52	0.9762	0.0335
	0.53	0.9729	0.0435
	0.54	0.9693	0.0535
	0.55	0.9655	0.0635
	0.56	0.9615	0.0735
	0.57	0.9573	0.0834
	0.58	0.9529	0.0931
	0.59	0.9483	0.1027
	0.60	0.9437	0.1120
	0.61	0.9389	0.1212
	0.62	0.9339	0.1302
	0.63	0.9289	0.1389
	0.64	0.9238	0.1474
	0.65	0.9187	0.1557
	0.66	0.9135	0.1637
	0.67	0.9083	0.1714
	0.68	0.9030	0.1789
	0.69	0.8977	0.1861
	0.70	0.8925	0.1931
	0.71	0.8872	0.1999
	0.72	0.8820	0.2064
	0.73	0.8768	0.2127
	0.74	0.8716	0.2188
	0.75	0.8665	0.2247
	0.76	0.8614	0.2304
	0.77	0.8564	0.2359
	0.78	0.8514	0.2412
	0.79	0.8464	0.2464
	0.80	0.8416	0.2513
	0.81	0.8368	0.2562
	0.82	0.8320	0.2608
	0.83	0.8273	0.2653
	0.84	0.8227	0.2697
	0.85	0.8182	0.2740
	0.86	0.8137	0.2781
	0.87	0.8093	0.2821
	0.88	0.8049	0.2860
	0.89	0.8006	0.2898
	0.90	0.7964	0.2934
	0.91	0.7923	0.2970
	0.92	0.7882	0.3005
	0.93	0.7841	0.3038
	0.94	0.7801	0.3071
	0.95	0.7762	0.3103
	0.96	0.7724	0.3134
	0.97	0.7686	0.3165
	0.98	0.7648	0.3194
	0.99	0.7611	0.3223
	1.00	0.7575	0.3251
	1.01	0.7539	0.3279
	1.02	0.7504	0.3306
	1.03	0.7469	0.3332
	1.04	0.7435	0.3358
	1.05	0.7401	0.3383
	1.06	0.7368	0.3408
	1.07	0.7335	0.3432
	1.08	0.7302	0.3455
	1.09	0.7270	0.3478
	1.10	0.7238	0.3501
	1.11	0.7207	0.3523
	1.12	0.7176	0.3545
	1.13	0.7146	0.3566
	1.14	0.7116	0.3587
	1.15	0.7086	0.3607
	1.16	0.7057	0.3627
	1.17	0.7028	0.3647
	1.18	0.6999	0.3666
	1.19	0.6971	0.3685
	1.20	0.6943	0.3703
	1.21	0.6916	0.3722
	1.22	0.6888	0.3739
	1.23	0.6861	0.3757
	1.24	0.6835	0.3774
	1.25	0.6808	0.3791
	1.26	0.6782	0.3808
	1.27	0.6756	0.3824
	1.28	0.6731	0.3841
	1.29	0.6706	0.3856
	1.30	0.6681	0.3872
	1.31	0.6656	0.3887
	1.32	0.6632	0.3903
	1.33	0.6607	0.3917
	1.34	0.6583	0.3932
	1.35	0.6560	0.3946
	1.36	0.6536	0.3961
	1.37	0.6513	0.3975
	1.38	0.6490	0.3989
	1.39	0.6467	0.4002
	1.40	0.6445	0.4015
	1.41	0.6422	0.4029
	1.42	0.6400	0.4042
	1.43	0.6378	0.4054
	1.44	0.6357	0.4067
	1.45	0.6335	0.4079
	1.46	0.6314	0.4092
	1.47	0.6293	0.4104
	1.48	0.6272	0.4116
	1.49	0.6251	0.4128
	1.50	0.6231	0.4139
	1.51	0.6210	0.4151
	1.52	0.6190	0.4162
	1.53	0.6170	0.4173
	1.54	0.6151	0.4184
	1.55	0.6131	0.4195
	1.56	0.6111	0.4206
	1.57	0.6092	0.4216
	1.58	0.6073	0.4227
	1.59	0.6054	0.4237
	1.60	0.6035	0.4247
	1.61	0.6017	0.4257
	1.62	0.5998	0.4267
	1.63	0.5980	0.4277
	1.64	0.5962	0.4287
	1.65	0.5944	0.4296
	1.66	0.5926	0.4306
	1.67	0.5908	0.4315
	1.68	0.5891	0.4324
	1.69	0.5873	0.4333
	1.70	0.5856	0.4343
	1.71	0.5839	0.4351
	1.72	0.5822	0.4360
	1.73	0.5805	0.4369
	1.74	0.5788	0.4378
	1.75	0.5772	0.4386
	1.76	0.5755	0.4395
	1.77	0.5739	0.4403
	1.78	0.5723	0.4411
	1.79	0.5706	0.4419
	1.80	0.5691	0.4427
	1.81	0.5675	0.4435
	1.82	0.5659	0.4443
	1.83	0.5643	0.4451
	1.84	0.5628	0.4459
	1.85	0.5612	0.4466
	1.86	0.5597	0.4474
	1.87	0.5582	0.4481
	1.88	0.5567	0.4489
	1.89	0.5552	0.4496
	1.90	0.5537	0.4503
	1.91	0.5523	0.4511
	1.92	0.5508	0.4518
	1.93	0.5494	0.4525
	1.94	0.5479	0.4532
	1.95	0.5465	0.4539
	1.96	0.5451	0.4545
	1.97	0.5437	0.4552
	1.98	0.5423	0.4559
	1.99	0.5409	0.4565
	2.00	0.5395	0.4572
	2.01	0.5381	0.4579
	2.02	0.5368	0.4585
	2.03	0.5354	0.4591
	2.04	0.5341	0.4598
	2.05	0.5328	0.4604
	2.06	0.5314	0.4610
	2.07	0.5301	0.4616
	2.08	0.5288	0.4622
	2.09	0.5275	0.4628
	2.10	0.5262	0.4634
	2.11	0.5250	0.4640
	2.12	0.5237	0.4646
	2.13	0.5224	0.4652
	2.14	0.5212	0.4658
	2.15	0.5199	0.4663
	2.16	0.5187	0.4669
	2.17	0.5175	0.4675
	2.18	0.5162	0.4680
	2.19	0.5150	0.4686
	2.20	0.5138	0.4691
	2.21	0.5126	0.4697
	2.22	0.5115	0.4702
	2.23	0.5103	0.4707
	2.24	0.5091	0.4713
	2.25	0.5079	0.4718
	2.26	0.5068	0.4723
	2.27	0.5056	0.4728
	2.28	0.5045	0.4733
	2.29	0.5034	0.4738
	2.30	0.5022	0.4743
	2.31	0.5011	0.4748
	2.32	0.5000	0.4753
	2.33	0.4989	0.4758
	2.34	0.4978	0.4763
	2.35	0.4967	0.4768
	2.36	0.4956	0.4772
	2.37	0.4945	0.4777
	2.38	0.4935	0.4782
	2.39	0.4924	0.4787
	2.40	0.4913	0.4791
	2.41	0.4903	0.4796
	2.42	0.4892	0.4800
	2.43	0.4882	0.4805
	2.44	0.4872	0.4809
	2.45	0.4861	0.4814
	2.46	0.4851	0.4818
	2.47	0.4841	0.4822
	2.48	0.4831	0.4827
	2.49	0.4821	0.4831
	2.50	0.4811	0.4835
	2.51	0.4801	0.4840
	2.52	0.4791	0.4844
	2.53	0.4781	0.4848
	2.54	0.4772	0.4852
	2.55	0.4762	0.4856
	2.56	0.4752	0.4860
	2.57	0.4743	0.4864
	2.58	0.4733	0.4868
	2.59	0.4724	0.4872
	2.60	0.4714	0.4876
	2.61	0.4705	0.4880
	2.62	0.4696	0.4884
	2.63	0.4686	0.4888
	2.64	0.4677	0.4892
	2.65	0.4668	0.4896
	2.66	0.4659	0.4900
	2.67	0.4650	0.4903
	2.68	0.4641	0.4907
	2.69	0.4632	0.4911
	2.70	0.4623	0.4914
	2.71	0.4615	0.4918
	2.72	0.4606	0.4922
	2.73	0.4597	0.4925
	2.74	0.4588	0.4929
	2.75	0.4580	0.4932
	2.76	0.4571	0.4936
	2.77	0.4563	0.4940
	2.78	0.4554	0.4943
	2.79	0.4546	0.4946
	2.80	0.4537	0.4950
	2.81	0.4529	0.4953
	2.82	0.4521	0.4957
	2.83	0.4513	0.4960
	2.84	0.4504	0.4963
	2.85	0.4496	0.4967
	2.86	0.4488	0.4970
	2.87	0.4480	0.4973
	2.88	0.4472	0.4977
	2.89	0.4464	0.4980
	2.90	0.4456	0.4983
	2.91	0.4448	0.4986
	2.92	0.4440	0.4989
	2.93	0.4432	0.4993
	2.94	0.4425	0.4996
	2.95	0.4417	0.4999
	2.96	0.4409	0.5002
	2.97	0.4402	0.5005
	2.98	0.4394	0.5008
	2.99	0.4386	0.5011
	3.00	0.4379	0.5014
	3.01	0.4371	0.5017
	3.02	0.4364	0.5020
	3.03	0.4357	0.5023
	3.04	0.4349	0.5026
	3.05	0.4342	0.5029
	3.06	0.4335	0.5032
	3.07	0.4327	0.5035
	3.08	0.4320	0.5038
	3.09	0.4313	0.5041
	3.10	0.4306	0.5043
	3.11	0.4299	0.5046
	3.12	0.4292	0.5049
	3.13	0.4285	0.5052
	3.14	0.4278	0.5055
	3.15	0.4271	0.5057
	3.16	0.4264	0.5060
	3.17	0.4257	0.5063
	3.18	0.4250	0.5066
	3.19	0.4243	0.5068
	3.20	0.4236	0.5071
	3.21	0.4230	0.5074
	3.22	0.4223	0.5076
	3.23	0.4216	0.5079
	3.24	0.4209	0.5081
	3.25	0.4203	0.5084
	3.26	0.4196	0.5087
	3.27	0.4190	0.5089
	3.28	0.4183	0.5092
	3.29	0.4177	0.5094
	3.30	0.4170	0.5097
	3.31	0.4164	0.5099
	3.32	0.4157	0.5102
	3.33	0.4151	0.5104
	3.34	0.4144	0.5107
	3.35	0.4138	0.5109
	3.36	0.4132	0.5112
	3.37	0.4126	0.5114
	3.38	0.4119	0.5117
	3.39	0.4113	0.5119
	3.40	0.4107	0.5121
	3.41	0.4101	0.5124
	3.42	0.4095	0.5126
	3.43	0.4089	0.5128
	3.44	0.4082	0.5131
	3.45	0.4076	0.5133
	3.46	0.4070	0.5135
	3.47	0.4064	0.5138
	3.48	0.4058	0.5140
	3.49	0.4053	0.5142
	3.50	0.4047	0.5145
	3.51	0.4041	0.5147
	3.52	0.4035	0.5149
	3.53	0.4029	0.5151
	3.54	0.4023	0.5154
	3.55	0.4017	0.5156
	3.56	0.4012	0.5158
	3.57	0.4006	0.5160
	3.58	0.4000	0.5162
	3.59	0.3995	0.5164
	3.60	0.3989	0.5167
	3.61	0.3983	0.5169
	3.62	0.3978	0.5171
	3.63	0.3972	0.5173
	3.64	0.3967	0.5175
	3.65	0.3961	0.5177
	3.66	0.3955	0.5179
	3.67	0.3950	0.5181
	3.68	0.3944	0.5184
	3.69	0.3939	0.5186
	3.70	0.3934	0.5188
	3.71	0.3928	0.5190
	3.72	0.3923	0.5192
	3.73	0.3917	0.5194
	3.74	0.3912	0.5196
	3.75	0.3907	0.5198
	3.76	0.3902	0.5200
	3.77	0.3896	0.5202
	3.78	0.3891	0.5204
	3.79	0.3886	0.5206
	3.80	0.3881	0.5208
	3.81	0.3875	0.5210
	3.82	0.3870	0.5211
	3.83	0.3865	0.5213
	3.84	0.3860	0.5215
	3.85	0.3855	0.5217
	3.86	0.3850	0.5219
	3.87	0.3845	0.5221
	3.88	0.3840	0.5223
	3.89	0.3835	0.5225
	3.90	0.3830	0.5227
	3.91	0.3825	0.5229
	3.92	0.3820	0.5230
	3.93	0.3815	0.5232
	3.94	0.3810	0.5234
	3.95	0.3805	0.5236
	3.96	0.3800	0.5238
	3.97	0.3795	0.5239
	3.98	0.3790	0.5241
	3.99	0.3785	0.5243
	4.00	0.3780	0.5245
	4.01	0.3776	0.5247
	4.02	0.3771	0.5248
	4.03	0.3766	0.5250
	4.04	0.3761	0.5252
	4.05	0.3757	0.5254
	4.06	0.3752	0.5255
	4.07	0.3747	0.5257
	4.08	0.3742	0.5259
	4.09	0.3738	0.5261
}\tableCorrSiI

\pgfplotstableread{
	sigma	corr	si
	0.01	1.0000	0.0833
	0.02	1.0000	0.0833
	0.03	1.0000	0.0833
	0.04	1.0000	0.0833
	0.05	1.0000	0.0833
	0.06	1.0000	0.0833
	0.07	1.0000	0.0833
	0.08	1.0000	0.0833
	0.09	1.0000	0.0833
	0.10	1.0000	0.0833
	0.11	1.0000	0.0833
	0.12	1.0000	0.0833
	0.13	1.0000	0.0833
	0.14	1.0000	0.0833
	0.15	1.0000	0.0833
	0.16	1.0000	0.0833
	0.17	1.0000	0.0833
	0.18	1.0000	0.0833
	0.19	1.0000	0.0833
	0.20	1.0000	0.0833
	0.21	1.0000	0.0833
	0.22	1.0000	0.0833
	0.23	1.0000	0.0833
	0.24	1.0000	0.0833
	0.25	1.0000	0.0832
	0.26	1.0000	0.0831
	0.27	1.0000	0.0830
	0.28	1.0000	0.0828
	0.29	1.0000	0.0825
	0.30	1.0000	0.0820
	0.31	1.0000	0.0814
	0.32	0.9999	0.0806
	0.33	0.9999	0.0795
	0.34	0.9998	0.0781
	0.35	0.9997	0.0763
	0.36	0.9996	0.0740
	0.37	0.9993	0.0712
	0.38	0.9990	0.0678
	0.39	0.9986	0.0636
	0.40	0.9981	0.0586
	0.41	0.9974	0.0528
	0.42	0.9966	0.0461
	0.43	0.9955	0.0386
	0.44	0.9943	0.0304
	0.45	0.9929	0.0215
	0.46	0.9912	0.0119
	0.47	0.9893	0.0018
	0.48	0.9872	0.0088
	0.49	0.9848	0.0198
	0.50	0.9821	0.0311
	0.51	0.9792	0.0426
	0.52	0.9760	0.0543
	0.53	0.9726	0.0661
	0.54	0.9689	0.0779
	0.55	0.9651	0.0896
	0.56	0.9610	0.1012
	0.57	0.9567	0.1126
	0.58	0.9522	0.1239
	0.59	0.9475	0.1348
	0.60	0.9427	0.1455
	0.61	0.9378	0.1559
	0.62	0.9328	0.1660
	0.63	0.9276	0.1758
	0.64	0.9224	0.1853
	0.65	0.9171	0.1944
	0.66	0.9117	0.2032
	0.67	0.9063	0.2117
	0.68	0.9009	0.2199
	0.69	0.8955	0.2278
	0.70	0.8900	0.2354
	0.71	0.8846	0.2428
	0.72	0.8792	0.2498
	0.73	0.8738	0.2566
	0.74	0.8684	0.2632
	0.75	0.8631	0.2695
	0.76	0.8578	0.2755
	0.77	0.8526	0.2814
	0.78	0.8474	0.2870
	0.79	0.8423	0.2925
	0.80	0.8372	0.2977
	0.81	0.8323	0.3028
	0.82	0.8273	0.3077
	0.83	0.8225	0.3124
	0.84	0.8177	0.3169
	0.85	0.8129	0.3214
	0.86	0.8083	0.3256
	0.87	0.8037	0.3298
	0.88	0.7991	0.3338
	0.89	0.7947	0.3377
	0.90	0.7903	0.3414
	0.91	0.7859	0.3451
	0.92	0.7817	0.3487
	0.93	0.7775	0.3521
	0.94	0.7733	0.3555
	0.95	0.7693	0.3587
	0.96	0.7652	0.3619
	0.97	0.7613	0.3650
	0.98	0.7574	0.3680
	0.99	0.7535	0.3709
	1.00	0.7497	0.3738
	1.01	0.7460	0.3765
	1.02	0.7423	0.3792
	1.03	0.7387	0.3819
	1.04	0.7351	0.3845
	1.05	0.7316	0.3870
	1.06	0.7281	0.3894
	1.07	0.7246	0.3918
	1.08	0.7213	0.3942
	1.09	0.7179	0.3965
	1.10	0.7146	0.3987
	1.11	0.7114	0.4009
	1.12	0.7081	0.4031
	1.13	0.7050	0.4052
	1.14	0.7018	0.4072
	1.15	0.6987	0.4092
	1.16	0.6957	0.4112
	1.17	0.6927	0.4131
	1.18	0.6897	0.4150
	1.19	0.6867	0.4169
	1.20	0.6838	0.4187
	1.21	0.6810	0.4205
	1.22	0.6781	0.4223
	1.23	0.6753	0.4240
	1.24	0.6725	0.4257
	1.25	0.6698	0.4273
	1.26	0.6671	0.4289
	1.27	0.6644	0.4305
	1.28	0.6618	0.4321
	1.29	0.6592	0.4336
	1.30	0.6566	0.4352
	1.31	0.6540	0.4367
	1.32	0.6515	0.4381
	1.33	0.6490	0.4395
	1.34	0.6465	0.4410
	1.35	0.6440	0.4424
	1.36	0.6416	0.4437
	1.37	0.6392	0.4451
	1.38	0.6368	0.4464
	1.39	0.6345	0.4477
	1.40	0.6322	0.4490
	1.41	0.6299	0.4502
	1.42	0.6276	0.4515
	1.43	0.6253	0.4527
	1.44	0.6231	0.4539
	1.45	0.6209	0.4551
	1.46	0.6187	0.4563
	1.47	0.6166	0.4574
	1.48	0.6144	0.4585
	1.49	0.6123	0.4597
	1.50	0.6102	0.4608
	1.51	0.6081	0.4618
	1.52	0.6061	0.4629
	1.53	0.6040	0.4640
	1.54	0.6020	0.4650
	1.55	0.6000	0.4660
	1.56	0.5980	0.4670
	1.57	0.5961	0.4680
	1.58	0.5941	0.4690
	1.59	0.5922	0.4700
	1.60	0.5903	0.4709
	1.61	0.5884	0.4719
	1.62	0.5865	0.4728
	1.63	0.5847	0.4737
	1.64	0.5828	0.4747
	1.65	0.5810	0.4755
	1.66	0.5792	0.4764
	1.67	0.5774	0.4773
	1.68	0.5757	0.4782
	1.69	0.5739	0.4790
	1.70	0.5722	0.4799
	1.71	0.5704	0.4807
	1.72	0.5687	0.4815
	1.73	0.5670	0.4823
	1.74	0.5654	0.4831
	1.75	0.5637	0.4839
	1.76	0.5620	0.4847
	1.77	0.5604	0.4855
	1.78	0.5588	0.4862
	1.79	0.5572	0.4870
	1.80	0.5556	0.4877
	1.81	0.5540	0.4885
	1.82	0.5524	0.4892
	1.83	0.5509	0.4899
	1.84	0.5494	0.4906
	1.85	0.5478	0.4913
	1.86	0.5463	0.4920
	1.87	0.5448	0.4927
	1.88	0.5433	0.4934
	1.89	0.5418	0.4941
	1.90	0.5404	0.4947
	1.91	0.5389	0.4954
	1.92	0.5375	0.4960
	1.93	0.5361	0.4967
	1.94	0.5346	0.4973
	1.95	0.5332	0.4979
	1.96	0.5318	0.4986
	1.97	0.5305	0.4992
	1.98	0.5291	0.4998
	1.99	0.5277	0.5004
	2.00	0.5264	0.5010
	2.01	0.5250	0.5016
	2.02	0.5237	0.5022
	2.03	0.5224	0.5027
	2.04	0.5211	0.5033
	2.05	0.5198	0.5039
	2.06	0.5185	0.5044
	2.07	0.5172	0.5050
	2.08	0.5159	0.5056
	2.09	0.5147	0.5061
	2.10	0.5134	0.5066
	2.11	0.5122	0.5072
	2.12	0.5109	0.5077
	2.13	0.5097	0.5082
	2.14	0.5085	0.5088
	2.15	0.5073	0.5093
	2.16	0.5061	0.5098
	2.17	0.5049	0.5103
	2.18	0.5037	0.5108
	2.19	0.5026	0.5113
	2.20	0.5014	0.5118
	2.21	0.5003	0.5123
	2.22	0.4991	0.5128
	2.23	0.4980	0.5132
	2.24	0.4968	0.5137
	2.25	0.4957	0.5142
	2.26	0.4946	0.5146
	2.27	0.4935	0.5151
	2.28	0.4924	0.5156
	2.29	0.4913	0.5160
	2.30	0.4902	0.5165
	2.31	0.4892	0.5169
	2.32	0.4881	0.5174
	2.33	0.4871	0.5178
	2.34	0.4860	0.5182
	2.35	0.4850	0.5187
	2.36	0.4839	0.5191
	2.37	0.4829	0.5195
	2.38	0.4819	0.5199
	2.39	0.4809	0.5204
	2.40	0.4798	0.5208
	2.41	0.4788	0.5212
	2.42	0.4779	0.5216
	2.43	0.4769	0.5220
	2.44	0.4759	0.5224
	2.45	0.4749	0.5228
	2.46	0.4739	0.5232
	2.47	0.4730	0.5236
	2.48	0.4720	0.5240
	2.49	0.4711	0.5243
	2.50	0.4701	0.5247
	2.51	0.4692	0.5251
	2.52	0.4683	0.5255
	2.53	0.4673	0.5258
	2.54	0.4664	0.5262
	2.55	0.4655	0.5266
	2.56	0.4646	0.5269
	2.57	0.4637	0.5273
	2.58	0.4628	0.5277
	2.59	0.4619	0.5280
	2.60	0.4610	0.5284
	2.61	0.4602	0.5287
	2.62	0.4593	0.5291
	2.63	0.4584	0.5294
	2.64	0.4576	0.5298
	2.65	0.4567	0.5301
	2.66	0.4558	0.5304
	2.67	0.4550	0.5308
	2.68	0.4542	0.5311
	2.69	0.4533	0.5314
	2.70	0.4525	0.5318
	2.71	0.4517	0.5321
	2.72	0.4509	0.5324
	2.73	0.4500	0.5327
	2.74	0.4492	0.5331
	2.75	0.4484	0.5334
	2.76	0.4476	0.5337
	2.77	0.4468	0.5340
	2.78	0.4460	0.5343
	2.79	0.4453	0.5346
	2.80	0.4445	0.5349
	2.81	0.4437	0.5352
	2.82	0.4429	0.5355
	2.83	0.4422	0.5358
	2.84	0.4414	0.5361
	2.85	0.4406	0.5364
	2.86	0.4399	0.5367
	2.87	0.4391	0.5370
	2.88	0.4384	0.5373
	2.89	0.4376	0.5376
	2.90	0.4369	0.5379
	2.91	0.4362	0.5382
	2.92	0.4354	0.5384
	2.93	0.4347	0.5387
	2.94	0.4340	0.5390
	2.95	0.4333	0.5393
	2.96	0.4326	0.5395
	2.97	0.4319	0.5398
	2.98	0.4312	0.5401
	2.99	0.4305	0.5404
	3.00	0.4298	0.5406
	3.01	0.4291	0.5409
	3.02	0.4284	0.5412
	3.03	0.4277	0.5414
	3.04	0.4270	0.5417
	3.05	0.4263	0.5419
	3.06	0.4257	0.5422
	3.07	0.4250	0.5425
	3.08	0.4243	0.5427
	3.09	0.4237	0.5430
	3.10	0.4230	0.5432
	3.11	0.4224	0.5435
	3.12	0.4217	0.5437
	3.13	0.4211	0.5440
	3.14	0.4204	0.5442
	3.15	0.4198	0.5445
	3.16	0.4191	0.5447
	3.17	0.4185	0.5449
	3.18	0.4179	0.5452
	3.19	0.4172	0.5454
	3.20	0.4166	0.5457
	3.21	0.4160	0.5459
	3.22	0.4154	0.5461
	3.23	0.4148	0.5464
	3.24	0.4141	0.5466
	3.25	0.4135	0.5468
	3.26	0.4129	0.5471
	3.27	0.4123	0.5473
	3.28	0.4117	0.5475
	3.29	0.4111	0.5477
	3.30	0.4105	0.5480
	3.31	0.4099	0.5482
	3.32	0.4094	0.5484
	3.33	0.4088	0.5486
	3.34	0.4082	0.5488
	3.35	0.4076	0.5491
	3.36	0.4070	0.5493
	3.37	0.4065	0.5495
	3.38	0.4059	0.5497
	3.39	0.4053	0.5499
	3.40	0.4047	0.5501
	3.41	0.4042	0.5504
	3.42	0.4036	0.5506
	3.43	0.4031	0.5508
	3.44	0.4025	0.5510
	3.45	0.4020	0.5512
	3.46	0.4014	0.5514
	3.47	0.4009	0.5516
	3.48	0.4003	0.5518
	3.49	0.3998	0.5520
	3.50	0.3992	0.5522
	3.51	0.3987	0.5524
	3.52	0.3982	0.5526
	3.53	0.3976	0.5528
	3.54	0.3971	0.5530
	3.55	0.3966	0.5532
	3.56	0.3960	0.5534
	3.57	0.3955	0.5536
	3.58	0.3950	0.5538
	3.59	0.3945	0.5540
	3.60	0.3940	0.5542
	3.61	0.3934	0.5544
	3.62	0.3929	0.5545
	3.63	0.3924	0.5547
	3.64	0.3919	0.5549
	3.65	0.3914	0.5551
	3.66	0.3909	0.5553
	3.67	0.3904	0.5555
	3.68	0.3899	0.5557
	3.69	0.3894	0.5559
	3.70	0.3889	0.5560
	3.71	0.3884	0.5562
	3.72	0.3879	0.5564
	3.73	0.3874	0.5566
	3.74	0.3870	0.5568
	3.75	0.3865	0.5569
	3.76	0.3860	0.5571
	3.77	0.3855	0.5573
	3.78	0.3850	0.5575
	3.79	0.3846	0.5576
	3.80	0.3841	0.5578
	3.81	0.3836	0.5580
	3.82	0.3831	0.5582
	3.83	0.3827	0.5583
	3.84	0.3822	0.5585
	3.85	0.3817	0.5587
	3.86	0.3813	0.5588
	3.87	0.3808	0.5590
	3.88	0.3804	0.5592
	3.89	0.3799	0.5593
	3.90	0.3794	0.5595
	3.91	0.3790	0.5597
	3.92	0.3785	0.5598
	3.93	0.3781	0.5600
	3.94	0.3776	0.5602
	3.95	0.3772	0.5603
	3.96	0.3767	0.5605
	3.97	0.3763	0.5607
	3.98	0.3759	0.5608
	3.99	0.3754	0.5610
	4.00	0.3750	0.5611
	4.01	0.3745	0.5613
	4.02	0.3741	0.5615
	4.03	0.3737	0.5616
	4.04	0.3732	0.5618
	4.05	0.3728	0.5619
	4.06	0.3724	0.5621
	4.07	0.3719	0.5622
	4.08	0.3715	0.5624
	4.09	0.3711	0.5625
}\tableCorrSiJ

\pgfplotstableread{
	sigma	corr	si
	0.01	1.0000	0.0833
	0.02	1.0000	0.0833
	0.03	1.0000	0.0833
	0.04	1.0000	0.0833
	0.05	1.0000	0.0833
	0.06	1.0000	0.0833
	0.07	1.0000	0.0833
	0.08	1.0000	0.0833
	0.09	1.0000	0.0833
	0.10	1.0000	0.0833
	0.11	1.0000	0.0833
	0.12	1.0000	0.0833
	0.13	1.0000	0.0833
	0.14	1.0000	0.0833
	0.15	1.0000	0.0833
	0.16	1.0000	0.0833
	0.17	1.0000	0.0833
	0.18	1.0000	0.0833
	0.19	1.0000	0.0833
	0.20	1.0000	0.0833
	0.21	1.0000	0.0833
	0.22	1.0000	0.0833
	0.23	1.0000	0.0833
	0.24	1.0000	0.0833
	0.25	1.0000	0.0832
	0.26	1.0000	0.0832
	0.27	1.0000	0.0830
	0.28	1.0000	0.0828
	0.29	1.0000	0.0826
	0.30	1.0000	0.0822
	0.31	1.0000	0.0816
	0.32	0.9999	0.0809
	0.33	0.9999	0.0799
	0.34	0.9998	0.0787
	0.35	0.9997	0.0771
	0.36	0.9995	0.0751
	0.37	0.9993	0.0726
	0.38	0.9990	0.0696
	0.39	0.9986	0.0660
	0.40	0.9980	0.0617
	0.41	0.9974	0.0567
	0.42	0.9965	0.0509
	0.43	0.9955	0.0444
	0.44	0.9942	0.0373
	0.45	0.9928	0.0296
	0.46	0.9911	0.0214
	0.47	0.9892	0.0127
	0.48	0.9870	0.0035
	0.49	0.9846	0.0060
	0.50	0.9819	0.0157
	0.51	0.9790	0.0257
	0.52	0.9758	0.0358
	0.53	0.9723	0.0460
	0.54	0.9687	0.0562
	0.55	0.9648	0.0664
	0.56	0.9607	0.0765
	0.57	0.9564	0.0865
	0.58	0.9519	0.0964
	0.59	0.9472	0.1060
	0.60	0.9425	0.1155
	0.61	0.9375	0.1247
	0.62	0.9325	0.1337
	0.63	0.9274	0.1425
	0.64	0.9222	0.1509
	0.65	0.9169	0.1591
	0.66	0.9116	0.1671
	0.67	0.9062	0.1748
	0.68	0.9008	0.1822
	0.69	0.8954	0.1894
	0.70	0.8901	0.1964
	0.71	0.8847	0.2031
	0.72	0.8793	0.2095
	0.73	0.8740	0.2158
	0.74	0.8687	0.2218
	0.75	0.8635	0.2276
	0.76	0.8583	0.2333
	0.77	0.8532	0.2387
	0.78	0.8481	0.2439
	0.79	0.8430	0.2490
	0.80	0.8381	0.2539
	0.81	0.8332	0.2587
	0.82	0.8284	0.2632
	0.83	0.8236	0.2677
	0.84	0.8189	0.2720
	0.85	0.8143	0.2762
	0.86	0.8097	0.2802
	0.87	0.8052	0.2841
	0.88	0.8008	0.2879
	0.89	0.7965	0.2916
	0.90	0.7922	0.2952
	0.91	0.7880	0.2987
	0.92	0.7838	0.3021
	0.93	0.7798	0.3054
	0.94	0.7757	0.3086
	0.95	0.7718	0.3117
	0.96	0.7679	0.3147
	0.97	0.7641	0.3177
	0.98	0.7603	0.3206
	0.99	0.7566	0.3234
	1.00	0.7529	0.3261
	1.01	0.7493	0.3288
	1.02	0.7457	0.3314
	1.03	0.7422	0.3340
	1.04	0.7388	0.3365
	1.05	0.7354	0.3389
	1.06	0.7320	0.3413
	1.07	0.7287	0.3436
	1.08	0.7254	0.3459
	1.09	0.7222	0.3482
	1.10	0.7190	0.3503
	1.11	0.7159	0.3525
	1.12	0.7128	0.3546
	1.13	0.7097	0.3566
	1.14	0.7067	0.3587
	1.15	0.7037	0.3606
	1.16	0.7008	0.3626
	1.17	0.6979	0.3645
	1.18	0.6950	0.3663
	1.19	0.6922	0.3682
	1.20	0.6894	0.3700
	1.21	0.6867	0.3717
	1.22	0.6839	0.3735
	1.23	0.6812	0.3752
	1.24	0.6786	0.3768
	1.25	0.6759	0.3785
	1.26	0.6733	0.3801
	1.27	0.6708	0.3817
	1.28	0.6682	0.3832
	1.29	0.6657	0.3848
	1.30	0.6632	0.3863
	1.31	0.6608	0.3877
	1.32	0.6583	0.3892
	1.33	0.6559	0.3906
	1.34	0.6535	0.3920
	1.35	0.6512	0.3934
	1.36	0.6488	0.3948
	1.37	0.6465	0.3961
	1.38	0.6443	0.3975
	1.39	0.6420	0.3988
	1.40	0.6398	0.4001
	1.41	0.6375	0.4013
	1.42	0.6354	0.4026
	1.43	0.6332	0.4038
	1.44	0.6310	0.4050
	1.45	0.6289	0.4062
	1.46	0.6268	0.4074
	1.47	0.6247	0.4086
	1.48	0.6227	0.4097
	1.49	0.6206	0.4108
	1.50	0.6186	0.4120
	1.51	0.6166	0.4131
	1.52	0.6146	0.4141
	1.53	0.6126	0.4152
	1.54	0.6107	0.4163
	1.55	0.6087	0.4173
	1.56	0.6068	0.4183
	1.57	0.6049	0.4193
	1.58	0.6031	0.4203
	1.59	0.6012	0.4213
	1.60	0.5994	0.4223
	1.61	0.5975	0.4233
	1.62	0.5957	0.4242
	1.63	0.5939	0.4252
	1.64	0.5921	0.4261
	1.65	0.5904	0.4270
	1.66	0.5886	0.4279
	1.67	0.5869	0.4288
	1.68	0.5852	0.4297
	1.69	0.5835	0.4306
	1.70	0.5818	0.4314
	1.71	0.5801	0.4323
	1.72	0.5784	0.4331
	1.73	0.5768	0.4340
	1.74	0.5752	0.4348
	1.75	0.5735	0.4356
	1.76	0.5719	0.4364
	1.77	0.5703	0.4372
	1.78	0.5688	0.4380
	1.79	0.5672	0.4388
	1.80	0.5656	0.4395
	1.81	0.5641	0.4403
	1.82	0.5626	0.4410
	1.83	0.5610	0.4418
	1.84	0.5595	0.4425
	1.85	0.5580	0.4432
	1.86	0.5566	0.4440
	1.87	0.5551	0.4447
	1.88	0.5536	0.4454
	1.89	0.5522	0.4461
	1.90	0.5507	0.4468
	1.91	0.5493	0.4475
	1.92	0.5479	0.4481
	1.93	0.5465	0.4488
	1.94	0.5451	0.4495
	1.95	0.5437	0.4501
	1.96	0.5423	0.4508
	1.97	0.5410	0.4514
	1.98	0.5396	0.4521
	1.99	0.5383	0.4527
	2.00	0.5369	0.4533
	2.01	0.5356	0.4539
	2.02	0.5343	0.4545
	2.03	0.5330	0.4551
	2.04	0.5317	0.4558
	2.05	0.5304	0.4563
	2.06	0.5291	0.4569
	2.07	0.5279	0.4575
	2.08	0.5266	0.4581
	2.09	0.5253	0.4587
	2.10	0.5241	0.4592
	2.11	0.5229	0.4598
	2.12	0.5216	0.4604
	2.13	0.5204	0.4609
	2.14	0.5192	0.4615
	2.15	0.5180	0.4620
	2.16	0.5168	0.4626
	2.17	0.5156	0.4631
	2.18	0.5144	0.4636
	2.19	0.5133	0.4641
	2.20	0.5121	0.4647
	2.21	0.5109	0.4652
	2.22	0.5098	0.4657
	2.23	0.5087	0.4662
	2.24	0.5075	0.4667
	2.25	0.5064	0.4672
	2.26	0.5053	0.4677
	2.27	0.5042	0.4682
	2.28	0.5031	0.4687
	2.29	0.5020	0.4691
	2.30	0.5009	0.4696
	2.31	0.4998	0.4701
	2.32	0.4987	0.4706
	2.33	0.4976	0.4710
	2.34	0.4966	0.4715
	2.35	0.4955	0.4720
	2.36	0.4945	0.4724
	2.37	0.4934	0.4729
	2.38	0.4924	0.4733
	2.39	0.4913	0.4738
	2.40	0.4903	0.4742
	2.41	0.4893	0.4746
	2.42	0.4883	0.4751
	2.43	0.4873	0.4755
	2.44	0.4863	0.4759
	2.45	0.4853	0.4764
	2.46	0.4843	0.4768
	2.47	0.4833	0.4772
	2.48	0.4823	0.4776
	2.49	0.4813	0.4780
	2.50	0.4804	0.4784
	2.51	0.4794	0.4788
	2.52	0.4784	0.4792
	2.53	0.4775	0.4796
	2.54	0.4765	0.4800
	2.55	0.4756	0.4804
	2.56	0.4747	0.4808
	2.57	0.4737	0.4812
	2.58	0.4728	0.4816
	2.59	0.4719	0.4820
	2.60	0.4710	0.4824
	2.61	0.4701	0.4827
	2.62	0.4692	0.4831
	2.63	0.4683	0.4835
	2.64	0.4674	0.4839
	2.65	0.4665	0.4842
	2.66	0.4656	0.4846
	2.67	0.4647	0.4849
	2.68	0.4639	0.4853
	2.69	0.4630	0.4857
	2.70	0.4621	0.4860
	2.71	0.4613	0.4864
	2.72	0.4604	0.4867
	2.73	0.4595	0.4871
	2.74	0.4587	0.4874
	2.75	0.4579	0.4878
	2.76	0.4570	0.4881
	2.77	0.4562	0.4884
	2.78	0.4554	0.4888
	2.79	0.4545	0.4891
	2.80	0.4537	0.4894
	2.81	0.4529	0.4898
	2.82	0.4521	0.4901
	2.83	0.4513	0.4904
	2.84	0.4505	0.4907
	2.85	0.4497	0.4911
	2.86	0.4489	0.4914
	2.87	0.4481	0.4917
	2.88	0.4473	0.4920
	2.89	0.4465	0.4923
	2.90	0.4458	0.4926
	2.91	0.4450	0.4929
	2.92	0.4442	0.4932
	2.93	0.4434	0.4935
	2.94	0.4427	0.4938
	2.95	0.4419	0.4941
	2.96	0.4412	0.4944
	2.97	0.4404	0.4947
	2.98	0.4397	0.4950
	2.99	0.4389	0.4953
	3.00	0.4382	0.4956
	3.01	0.4375	0.4959
	3.02	0.4367	0.4962
	3.03	0.4360	0.4965
	3.04	0.4353	0.4968
	3.05	0.4346	0.4971
	3.06	0.4338	0.4973
	3.07	0.4331	0.4976
	3.08	0.4324	0.4979
	3.09	0.4317	0.4982
	3.10	0.4310	0.4984
	3.11	0.4303	0.4987
	3.12	0.4296	0.4990
	3.13	0.4289	0.4993
	3.14	0.4282	0.4995
	3.15	0.4276	0.4998
	3.16	0.4269	0.5001
	3.17	0.4262	0.5003
	3.18	0.4255	0.5006
	3.19	0.4249	0.5008
	3.20	0.4242	0.5011
	3.21	0.4235	0.5014
	3.22	0.4229	0.5016
	3.23	0.4222	0.5019
	3.24	0.4215	0.5021
	3.25	0.4209	0.5024
	3.26	0.4202	0.5026
	3.27	0.4196	0.5029
	3.28	0.4189	0.5031
	3.29	0.4183	0.5034
	3.30	0.4177	0.5036
	3.31	0.4170	0.5039
	3.32	0.4164	0.5041
	3.33	0.4158	0.5043
	3.34	0.4152	0.5046
	3.35	0.4145	0.5048
	3.36	0.4139	0.5051
	3.37	0.4133	0.5053
	3.38	0.4127	0.5055
	3.39	0.4121	0.5058
	3.40	0.4115	0.5060
	3.41	0.4108	0.5062
	3.42	0.4102	0.5065
	3.43	0.4096	0.5067
	3.44	0.4090	0.5069
	3.45	0.4085	0.5071
	3.46	0.4079	0.5074
	3.47	0.4073	0.5076
	3.48	0.4067	0.5078
	3.49	0.4061	0.5080
	3.50	0.4055	0.5082
	3.51	0.4049	0.5085
	3.52	0.4044	0.5087
	3.53	0.4038	0.5089
	3.54	0.4032	0.5091
	3.55	0.4027	0.5093
	3.56	0.4021	0.5095
	3.57	0.4015	0.5098
	3.58	0.4010	0.5100
	3.59	0.4004	0.5102
	3.60	0.3998	0.5104
	3.61	0.3993	0.5106
	3.62	0.3987	0.5108
	3.63	0.3982	0.5110
	3.64	0.3977	0.5112
	3.65	0.3971	0.5114
	3.66	0.3966	0.5116
	3.67	0.3960	0.5118
	3.68	0.3955	0.5120
	3.69	0.3950	0.5122
	3.70	0.3944	0.5124
	3.71	0.3939	0.5126
	3.72	0.3934	0.5128
	3.73	0.3928	0.5130
	3.74	0.3923	0.5132
	3.75	0.3918	0.5134
	3.76	0.3913	0.5136
	3.77	0.3908	0.5138
	3.78	0.3902	0.5140
	3.79	0.3897	0.5142
	3.80	0.3892	0.5144
	3.81	0.3887	0.5145
	3.82	0.3882	0.5147
	3.83	0.3877	0.5149
	3.84	0.3872	0.5151
	3.85	0.3867	0.5153
	3.86	0.3862	0.5155
	3.87	0.3857	0.5157
	3.88	0.3852	0.5158
	3.89	0.3847	0.5160
	3.90	0.3842	0.5162
	3.91	0.3837	0.5164
	3.92	0.3833	0.5166
	3.93	0.3828	0.5167
	3.94	0.3823	0.5169
	3.95	0.3818	0.5171
	3.96	0.3813	0.5173
	3.97	0.3809	0.5174
	3.98	0.3804	0.5176
	3.99	0.3799	0.5178
	4.00	0.3794	0.5180
	4.01	0.3790	0.5181
	4.02	0.3785	0.5183
	4.03	0.3780	0.5185
	4.04	0.3776	0.5186
	4.05	0.3771	0.5188
	4.06	0.3767	0.5190
	4.07	0.3762	0.5192
	4.08	0.3757	0.5193
	4.09	0.3753	0.5195
}\tableCorrSiK

\pgfplotstableread{
	sigma	corr	si
	0.01	1.0000	0.0833
	0.02	1.0000	0.0833
	0.03	1.0000	0.0833
	0.04	1.0000	0.0833
	0.05	1.0000	0.0833
	0.06	1.0000	0.0833
	0.07	1.0000	0.0833
	0.08	1.0000	0.0833
	0.09	1.0000	0.0833
	0.10	1.0000	0.0833
	0.11	1.0000	0.0833
	0.12	1.0000	0.0833
	0.13	1.0000	0.0833
	0.14	1.0000	0.0833
	0.15	1.0000	0.0833
	0.16	1.0000	0.0833
	0.17	1.0000	0.0833
	0.18	1.0000	0.0833
	0.19	1.0000	0.0833
	0.20	1.0000	0.0833
	0.21	1.0000	0.0833
	0.22	1.0000	0.0833
	0.23	1.0000	0.0833
	0.24	1.0000	0.0833
	0.25	1.0000	0.0832
	0.26	1.0000	0.0831
	0.27	1.0000	0.0830
	0.28	1.0000	0.0828
	0.29	1.0000	0.0824
	0.30	1.0000	0.0820
	0.31	1.0000	0.0813
	0.32	0.9999	0.0805
	0.33	0.9999	0.0793
	0.34	0.9998	0.0778
	0.35	0.9997	0.0759
	0.36	0.9996	0.0735
	0.37	0.9993	0.0704
	0.38	0.9990	0.0667
	0.39	0.9986	0.0623
	0.40	0.9981	0.0569
	0.41	0.9974	0.0506
	0.42	0.9966	0.0434
	0.43	0.9956	0.0354
	0.44	0.9944	0.0265
	0.45	0.9930	0.0169
	0.46	0.9913	0.0066
	0.47	0.9894	0.0043
	0.48	0.9873	0.0157
	0.49	0.9849	0.0275
	0.50	0.9822	0.0396
	0.51	0.9793	0.0520
	0.52	0.9762	0.0645
	0.53	0.9727	0.0772
	0.54	0.9691	0.0898
	0.55	0.9652	0.1023
	0.56	0.9611	0.1147
	0.57	0.9568	0.1269
	0.58	0.9523	0.1388
	0.59	0.9477	0.1505
	0.60	0.9428	0.1618
	0.61	0.9379	0.1728
	0.62	0.9328	0.1834
	0.63	0.9276	0.1938
	0.64	0.9223	0.2037
	0.65	0.9170	0.2133
	0.66	0.9115	0.2225
	0.67	0.9061	0.2314
	0.68	0.9006	0.2400
	0.69	0.8951	0.2482
	0.70	0.8895	0.2561
	0.71	0.8840	0.2637
	0.72	0.8785	0.2710
	0.73	0.8730	0.2780
	0.74	0.8675	0.2847
	0.75	0.8621	0.2912
	0.76	0.8567	0.2975
	0.77	0.8513	0.3035
	0.78	0.8460	0.3093
	0.79	0.8408	0.3149
	0.80	0.8356	0.3202
	0.81	0.8305	0.3254
	0.82	0.8254	0.3304
	0.83	0.8204	0.3353
	0.84	0.8154	0.3399
	0.85	0.8105	0.3445
	0.86	0.8057	0.3488
	0.87	0.8010	0.3530
	0.88	0.7963	0.3571
	0.89	0.7916	0.3611
	0.90	0.7871	0.3649
	0.91	0.7826	0.3687
	0.92	0.7782	0.3723
	0.93	0.7738	0.3758
	0.94	0.7695	0.3792
	0.95	0.7653	0.3825
	0.96	0.7611	0.3857
	0.97	0.7570	0.3888
	0.98	0.7529	0.3918
	0.99	0.7490	0.3948
	1.00	0.7450	0.3976
	1.01	0.7411	0.4004
	1.02	0.7373	0.4032
	1.03	0.7336	0.4058
	1.04	0.7299	0.4084
	1.05	0.7262	0.4109
	1.06	0.7226	0.4134
	1.07	0.7191	0.4158
	1.08	0.7156	0.4181
	1.09	0.7121	0.4204
	1.10	0.7087	0.4226
	1.11	0.7054	0.4248
	1.12	0.7021	0.4270
	1.13	0.6988	0.4291
	1.14	0.6956	0.4311
	1.15	0.6924	0.4331
	1.16	0.6893	0.4350
	1.17	0.6862	0.4369
	1.18	0.6832	0.4388
	1.19	0.6802	0.4406
	1.20	0.6772	0.4424
	1.21	0.6743	0.4442
	1.22	0.6714	0.4459
	1.23	0.6686	0.4476
	1.24	0.6658	0.4493
	1.25	0.6630	0.4509
	1.26	0.6603	0.4525
	1.27	0.6576	0.4540
	1.28	0.6549	0.4556
	1.29	0.6523	0.4571
	1.30	0.6497	0.4585
	1.31	0.6471	0.4600
	1.32	0.6446	0.4614
	1.33	0.6421	0.4628
	1.34	0.6396	0.4642
	1.35	0.6372	0.4655
	1.36	0.6348	0.4668
	1.37	0.6324	0.4681
	1.38	0.6301	0.4694
	1.39	0.6278	0.4707
	1.40	0.6255	0.4719
	1.41	0.6232	0.4731
	1.42	0.6210	0.4743
	1.43	0.6187	0.4755
	1.44	0.6166	0.4767
	1.45	0.6144	0.4778
	1.46	0.6123	0.4789
	1.47	0.6101	0.4800
	1.48	0.6081	0.4811
	1.49	0.6060	0.4822
	1.50	0.6039	0.4832
	1.51	0.6019	0.4843
	1.52	0.5999	0.4853
	1.53	0.5979	0.4863
	1.54	0.5960	0.4873
	1.55	0.5941	0.4883
	1.56	0.5921	0.4893
	1.57	0.5902	0.4902
	1.58	0.5884	0.4911
	1.59	0.5865	0.4921
	1.60	0.5847	0.4930
	1.61	0.5829	0.4939
	1.62	0.5811	0.4948
	1.63	0.5793	0.4957
	1.64	0.5775	0.4965
	1.65	0.5758	0.4974
	1.66	0.5741	0.4982
	1.67	0.5723	0.4991
	1.68	0.5707	0.4999
	1.69	0.5690	0.5007
	1.70	0.5673	0.5015
	1.71	0.5657	0.5023
	1.72	0.5640	0.5031
	1.73	0.5624	0.5038
	1.74	0.5608	0.5046
	1.75	0.5593	0.5053
	1.76	0.5577	0.5061
	1.77	0.5561	0.5068
	1.78	0.5546	0.5075
	1.79	0.5531	0.5083
	1.80	0.5516	0.5090
	1.81	0.5501	0.5097
	1.82	0.5486	0.5104
	1.83	0.5471	0.5110
	1.84	0.5456	0.5117
	1.85	0.5442	0.5124
	1.86	0.5428	0.5131
	1.87	0.5413	0.5137
	1.88	0.5399	0.5144
	1.89	0.5385	0.5150
	1.90	0.5371	0.5156
	1.91	0.5358	0.5163
	1.92	0.5344	0.5169
	1.93	0.5331	0.5175
	1.94	0.5317	0.5181
	1.95	0.5304	0.5187
	1.96	0.5291	0.5193
	1.97	0.5278	0.5199
	1.98	0.5265	0.5205
	1.99	0.5252	0.5210
	2.00	0.5239	0.5216
	2.01	0.5226	0.5222
	2.02	0.5214	0.5227
	2.03	0.5201	0.5233
	2.04	0.5189	0.5238
	2.05	0.5177	0.5244
	2.06	0.5164	0.5249
	2.07	0.5152	0.5255
	2.08	0.5140	0.5260
	2.09	0.5128	0.5265
	2.10	0.5117	0.5270
	2.11	0.5105	0.5275
	2.12	0.5093	0.5280
	2.13	0.5082	0.5286
	2.14	0.5070	0.5291
	2.15	0.5059	0.5295
	2.16	0.5047	0.5300
	2.17	0.5036	0.5305
	2.18	0.5025	0.5310
	2.19	0.5014	0.5315
	2.20	0.5003	0.5320
	2.21	0.4992	0.5324
	2.22	0.4981	0.5329
	2.23	0.4970	0.5333
	2.24	0.4960	0.5338
	2.25	0.4949	0.5343
	2.26	0.4939	0.5347
	2.27	0.4928	0.5351
	2.28	0.4918	0.5356
	2.29	0.4907	0.5360
	2.30	0.4897	0.5365
	2.31	0.4887	0.5369
	2.32	0.4877	0.5373
	2.33	0.4867	0.5377
	2.34	0.4857	0.5382
	2.35	0.4847	0.5386
	2.36	0.4837	0.5390
	2.37	0.4827	0.5394
	2.38	0.4817	0.5398
	2.39	0.4807	0.5402
	2.40	0.4798	0.5406
	2.41	0.4788	0.5410
	2.42	0.4779	0.5414
	2.43	0.4769	0.5418
	2.44	0.4760	0.5422
	2.45	0.4751	0.5426
	2.46	0.4741	0.5429
	2.47	0.4732	0.5433
	2.48	0.4723	0.5437
	2.49	0.4714	0.5441
	2.50	0.4705	0.5444
	2.51	0.4696	0.5448
	2.52	0.4687	0.5452
	2.53	0.4678	0.5455
	2.54	0.4669	0.5459
	2.55	0.4660	0.5462
	2.56	0.4651	0.5466
	2.57	0.4643	0.5469
	2.58	0.4634	0.5473
	2.59	0.4625	0.5476
	2.60	0.4617	0.5480
	2.61	0.4608	0.5483
	2.62	0.4600	0.5487
	2.63	0.4592	0.5490
	2.64	0.4583	0.5493
	2.65	0.4575	0.5497
	2.66	0.4567	0.5500
	2.67	0.4558	0.5503
	2.68	0.4550	0.5506
	2.69	0.4542	0.5510
	2.70	0.4534	0.5513
	2.71	0.4526	0.5516
	2.72	0.4518	0.5519
	2.73	0.4510	0.5522
	2.74	0.4502	0.5525
	2.75	0.4494	0.5528
	2.76	0.4487	0.5532
	2.77	0.4479	0.5535
	2.78	0.4471	0.5538
	2.79	0.4463	0.5541
	2.80	0.4456	0.5544
	2.81	0.4448	0.5547
	2.82	0.4440	0.5550
	2.83	0.4433	0.5553
	2.84	0.4425	0.5555
	2.85	0.4418	0.5558
	2.86	0.4411	0.5561
	2.87	0.4403	0.5564
	2.88	0.4396	0.5567
	2.89	0.4388	0.5570
	2.90	0.4381	0.5573
	2.91	0.4374	0.5575
	2.92	0.4367	0.5578
	2.93	0.4360	0.5581
	2.94	0.4352	0.5584
	2.95	0.4345	0.5586
	2.96	0.4338	0.5589
	2.97	0.4331	0.5592
	2.98	0.4324	0.5594
	2.99	0.4317	0.5597
	3.00	0.4310	0.5600
	3.01	0.4303	0.5602
	3.02	0.4297	0.5605
	3.03	0.4290	0.5607
	3.04	0.4283	0.5610
	3.05	0.4276	0.5613
	3.06	0.4270	0.5615
	3.07	0.4263	0.5618
	3.08	0.4256	0.5620
	3.09	0.4250	0.5623
	3.10	0.4243	0.5625
	3.11	0.4236	0.5628
	3.12	0.4230	0.5630
	3.13	0.4223	0.5633
	3.14	0.4217	0.5635
	3.15	0.4210	0.5637
	3.16	0.4204	0.5640
	3.17	0.4198	0.5642
	3.18	0.4191	0.5645
	3.19	0.4185	0.5647
	3.20	0.4179	0.5649
	3.21	0.4172	0.5652
	3.22	0.4166	0.5654
	3.23	0.4160	0.5656
	3.24	0.4154	0.5659
	3.25	0.4147	0.5661
	3.26	0.4141	0.5663
	3.27	0.4135	0.5665
	3.28	0.4129	0.5668
	3.29	0.4123	0.5670
	3.30	0.4117	0.5672
	3.31	0.4111	0.5674
	3.32	0.4105	0.5676
	3.33	0.4099	0.5679
	3.34	0.4093	0.5681
	3.35	0.4087	0.5683
	3.36	0.4081	0.5685
	3.37	0.4076	0.5687
	3.38	0.4070	0.5689
	3.39	0.4064	0.5692
	3.40	0.4058	0.5694
	3.41	0.4052	0.5696
	3.42	0.4047	0.5698
	3.43	0.4041	0.5700
	3.44	0.4035	0.5702
	3.45	0.4030	0.5704
	3.46	0.4024	0.5706
	3.47	0.4018	0.5708
	3.48	0.4013	0.5710
	3.49	0.4007	0.5712
	3.50	0.4002	0.5714
	3.51	0.3996	0.5716
	3.52	0.3991	0.5718
	3.53	0.3985	0.5720
	3.54	0.3980	0.5722
	3.55	0.3974	0.5724
	3.56	0.3969	0.5726
	3.57	0.3963	0.5728
	3.58	0.3958	0.5730
	3.59	0.3953	0.5732
	3.60	0.3947	0.5734
	3.61	0.3942	0.5735
	3.62	0.3937	0.5737
	3.63	0.3931	0.5739
	3.64	0.3926	0.5741
	3.65	0.3921	0.5743
	3.66	0.3916	0.5745
	3.67	0.3911	0.5747
	3.68	0.3905	0.5748
	3.69	0.3900	0.5750
	3.70	0.3895	0.5752
	3.71	0.3890	0.5754
	3.72	0.3885	0.5756
	3.73	0.3880	0.5757
	3.74	0.3875	0.5759
	3.75	0.3870	0.5761
	3.76	0.3865	0.5763
	3.77	0.3860	0.5765
	3.78	0.3855	0.5766
	3.79	0.3850	0.5768
	3.80	0.3845	0.5770
	3.81	0.3840	0.5771
	3.82	0.3835	0.5773
	3.83	0.3830	0.5775
	3.84	0.3825	0.5777
	3.85	0.3820	0.5778
	3.86	0.3816	0.5780
	3.87	0.3811	0.5782
	3.88	0.3806	0.5783
	3.89	0.3801	0.5785
	3.90	0.3796	0.5787
	3.91	0.3792	0.5788
	3.92	0.3787	0.5790
	3.93	0.3782	0.5792
	3.94	0.3777	0.5793
	3.95	0.3773	0.5795
	3.96	0.3768	0.5796
	3.97	0.3764	0.5798
	3.98	0.3759	0.5800
	3.99	0.3754	0.5801
	4.00	0.3750	0.5803
	4.01	0.3745	0.5804
	4.02	0.3741	0.5806
	4.03	0.3736	0.5807
	4.04	0.3731	0.5809
	4.05	0.3727	0.5811
	4.06	0.3722	0.5812
	4.07	0.3718	0.5814
	4.08	0.3713	0.5815
	4.09	0.3709	0.5817
}\tableCorrSiL

\pgfplotstableread{
	sigma	corr	si
	0.01	1.0000	0.0833
	0.02	1.0000	0.0833
	0.03	1.0000	0.0833
	0.04	1.0000	0.0833
	0.05	1.0000	0.0833
	0.06	1.0000	0.0833
	0.07	1.0000	0.0833
	0.08	1.0000	0.0833
	0.09	1.0000	0.0833
	0.10	1.0000	0.0833
	0.11	1.0000	0.0833
	0.12	1.0000	0.0833
	0.13	1.0000	0.0833
	0.14	1.0000	0.0833
	0.15	1.0000	0.0833
	0.16	1.0000	0.0833
	0.17	1.0000	0.0833
	0.18	1.0000	0.0833
	0.19	1.0000	0.0833
	0.20	1.0000	0.0833
	0.21	1.0000	0.0833
	0.22	1.0000	0.0833
	0.23	1.0000	0.0833
	0.24	1.0000	0.0833
	0.25	1.0000	0.0832
	0.26	1.0000	0.0831
	0.27	1.0000	0.0830
	0.28	1.0000	0.0827
	0.29	1.0000	0.0824
	0.30	1.0000	0.0819
	0.31	1.0000	0.0812
	0.32	0.9999	0.0803
	0.33	0.9999	0.0792
	0.34	0.9998	0.0776
	0.35	0.9997	0.0756
	0.36	0.9995	0.0730
	0.37	0.9993	0.0699
	0.38	0.9990	0.0660
	0.39	0.9986	0.0612
	0.40	0.9980	0.0556
	0.41	0.9973	0.0490
	0.42	0.9965	0.0414
	0.43	0.9955	0.0329
	0.44	0.9942	0.0236
	0.45	0.9928	0.0134
	0.46	0.9911	0.0026
	0.47	0.9891	0.0088
	0.48	0.9870	0.0208
	0.49	0.9845	0.0333
	0.50	0.9818	0.0460
	0.51	0.9788	0.0590
	0.52	0.9756	0.0722
	0.53	0.9722	0.0854
	0.54	0.9685	0.0985
	0.55	0.9645	0.1115
	0.56	0.9604	0.1244
	0.57	0.9560	0.1369
	0.58	0.9515	0.1493
	0.59	0.9468	0.1613
	0.60	0.9419	0.1729
	0.61	0.9370	0.1842
	0.62	0.9319	0.1951
	0.63	0.9267	0.2057
	0.64	0.9214	0.2159
	0.65	0.9160	0.2256
	0.66	0.9106	0.2350
	0.67	0.9052	0.2441
	0.68	0.8997	0.2527
	0.69	0.8943	0.2611
	0.70	0.8888	0.2690
	0.71	0.8833	0.2767
	0.72	0.8779	0.2841
	0.73	0.8725	0.2911
	0.74	0.8671	0.2979
	0.75	0.8618	0.3044
	0.76	0.8565	0.3107
	0.77	0.8513	0.3167
	0.78	0.8461	0.3225
	0.79	0.8410	0.3281
	0.80	0.8359	0.3335
	0.81	0.8309	0.3386
	0.82	0.8260	0.3436
	0.83	0.8212	0.3484
	0.84	0.8164	0.3531
	0.85	0.8117	0.3575
	0.86	0.8070	0.3619
	0.87	0.8025	0.3661
	0.88	0.7980	0.3701
	0.89	0.7935	0.3740
	0.90	0.7892	0.3778
	0.91	0.7849	0.3815
	0.92	0.7807	0.3851
	0.93	0.7765	0.3885
	0.94	0.7724	0.3919
	0.95	0.7684	0.3952
	0.96	0.7644	0.3983
	0.97	0.7605	0.4014
	0.98	0.7567	0.4044
	0.99	0.7529	0.4073
	1.00	0.7491	0.4101
	1.01	0.7455	0.4129
	1.02	0.7419	0.4156
	1.03	0.7383	0.4182
	1.04	0.7348	0.4208
	1.05	0.7313	0.4233
	1.06	0.7279	0.4257
	1.07	0.7246	0.4281
	1.08	0.7213	0.4304
	1.09	0.7180	0.4326
	1.10	0.7148	0.4349
	1.11	0.7116	0.4370
	1.12	0.7085	0.4391
	1.13	0.7054	0.4412
	1.14	0.7023	0.4432
	1.15	0.6993	0.4452
	1.16	0.6964	0.4471
	1.17	0.6935	0.4490
	1.18	0.6906	0.4509
	1.19	0.6877	0.4527
	1.20	0.6849	0.4545
	1.21	0.6821	0.4563
	1.22	0.6794	0.4580
	1.23	0.6767	0.4597
	1.24	0.6740	0.4613
	1.25	0.6714	0.4629
	1.26	0.6688	0.4645
	1.27	0.6662	0.4661
	1.28	0.6636	0.4676
	1.29	0.6611	0.4691
	1.30	0.6586	0.4706
	1.31	0.6562	0.4720
	1.32	0.6537	0.4734
	1.33	0.6513	0.4748
	1.34	0.6489	0.4762
	1.35	0.6466	0.4776
	1.36	0.6443	0.4789
	1.37	0.6420	0.4802
	1.38	0.6397	0.4815
	1.39	0.6374	0.4828
	1.40	0.6352	0.4840
	1.41	0.6330	0.4852
	1.42	0.6308	0.4864
	1.43	0.6287	0.4876
	1.44	0.6266	0.4888
	1.45	0.6245	0.4900
	1.46	0.6224	0.4911
	1.47	0.6203	0.4922
	1.48	0.6183	0.4933
	1.49	0.6162	0.4944
	1.50	0.6142	0.4955
	1.51	0.6122	0.4965
	1.52	0.6103	0.4975
	1.53	0.6083	0.4986
	1.54	0.6064	0.4996
	1.55	0.6045	0.5006
	1.56	0.6026	0.5016
	1.57	0.6007	0.5025
	1.58	0.5989	0.5035
	1.59	0.5971	0.5044
	1.60	0.5952	0.5053
	1.61	0.5934	0.5063
	1.62	0.5917	0.5072
	1.63	0.5899	0.5081
	1.64	0.5881	0.5089
	1.65	0.5864	0.5098
	1.66	0.5847	0.5107
	1.67	0.5830	0.5115
	1.68	0.5813	0.5123
	1.69	0.5796	0.5132
	1.70	0.5780	0.5140
	1.71	0.5763	0.5148
	1.72	0.5747	0.5156
	1.73	0.5731	0.5164
	1.74	0.5715	0.5171
	1.75	0.5699	0.5179
	1.76	0.5683	0.5187
	1.77	0.5667	0.5194
	1.78	0.5652	0.5201
	1.79	0.5637	0.5209
	1.80	0.5621	0.5216
	1.81	0.5606	0.5223
	1.82	0.5591	0.5230
	1.83	0.5576	0.5237
	1.84	0.5562	0.5244
	1.85	0.5547	0.5251
	1.86	0.5533	0.5258
	1.87	0.5518	0.5264
	1.88	0.5504	0.5271
	1.89	0.5490	0.5277
	1.90	0.5476	0.5284
	1.91	0.5462	0.5290
	1.92	0.5448	0.5297
	1.93	0.5434	0.5303
	1.94	0.5421	0.5309
	1.95	0.5407	0.5315
	1.96	0.5394	0.5321
	1.97	0.5381	0.5327
	1.98	0.5368	0.5333
	1.99	0.5355	0.5339
	2.00	0.5342	0.5345
	2.01	0.5329	0.5351
	2.02	0.5316	0.5356
	2.03	0.5303	0.5362
	2.04	0.5291	0.5367
	2.05	0.5278	0.5373
	2.06	0.5266	0.5378
	2.07	0.5253	0.5384
	2.08	0.5241	0.5389
	2.09	0.5229	0.5395
	2.10	0.5217	0.5400
	2.11	0.5205	0.5405
	2.12	0.5193	0.5410
	2.13	0.5181	0.5415
	2.14	0.5170	0.5420
	2.15	0.5158	0.5425
	2.16	0.5146	0.5430
	2.17	0.5135	0.5435
	2.18	0.5124	0.5440
	2.19	0.5112	0.5445
	2.20	0.5101	0.5450
	2.21	0.5090	0.5455
	2.22	0.5079	0.5459
	2.23	0.5068	0.5464
	2.24	0.5057	0.5469
	2.25	0.5046	0.5473
	2.26	0.5035	0.5478
	2.27	0.5025	0.5482
	2.28	0.5014	0.5487
	2.29	0.5004	0.5491
	2.30	0.4993	0.5496
	2.31	0.4983	0.5500
	2.32	0.4972	0.5504
	2.33	0.4962	0.5508
	2.34	0.4952	0.5513
	2.35	0.4942	0.5517
	2.36	0.4932	0.5521
	2.37	0.4922	0.5525
	2.38	0.4912	0.5529
	2.39	0.4902	0.5533
	2.40	0.4892	0.5538
	2.41	0.4882	0.5542
	2.42	0.4873	0.5545
	2.43	0.4863	0.5549
	2.44	0.4853	0.5553
	2.45	0.4844	0.5557
	2.46	0.4834	0.5561
	2.47	0.4825	0.5565
	2.48	0.4816	0.5569
	2.49	0.4806	0.5572
	2.50	0.4797	0.5576
	2.51	0.4788	0.5580
	2.52	0.4779	0.5583
	2.53	0.4770	0.5587
	2.54	0.4761	0.5591
	2.55	0.4752	0.5594
	2.56	0.4743	0.5598
	2.57	0.4734	0.5601
	2.58	0.4726	0.5605
	2.59	0.4717	0.5608
	2.60	0.4708	0.5612
	2.61	0.4700	0.5615
	2.62	0.4691	0.5619
	2.63	0.4682	0.5622
	2.64	0.4674	0.5625
	2.65	0.4666	0.5629
	2.66	0.4657	0.5632
	2.67	0.4649	0.5635
	2.68	0.4641	0.5639
	2.69	0.4632	0.5642
	2.70	0.4624	0.5645
	2.71	0.4616	0.5648
	2.72	0.4608	0.5651
	2.73	0.4600	0.5655
	2.74	0.4592	0.5658
	2.75	0.4584	0.5661
	2.76	0.4576	0.5664
	2.77	0.4568	0.5667
	2.78	0.4561	0.5670
	2.79	0.4553	0.5673
	2.80	0.4545	0.5676
	2.81	0.4537	0.5679
	2.82	0.4530	0.5682
	2.83	0.4522	0.5685
	2.84	0.4515	0.5688
	2.85	0.4507	0.5691
	2.86	0.4500	0.5693
	2.87	0.4492	0.5696
	2.88	0.4485	0.5699
	2.89	0.4478	0.5702
	2.90	0.4470	0.5705
	2.91	0.4463	0.5708
	2.92	0.4456	0.5710
	2.93	0.4449	0.5713
	2.94	0.4441	0.5716
	2.95	0.4434	0.5719
	2.96	0.4427	0.5721
	2.97	0.4420	0.5724
	2.98	0.4413	0.5727
	2.99	0.4406	0.5729
	3.00	0.4399	0.5732
	3.01	0.4392	0.5734
	3.02	0.4385	0.5737
	3.03	0.4379	0.5740
	3.04	0.4372	0.5742
	3.05	0.4365	0.5745
	3.06	0.4358	0.5747
	3.07	0.4352	0.5750
	3.08	0.4345	0.5752
	3.09	0.4338	0.5755
	3.10	0.4332	0.5757
	3.11	0.4325	0.5760
	3.12	0.4319	0.5762
	3.13	0.4312	0.5765
	3.14	0.4306	0.5767
	3.15	0.4299	0.5769
	3.16	0.4293	0.5772
	3.17	0.4287	0.5774
	3.18	0.4280	0.5776
	3.19	0.4274	0.5779
	3.20	0.4268	0.5781
	3.21	0.4261	0.5783
	3.22	0.4255	0.5786
	3.23	0.4249	0.5788
	3.24	0.4243	0.5790
	3.25	0.4237	0.5792
	3.26	0.4231	0.5795
	3.27	0.4224	0.5797
	3.28	0.4218	0.5799
	3.29	0.4212	0.5801
	3.30	0.4206	0.5804
	3.31	0.4200	0.5806
	3.32	0.4194	0.5808
	3.33	0.4189	0.5810
	3.34	0.4183	0.5812
	3.35	0.4177	0.5814
	3.36	0.4171	0.5816
	3.37	0.4165	0.5819
	3.38	0.4159	0.5821
	3.39	0.4154	0.5823
	3.40	0.4148	0.5825
	3.41	0.4142	0.5827
	3.42	0.4136	0.5829
	3.43	0.4131	0.5831
	3.44	0.4125	0.5833
	3.45	0.4119	0.5835
	3.46	0.4114	0.5837
	3.47	0.4108	0.5839
	3.48	0.4103	0.5841
	3.49	0.4097	0.5843
	3.50	0.4092	0.5845
	3.51	0.4086	0.5847
	3.52	0.4081	0.5849
	3.53	0.4075	0.5851
	3.54	0.4070	0.5853
	3.55	0.4064	0.5855
	3.56	0.4059	0.5857
	3.57	0.4054	0.5859
	3.58	0.4048	0.5860
	3.59	0.4043	0.5862
	3.60	0.4038	0.5864
	3.61	0.4032	0.5866
	3.62	0.4027	0.5868
	3.63	0.4022	0.5870
	3.64	0.4017	0.5872
	3.65	0.4012	0.5873
	3.66	0.4006	0.5875
	3.67	0.4001	0.5877
	3.68	0.3996	0.5879
	3.69	0.3991	0.5881
	3.70	0.3986	0.5882
	3.71	0.3981	0.5884
	3.72	0.3976	0.5886
	3.73	0.3971	0.5888
	3.74	0.3966	0.5889
	3.75	0.3961	0.5891
	3.76	0.3956	0.5893
	3.77	0.3951	0.5895
	3.78	0.3946	0.5896
	3.79	0.3941	0.5898
	3.80	0.3936	0.5900
	3.81	0.3931	0.5901
	3.82	0.3926	0.5903
	3.83	0.3921	0.5905
	3.84	0.3916	0.5906
	3.85	0.3912	0.5908
	3.86	0.3907	0.5910
	3.87	0.3902	0.5911
	3.88	0.3897	0.5913
	3.89	0.3892	0.5915
	3.90	0.3888	0.5916
	3.91	0.3883	0.5918
	3.92	0.3878	0.5920
	3.93	0.3874	0.5921
	3.94	0.3869	0.5923
	3.95	0.3864	0.5924
	3.96	0.3860	0.5926
	3.97	0.3855	0.5927
	3.98	0.3850	0.5929
	3.99	0.3846	0.5931
	4.00	0.3841	0.5932
	4.01	0.3837	0.5934
	4.02	0.3832	0.5935
	4.03	0.3827	0.5937
	4.04	0.3823	0.5938
	4.05	0.3818	0.5940
	4.06	0.3814	0.5941
	4.07	0.3809	0.5943
	4.08	0.3805	0.5944
	4.09	0.3800	0.5946
}\tableCorrSiM

\pgfplotstableread{
	sigma	corr	si
	0.01	1.0000	0.0833
	0.02	1.0000	0.0833
	0.03	1.0000	0.0833
	0.04	1.0000	0.0833
	0.05	1.0000	0.0833
	0.06	1.0000	0.0833
	0.07	1.0000	0.0833
	0.08	1.0000	0.0833
	0.09	1.0000	0.0833
	0.10	1.0000	0.0833
	0.11	1.0000	0.0833
	0.12	1.0000	0.0833
	0.13	1.0000	0.0833
	0.14	1.0000	0.0833
	0.15	1.0000	0.0833
	0.16	1.0000	0.0833
	0.17	1.0000	0.0833
	0.18	1.0000	0.0833
	0.19	1.0000	0.0833
	0.20	1.0000	0.0833
	0.21	1.0000	0.0833
	0.22	1.0000	0.0833
	0.23	1.0000	0.0833
	0.24	1.0000	0.0833
	0.25	1.0000	0.0832
	0.26	1.0000	0.0831
	0.27	1.0000	0.0829
	0.28	1.0000	0.0827
	0.29	1.0000	0.0823
	0.30	1.0000	0.0818
	0.31	1.0000	0.0810
	0.32	0.9999	0.0800
	0.33	0.9999	0.0787
	0.34	0.9998	0.0769
	0.35	0.9997	0.0746
	0.36	0.9995	0.0717
	0.37	0.9993	0.0680
	0.38	0.9990	0.0635
	0.39	0.9986	0.0580
	0.40	0.9980	0.0515
	0.41	0.9973	0.0438
	0.42	0.9965	0.0350
	0.43	0.9954	0.0252
	0.44	0.9942	0.0144
	0.45	0.9927	0.0027
	0.46	0.9910	0.0099
	0.47	0.9890	0.0231
	0.48	0.9868	0.0369
	0.49	0.9843	0.0513
	0.50	0.9816	0.0659
	0.51	0.9785	0.0807
	0.52	0.9753	0.0957
	0.53	0.9717	0.1105
	0.54	0.9679	0.1253
	0.55	0.9638	0.1397
	0.56	0.9596	0.1539
	0.57	0.9551	0.1677
	0.58	0.9504	0.1811
	0.59	0.9455	0.1941
	0.60	0.9405	0.2066
	0.61	0.9353	0.2187
	0.62	0.9300	0.2303
	0.63	0.9246	0.2415
	0.64	0.9190	0.2522
	0.65	0.9135	0.2625
	0.66	0.9078	0.2723
	0.67	0.9021	0.2817
	0.68	0.8964	0.2907
	0.69	0.8907	0.2993
	0.70	0.8849	0.3075
	0.71	0.8792	0.3154
	0.72	0.8735	0.3229
	0.73	0.8678	0.3301
	0.74	0.8621	0.3370
	0.75	0.8565	0.3437
	0.76	0.8509	0.3500
	0.77	0.8454	0.3561
	0.78	0.8400	0.3619
	0.79	0.8346	0.3675
	0.80	0.8293	0.3729
	0.81	0.8240	0.3781
	0.82	0.8188	0.3831
	0.83	0.8137	0.3879
	0.84	0.8087	0.3925
	0.85	0.8037	0.3969
	0.86	0.7988	0.4012
	0.87	0.7940	0.4054
	0.88	0.7893	0.4094
	0.89	0.7846	0.4132
	0.90	0.7800	0.4169
	0.91	0.7755	0.4206
	0.92	0.7711	0.4240
	0.93	0.7667	0.4274
	0.94	0.7624	0.4307
	0.95	0.7582	0.4339
	0.96	0.7541	0.4369
	0.97	0.7500	0.4399
	0.98	0.7459	0.4428
	0.99	0.7420	0.4456
	1.00	0.7381	0.4484
	1.01	0.7343	0.4510
	1.02	0.7305	0.4536
	1.03	0.7268	0.4561
	1.04	0.7231	0.4586
	1.05	0.7195	0.4610
	1.06	0.7160	0.4633
	1.07	0.7125	0.4655
	1.08	0.7091	0.4677
	1.09	0.7057	0.4699
	1.10	0.7023	0.4720
	1.11	0.6991	0.4741
	1.12	0.6958	0.4761
	1.13	0.6926	0.4780
	1.14	0.6895	0.4799
	1.15	0.6864	0.4818
	1.16	0.6833	0.4836
	1.17	0.6803	0.4854
	1.18	0.6774	0.4872
	1.19	0.6744	0.4889
	1.20	0.6715	0.4905
	1.21	0.6687	0.4922
	1.22	0.6659	0.4938
	1.23	0.6631	0.4954
	1.24	0.6604	0.4969
	1.25	0.6577	0.4984
	1.26	0.6550	0.4999
	1.27	0.6524	0.5013
	1.28	0.6498	0.5028
	1.29	0.6472	0.5042
	1.30	0.6447	0.5055
	1.31	0.6422	0.5069
	1.32	0.6397	0.5082
	1.33	0.6373	0.5095
	1.34	0.6349	0.5108
	1.35	0.6325	0.5120
	1.36	0.6301	0.5132
	1.37	0.6278	0.5145
	1.38	0.6255	0.5157
	1.39	0.6232	0.5168
	1.40	0.6210	0.5180
	1.41	0.6187	0.5191
	1.42	0.6165	0.5202
	1.43	0.6144	0.5213
	1.44	0.6122	0.5224
	1.45	0.6101	0.5234
	1.46	0.6080	0.5245
	1.47	0.6059	0.5255
	1.48	0.6038	0.5265
	1.49	0.6018	0.5275
	1.50	0.5998	0.5285
	1.51	0.5978	0.5294
	1.52	0.5958	0.5304
	1.53	0.5939	0.5313
	1.54	0.5919	0.5323
	1.55	0.5900	0.5332
	1.56	0.5881	0.5341
	1.57	0.5862	0.5349
	1.58	0.5844	0.5358
	1.59	0.5825	0.5367
	1.60	0.5807	0.5375
	1.61	0.5789	0.5384
	1.62	0.5771	0.5392
	1.63	0.5754	0.5400
	1.64	0.5736	0.5408
	1.65	0.5719	0.5416
	1.66	0.5701	0.5424
	1.67	0.5684	0.5431
	1.68	0.5668	0.5439
	1.69	0.5651	0.5446
	1.70	0.5634	0.5454
	1.71	0.5618	0.5461
	1.72	0.5601	0.5468
	1.73	0.5585	0.5476
	1.74	0.5569	0.5483
	1.75	0.5553	0.5490
	1.76	0.5538	0.5496
	1.77	0.5522	0.5503
	1.78	0.5507	0.5510
	1.79	0.5491	0.5517
	1.80	0.5476	0.5523
	1.81	0.5461	0.5530
	1.82	0.5446	0.5536
	1.83	0.5431	0.5542
	1.84	0.5417	0.5548
	1.85	0.5402	0.5555
	1.86	0.5388	0.5561
	1.87	0.5373	0.5567
	1.88	0.5359	0.5573
	1.89	0.5345	0.5579
	1.90	0.5331	0.5585
	1.91	0.5317	0.5590
	1.92	0.5303	0.5596
	1.93	0.5290	0.5602
	1.94	0.5276	0.5607
	1.95	0.5263	0.5613
	1.96	0.5250	0.5618
	1.97	0.5236	0.5624
	1.98	0.5223	0.5629
	1.99	0.5210	0.5634
	2.00	0.5197	0.5639
	2.01	0.5185	0.5645
	2.02	0.5172	0.5650
	2.03	0.5159	0.5655
	2.04	0.5147	0.5660
	2.05	0.5134	0.5665
	2.06	0.5122	0.5670
	2.07	0.5110	0.5675
	2.08	0.5098	0.5679
	2.09	0.5086	0.5684
	2.10	0.5074	0.5689
	2.11	0.5062	0.5694
	2.12	0.5050	0.5698
	2.13	0.5038	0.5703
	2.14	0.5027	0.5707
	2.15	0.5015	0.5712
	2.16	0.5004	0.5716
	2.17	0.4993	0.5721
	2.18	0.4981	0.5725
	2.19	0.4970	0.5729
	2.20	0.4959	0.5734
	2.21	0.4948	0.5738
	2.22	0.4937	0.5742
	2.23	0.4926	0.5746
	2.24	0.4915	0.5751
	2.25	0.4905	0.5755
	2.26	0.4894	0.5759
	2.27	0.4884	0.5763
	2.28	0.4873	0.5767
	2.29	0.4863	0.5771
	2.30	0.4852	0.5775
	2.31	0.4842	0.5778
	2.32	0.4832	0.5782
	2.33	0.4822	0.5786
	2.34	0.4812	0.5790
	2.35	0.4802	0.5794
	2.36	0.4792	0.5797
	2.37	0.4782	0.5801
	2.38	0.4772	0.5805
	2.39	0.4763	0.5808
	2.40	0.4753	0.5812
	2.41	0.4743	0.5816
	2.42	0.4734	0.5819
	2.43	0.4724	0.5823
	2.44	0.4715	0.5826
	2.45	0.4706	0.5830
	2.46	0.4696	0.5833
	2.47	0.4687	0.5836
	2.48	0.4678	0.5840
	2.49	0.4669	0.5843
	2.50	0.4660	0.5846
	2.51	0.4651	0.5850
	2.52	0.4642	0.5853
	2.53	0.4633	0.5856
	2.54	0.4624	0.5859
	2.55	0.4616	0.5862
	2.56	0.4607	0.5866
	2.57	0.4598	0.5869
	2.58	0.4590	0.5872
	2.59	0.4581	0.5875
	2.60	0.4573	0.5878
	2.61	0.4564	0.5881
	2.62	0.4556	0.5884
	2.63	0.4548	0.5887
	2.64	0.4539	0.5890
	2.65	0.4531	0.5893
	2.66	0.4523	0.5896
	2.67	0.4515	0.5899
	2.68	0.4507	0.5902
	2.69	0.4499	0.5905
	2.70	0.4491	0.5907
	2.71	0.4483	0.5910
	2.72	0.4475	0.5913
	2.73	0.4467	0.5916
	2.74	0.4459	0.5919
	2.75	0.4451	0.5921
	2.76	0.4444	0.5924
	2.77	0.4436	0.5927
	2.78	0.4428	0.5930
	2.79	0.4421	0.5932
	2.80	0.4413	0.5935
	2.81	0.4406	0.5937
	2.82	0.4398	0.5940
	2.83	0.4391	0.5943
	2.84	0.4384	0.5945
	2.85	0.4376	0.5948
	2.86	0.4369	0.5950
	2.87	0.4362	0.5953
	2.88	0.4355	0.5955
	2.89	0.4347	0.5958
	2.90	0.4340	0.5960
	2.91	0.4333	0.5963
	2.92	0.4326	0.5965
	2.93	0.4319	0.5968
	2.94	0.4312	0.5970
	2.95	0.4305	0.5973
	2.96	0.4298	0.5975
	2.97	0.4291	0.5977
	2.98	0.4285	0.5980
	2.99	0.4278	0.5982
	3.00	0.4271	0.5984
	3.01	0.4264	0.5987
	3.02	0.4258	0.5989
	3.03	0.4251	0.5991
	3.04	0.4244	0.5993
	3.05	0.4238	0.5996
	3.06	0.4231	0.5998
	3.07	0.4225	0.6000
	3.08	0.4218	0.6002
	3.09	0.4212	0.6005
	3.10	0.4205	0.6007
	3.11	0.4199	0.6009
	3.12	0.4193	0.6011
	3.13	0.4186	0.6013
	3.14	0.4180	0.6015
	3.15	0.4174	0.6017
	3.16	0.4167	0.6020
	3.17	0.4161	0.6022
	3.18	0.4155	0.6024
	3.19	0.4149	0.6026
	3.20	0.4143	0.6028
	3.21	0.4137	0.6030
	3.22	0.4131	0.6032
	3.23	0.4124	0.6034
	3.24	0.4118	0.6036
	3.25	0.4112	0.6038
	3.26	0.4107	0.6040
	3.27	0.4101	0.6042
	3.28	0.4095	0.6044
	3.29	0.4089	0.6046
	3.30	0.4083	0.6048
	3.31	0.4077	0.6050
	3.32	0.4071	0.6052
	3.33	0.4066	0.6054
	3.34	0.4060	0.6055
	3.35	0.4054	0.6057
	3.36	0.4048	0.6059
	3.37	0.4043	0.6061
	3.38	0.4037	0.6063
	3.39	0.4031	0.6065
	3.40	0.4026	0.6067
	3.41	0.4020	0.6069
	3.42	0.4015	0.6070
	3.43	0.4009	0.6072
	3.44	0.4003	0.6074
	3.45	0.3998	0.6076
	3.46	0.3993	0.6078
	3.47	0.3987	0.6079
	3.48	0.3982	0.6081
	3.49	0.3976	0.6083
	3.50	0.3971	0.6085
	3.51	0.3965	0.6086
	3.52	0.3960	0.6088
	3.53	0.3955	0.6090
	3.54	0.3950	0.6091
	3.55	0.3944	0.6093
	3.56	0.3939	0.6095
	3.57	0.3934	0.6097
	3.58	0.3929	0.6098
	3.59	0.3923	0.6100
	3.60	0.3918	0.6102
	3.61	0.3913	0.6103
	3.62	0.3908	0.6105
	3.63	0.3903	0.6107
	3.64	0.3898	0.6108
	3.65	0.3893	0.6110
	3.66	0.3888	0.6111
	3.67	0.3882	0.6113
	3.68	0.3877	0.6115
	3.69	0.3872	0.6116
	3.70	0.3867	0.6118
	3.71	0.3862	0.6119
	3.72	0.3858	0.6121
	3.73	0.3853	0.6122
	3.74	0.3848	0.6124
	3.75	0.3843	0.6125
	3.76	0.3838	0.6127
	3.77	0.3833	0.6129
	3.78	0.3828	0.6130
	3.79	0.3823	0.6132
	3.80	0.3819	0.6133
	3.81	0.3814	0.6135
	3.82	0.3809	0.6136
	3.83	0.3804	0.6138
	3.84	0.3800	0.6139
	3.85	0.3795	0.6141
	3.86	0.3790	0.6142
	3.87	0.3785	0.6143
	3.88	0.3781	0.6145
	3.89	0.3776	0.6146
	3.90	0.3771	0.6148
	3.91	0.3767	0.6149
	3.92	0.3762	0.6151
	3.93	0.3758	0.6152
	3.94	0.3753	0.6153
	3.95	0.3748	0.6155
	3.96	0.3744	0.6156
	3.97	0.3739	0.6158
	3.98	0.3735	0.6159
	3.99	0.3730	0.6160
	4.00	0.3726	0.6162
	4.01	0.3721	0.6163
	4.02	0.3717	0.6165
	4.03	0.3713	0.6166
	4.04	0.3708	0.6167
	4.05	0.3704	0.6169
	4.06	0.3699	0.6170
	4.07	0.3695	0.6171
	4.08	0.3691	0.6173
	4.09	0.3686	0.6174
}\tableCorrSiN

\pgfplotstableread{
	sigma	corr	si
	0.01	1.0000	0.0833
	0.02	1.0000	0.0833
	0.03	1.0000	0.0833
	0.04	1.0000	0.0833
	0.05	1.0000	0.0833
	0.06	1.0000	0.0833
	0.07	1.0000	0.0833
	0.08	1.0000	0.0833
	0.09	1.0000	0.0833
	0.10	1.0000	0.0833
	0.11	1.0000	0.0833
	0.12	1.0000	0.0833
	0.13	1.0000	0.0833
	0.14	1.0000	0.0833
	0.15	1.0000	0.0833
	0.16	1.0000	0.0833
	0.17	1.0000	0.0833
	0.18	1.0000	0.0833
	0.19	1.0000	0.0833
	0.20	1.0000	0.0833
	0.21	1.0000	0.0833
	0.22	1.0000	0.0833
	0.23	1.0000	0.0833
	0.24	1.0000	0.0833
	0.25	1.0000	0.0833
	0.26	1.0000	0.0832
	0.27	1.0000	0.0831
	0.28	1.0000	0.0829
	0.29	1.0000	0.0827
	0.30	1.0000	0.0824
	0.31	1.0000	0.0819
	0.32	0.9999	0.0814
	0.33	0.9999	0.0806
	0.34	0.9998	0.0797
	0.35	0.9997	0.0785
	0.36	0.9996	0.0770
	0.37	0.9993	0.0751
	0.38	0.9990	0.0729
	0.39	0.9986	0.0703
	0.40	0.9981	0.0672
	0.41	0.9974	0.0636
	0.42	0.9966	0.0595
	0.43	0.9956	0.0548
	0.44	0.9943	0.0497
	0.45	0.9929	0.0442
	0.46	0.9913	0.0383
	0.47	0.9894	0.0321
	0.48	0.9872	0.0256
	0.49	0.9848	0.0189
	0.50	0.9822	0.0119
	0.51	0.9793	0.0048
	0.52	0.9761	0.0024
	0.53	0.9727	0.0097
	0.54	0.9691	0.0170
	0.55	0.9652	0.0243
	0.56	0.9611	0.0316
	0.57	0.9568	0.0389
	0.58	0.9524	0.0460
	0.59	0.9477	0.0531
	0.60	0.9430	0.0601
	0.61	0.9380	0.0669
	0.62	0.9330	0.0736
	0.63	0.9279	0.0801
	0.64	0.9227	0.0865
	0.65	0.9174	0.0928
	0.66	0.9120	0.0989
	0.67	0.9066	0.1048
	0.68	0.9012	0.1106
	0.69	0.8958	0.1162
	0.70	0.8904	0.1217
	0.71	0.8849	0.1270
	0.72	0.8795	0.1321
	0.73	0.8741	0.1372
	0.74	0.8688	0.1421
	0.75	0.8634	0.1468
	0.76	0.8581	0.1514
	0.77	0.8529	0.1559
	0.78	0.8477	0.1603
	0.79	0.8426	0.1646
	0.80	0.8375	0.1687
	0.81	0.8325	0.1728
	0.82	0.8275	0.1767
	0.83	0.8226	0.1805
	0.84	0.8178	0.1842
	0.85	0.8131	0.1879
	0.86	0.8084	0.1914
	0.87	0.8037	0.1948
	0.88	0.7992	0.1982
	0.89	0.7947	0.2015
	0.90	0.7903	0.2047
	0.91	0.7859	0.2078
	0.92	0.7816	0.2109
	0.93	0.7774	0.2138
	0.94	0.7732	0.2167
	0.95	0.7691	0.2196
	0.96	0.7650	0.2224
	0.97	0.7611	0.2251
	0.98	0.7571	0.2277
	0.99	0.7533	0.2303
	1.00	0.7495	0.2329
	1.01	0.7457	0.2354
	1.02	0.7420	0.2378
	1.03	0.7383	0.2402
	1.04	0.7347	0.2425
	1.05	0.7312	0.2448
	1.06	0.7277	0.2470
	1.07	0.7243	0.2492
	1.08	0.7209	0.2514
	1.09	0.7175	0.2535
	1.10	0.7142	0.2556
	1.11	0.7110	0.2576
	1.12	0.7077	0.2596
	1.13	0.7046	0.2616
	1.14	0.7014	0.2635
	1.15	0.6984	0.2654
	1.16	0.6953	0.2672
	1.17	0.6923	0.2691
	1.18	0.6893	0.2708
	1.19	0.6864	0.2726
	1.20	0.6835	0.2743
	1.21	0.6807	0.2760
	1.22	0.6779	0.2777
	1.23	0.6751	0.2793
	1.24	0.6723	0.2810
	1.25	0.6696	0.2825
	1.26	0.6669	0.2841
	1.27	0.6643	0.2856
	1.28	0.6617	0.2872
	1.29	0.6591	0.2887
	1.30	0.6565	0.2901
	1.31	0.6540	0.2916
	1.32	0.6515	0.2930
	1.33	0.6491	0.2944
	1.34	0.6466	0.2958
	1.35	0.6442	0.2971
	1.36	0.6418	0.2985
	1.37	0.6395	0.2998
	1.38	0.6371	0.3011
	1.39	0.6348	0.3024
	1.40	0.6326	0.3037
	1.41	0.6303	0.3049
	1.42	0.6281	0.3061
	1.43	0.6259	0.3074
	1.44	0.6237	0.3086
	1.45	0.6215	0.3097
	1.46	0.6194	0.3109
	1.47	0.6173	0.3121
	1.48	0.6152	0.3132
	1.49	0.6131	0.3143
	1.50	0.6111	0.3154
	1.51	0.6090	0.3165
	1.52	0.6070	0.3176
	1.53	0.6050	0.3187
	1.54	0.6031	0.3197
	1.55	0.6011	0.3207
	1.56	0.5992	0.3218
	1.57	0.5973	0.3228
	1.58	0.5954	0.3238
	1.59	0.5935	0.3248
	1.60	0.5917	0.3257
	1.61	0.5898	0.3267
	1.62	0.5880	0.3277
	1.63	0.5862	0.3286
	1.64	0.5844	0.3296
	1.65	0.5827	0.3305
	1.66	0.5809	0.3314
	1.67	0.5792	0.3323
	1.68	0.5775	0.3332
	1.69	0.5757	0.3341
	1.70	0.5741	0.3349
	1.71	0.5724	0.3358
	1.72	0.5707	0.3366
	1.73	0.5691	0.3375
	1.74	0.5675	0.3383
	1.75	0.5658	0.3391
	1.76	0.5642	0.3400
	1.77	0.5627	0.3408
	1.78	0.5611	0.3416
	1.79	0.5595	0.3423
	1.80	0.5580	0.3431
	1.81	0.5564	0.3439
	1.82	0.5549	0.3447
	1.83	0.5534	0.3454
	1.84	0.5519	0.3462
	1.85	0.5504	0.3469
	1.86	0.5490	0.3476
	1.87	0.5475	0.3484
	1.88	0.5461	0.3491
	1.89	0.5446	0.3498
	1.90	0.5432	0.3505
	1.91	0.5418	0.3512
	1.92	0.5404	0.3519
	1.93	0.5390	0.3526
	1.94	0.5376	0.3533
	1.95	0.5363	0.3539
	1.96	0.5349	0.3546
	1.97	0.5336	0.3553
	1.98	0.5322	0.3559
	1.99	0.5309	0.3566
	2.00	0.5296	0.3572
	2.01	0.5283	0.3578
	2.02	0.5270	0.3585
	2.03	0.5257	0.3591
	2.04	0.5245	0.3597
	2.05	0.5232	0.3603
	2.06	0.5219	0.3609
	2.07	0.5207	0.3615
	2.08	0.5195	0.3621
	2.09	0.5182	0.3627
	2.10	0.5170	0.3633
	2.11	0.5158	0.3639
	2.12	0.5146	0.3645
	2.13	0.5134	0.3650
	2.14	0.5123	0.3656
	2.15	0.5111	0.3662
	2.16	0.5099	0.3667
	2.17	0.5088	0.3673
	2.18	0.5076	0.3678
	2.19	0.5065	0.3684
	2.20	0.5054	0.3689
	2.21	0.5042	0.3694
	2.22	0.5031	0.3700
	2.23	0.5020	0.3705
	2.24	0.5009	0.3710
	2.25	0.4998	0.3715
	2.26	0.4988	0.3720
	2.27	0.4977	0.3725
	2.28	0.4966	0.3730
	2.29	0.4956	0.3735
	2.30	0.4945	0.3740
	2.31	0.4935	0.3745
	2.32	0.4924	0.3750
	2.33	0.4914	0.3755
	2.34	0.4904	0.3760
	2.35	0.4894	0.3764
	2.36	0.4883	0.3769
	2.37	0.4873	0.3774
	2.38	0.4863	0.3778
	2.39	0.4854	0.3783
	2.40	0.4844	0.3788
	2.41	0.4834	0.3792
	2.42	0.4824	0.3797
	2.43	0.4815	0.3801
	2.44	0.4805	0.3806
	2.45	0.4796	0.3810
	2.46	0.4786	0.3814
	2.47	0.4777	0.3819
	2.48	0.4768	0.3823
	2.49	0.4758	0.3827
	2.50	0.4749	0.3832
	2.51	0.4740	0.3836
	2.52	0.4731	0.3840
	2.53	0.4722	0.3844
	2.54	0.4713	0.3848
	2.55	0.4704	0.3852
	2.56	0.4695	0.3856
	2.57	0.4686	0.3860
	2.58	0.4678	0.3864
	2.59	0.4669	0.3868
	2.60	0.4660	0.3872
	2.61	0.4652	0.3876
	2.62	0.4643	0.3880
	2.63	0.4635	0.3884
	2.64	0.4626	0.3888
	2.65	0.4618	0.3892
	2.66	0.4610	0.3895
	2.67	0.4602	0.3899
	2.68	0.4593	0.3903
	2.69	0.4585	0.3907
	2.70	0.4577	0.3910
	2.71	0.4569	0.3914
	2.72	0.4561	0.3918
	2.73	0.4553	0.3921
	2.74	0.4545	0.3925
	2.75	0.4537	0.3928
	2.76	0.4530	0.3932
	2.77	0.4522	0.3935
	2.78	0.4514	0.3939
	2.79	0.4506	0.3942
	2.80	0.4499	0.3946
	2.81	0.4491	0.3949
	2.82	0.4484	0.3953
	2.83	0.4476	0.3956
	2.84	0.4469	0.3959
	2.85	0.4461	0.3963
	2.86	0.4454	0.3966
	2.87	0.4447	0.3969
	2.88	0.4439	0.3973
	2.89	0.4432	0.3976
	2.90	0.4425	0.3979
	2.91	0.4418	0.3982
	2.92	0.4411	0.3986
	2.93	0.4403	0.3989
	2.94	0.4396	0.3992
	2.95	0.4389	0.3995
	2.96	0.4382	0.3998
	2.97	0.4376	0.4001
	2.98	0.4369	0.4004
	2.99	0.4362	0.4007
	3.00	0.4355	0.4011
	3.01	0.4348	0.4014
	3.02	0.4341	0.4017
	3.03	0.4335	0.4020
	3.04	0.4328	0.4023
	3.05	0.4321	0.4026
	3.06	0.4315	0.4028
	3.07	0.4308	0.4031
	3.08	0.4302	0.4034
	3.09	0.4295	0.4037
	3.10	0.4289	0.4040
	3.11	0.4282	0.4043
	3.12	0.4276	0.4046
	3.13	0.4270	0.4049
	3.14	0.4263	0.4051
	3.15	0.4257	0.4054
	3.16	0.4251	0.4057
	3.17	0.4245	0.4060
	3.18	0.4238	0.4063
	3.19	0.4232	0.4065
	3.20	0.4226	0.4068
	3.21	0.4220	0.4071
	3.22	0.4214	0.4073
	3.23	0.4208	0.4076
	3.24	0.4202	0.4079
	3.25	0.4196	0.4081
	3.26	0.4190	0.4084
	3.27	0.4184	0.4087
	3.28	0.4178	0.4089
	3.29	0.4172	0.4092
	3.30	0.4166	0.4094
	3.31	0.4161	0.4097
	3.32	0.4155	0.4099
	3.33	0.4149	0.4102
	3.34	0.4143	0.4105
	3.35	0.4138	0.4107
	3.36	0.4132	0.4110
	3.37	0.4126	0.4112
	3.38	0.4121	0.4115
	3.39	0.4115	0.4117
	3.40	0.4110	0.4119
	3.41	0.4104	0.4122
	3.42	0.4099	0.4124
	3.43	0.4093	0.4127
	3.44	0.4088	0.4129
	3.45	0.4082	0.4132
	3.46	0.4077	0.4134
	3.47	0.4071	0.4136
	3.48	0.4066	0.4139
	3.49	0.4061	0.4141
	3.50	0.4055	0.4143
	3.51	0.4050	0.4146
	3.52	0.4045	0.4148
	3.53	0.4039	0.4150
	3.54	0.4034	0.4152
	3.55	0.4029	0.4155
	3.56	0.4024	0.4157
	3.57	0.4019	0.4159
	3.58	0.4014	0.4162
	3.59	0.4008	0.4164
	3.60	0.4003	0.4166
	3.61	0.3998	0.4168
	3.62	0.3993	0.4170
	3.63	0.3988	0.4173
	3.64	0.3983	0.4175
	3.65	0.3978	0.4177
	3.66	0.3973	0.4179
	3.67	0.3968	0.4181
	3.68	0.3963	0.4183
	3.69	0.3958	0.4186
	3.70	0.3954	0.4188
	3.71	0.3949	0.4190
	3.72	0.3944	0.4192
	3.73	0.3939	0.4194
	3.74	0.3934	0.4196
	3.75	0.3929	0.4198
	3.76	0.3925	0.4200
	3.77	0.3920	0.4202
	3.78	0.3915	0.4204
	3.79	0.3910	0.4206
	3.80	0.3906	0.4208
	3.81	0.3901	0.4210
	3.82	0.3896	0.4212
	3.83	0.3892	0.4214
	3.84	0.3887	0.4216
	3.85	0.3882	0.4218
	3.86	0.3878	0.4220
	3.87	0.3873	0.4222
	3.88	0.3869	0.4224
	3.89	0.3864	0.4226
	3.90	0.3860	0.4228
	3.91	0.3855	0.4230
	3.92	0.3851	0.4232
	3.93	0.3846	0.4234
	3.94	0.3842	0.4236
	3.95	0.3837	0.4238
	3.96	0.3833	0.4240
	3.97	0.3828	0.4242
	3.98	0.3824	0.4243
	3.99	0.3820	0.4245
	4.00	0.3815	0.4247
	4.01	0.3811	0.4249
	4.02	0.3806	0.4251
	4.03	0.3802	0.4253
	4.04	0.3798	0.4255
	4.05	0.3794	0.4256
	4.06	0.3789	0.4258
	4.07	0.3785	0.4260
	4.08	0.3781	0.4262
	4.09	0.3776	0.4264
}\tableCorrSiO

\pgfplotstableread{
	sigma	corr	si
	0.01	1.0000	0.0833
	0.02	1.0000	0.0833
	0.03	1.0000	0.0833
	0.04	1.0000	0.0833
	0.05	1.0000	0.0833
	0.06	1.0000	0.0833
	0.07	1.0000	0.0833
	0.08	1.0000	0.0833
	0.09	1.0000	0.0833
	0.10	1.0000	0.0833
	0.11	1.0000	0.0833
	0.12	1.0000	0.0833
	0.13	1.0000	0.0833
	0.14	1.0000	0.0833
	0.15	1.0000	0.0833
	0.16	1.0000	0.0833
	0.17	1.0000	0.0833
	0.18	1.0000	0.0833
	0.19	1.0000	0.0833
	0.20	1.0000	0.0833
	0.21	1.0000	0.0833
	0.22	1.0000	0.0833
	0.23	1.0000	0.0833
	0.24	1.0000	0.0833
	0.25	1.0000	0.0832
	0.26	1.0000	0.0832
	0.27	1.0000	0.0830
	0.28	1.0000	0.0828
	0.29	1.0000	0.0825
	0.30	1.0000	0.0821
	0.31	1.0000	0.0816
	0.32	0.9999	0.0809
	0.33	0.9999	0.0799
	0.34	0.9998	0.0787
	0.35	0.9997	0.0771
	0.36	0.9996	0.0751
	0.37	0.9993	0.0726
	0.38	0.9990	0.0696
	0.39	0.9986	0.0660
	0.40	0.9981	0.0617
	0.41	0.9974	0.0567
	0.42	0.9966	0.0509
	0.43	0.9956	0.0444
	0.44	0.9943	0.0373
	0.45	0.9929	0.0296
	0.46	0.9913	0.0213
	0.47	0.9894	0.0126
	0.48	0.9872	0.0034
	0.49	0.9848	0.0061
	0.50	0.9822	0.0159
	0.51	0.9793	0.0259
	0.52	0.9761	0.0360
	0.53	0.9727	0.0462
	0.54	0.9691	0.0565
	0.55	0.9652	0.0667
	0.56	0.9612	0.0769
	0.57	0.9569	0.0869
	0.58	0.9524	0.0968
	0.59	0.9478	0.1066
	0.60	0.9431	0.1161
	0.61	0.9382	0.1254
	0.62	0.9331	0.1345
	0.63	0.9280	0.1433
	0.64	0.9228	0.1519
	0.65	0.9175	0.1603
	0.66	0.9122	0.1684
	0.67	0.9068	0.1762
	0.68	0.9014	0.1838
	0.69	0.8960	0.1911
	0.70	0.8906	0.1982
	0.71	0.8852	0.2050
	0.72	0.8798	0.2117
	0.73	0.8744	0.2180
	0.74	0.8691	0.2242
	0.75	0.8637	0.2302
	0.76	0.8585	0.2359
	0.77	0.8533	0.2415
	0.78	0.8481	0.2469
	0.79	0.8430	0.2521
	0.80	0.8379	0.2571
	0.81	0.8329	0.2620
	0.82	0.8280	0.2667
	0.83	0.8231	0.2712
	0.84	0.8183	0.2756
	0.85	0.8136	0.2799
	0.86	0.8089	0.2841
	0.87	0.8043	0.2881
	0.88	0.7997	0.2920
	0.89	0.7952	0.2958
	0.90	0.7908	0.2995
	0.91	0.7865	0.3031
	0.92	0.7822	0.3066
	0.93	0.7780	0.3100
	0.94	0.7738	0.3133
	0.95	0.7697	0.3165
	0.96	0.7657	0.3196
	0.97	0.7617	0.3227
	0.98	0.7578	0.3257
	0.99	0.7539	0.3285
	1.00	0.7501	0.3314
	1.01	0.7464	0.3341
	1.02	0.7427	0.3368
	1.03	0.7390	0.3394
	1.04	0.7355	0.3420
	1.05	0.7319	0.3445
	1.06	0.7284	0.3470
	1.07	0.7250	0.3494
	1.08	0.7216	0.3517
	1.09	0.7182	0.3540
	1.10	0.7149	0.3562
	1.11	0.7117	0.3584
	1.12	0.7085	0.3606
	1.13	0.7053	0.3627
	1.14	0.7022	0.3647
	1.15	0.6991	0.3668
	1.16	0.6960	0.3687
	1.17	0.6930	0.3707
	1.18	0.6901	0.3726
	1.19	0.6871	0.3744
	1.20	0.6842	0.3763
	1.21	0.6814	0.3781
	1.22	0.6786	0.3798
	1.23	0.6758	0.3815
	1.24	0.6730	0.3832
	1.25	0.6703	0.3849
	1.26	0.6676	0.3865
	1.27	0.6650	0.3881
	1.28	0.6624	0.3897
	1.29	0.6598	0.3913
	1.30	0.6572	0.3928
	1.31	0.6547	0.3943
	1.32	0.6522	0.3958
	1.33	0.6497	0.3972
	1.34	0.6472	0.3986
	1.35	0.6448	0.4000
	1.36	0.6424	0.4014
	1.37	0.6401	0.4028
	1.38	0.6377	0.4041
	1.39	0.6354	0.4054
	1.40	0.6331	0.4067
	1.41	0.6309	0.4080
	1.42	0.6286	0.4092
	1.43	0.6264	0.4105
	1.44	0.6242	0.4117
	1.45	0.6221	0.4129
	1.46	0.6199	0.4141
	1.47	0.6178	0.4152
	1.48	0.6157	0.4164
	1.49	0.6136	0.4175
	1.50	0.6115	0.4186
	1.51	0.6095	0.4197
	1.52	0.6075	0.4208
	1.53	0.6055	0.4218
	1.54	0.6035	0.4229
	1.55	0.6015	0.4239
	1.56	0.5996	0.4250
	1.57	0.5977	0.4260
	1.58	0.5958	0.4270
	1.59	0.5939	0.4280
	1.60	0.5920	0.4289
	1.61	0.5902	0.4299
	1.62	0.5883	0.4308
	1.63	0.5865	0.4318
	1.64	0.5847	0.4327
	1.65	0.5829	0.4336
	1.66	0.5812	0.4345
	1.67	0.5794	0.4354
	1.68	0.5777	0.4363
	1.69	0.5759	0.4371
	1.70	0.5742	0.4380
	1.71	0.5725	0.4388
	1.72	0.5709	0.4397
	1.73	0.5692	0.4405
	1.74	0.5676	0.4413
	1.75	0.5659	0.4421
	1.76	0.5643	0.4429
	1.77	0.5627	0.4437
	1.78	0.5611	0.4445
	1.79	0.5595	0.4452
	1.80	0.5580	0.4460
	1.81	0.5564	0.4468
	1.82	0.5549	0.4475
	1.83	0.5533	0.4482
	1.84	0.5518	0.4490
	1.85	0.5503	0.4497
	1.86	0.5488	0.4504
	1.87	0.5473	0.4511
	1.88	0.5459	0.4518
	1.89	0.5444	0.4525
	1.90	0.5430	0.4532
	1.91	0.5415	0.4539
	1.92	0.5401	0.4545
	1.93	0.5387	0.4552
	1.94	0.5373	0.4558
	1.95	0.5359	0.4565
	1.96	0.5345	0.4571
	1.97	0.5332	0.4578
	1.98	0.5318	0.4584
	1.99	0.5304	0.4590
	2.00	0.5291	0.4596
	2.01	0.5278	0.4602
	2.02	0.5265	0.4608
	2.03	0.5252	0.4614
	2.04	0.5239	0.4620
	2.05	0.5226	0.4626
	2.06	0.5213	0.4632
	2.07	0.5200	0.4638
	2.08	0.5188	0.4643
	2.09	0.5175	0.4649
	2.10	0.5163	0.4655
	2.11	0.5150	0.4660
	2.12	0.5138	0.4666
	2.13	0.5126	0.4671
	2.14	0.5114	0.4677
	2.15	0.5102	0.4682
	2.16	0.5090	0.4687
	2.17	0.5078	0.4692
	2.18	0.5066	0.4698
	2.19	0.5055	0.4703
	2.20	0.5043	0.4708
	2.21	0.5032	0.4713
	2.22	0.5020	0.4718
	2.23	0.5009	0.4723
	2.24	0.4998	0.4728
	2.25	0.4986	0.4733
	2.26	0.4975	0.4738
	2.27	0.4964	0.4742
	2.28	0.4953	0.4747
	2.29	0.4942	0.4752
	2.30	0.4932	0.4756
	2.31	0.4921	0.4761
	2.32	0.4910	0.4766
	2.33	0.4900	0.4770
	2.34	0.4889	0.4775
	2.35	0.4879	0.4779
	2.36	0.4868	0.4784
	2.37	0.4858	0.4788
	2.38	0.4848	0.4793
	2.39	0.4838	0.4797
	2.40	0.4827	0.4801
	2.41	0.4817	0.4805
	2.42	0.4807	0.4810
	2.43	0.4797	0.4814
	2.44	0.4788	0.4818
	2.45	0.4778	0.4822
	2.46	0.4768	0.4826
	2.47	0.4758	0.4830
	2.48	0.4749	0.4834
	2.49	0.4739	0.4838
	2.50	0.4730	0.4842
	2.51	0.4720	0.4846
	2.52	0.4711	0.4850
	2.53	0.4702	0.4854
	2.54	0.4693	0.4858
	2.55	0.4683	0.4862
	2.56	0.4674	0.4866
	2.57	0.4665	0.4869
	2.58	0.4656	0.4873
	2.59	0.4647	0.4877
	2.60	0.4638	0.4880
	2.61	0.4629	0.4884
	2.62	0.4621	0.4888
	2.63	0.4612	0.4891
	2.64	0.4603	0.4895
	2.65	0.4595	0.4899
	2.66	0.4586	0.4902
	2.67	0.4577	0.4906
	2.68	0.4569	0.4909
	2.69	0.4561	0.4912
	2.70	0.4552	0.4916
	2.71	0.4544	0.4919
	2.72	0.4536	0.4923
	2.73	0.4527	0.4926
	2.74	0.4519	0.4929
	2.75	0.4511	0.4933
	2.76	0.4503	0.4936
	2.77	0.4495	0.4939
	2.78	0.4487	0.4942
	2.79	0.4479	0.4946
	2.80	0.4471	0.4949
	2.81	0.4463	0.4952
	2.82	0.4456	0.4955
	2.83	0.4448	0.4958
	2.84	0.4440	0.4961
	2.85	0.4432	0.4965
	2.86	0.4425	0.4968
	2.87	0.4417	0.4971
	2.88	0.4410	0.4974
	2.89	0.4402	0.4977
	2.90	0.4395	0.4980
	2.91	0.4387	0.4983
	2.92	0.4380	0.4986
	2.93	0.4373	0.4989
	2.94	0.4365	0.4992
	2.95	0.4358	0.4994
	2.96	0.4351	0.4997
	2.97	0.4344	0.5000
	2.98	0.4337	0.5003
	2.99	0.4330	0.5006
	3.00	0.4323	0.5009
	3.01	0.4316	0.5011
	3.02	0.4309	0.5014
	3.03	0.4302	0.5017
	3.04	0.4295	0.5020
	3.05	0.4288	0.5022
	3.06	0.4281	0.5025
	3.07	0.4274	0.5028
	3.08	0.4268	0.5030
	3.09	0.4261	0.5033
	3.10	0.4254	0.5036
	3.11	0.4248	0.5038
	3.12	0.4241	0.5041
	3.13	0.4234	0.5044
	3.14	0.4228	0.5046
	3.15	0.4221	0.5049
	3.16	0.4215	0.5051
	3.17	0.4209	0.5054
	3.18	0.4202	0.5056
	3.19	0.4196	0.5059
	3.20	0.4189	0.5061
	3.21	0.4183	0.5064
	3.22	0.4177	0.5066
	3.23	0.4171	0.5069
	3.24	0.4164	0.5071
	3.25	0.4158	0.5074
	3.26	0.4152	0.5076
	3.27	0.4146	0.5078
	3.28	0.4140	0.5081
	3.29	0.4134	0.5083
	3.30	0.4128	0.5085
	3.31	0.4122	0.5088
	3.32	0.4116	0.5090
	3.33	0.4110	0.5092
	3.34	0.4104	0.5095
	3.35	0.4098	0.5097
	3.36	0.4092	0.5099
	3.37	0.4087	0.5101
	3.38	0.4081	0.5104
	3.39	0.4075	0.5106
	3.40	0.4069	0.5108
	3.41	0.4064	0.5110
	3.42	0.4058	0.5113
	3.43	0.4052	0.5115
	3.44	0.4047	0.5117
	3.45	0.4041	0.5119
	3.46	0.4036	0.5121
	3.47	0.4030	0.5123
	3.48	0.4024	0.5126
	3.49	0.4019	0.5128
	3.50	0.4013	0.5130
	3.51	0.4008	0.5132
	3.52	0.4003	0.5134
	3.53	0.3997	0.5136
	3.54	0.3992	0.5138
	3.55	0.3986	0.5140
	3.56	0.3981	0.5142
	3.57	0.3976	0.5144
	3.58	0.3971	0.5146
	3.59	0.3965	0.5148
	3.60	0.3960	0.5150
	3.61	0.3955	0.5152
	3.62	0.3950	0.5154
	3.63	0.3945	0.5156
	3.64	0.3939	0.5158
	3.65	0.3934	0.5160
	3.66	0.3929	0.5162
	3.67	0.3924	0.5164
	3.68	0.3919	0.5166
	3.69	0.3914	0.5168
	3.70	0.3909	0.5170
	3.71	0.3904	0.5172
	3.72	0.3899	0.5174
	3.73	0.3894	0.5176
	3.74	0.3889	0.5177
	3.75	0.3884	0.5179
	3.76	0.3879	0.5181
	3.77	0.3875	0.5183
	3.78	0.3870	0.5185
	3.79	0.3865	0.5187
	3.80	0.3860	0.5189
	3.81	0.3855	0.5190
	3.82	0.3851	0.5192
	3.83	0.3846	0.5194
	3.84	0.3841	0.5196
	3.85	0.3836	0.5198
	3.86	0.3832	0.5199
	3.87	0.3827	0.5201
	3.88	0.3822	0.5203
	3.89	0.3818	0.5205
	3.90	0.3813	0.5206
	3.91	0.3809	0.5208
	3.92	0.3804	0.5210
	3.93	0.3799	0.5212
	3.94	0.3795	0.5213
	3.95	0.3790	0.5215
	3.96	0.3786	0.5217
	3.97	0.3781	0.5218
	3.98	0.3777	0.5220
	3.99	0.3772	0.5222
	4.00	0.3768	0.5223
	4.01	0.3764	0.5225
	4.02	0.3759	0.5227
	4.03	0.3755	0.5228
	4.04	0.3750	0.5230
	4.05	0.3746	0.5232
	4.06	0.3742	0.5233
	4.07	0.3737	0.5235
	4.08	0.3733	0.5236
	4.09	0.3729	0.5238
}\tableCorrSiP

\pgfplotstableread{
	sigma	corr	si
	0.01	1.0000	0.0833
	0.02	1.0000	0.0833
	0.03	1.0000	0.0833
	0.04	1.0000	0.0833
	0.05	1.0000	0.0833
	0.06	1.0000	0.0833
	0.07	1.0000	0.0833
	0.08	1.0000	0.0833
	0.09	1.0000	0.0833
	0.10	1.0000	0.0833
	0.11	1.0000	0.0833
	0.12	1.0000	0.0833
	0.13	1.0000	0.0833
	0.14	1.0000	0.0833
	0.15	1.0000	0.0833
	0.16	1.0000	0.0833
	0.17	1.0000	0.0833
	0.18	1.0000	0.0833
	0.19	1.0000	0.0833
	0.20	1.0000	0.0833
	0.21	1.0000	0.0833
	0.22	1.0000	0.0833
	0.23	1.0000	0.0833
	0.24	1.0000	0.0833
	0.25	1.0000	0.0832
	0.26	1.0000	0.0831
	0.27	1.0000	0.0829
	0.28	1.0000	0.0827
	0.29	1.0000	0.0823
	0.30	1.0000	0.0818
	0.31	1.0000	0.0810
	0.32	0.9999	0.0801
	0.33	0.9999	0.0787
	0.34	0.9998	0.0770
	0.35	0.9997	0.0747
	0.36	0.9995	0.0719
	0.37	0.9993	0.0683
	0.38	0.9990	0.0639
	0.39	0.9986	0.0585
	0.40	0.9980	0.0521
	0.41	0.9973	0.0445
	0.42	0.9964	0.0359
	0.43	0.9954	0.0263
	0.44	0.9941	0.0157
	0.45	0.9926	0.0042
	0.46	0.9909	0.0081
	0.47	0.9890	0.0211
	0.48	0.9867	0.0347
	0.49	0.9842	0.0487
	0.50	0.9815	0.0631
	0.51	0.9785	0.0777
	0.52	0.9752	0.0924
	0.53	0.9717	0.1071
	0.54	0.9679	0.1217
	0.55	0.9639	0.1361
	0.56	0.9597	0.1503
	0.57	0.9553	0.1640
	0.58	0.9507	0.1774
	0.59	0.9459	0.1904
	0.60	0.9410	0.2029
	0.61	0.9359	0.2150
	0.62	0.9307	0.2266
	0.63	0.9254	0.2378
	0.64	0.9201	0.2485
	0.65	0.9146	0.2588
	0.66	0.9092	0.2686
	0.67	0.9036	0.2780
	0.68	0.8981	0.2870
	0.69	0.8926	0.2956
	0.70	0.8870	0.3038
	0.71	0.8815	0.3117
	0.72	0.8760	0.3193
	0.73	0.8705	0.3265
	0.74	0.8651	0.3334
	0.75	0.8597	0.3400
	0.76	0.8543	0.3464
	0.77	0.8491	0.3525
	0.78	0.8438	0.3584
	0.79	0.8387	0.3640
	0.80	0.8336	0.3694
	0.81	0.8286	0.3746
	0.82	0.8236	0.3796
	0.83	0.8187	0.3844
	0.84	0.8139	0.3890
	0.85	0.8092	0.3935
	0.86	0.8046	0.3978
	0.87	0.8000	0.4020
	0.88	0.7955	0.4060
	0.89	0.7911	0.4099
	0.90	0.7867	0.4136
	0.91	0.7824	0.4173
	0.92	0.7782	0.4208
	0.93	0.7740	0.4242
	0.94	0.7700	0.4275
	0.95	0.7660	0.4307
	0.96	0.7620	0.4338
	0.97	0.7581	0.4368
	0.98	0.7543	0.4397
	0.99	0.7506	0.4426
	1.00	0.7469	0.4453
	1.01	0.7433	0.4480
	1.02	0.7397	0.4506
	1.03	0.7362	0.4532
	1.04	0.7327	0.4557
	1.05	0.7293	0.4581
	1.06	0.7260	0.4604
	1.07	0.7227	0.4627
	1.08	0.7194	0.4650
	1.09	0.7162	0.4672
	1.10	0.7130	0.4693
	1.11	0.7099	0.4714
	1.12	0.7069	0.4734
	1.13	0.7039	0.4754
	1.14	0.7009	0.4774
	1.15	0.6980	0.4793
	1.16	0.6951	0.4811
	1.17	0.6922	0.4830
	1.18	0.6894	0.4848
	1.19	0.6866	0.4865
	1.20	0.6839	0.4882
	1.21	0.6812	0.4899
	1.22	0.6785	0.4915
	1.23	0.6759	0.4931
	1.24	0.6733	0.4947
	1.25	0.6707	0.4963
	1.26	0.6682	0.4978
	1.27	0.6657	0.4993
	1.28	0.6632	0.5007
	1.29	0.6608	0.5022
	1.30	0.6584	0.5036
	1.31	0.6560	0.5050
	1.32	0.6537	0.5063
	1.33	0.6514	0.5077
	1.34	0.6491	0.5090
	1.35	0.6468	0.5103
	1.36	0.6446	0.5115
	1.37	0.6424	0.5128
	1.38	0.6402	0.5140
	1.39	0.6380	0.5152
	1.40	0.6359	0.5164
	1.41	0.6337	0.5176
	1.42	0.6316	0.5187
	1.43	0.6296	0.5199
	1.44	0.6275	0.5210
	1.45	0.6255	0.5221
	1.46	0.6235	0.5231
	1.47	0.6215	0.5242
	1.48	0.6195	0.5253
	1.49	0.6176	0.5263
	1.50	0.6157	0.5273
	1.51	0.6138	0.5283
	1.52	0.6119	0.5293
	1.53	0.6100	0.5303
	1.54	0.6082	0.5312
	1.55	0.6063	0.5322
	1.56	0.6045	0.5331
	1.57	0.6027	0.5340
	1.58	0.6009	0.5349
	1.59	0.5992	0.5358
	1.60	0.5974	0.5367
	1.61	0.5957	0.5376
	1.62	0.5940	0.5384
	1.63	0.5923	0.5393
	1.64	0.5906	0.5401
	1.65	0.5889	0.5410
	1.66	0.5873	0.5418
	1.67	0.5856	0.5426
	1.68	0.5840	0.5434
	1.69	0.5824	0.5442
	1.70	0.5808	0.5450
	1.71	0.5792	0.5457
	1.72	0.5777	0.5465
	1.73	0.5761	0.5472
	1.74	0.5746	0.5480
	1.75	0.5730	0.5487
	1.76	0.5715	0.5494
	1.77	0.5700	0.5501
	1.78	0.5685	0.5508
	1.79	0.5670	0.5515
	1.80	0.5656	0.5522
	1.81	0.5641	0.5529
	1.82	0.5627	0.5536
	1.83	0.5612	0.5543
	1.84	0.5598	0.5549
	1.85	0.5584	0.5556
	1.86	0.5570	0.5562
	1.87	0.5556	0.5569
	1.88	0.5542	0.5575
	1.89	0.5528	0.5581
	1.90	0.5515	0.5587
	1.91	0.5501	0.5593
	1.92	0.5488	0.5600
	1.93	0.5475	0.5606
	1.94	0.5461	0.5611
	1.95	0.5448	0.5617
	1.96	0.5435	0.5623
	1.97	0.5422	0.5629
	1.98	0.5410	0.5635
	1.99	0.5397	0.5640
	2.00	0.5384	0.5646
	2.01	0.5372	0.5651
	2.02	0.5359	0.5657
	2.03	0.5347	0.5662
	2.04	0.5334	0.5668
	2.05	0.5322	0.5673
	2.06	0.5310	0.5678
	2.07	0.5298	0.5683
	2.08	0.5286	0.5689
	2.09	0.5274	0.5694
	2.10	0.5262	0.5699
	2.11	0.5250	0.5704
	2.12	0.5239	0.5709
	2.13	0.5227	0.5714
	2.14	0.5216	0.5719
	2.15	0.5204	0.5723
	2.16	0.5193	0.5728
	2.17	0.5182	0.5733
	2.18	0.5170	0.5738
	2.19	0.5159	0.5742
	2.20	0.5148	0.5747
	2.21	0.5137	0.5752
	2.22	0.5126	0.5756
	2.23	0.5115	0.5761
	2.24	0.5104	0.5765
	2.25	0.5094	0.5769
	2.26	0.5083	0.5774
	2.27	0.5072	0.5778
	2.28	0.5062	0.5783
	2.29	0.5051	0.5787
	2.30	0.5041	0.5791
	2.31	0.5031	0.5795
	2.32	0.5020	0.5799
	2.33	0.5010	0.5804
	2.34	0.5000	0.5808
	2.35	0.4990	0.5812
	2.36	0.4980	0.5816
	2.37	0.4970	0.5820
	2.38	0.4960	0.5824
	2.39	0.4950	0.5828
	2.40	0.4940	0.5832
	2.41	0.4930	0.5835
	2.42	0.4921	0.5839
	2.43	0.4911	0.5843
	2.44	0.4901	0.5847
	2.45	0.4892	0.5851
	2.46	0.4882	0.5854
	2.47	0.4873	0.5858
	2.48	0.4863	0.5862
	2.49	0.4854	0.5865
	2.50	0.4845	0.5869
	2.51	0.4836	0.5873
	2.52	0.4826	0.5876
	2.53	0.4817	0.5880
	2.54	0.4808	0.5883
	2.55	0.4799	0.5887
	2.56	0.4790	0.5890
	2.57	0.4781	0.5894
	2.58	0.4772	0.5897
	2.59	0.4764	0.5900
	2.60	0.4755	0.5904
	2.61	0.4746	0.5907
	2.62	0.4737	0.5910
	2.63	0.4729	0.5914
	2.64	0.4720	0.5917
	2.65	0.4712	0.5920
	2.66	0.4703	0.5923
	2.67	0.4695	0.5926
	2.68	0.4686	0.5930
	2.69	0.4678	0.5933
	2.70	0.4670	0.5936
	2.71	0.4661	0.5939
	2.72	0.4653	0.5942
	2.73	0.4645	0.5945
	2.74	0.4637	0.5948
	2.75	0.4629	0.5951
	2.76	0.4620	0.5954
	2.77	0.4612	0.5957
	2.78	0.4604	0.5960
	2.79	0.4597	0.5963
	2.80	0.4589	0.5966
	2.81	0.4581	0.5969
	2.82	0.4573	0.5972
	2.83	0.4565	0.5974
	2.84	0.4557	0.5977
	2.85	0.4550	0.5980
	2.86	0.4542	0.5983
	2.87	0.4534	0.5986
	2.88	0.4527	0.5988
	2.89	0.4519	0.5991
	2.90	0.4512	0.5994
	2.91	0.4504	0.5997
	2.92	0.4497	0.5999
	2.93	0.4489	0.6002
	2.94	0.4482	0.6005
	2.95	0.4475	0.6007
	2.96	0.4467	0.6010
	2.97	0.4460	0.6012
	2.98	0.4453	0.6015
	2.99	0.4446	0.6018
	3.00	0.4439	0.6020
	3.01	0.4432	0.6023
	3.02	0.4424	0.6025
	3.03	0.4417	0.6028
	3.04	0.4410	0.6030
	3.05	0.4403	0.6033
	3.06	0.4396	0.6035
	3.07	0.4390	0.6037
	3.08	0.4383	0.6040
	3.09	0.4376	0.6042
	3.10	0.4369	0.6045
	3.11	0.4362	0.6047
	3.12	0.4355	0.6049
	3.13	0.4349	0.6052
	3.14	0.4342	0.6054
	3.15	0.4335	0.6056
	3.16	0.4329	0.6059
	3.17	0.4322	0.6061
	3.18	0.4316	0.6063
	3.19	0.4309	0.6066
	3.20	0.4303	0.6068
	3.21	0.4296	0.6070
	3.22	0.4290	0.6072
	3.23	0.4283	0.6075
	3.24	0.4277	0.6077
	3.25	0.4271	0.6079
	3.26	0.4264	0.6081
	3.27	0.4258	0.6083
	3.28	0.4252	0.6085
	3.29	0.4245	0.6088
	3.30	0.4239	0.6090
	3.31	0.4233	0.6092
	3.32	0.4227	0.6094
	3.33	0.4221	0.6096
	3.34	0.4215	0.6098
	3.35	0.4208	0.6100
	3.36	0.4202	0.6102
	3.37	0.4196	0.6104
	3.38	0.4190	0.6106
	3.39	0.4184	0.6108
	3.40	0.4178	0.6110
	3.41	0.4173	0.6112
	3.42	0.4167	0.6114
	3.43	0.4161	0.6116
	3.44	0.4155	0.6118
	3.45	0.4149	0.6120
	3.46	0.4143	0.6122
	3.47	0.4137	0.6124
	3.48	0.4132	0.6126
	3.49	0.4126	0.6128
	3.50	0.4120	0.6130
	3.51	0.4115	0.6132
	3.52	0.4109	0.6134
	3.53	0.4103	0.6136
	3.54	0.4098	0.6137
	3.55	0.4092	0.6139
	3.56	0.4087	0.6141
	3.57	0.4081	0.6143
	3.58	0.4075	0.6145
	3.59	0.4070	0.6147
	3.60	0.4065	0.6148
	3.61	0.4059	0.6150
	3.62	0.4054	0.6152
	3.63	0.4048	0.6154
	3.64	0.4043	0.6156
	3.65	0.4037	0.6157
	3.66	0.4032	0.6159
	3.67	0.4027	0.6161
	3.68	0.4021	0.6163
	3.69	0.4016	0.6164
	3.70	0.4011	0.6166
	3.71	0.4006	0.6168
	3.72	0.4000	0.6169
	3.73	0.3995	0.6171
	3.74	0.3990	0.6173
	3.75	0.3985	0.6175
	3.76	0.3980	0.6176
	3.77	0.3975	0.6178
	3.78	0.3970	0.6179
	3.79	0.3964	0.6181
	3.80	0.3959	0.6183
	3.81	0.3954	0.6184
	3.82	0.3949	0.6186
	3.83	0.3944	0.6188
	3.84	0.3939	0.6189
	3.85	0.3934	0.6191
	3.86	0.3929	0.6192
	3.87	0.3925	0.6194
	3.88	0.3920	0.6196
	3.89	0.3915	0.6197
	3.90	0.3910	0.6199
	3.91	0.3905	0.6200
	3.92	0.3900	0.6202
	3.93	0.3895	0.6203
	3.94	0.3891	0.6205
	3.95	0.3886	0.6206
	3.96	0.3881	0.6208
	3.97	0.3876	0.6209
	3.98	0.3871	0.6211
	3.99	0.3867	0.6212
	4.00	0.3862	0.6214
	4.01	0.3857	0.6215
	4.02	0.3853	0.6217
	4.03	0.3848	0.6218
	4.04	0.3843	0.6220
	4.05	0.3839	0.6221
	4.06	0.3834	0.6223
	4.07	0.3830	0.6224
	4.08	0.3825	0.6226
	4.09	0.3820	0.6227
}\tableCorrSiQ

\pgfplotstableread{
	sigma	corr	si
	0.01	1.0000	0.0833
	0.02	1.0000	0.0833
	0.03	1.0000	0.0833
	0.04	1.0000	0.0833
	0.05	1.0000	0.0833
	0.06	1.0000	0.0833
	0.07	1.0000	0.0833
	0.08	1.0000	0.0833
	0.09	1.0000	0.0833
	0.10	1.0000	0.0833
	0.11	1.0000	0.0833
	0.12	1.0000	0.0833
	0.13	1.0000	0.0833
	0.14	1.0000	0.0833
	0.15	1.0000	0.0833
	0.16	1.0000	0.0833
	0.17	1.0000	0.0833
	0.18	1.0000	0.0833
	0.19	1.0000	0.0833
	0.20	1.0000	0.0833
	0.21	1.0000	0.0833
	0.22	1.0000	0.0833
	0.23	1.0000	0.0833
	0.24	1.0000	0.0833
	0.25	1.0000	0.0833
	0.26	1.0000	0.0832
	0.27	1.0000	0.0831
	0.28	1.0000	0.0830
	0.29	1.0000	0.0827
	0.30	1.0000	0.0825
	0.31	1.0000	0.0821
	0.32	0.9999	0.0815
	0.33	0.9999	0.0808
	0.34	0.9998	0.0800
	0.35	0.9997	0.0789
	0.36	0.9996	0.0775
	0.37	0.9994	0.0759
	0.38	0.9991	0.0739
	0.39	0.9987	0.0715
	0.40	0.9981	0.0688
	0.41	0.9975	0.0656
	0.42	0.9967	0.0619
	0.43	0.9957	0.0578
	0.44	0.9945	0.0533
	0.45	0.9931	0.0484
	0.46	0.9915	0.0432
	0.47	0.9897	0.0377
	0.48	0.9876	0.0320
	0.49	0.9852	0.0261
	0.50	0.9827	0.0200
	0.51	0.9798	0.0137
	0.52	0.9767	0.0074
	0.53	0.9734	0.0010
	0.54	0.9698	0.0054
	0.55	0.9660	0.0119
	0.56	0.9620	0.0183
	0.57	0.9578	0.0247
	0.58	0.9534	0.0310
	0.59	0.9489	0.0372
	0.60	0.9442	0.0433
	0.61	0.9393	0.0493
	0.62	0.9344	0.0552
	0.63	0.9293	0.0610
	0.64	0.9241	0.0667
	0.65	0.9189	0.0722
	0.66	0.9136	0.0776
	0.67	0.9082	0.0828
	0.68	0.9029	0.0879
	0.69	0.8975	0.0929
	0.70	0.8921	0.0978
	0.71	0.8867	0.1025
	0.72	0.8813	0.1072
	0.73	0.8759	0.1117
	0.74	0.8706	0.1160
	0.75	0.8652	0.1203
	0.76	0.8600	0.1244
	0.77	0.8547	0.1285
	0.78	0.8495	0.1324
	0.79	0.8444	0.1363
	0.80	0.8393	0.1400
	0.81	0.8343	0.1437
	0.82	0.8293	0.1472
	0.83	0.8244	0.1507
	0.84	0.8195	0.1541
	0.85	0.8147	0.1574
	0.86	0.8100	0.1606
	0.87	0.8053	0.1638
	0.88	0.8007	0.1669
	0.89	0.7962	0.1699
	0.90	0.7917	0.1729
	0.91	0.7873	0.1758
	0.92	0.7829	0.1786
	0.93	0.7786	0.1814
	0.94	0.7744	0.1841
	0.95	0.7702	0.1867
	0.96	0.7661	0.1893
	0.97	0.7620	0.1919
	0.98	0.7580	0.1944
	0.99	0.7540	0.1968
	1.00	0.7501	0.1992
	1.01	0.7463	0.2016
	1.02	0.7425	0.2039
	1.03	0.7387	0.2062
	1.04	0.7350	0.2084
	1.05	0.7314	0.2106
	1.06	0.7278	0.2127
	1.07	0.7242	0.2148
	1.08	0.7207	0.2169
	1.09	0.7172	0.2189
	1.10	0.7138	0.2209
	1.11	0.7104	0.2229
	1.12	0.7070	0.2248
	1.13	0.7037	0.2267
	1.14	0.7005	0.2286
	1.15	0.6973	0.2304
	1.16	0.6941	0.2322
	1.17	0.6909	0.2340
	1.18	0.6878	0.2357
	1.19	0.6847	0.2375
	1.20	0.6817	0.2392
	1.21	0.6787	0.2408
	1.22	0.6757	0.2425
	1.23	0.6728	0.2441
	1.24	0.6699	0.2457
	1.25	0.6670	0.2472
	1.26	0.6641	0.2488
	1.27	0.6613	0.2503
	1.28	0.6585	0.2518
	1.29	0.6558	0.2533
	1.30	0.6531	0.2547
	1.31	0.6504	0.2562
	1.32	0.6477	0.2576
	1.33	0.6451	0.2590
	1.34	0.6424	0.2604
	1.35	0.6399	0.2617
	1.36	0.6373	0.2631
	1.37	0.6348	0.2644
	1.38	0.6323	0.2657
	1.39	0.6298	0.2670
	1.40	0.6273	0.2682
	1.41	0.6249	0.2695
	1.42	0.6225	0.2707
	1.43	0.6201	0.2720
	1.44	0.6177	0.2732
	1.45	0.6154	0.2743
	1.46	0.6131	0.2755
	1.47	0.6108	0.2767
	1.48	0.6085	0.2778
	1.49	0.6062	0.2789
	1.50	0.6040	0.2801
	1.51	0.6018	0.2812
	1.52	0.5996	0.2822
	1.53	0.5974	0.2833
	1.54	0.5953	0.2844
	1.55	0.5931	0.2854
	1.56	0.5910	0.2865
	1.57	0.5889	0.2875
	1.58	0.5869	0.2885
	1.59	0.5848	0.2895
	1.60	0.5828	0.2905
	1.61	0.5808	0.2915
	1.62	0.5788	0.2924
	1.63	0.5768	0.2934
	1.64	0.5748	0.2943
	1.65	0.5729	0.2953
	1.66	0.5709	0.2962
	1.67	0.5690	0.2971
	1.68	0.5671	0.2980
	1.69	0.5652	0.2989
	1.70	0.5634	0.2998
	1.71	0.5615	0.3007
	1.72	0.5597	0.3015
	1.73	0.5579	0.3024
	1.74	0.5561	0.3032
	1.75	0.5543	0.3041
	1.76	0.5525	0.3049
	1.77	0.5508	0.3057
	1.78	0.5490	0.3065
	1.79	0.5473	0.3073
	1.80	0.5456	0.3081
	1.81	0.5439	0.3089
	1.82	0.5422	0.3097
	1.83	0.5405	0.3104
	1.84	0.5389	0.3112
	1.85	0.5372	0.3119
	1.86	0.5356	0.3127
	1.87	0.5340	0.3134
	1.88	0.5324	0.3142
	1.89	0.5308	0.3149
	1.90	0.5292	0.3156
	1.91	0.5277	0.3163
	1.92	0.5261	0.3170
	1.93	0.5246	0.3177
	1.94	0.5230	0.3184
	1.95	0.5215	0.3191
	1.96	0.5200	0.3197
	1.97	0.5185	0.3204
	1.98	0.5170	0.3211
	1.99	0.5156	0.3217
	2.00	0.5141	0.3224
	2.01	0.5127	0.3230
	2.02	0.5112	0.3237
	2.03	0.5098	0.3243
	2.04	0.5084	0.3249
	2.05	0.5070	0.3255
	2.06	0.5056	0.3261
	2.07	0.5042	0.3268
	2.08	0.5029	0.3274
	2.09	0.5015	0.3280
	2.10	0.5002	0.3285
	2.11	0.4988	0.3291
	2.12	0.4975	0.3297
	2.13	0.4962	0.3303
	2.14	0.4949	0.3309
	2.15	0.4936	0.3314
	2.16	0.4923	0.3320
	2.17	0.4910	0.3326
	2.18	0.4898	0.3331
	2.19	0.4885	0.3337
	2.20	0.4872	0.3342
	2.21	0.4860	0.3347
	2.22	0.4848	0.3353
	2.23	0.4835	0.3358
	2.24	0.4823	0.3363
	2.25	0.4811	0.3368
	2.26	0.4799	0.3374
	2.27	0.4787	0.3379
	2.28	0.4776	0.3384
	2.29	0.4764	0.3389
	2.30	0.4752	0.3394
	2.31	0.4741	0.3399
	2.32	0.4729	0.3404
	2.33	0.4718	0.3409
	2.34	0.4707	0.3413
	2.35	0.4695	0.3418
	2.36	0.4684	0.3423
	2.37	0.4673	0.3428
	2.38	0.4662	0.3432
	2.39	0.4651	0.3437
	2.40	0.4640	0.3442
	2.41	0.4630	0.3446
	2.42	0.4619	0.3451
	2.43	0.4608	0.3455
	2.44	0.4598	0.3460
	2.45	0.4587	0.3464
	2.46	0.4577	0.3469
	2.47	0.4567	0.3473
	2.48	0.4556	0.3477
	2.49	0.4546	0.3482
	2.50	0.4536	0.3486
	2.51	0.4526	0.3490
	2.52	0.4516	0.3494
	2.53	0.4506	0.3499
	2.54	0.4496	0.3503
	2.55	0.4486	0.3507
	2.56	0.4477	0.3511
	2.57	0.4467	0.3515
	2.58	0.4457	0.3519
	2.59	0.4448	0.3523
	2.60	0.4438	0.3527
	2.61	0.4429	0.3531
	2.62	0.4420	0.3535
	2.63	0.4410	0.3539
	2.64	0.4401	0.3543
	2.65	0.4392	0.3546
	2.66	0.4383	0.3550
	2.67	0.4374	0.3554
	2.68	0.4365	0.3558
	2.69	0.4356	0.3561
	2.70	0.4347	0.3565
	2.71	0.4338	0.3569
	2.72	0.4329	0.3573
	2.73	0.4321	0.3576
	2.74	0.4312	0.3580
	2.75	0.4303	0.3583
	2.76	0.4295	0.3587
	2.77	0.4286	0.3590
	2.78	0.4278	0.3594
	2.79	0.4269	0.3597
	2.80	0.4261	0.3601
	2.81	0.4253	0.3604
	2.82	0.4245	0.3608
	2.83	0.4236	0.3611
	2.84	0.4228	0.3614
	2.85	0.4220	0.3618
	2.86	0.4212	0.3621
	2.87	0.4204	0.3624
	2.88	0.4196	0.3628
	2.89	0.4188	0.3631
	2.90	0.4180	0.3634
	2.91	0.4172	0.3638
	2.92	0.4165	0.3641
	2.93	0.4157	0.3644
	2.94	0.4149	0.3647
	2.95	0.4142	0.3650
	2.96	0.4134	0.3653
	2.97	0.4126	0.3656
	2.98	0.4119	0.3660
	2.99	0.4111	0.3663
	3.00	0.4104	0.3666
	3.01	0.4097	0.3669
	3.02	0.4089	0.3672
	3.03	0.4082	0.3675
	3.04	0.4075	0.3678
	3.05	0.4068	0.3681
	3.06	0.4060	0.3684
	3.07	0.4053	0.3687
	3.08	0.4046	0.3689
	3.09	0.4039	0.3692
	3.10	0.4032	0.3695
	3.11	0.4025	0.3698
	3.12	0.4018	0.3701
	3.13	0.4011	0.3704
	3.14	0.4005	0.3707
	3.15	0.3998	0.3709
	3.16	0.3991	0.3712
	3.17	0.3984	0.3715
	3.18	0.3978	0.3718
	3.19	0.3971	0.3720
	3.20	0.3964	0.3723
	3.21	0.3958	0.3726
	3.22	0.3951	0.3728
	3.23	0.3945	0.3731
	3.24	0.3938	0.3734
	3.25	0.3932	0.3736
	3.26	0.3925	0.3739
	3.27	0.3919	0.3742
	3.28	0.3912	0.3744
	3.29	0.3906	0.3747
	3.30	0.3900	0.3749
	3.31	0.3894	0.3752
	3.32	0.3887	0.3754
	3.33	0.3881	0.3757
	3.34	0.3875	0.3759
	3.35	0.3869	0.3762
	3.36	0.3863	0.3764
	3.37	0.3857	0.3767
	3.38	0.3851	0.3769
	3.39	0.3845	0.3772
	3.40	0.3839	0.3774
	3.41	0.3833	0.3777
	3.42	0.3827	0.3779
	3.43	0.3821	0.3781
	3.44	0.3815	0.3784
	3.45	0.3809	0.3786
	3.46	0.3804	0.3788
	3.47	0.3798	0.3791
	3.48	0.3792	0.3793
	3.49	0.3786	0.3795
	3.50	0.3781	0.3798
	3.51	0.3775	0.3800
	3.52	0.3770	0.3802
	3.53	0.3764	0.3805
	3.54	0.3758	0.3807
	3.55	0.3753	0.3809
	3.56	0.3747	0.3811
	3.57	0.3742	0.3813
	3.58	0.3736	0.3816
	3.59	0.3731	0.3818
	3.60	0.3726	0.3820
	3.61	0.3720	0.3822
	3.62	0.3715	0.3824
	3.63	0.3710	0.3827
	3.64	0.3704	0.3829
	3.65	0.3699	0.3831
	3.66	0.3694	0.3833
	3.67	0.3689	0.3835
	3.68	0.3683	0.3837
	3.69	0.3678	0.3839
	3.70	0.3673	0.3841
	3.71	0.3668	0.3843
	3.72	0.3663	0.3846
	3.73	0.3658	0.3848
	3.74	0.3653	0.3850
	3.75	0.3648	0.3852
	3.76	0.3643	0.3854
	3.77	0.3638	0.3856
	3.78	0.3633	0.3858
	3.79	0.3628	0.3860
	3.80	0.3623	0.3862
	3.81	0.3618	0.3864
	3.82	0.3613	0.3866
	3.83	0.3608	0.3868
	3.84	0.3603	0.3870
	3.85	0.3599	0.3871
	3.86	0.3594	0.3873
	3.87	0.3589	0.3875
	3.88	0.3584	0.3877
	3.89	0.3580	0.3879
	3.90	0.3575	0.3881
	3.91	0.3570	0.3883
	3.92	0.3565	0.3885
	3.93	0.3561	0.3887
	3.94	0.3556	0.3889
	3.95	0.3552	0.3890
	3.96	0.3547	0.3892
	3.97	0.3542	0.3894
	3.98	0.3538	0.3896
	3.99	0.3533	0.3898
	4.00	0.3529	0.3900
	4.01	0.3524	0.3901
	4.02	0.3520	0.3903
	4.03	0.3515	0.3905
	4.04	0.3511	0.3907
	4.05	0.3507	0.3908
	4.06	0.3502	0.3910
	4.07	0.3498	0.3912
	4.08	0.3493	0.3914
	4.09	0.3489	0.3915
}\tableCorrSiR

\pgfplotstableread{
	sigma	corr	si
	0.01	1.0000	0.0833
	0.02	1.0000	0.0833
	0.03	1.0000	0.0833
	0.04	1.0000	0.0833
	0.05	1.0000	0.0833
	0.06	1.0000	0.0833
	0.07	1.0000	0.0833
	0.08	1.0000	0.0833
	0.09	1.0000	0.0833
	0.10	1.0000	0.0833
	0.11	1.0000	0.0833
	0.12	1.0000	0.0833
	0.13	1.0000	0.0833
	0.14	1.0000	0.0833
	0.15	1.0000	0.0833
	0.16	1.0000	0.0833
	0.17	1.0000	0.0833
	0.18	1.0000	0.0833
	0.19	1.0000	0.0833
	0.20	1.0000	0.0833
	0.21	1.0000	0.0833
	0.22	1.0000	0.0833
	0.23	1.0000	0.0833
	0.24	1.0000	0.0833
	0.25	1.0000	0.0832
	0.26	1.0000	0.0831
	0.27	1.0000	0.0830
	0.28	1.0000	0.0828
	0.29	1.0000	0.0825
	0.30	1.0000	0.0821
	0.31	1.0000	0.0815
	0.32	0.9999	0.0807
	0.33	0.9999	0.0797
	0.34	0.9998	0.0783
	0.35	0.9997	0.0766
	0.36	0.9995	0.0744
	0.37	0.9993	0.0717
	0.38	0.9990	0.0684
	0.39	0.9986	0.0645
	0.40	0.9980	0.0597
	0.41	0.9973	0.0542
	0.42	0.9965	0.0478
	0.43	0.9954	0.0407
	0.44	0.9942	0.0329
	0.45	0.9927	0.0244
	0.46	0.9910	0.0153
	0.47	0.9891	0.0057
	0.48	0.9869	0.0044
	0.49	0.9844	0.0149
	0.50	0.9817	0.0256
	0.51	0.9787	0.0366
	0.52	0.9754	0.0477
	0.53	0.9719	0.0589
	0.54	0.9681	0.0701
	0.55	0.9642	0.0813
	0.56	0.9599	0.0923
	0.57	0.9555	0.1032
	0.58	0.9509	0.1139
	0.59	0.9461	0.1243
	0.60	0.9412	0.1346
	0.61	0.9361	0.1445
	0.62	0.9309	0.1542
	0.63	0.9256	0.1636
	0.64	0.9203	0.1726
	0.65	0.9148	0.1814
	0.66	0.9093	0.1898
	0.67	0.9038	0.1980
	0.68	0.8982	0.2058
	0.69	0.8926	0.2134
	0.70	0.8870	0.2207
	0.71	0.8815	0.2278
	0.72	0.8759	0.2345
	0.73	0.8704	0.2411
	0.74	0.8649	0.2473
	0.75	0.8595	0.2534
	0.76	0.8541	0.2592
	0.77	0.8488	0.2649
	0.78	0.8435	0.2703
	0.79	0.8383	0.2756
	0.80	0.8331	0.2806
	0.81	0.8281	0.2855
	0.82	0.8231	0.2902
	0.83	0.8181	0.2948
	0.84	0.8133	0.2992
	0.85	0.8085	0.3035
	0.86	0.8038	0.3076
	0.87	0.7991	0.3116
	0.88	0.7945	0.3155
	0.89	0.7900	0.3192
	0.90	0.7856	0.3229
	0.91	0.7813	0.3264
	0.92	0.7770	0.3298
	0.93	0.7728	0.3332
	0.94	0.7686	0.3364
	0.95	0.7646	0.3396
	0.96	0.7605	0.3426
	0.97	0.7566	0.3456
	0.98	0.7527	0.3485
	0.99	0.7489	0.3513
	1.00	0.7451	0.3541
	1.01	0.7414	0.3568
	1.02	0.7378	0.3594
	1.03	0.7342	0.3620
	1.04	0.7306	0.3645
	1.05	0.7271	0.3669
	1.06	0.7237	0.3693
	1.07	0.7203	0.3716
	1.08	0.7170	0.3739
	1.09	0.7137	0.3761
	1.10	0.7105	0.3783
	1.11	0.7073	0.3804
	1.12	0.7041	0.3825
	1.13	0.7010	0.3845
	1.14	0.6980	0.3865
	1.15	0.6950	0.3884
	1.16	0.6920	0.3904
	1.17	0.6891	0.3922
	1.18	0.6862	0.3941
	1.19	0.6833	0.3959
	1.20	0.6805	0.3976
	1.21	0.6777	0.3994
	1.22	0.6749	0.4011
	1.23	0.6722	0.4027
	1.24	0.6695	0.4044
	1.25	0.6669	0.4060
	1.26	0.6643	0.4075
	1.27	0.6617	0.4091
	1.28	0.6591	0.4106
	1.29	0.6566	0.4121
	1.30	0.6541	0.4136
	1.31	0.6516	0.4150
	1.32	0.6492	0.4165
	1.33	0.6468	0.4179
	1.34	0.6444	0.4192
	1.35	0.6420	0.4206
	1.36	0.6397	0.4219
	1.37	0.6374	0.4232
	1.38	0.6351	0.4245
	1.39	0.6328	0.4258
	1.40	0.6306	0.4270
	1.41	0.6284	0.4283
	1.42	0.6262	0.4295
	1.43	0.6240	0.4307
	1.44	0.6219	0.4319
	1.45	0.6198	0.4330
	1.46	0.6177	0.4342
	1.47	0.6156	0.4353
	1.48	0.6135	0.4364
	1.49	0.6115	0.4375
	1.50	0.6095	0.4386
	1.51	0.6075	0.4396
	1.52	0.6055	0.4407
	1.53	0.6035	0.4417
	1.54	0.6016	0.4428
	1.55	0.5997	0.4438
	1.56	0.5978	0.4448
	1.57	0.5959	0.4457
	1.58	0.5940	0.4467
	1.59	0.5922	0.4477
	1.60	0.5903	0.4486
	1.61	0.5885	0.4495
	1.62	0.5867	0.4505
	1.63	0.5849	0.4514
	1.64	0.5831	0.4523
	1.65	0.5814	0.4532
	1.66	0.5796	0.4540
	1.67	0.5779	0.4549
	1.68	0.5762	0.4558
	1.69	0.5745	0.4566
	1.70	0.5728	0.4574
	1.71	0.5712	0.4583
	1.72	0.5695	0.4591
	1.73	0.5679	0.4599
	1.74	0.5662	0.4607
	1.75	0.5646	0.4615
	1.76	0.5630	0.4622
	1.77	0.5614	0.4630
	1.78	0.5599	0.4638
	1.79	0.5583	0.4645
	1.80	0.5567	0.4653
	1.81	0.5552	0.4660
	1.82	0.5537	0.4667
	1.83	0.5522	0.4674
	1.84	0.5507	0.4681
	1.85	0.5492	0.4688
	1.86	0.5477	0.4695
	1.87	0.5462	0.4702
	1.88	0.5448	0.4709
	1.89	0.5433	0.4716
	1.90	0.5419	0.4723
	1.91	0.5405	0.4729
	1.92	0.5391	0.4736
	1.93	0.5377	0.4742
	1.94	0.5363	0.4749
	1.95	0.5349	0.4755
	1.96	0.5335	0.4761
	1.97	0.5321	0.4767
	1.98	0.5308	0.4774
	1.99	0.5295	0.4780
	2.00	0.5281	0.4786
	2.01	0.5268	0.4792
	2.02	0.5255	0.4798
	2.03	0.5242	0.4803
	2.04	0.5229	0.4809
	2.05	0.5216	0.4815
	2.06	0.5203	0.4821
	2.07	0.5191	0.4826
	2.08	0.5178	0.4832
	2.09	0.5166	0.4837
	2.10	0.5153	0.4843
	2.11	0.5141	0.4848
	2.12	0.5129	0.4854
	2.13	0.5116	0.4859
	2.14	0.5104	0.4864
	2.15	0.5092	0.4869
	2.16	0.5081	0.4875
	2.17	0.5069	0.4880
	2.18	0.5057	0.4885
	2.19	0.5045	0.4890
	2.20	0.5034	0.4895
	2.21	0.5022	0.4900
	2.22	0.5011	0.4905
	2.23	0.4999	0.4910
	2.24	0.4988	0.4915
	2.25	0.4977	0.4919
	2.26	0.4966	0.4924
	2.27	0.4955	0.4929
	2.28	0.4944	0.4933
	2.29	0.4933	0.4938
	2.30	0.4922	0.4943
	2.31	0.4911	0.4947
	2.32	0.4901	0.4952
	2.33	0.4890	0.4956
	2.34	0.4879	0.4961
	2.35	0.4869	0.4965
	2.36	0.4858	0.4969
	2.37	0.4848	0.4974
	2.38	0.4838	0.4978
	2.39	0.4828	0.4982
	2.40	0.4817	0.4986
	2.41	0.4807	0.4991
	2.42	0.4797	0.4995
	2.43	0.4787	0.4999
	2.44	0.4777	0.5003
	2.45	0.4768	0.5007
	2.46	0.4758	0.5011
	2.47	0.4748	0.5015
	2.48	0.4738	0.5019
	2.49	0.4729	0.5023
	2.50	0.4719	0.5027
	2.51	0.4710	0.5031
	2.52	0.4700	0.5035
	2.53	0.4691	0.5038
	2.54	0.4682	0.5042
	2.55	0.4672	0.5046
	2.56	0.4663	0.5050
	2.57	0.4654	0.5053
	2.58	0.4645	0.5057
	2.59	0.4636	0.5061
	2.60	0.4627	0.5064
	2.61	0.4618	0.5068
	2.62	0.4609	0.5071
	2.63	0.4600	0.5075
	2.64	0.4592	0.5078
	2.65	0.4583	0.5082
	2.66	0.4574	0.5085
	2.67	0.4566	0.5089
	2.68	0.4557	0.5092
	2.69	0.4548	0.5096
	2.70	0.4540	0.5099
	2.71	0.4532	0.5102
	2.72	0.4523	0.5106
	2.73	0.4515	0.5109
	2.74	0.4507	0.5112
	2.75	0.4499	0.5115
	2.76	0.4490	0.5119
	2.77	0.4482	0.5122
	2.78	0.4474	0.5125
	2.79	0.4466	0.5128
	2.80	0.4458	0.5131
	2.81	0.4450	0.5134
	2.82	0.4442	0.5137
	2.83	0.4435	0.5140
	2.84	0.4427	0.5143
	2.85	0.4419	0.5146
	2.86	0.4411	0.5149
	2.87	0.4404	0.5152
	2.88	0.4396	0.5155
	2.89	0.4388	0.5158
	2.90	0.4381	0.5161
	2.91	0.4373	0.5164
	2.92	0.4366	0.5167
	2.93	0.4359	0.5170
	2.94	0.4351	0.5173
	2.95	0.4344	0.5176
	2.96	0.4337	0.5178
	2.97	0.4329	0.5181
	2.98	0.4322	0.5184
	2.99	0.4315	0.5187
	3.00	0.4308	0.5189
	3.01	0.4301	0.5192
	3.02	0.4294	0.5195
	3.03	0.4287	0.5198
	3.04	0.4280	0.5200
	3.05	0.4273	0.5203
	3.06	0.4266	0.5206
	3.07	0.4259	0.5208
	3.08	0.4252	0.5211
	3.09	0.4246	0.5213
	3.10	0.4239	0.5216
	3.11	0.4232	0.5218
	3.12	0.4225	0.5221
	3.13	0.4219	0.5224
	3.14	0.4212	0.5226
	3.15	0.4206	0.5229
	3.16	0.4199	0.5231
	3.17	0.4193	0.5233
	3.18	0.4186	0.5236
	3.19	0.4180	0.5238
	3.20	0.4173	0.5241
	3.21	0.4167	0.5243
	3.22	0.4161	0.5246
	3.23	0.4154	0.5248
	3.24	0.4148	0.5250
	3.25	0.4142	0.5253
	3.26	0.4136	0.5255
	3.27	0.4129	0.5257
	3.28	0.4123	0.5260
	3.29	0.4117	0.5262
	3.30	0.4111	0.5264
	3.31	0.4105	0.5266
	3.32	0.4099	0.5269
	3.33	0.4093	0.5271
	3.34	0.4087	0.5273
	3.35	0.4081	0.5275
	3.36	0.4075	0.5278
	3.37	0.4069	0.5280
	3.38	0.4064	0.5282
	3.39	0.4058	0.5284
	3.40	0.4052	0.5286
	3.41	0.4046	0.5288
	3.42	0.4041	0.5291
	3.43	0.4035	0.5293
	3.44	0.4029	0.5295
	3.45	0.4024	0.5297
	3.46	0.4018	0.5299
	3.47	0.4012	0.5301
	3.48	0.4007	0.5303
	3.49	0.4001	0.5305
	3.50	0.3996	0.5307
	3.51	0.3990	0.5309
	3.52	0.3985	0.5311
	3.53	0.3979	0.5313
	3.54	0.3974	0.5315
	3.55	0.3969	0.5317
	3.56	0.3963	0.5319
	3.57	0.3958	0.5321
	3.58	0.3953	0.5323
	3.59	0.3947	0.5325
	3.60	0.3942	0.5327
	3.61	0.3937	0.5329
	3.62	0.3932	0.5331
	3.63	0.3926	0.5333
	3.64	0.3921	0.5335
	3.65	0.3916	0.5337
	3.66	0.3911	0.5338
	3.67	0.3906	0.5340
	3.68	0.3901	0.5342
	3.69	0.3896	0.5344
	3.70	0.3891	0.5346
	3.71	0.3886	0.5348
	3.72	0.3881	0.5349
	3.73	0.3876	0.5351
	3.74	0.3871	0.5353
	3.75	0.3866	0.5355
	3.76	0.3861	0.5357
	3.77	0.3856	0.5358
	3.78	0.3851	0.5360
	3.79	0.3846	0.5362
	3.80	0.3842	0.5364
	3.81	0.3837	0.5365
	3.82	0.3832	0.5367
	3.83	0.3827	0.5369
	3.84	0.3823	0.5371
	3.85	0.3818	0.5372
	3.86	0.3813	0.5374
	3.87	0.3808	0.5376
	3.88	0.3804	0.5377
	3.89	0.3799	0.5379
	3.90	0.3795	0.5381
	3.91	0.3790	0.5382
	3.92	0.3785	0.5384
	3.93	0.3781	0.5386
	3.94	0.3776	0.5387
	3.95	0.3772	0.5389
	3.96	0.3767	0.5391
	3.97	0.3763	0.5392
	3.98	0.3758	0.5394
	3.99	0.3754	0.5395
	4.00	0.3749	0.5397
	4.01	0.3745	0.5399
	4.02	0.3741	0.5400
	4.03	0.3736	0.5402
	4.04	0.3732	0.5403
	4.05	0.3728	0.5405
	4.06	0.3723	0.5406
	4.07	0.3719	0.5408
	4.08	0.3715	0.5409
	4.09	0.3710	0.5411
}\tableCorrSiS

\pgfplotstableread{
	sigma	corr	si
	0.01	1.0000	0.0833
	0.02	1.0000	0.0833
	0.03	1.0000	0.0833
	0.04	1.0000	0.0833
	0.05	1.0000	0.0833
	0.06	1.0000	0.0833
	0.07	1.0000	0.0833
	0.08	1.0000	0.0833
	0.09	1.0000	0.0833
	0.10	1.0000	0.0833
	0.11	1.0000	0.0833
	0.12	1.0000	0.0833
	0.13	1.0000	0.0833
	0.14	1.0000	0.0833
	0.15	1.0000	0.0833
	0.16	1.0000	0.0833
	0.17	1.0000	0.0833
	0.18	1.0000	0.0833
	0.19	1.0000	0.0833
	0.20	1.0000	0.0833
	0.21	1.0000	0.0833
	0.22	1.0000	0.0833
	0.23	1.0000	0.0833
	0.24	1.0000	0.0833
	0.25	1.0000	0.0832
	0.26	1.0000	0.0832
	0.27	1.0000	0.0830
	0.28	1.0000	0.0829
	0.29	1.0000	0.0826
	0.30	1.0000	0.0822
	0.31	1.0000	0.0817
	0.32	0.9999	0.0811
	0.33	0.9999	0.0802
	0.34	0.9998	0.0791
	0.35	0.9997	0.0776
	0.36	0.9995	0.0758
	0.37	0.9993	0.0736
	0.38	0.9990	0.0709
	0.39	0.9986	0.0677
	0.40	0.9980	0.0639
	0.41	0.9974	0.0594
	0.42	0.9965	0.0543
	0.43	0.9955	0.0485
	0.44	0.9942	0.0422
	0.45	0.9928	0.0354
	0.46	0.9911	0.0281
	0.47	0.9892	0.0204
	0.48	0.9870	0.0123
	0.49	0.9846	0.0039
	0.50	0.9819	0.0048
	0.51	0.9790	0.0136
	0.52	0.9758	0.0226
	0.53	0.9724	0.0317
	0.54	0.9687	0.0408
	0.55	0.9649	0.0499
	0.56	0.9608	0.0590
	0.57	0.9565	0.0679
	0.58	0.9520	0.0768
	0.59	0.9474	0.0855
	0.60	0.9426	0.0940
	0.61	0.9377	0.1024
	0.62	0.9327	0.1106
	0.63	0.9276	0.1185
	0.64	0.9224	0.1263
	0.65	0.9171	0.1339
	0.66	0.9118	0.1413
	0.67	0.9065	0.1484
	0.68	0.9011	0.1554
	0.69	0.8957	0.1621
	0.70	0.8903	0.1686
	0.71	0.8849	0.1749
	0.72	0.8796	0.1810
	0.73	0.8743	0.1870
	0.74	0.8690	0.1927
	0.75	0.8637	0.1982
	0.76	0.8585	0.2036
	0.77	0.8534	0.2088
	0.78	0.8483	0.2139
	0.79	0.8432	0.2187
	0.80	0.8383	0.2235
	0.81	0.8333	0.2281
	0.82	0.8285	0.2325
	0.83	0.8237	0.2368
	0.84	0.8190	0.2410
	0.85	0.8143	0.2451
	0.86	0.8098	0.2490
	0.87	0.8052	0.2529
	0.88	0.8008	0.2566
	0.89	0.7964	0.2603
	0.90	0.7921	0.2638
	0.91	0.7879	0.2672
	0.92	0.7837	0.2706
	0.93	0.7796	0.2738
	0.94	0.7755	0.2770
	0.95	0.7715	0.2801
	0.96	0.7676	0.2831
	0.97	0.7638	0.2860
	0.98	0.7599	0.2889
	0.99	0.7562	0.2917
	1.00	0.7525	0.2944
	1.01	0.7489	0.2971
	1.02	0.7453	0.2997
	1.03	0.7418	0.3022
	1.04	0.7383	0.3047
	1.05	0.7349	0.3072
	1.06	0.7315	0.3095
	1.07	0.7282	0.3119
	1.08	0.7249	0.3141
	1.09	0.7217	0.3164
	1.10	0.7185	0.3186
	1.11	0.7154	0.3207
	1.12	0.7123	0.3228
	1.13	0.7092	0.3249
	1.14	0.7062	0.3269
	1.15	0.7032	0.3289
	1.16	0.7003	0.3308
	1.17	0.6974	0.3327
	1.18	0.6946	0.3346
	1.19	0.6917	0.3364
	1.20	0.6890	0.3382
	1.21	0.6862	0.3400
	1.22	0.6835	0.3417
	1.23	0.6808	0.3434
	1.24	0.6782	0.3451
	1.25	0.6756	0.3468
	1.26	0.6730	0.3484
	1.27	0.6705	0.3500
	1.28	0.6680	0.3515
	1.29	0.6655	0.3531
	1.30	0.6630	0.3546
	1.31	0.6606	0.3561
	1.32	0.6582	0.3575
	1.33	0.6558	0.3590
	1.34	0.6535	0.3604
	1.35	0.6512	0.3618
	1.36	0.6489	0.3632
	1.37	0.6467	0.3645
	1.38	0.6444	0.3659
	1.39	0.6422	0.3672
	1.40	0.6400	0.3685
	1.41	0.6379	0.3698
	1.42	0.6357	0.3710
	1.43	0.6336	0.3723
	1.44	0.6315	0.3735
	1.45	0.6294	0.3747
	1.46	0.6274	0.3759
	1.47	0.6254	0.3771
	1.48	0.6234	0.3782
	1.49	0.6214	0.3794
	1.50	0.6194	0.3805
	1.51	0.6175	0.3816
	1.52	0.6156	0.3827
	1.53	0.6137	0.3838
	1.54	0.6118	0.3848
	1.55	0.6099	0.3859
	1.56	0.6081	0.3869
	1.57	0.6062	0.3880
	1.58	0.6044	0.3890
	1.59	0.6026	0.3900
	1.60	0.6008	0.3910
	1.61	0.5991	0.3919
	1.62	0.5973	0.3929
	1.63	0.5956	0.3939
	1.64	0.5939	0.3948
	1.65	0.5922	0.3957
	1.66	0.5905	0.3967
	1.67	0.5889	0.3976
	1.68	0.5872	0.3985
	1.69	0.5856	0.3994
	1.70	0.5839	0.4002
	1.71	0.5823	0.4011
	1.72	0.5808	0.4020
	1.73	0.5792	0.4028
	1.74	0.5776	0.4036
	1.75	0.5761	0.4045
	1.76	0.5745	0.4053
	1.77	0.5730	0.4061
	1.78	0.5715	0.4069
	1.79	0.5700	0.4077
	1.80	0.5685	0.4085
	1.81	0.5670	0.4093
	1.82	0.5656	0.4100
	1.83	0.5641	0.4108
	1.84	0.5627	0.4115
	1.85	0.5613	0.4123
	1.86	0.5598	0.4130
	1.87	0.5584	0.4138
	1.88	0.5571	0.4145
	1.89	0.5557	0.4152
	1.90	0.5543	0.4159
	1.91	0.5530	0.4166
	1.92	0.5516	0.4173
	1.93	0.5503	0.4180
	1.94	0.5489	0.4187
	1.95	0.5476	0.4194
	1.96	0.5463	0.4200
	1.97	0.5450	0.4207
	1.98	0.5437	0.4213
	1.99	0.5425	0.4220
	2.00	0.5412	0.4226
	2.01	0.5399	0.4233
	2.02	0.5387	0.4239
	2.03	0.5374	0.4245
	2.04	0.5362	0.4251
	2.05	0.5350	0.4258
	2.06	0.5338	0.4264
	2.07	0.5326	0.4270
	2.08	0.5314	0.4276
	2.09	0.5302	0.4282
	2.10	0.5290	0.4287
	2.11	0.5278	0.4293
	2.12	0.5267	0.4299
	2.13	0.5255	0.4305
	2.14	0.5243	0.4311
	2.15	0.5232	0.4316
	2.16	0.5221	0.4322
	2.17	0.5209	0.4327
	2.18	0.5198	0.4333
	2.19	0.5187	0.4338
	2.20	0.5176	0.4344
	2.21	0.5165	0.4349
	2.22	0.5154	0.4354
	2.23	0.5143	0.4360
	2.24	0.5133	0.4365
	2.25	0.5122	0.4370
	2.26	0.5111	0.4375
	2.27	0.5101	0.4380
	2.28	0.5090	0.4385
	2.29	0.5080	0.4390
	2.30	0.5070	0.4395
	2.31	0.5059	0.4400
	2.32	0.5049	0.4405
	2.33	0.5039	0.4410
	2.34	0.5029	0.4415
	2.35	0.5019	0.4420
	2.36	0.5009	0.4424
	2.37	0.4999	0.4429
	2.38	0.4989	0.4434
	2.39	0.4979	0.4438
	2.40	0.4969	0.4443
	2.41	0.4960	0.4448
	2.42	0.4950	0.4452
	2.43	0.4940	0.4457
	2.44	0.4931	0.4461
	2.45	0.4922	0.4466
	2.46	0.4912	0.4470
	2.47	0.4903	0.4474
	2.48	0.4893	0.4479
	2.49	0.4884	0.4483
	2.50	0.4875	0.4487
	2.51	0.4866	0.4492
	2.52	0.4857	0.4496
	2.53	0.4848	0.4500
	2.54	0.4839	0.4504
	2.55	0.4830	0.4508
	2.56	0.4821	0.4513
	2.57	0.4812	0.4517
	2.58	0.4803	0.4521
	2.59	0.4794	0.4525
	2.60	0.4786	0.4529
	2.61	0.4777	0.4533
	2.62	0.4768	0.4537
	2.63	0.4760	0.4541
	2.64	0.4751	0.4545
	2.65	0.4743	0.4548
	2.66	0.4734	0.4552
	2.67	0.4726	0.4556
	2.68	0.4718	0.4560
	2.69	0.4709	0.4564
	2.70	0.4701	0.4567
	2.71	0.4693	0.4571
	2.72	0.4685	0.4575
	2.73	0.4677	0.4579
	2.74	0.4669	0.4582
	2.75	0.4661	0.4586
	2.76	0.4653	0.4589
	2.77	0.4645	0.4593
	2.78	0.4637	0.4597
	2.79	0.4629	0.4600
	2.80	0.4621	0.4604
	2.81	0.4613	0.4607
	2.82	0.4605	0.4611
	2.83	0.4598	0.4614
	2.84	0.4590	0.4617
	2.85	0.4582	0.4621
	2.86	0.4575	0.4624
	2.87	0.4567	0.4628
	2.88	0.4559	0.4631
	2.89	0.4552	0.4634
	2.90	0.4544	0.4638
	2.91	0.4537	0.4641
	2.92	0.4530	0.4644
	2.93	0.4522	0.4648
	2.94	0.4515	0.4651
	2.95	0.4508	0.4654
	2.96	0.4500	0.4657
	2.97	0.4493	0.4660
	2.98	0.4486	0.4663
	2.99	0.4479	0.4667
	3.00	0.4472	0.4670
	3.01	0.4465	0.4673
	3.02	0.4458	0.4676
	3.03	0.4450	0.4679
	3.04	0.4443	0.4682
	3.05	0.4437	0.4685
	3.06	0.4430	0.4688
	3.07	0.4423	0.4691
	3.08	0.4416	0.4694
	3.09	0.4409	0.4697
	3.10	0.4402	0.4700
	3.11	0.4395	0.4703
	3.12	0.4389	0.4706
	3.13	0.4382	0.4709
	3.14	0.4375	0.4712
	3.15	0.4369	0.4715
	3.16	0.4362	0.4717
	3.17	0.4355	0.4720
	3.18	0.4349	0.4723
	3.19	0.4342	0.4726
	3.20	0.4336	0.4729
	3.21	0.4329	0.4732
	3.22	0.4323	0.4734
	3.23	0.4316	0.4737
	3.24	0.4310	0.4740
	3.25	0.4303	0.4743
	3.26	0.4297	0.4745
	3.27	0.4291	0.4748
	3.28	0.4285	0.4751
	3.29	0.4278	0.4753
	3.30	0.4272	0.4756
	3.31	0.4266	0.4759
	3.32	0.4260	0.4761
	3.33	0.4253	0.4764
	3.34	0.4247	0.4766
	3.35	0.4241	0.4769
	3.36	0.4235	0.4772
	3.37	0.4229	0.4774
	3.38	0.4223	0.4777
	3.39	0.4217	0.4779
	3.40	0.4211	0.4782
	3.41	0.4205	0.4784
	3.42	0.4199	0.4787
	3.43	0.4193	0.4789
	3.44	0.4187	0.4792
	3.45	0.4181	0.4794
	3.46	0.4175	0.4797
	3.47	0.4170	0.4799
	3.48	0.4164	0.4802
	3.49	0.4158	0.4804
	3.50	0.4152	0.4806
	3.51	0.4147	0.4809
	3.52	0.4141	0.4811
	3.53	0.4135	0.4813
	3.54	0.4129	0.4816
	3.55	0.4124	0.4818
	3.56	0.4118	0.4821
	3.57	0.4113	0.4823
	3.58	0.4107	0.4825
	3.59	0.4101	0.4827
	3.60	0.4096	0.4830
	3.61	0.4090	0.4832
	3.62	0.4085	0.4834
	3.63	0.4079	0.4837
	3.64	0.4074	0.4839
	3.65	0.4068	0.4841
	3.66	0.4063	0.4843
	3.67	0.4058	0.4846
	3.68	0.4052	0.4848
	3.69	0.4047	0.4850
	3.70	0.4042	0.4852
	3.71	0.4036	0.4854
	3.72	0.4031	0.4857
	3.73	0.4026	0.4859
	3.74	0.4020	0.4861
	3.75	0.4015	0.4863
	3.76	0.4010	0.4865
	3.77	0.4005	0.4867
	3.78	0.3999	0.4869
	3.79	0.3994	0.4871
	3.80	0.3989	0.4874
	3.81	0.3984	0.4876
	3.82	0.3979	0.4878
	3.83	0.3974	0.4880
	3.84	0.3969	0.4882
	3.85	0.3964	0.4884
	3.86	0.3959	0.4886
	3.87	0.3954	0.4888
	3.88	0.3949	0.4890
	3.89	0.3944	0.4892
	3.90	0.3939	0.4894
	3.91	0.3934	0.4896
	3.92	0.3929	0.4898
	3.93	0.3924	0.4900
	3.94	0.3919	0.4902
	3.95	0.3914	0.4904
	3.96	0.3909	0.4906
	3.97	0.3904	0.4908
	3.98	0.3899	0.4910
	3.99	0.3895	0.4912
	4.00	0.3890	0.4914
	4.01	0.3885	0.4916
	4.02	0.3880	0.4917
	4.03	0.3875	0.4919
	4.04	0.3871	0.4921
	4.05	0.3866	0.4923
	4.06	0.3861	0.4925
	4.07	0.3857	0.4927
	4.08	0.3852	0.4929
	4.09	0.3847	0.4931
}\tableCorrSiT

\begin{document}
	\bstctlcite{IEEEexample:BSTcontrol}
	%
	
	\title{{Intrapersonal Parameter Optimization for Offline Handwritten Signature Augmentation}}
	
	
	%
	%
	%
	
	\author{Teruo~M.~Maruyama,
		Luiz~S.~Oliveira,
		Alceu~S.~Britto~Jr.,
		and~Robert~Sabourin,~\IEEEmembership{Member~IEEE}
		
		\thanks{
			The authors would like to thank Prof. Diego Bertolini of the Federal University of Technology - Paraná, Prof. Daniel Weingaertner of the Federal University of Paraná; Luiz Gustavo Hafemann and Rafael Menelau Oliveira e Cruz of École de Technologie Supérieure. This study was financed in part by the Coordenação de Aperfeiçoamento de Pessoal de Nível Superior - Brazil (CAPES) - Finance Code 001, and by the Government of Canada.}
		
		\thanks{T. M. Maruyama is with the Department of Informatics, Federal University of Paraná, Curitiba, 81531-990, Brazil (e-mail: shinigam8@gmail.com).}
		
		\thanks{L. S. Oliveira is with the Department of Informatics, Federal University of Paraná, Curitiba, 81531-990, Brazil (e-mail: luiz.oliveira@ufpr.br).}
		
		\thanks{A. S. Britto Jr. is with the Programa de Pós-Graduação em Informática (PPGIa), Pontifícia Universidade Católica do Paraná (PUCPR), Curitiba, 80215-901, Brazil (e-mail: alceu@ppgia.pucpr.br).}%
		
		\thanks{R. Sabourin is with the Laboratoire d'imagerie, de vision et d'inteligence artificielle, École de technologie supérieure, Université du Québec, Montreal, QC H3C 1K3, Canada (email: robert.sabourin@etsmtl.ca).}
	}
	
	%
	%
	
	\makeatletter
	\def\ps@IEEEtitlepagestyle{
		\def\@oddfoot{\mycopyrightnotice}
		\def\@evenfoot{}
	}
	\def\mycopyrightnotice{
		{\footnotesize
			\begin{minipage}{\textwidth}
				\centering
				\copyright2020 IEEE. Personal use of this material is permitted. However, republication/redistribution requires IEEE permission.
			\end{minipage}
		}
	}

	\markboth{}
	{Maruyama \MakeLowercase{\textit{et al.}}: Intrapersonal Parameter Optimization for Offline Handwritten Signature Augmentation}
	%



	\maketitle
	
	\begin{abstract}			
		Usually, in a real-world scenario, few signature samples are available to train an automatic signature verification system (ASVS). However, such systems do indeed need a lot of signatures to achieve an acceptable performance. Neuromotor signature duplication methods and feature space augmentation methods may be used to meet the need for an increase in the number of samples. Such techniques manually or empirically define a set of parameters to introduce a degree of writer variability.
		Therefore, in the present study, a method to automatically model the most common writer variability traits is proposed. The method is used to generate offline signatures in the image and the feature space and train an ASVS. We also introduce an alternative approach to evaluate the quality of samples considering their feature vectors. We evaluated the performance of an ASVS with the generated samples using three well-known offline signature datasets: GPDS, MCYT-75, and CEDAR. In GPDS-300, when the SVM classifier was trained using one genuine signature per writer and the duplicates generated in the image space, the Equal Error Rate (EER) decreased from 5.71\% to 1.08\%. Under the same conditions, the EER decreased to 1.04\% using the feature space augmentation technique.
		We also verified that the model that generates duplicates in the image space reproduces the most common writer variability traits in the three different datasets.
	\end{abstract}
	

	%
	\IEEEpeerreviewmaketitle

	\section{Introduction}
	%
	%
	%
	%
	\IEEEPARstart{T}{he} term \textit{signature} is derived from the Latin word \textit{signare} which means to put a mark, and it may be considered a special type of handwriting for several reasons. Generally, the signature is a handwritten representation of a person's name, which may be presented in several formats, ranging from a simple abbreviation to the complete name. It can also have a legible or a flourished format \cite[p.~309]{Mitra2017} \cite[p.~87]{Allen2016}. Furthermore, each writer has an individual behavior when signing. Several factors, such as culture \cite[p.~1368]{Li2015}, handwriting skill, age \cite{Tolosana2019}, health \cite{Impedovo2019}, and the physical and emotional state \cite[p.~26-27]{Koppenhaver2008} can affect this behavior. Due to this individual behavior, signatures produced by the same writer never have the exact same visual appearance. This is called intra-personal variability or writer variability. Since the handwritten signatures of two writers are never exactly alike either, their signatures can be used to distinguish individuals \cite[p.~7]{Koppenhaver2008}. This is known as interpersonal variability \cite{Diaz2019}. 
	
	Thanks to interpersonal variability, handwritten signatures are widely used and accepted in verifying a person's identity in legal, administrative, and financial endeavors \cite{Hafemann2018}. Even with the technological advancements that have occurred in the last few decades, handwritten signatures are still the most frequently used type of handwriting among certain writers \cite[p.~64]{Allen2016} \cite{Linden2018}.
	
	
	Like other behavioral biometric traits, the handwritten signature cannot be lost, stolen or forgotten \cite{Jain2016}. However, signatures can be forged \cite[p.~55]{Koppenhaver2008}. Forgeries can be classified as random, simple, or skilled. A forgery is considered \textit{random} when the forger does not know the target's name and they use their own signature instead. When the forger knows only the target's name, the forgery is \textit{simple}. The forgery is considered \textit{skilled} when the forger has access to the target's signature and trains to reproduce it. When a forger is trying to reproduce a signature, they try to copy the speed, pressure, and other visual features of the genuine signature. Consequently, the forger tries to reproduce the writer variability of their target \cite[p. ~204]{Zhang2000} \cite{Malik2012,Zois2019}.
	Given these elements, modeling the writer variability is a complex and challenging task.
	
	When discussing automatic handwritten signature verification \cite{Plamondon1989, Impedovo2008, Hafemann2017b, Diaz2019} two different forms of acquisition are considered: online (dynamic) and offline (static). {Online} signatures are acquired using a digital device, such as a digitizing table, and are represented as a sequence of values over time. These values can represent the pressure, speed, coordinate of the pen, etc. \cite{Diaz2019}. Offline signatures are acquired after signatures have been written on a piece of paper. The image of the signature is digitized using a scanner or a digital camera. Unlike online signatures, offline signatures are represented by color or gray level data \cite{Hafemann2017b}. This lack of complementary features makes it even more difficult to identify offline signature falsifications \cite{Hafemann2018}.
	
	Contributing to increase the difficulty of offline signature verification, in the real-world scenario, the number of genuine signature samples per writer is usually too small to train a machine learning model for automatic handwritten signature verification.
	One alternative is to use data augmentation techniques to increase the number of signatures in the feature and image space.
	When this process is performed in the image space, it is known as signature duplication \cite{Song2014}.
	Several signature duplication techniques have been proposed in the literature \cite{Diaz2017,Fang2002,Ferrer2013b,Frias2006,Galbally2009,Huang1997,Rabasse2008,Song2014}.
	In particular, techniques based on human behavior present more realistic duplicates than other approaches  \cite{Diaz2017,Ferrer2013b,Galbally2009}.
	Among these techniques, the method proposed by Diaz et al. \cite{Diaz2017} must be highlighted.
	
	In their work, Diaz et al. \cite{Diaz2017} argue that writer variability can be described by a global set of parameters. Even though each writer has a different behavior when they sign, some common behaviors can be shared by different writers during the writing process. Therefore, these common variability traits can be modeled by a global set of parameters. To support their hypothesis, the authors performed experiments on two different datasets and alphabetical systems using a set of predefined parameters. However, the parameters were defined empirically, and were thus not based on real writer variability. The parameters were manually optimized based on two factors: human-like aspect and the performance of an automatic signature verification system. The parameters that produced the most human-like duplicates and the best performance with the automatic signature verification system were selected.
	
	Most data augmentation techniques in the feature space have low computational complexity and are simple to implement. Some of them require at least two or three samples to increase the number of training samples \cite{Kumar2019}. Among the simple and low computational complexity techniques, the application of a Gaussian filter does not require the use of more than one sample. Nevertheless, it still needs to define a parameter to be applied, despite its simplicity. Generally, this parameter is determined manually \cite {Schluter2015} \cite{DeVries2017} \cite{Kumar2019}.  

	In this work, we advocate that writer variability can be better modeled based on real data. With that in mind, the main contributions of this work are: 1) a method to automatically model writer variability based on offline signatures; 2) for the first time, we present a method to generate synthetic offline signature samples in the feature space using a Gaussian filter; 3) we show how our model can be used to generate more realistic offline signature samples in the image and in the feature spaces; and 4) we propose a new approach to validate the writer variability of synthetic signature samples. Instead of selecting parameters manually, we adopt a real parameter black-box optimization strategy. As a consequence, we can build more robust signature verification systems using very few genuine signatures. Considering a discriminant feature descriptor, we hypothesize that the writer variability observed on the image space can be reflected in the feature space. Our method thus tries to model the writer variability, while considering just the feature space. Therefore, the visual appearance of our duplicates is not directly assessed in this work.
	
	To support our claims, we conducted experiments based on three well-known benchmarks, GPDS-960, CEDAR, and MCYT-75. Considering a real-world scenario, no more than three genuine signatures per writer were used to train the verification system. Duplicates generated using the default parameters proposed by Diaz et al. \cite{Diaz2017} and synthetic samples with our model were also compared. When the ASVS used the duplicates generated using the default parameters, it achieved EERs of 0.83, 0.70, and 3.04, for MCYT, GPDS, and CEDAR, respectively. When it used the duplicates generated by our method, it achieved EERs of 0.07, 0.24, and 2.16, for MCYT, GPDS, and CEDAR, respectively. When it used the synthetic feature vectors generated by our method, it achieved the lowest EERs of 0.01, 0.20, and 0.82, for MCYT, GPDS, and CEDAR, respectively. Thus, despite using fewer genuine signatures per writer than other state-of-the-art signature verification systems, the proposed method outperformed them. The model proved capable of generalizing the most common writer variability traits that are modeled using a GPDS subset, on different datasets. With the proposed method, automatic handwritten signature verification systems can even more closely approximate the real-world scenario.
		
	The remainder of this paper is organized as follows: Section \ref{sec:related_works} reviews related works on handwritten signature duplication and {feature space augmentation}. Section \ref{sec:method} describes the proposed method for modeling writer variability and associated concepts. In addition, it shows how the proposed method can be used by an automatic signature verification system. Section \ref{sec:results} shows how writer variability can be evaluated, considering signature features and the performance of a verification system. Lastly, Section \ref{sec:conclusion} presents the conclusions.

	\section{Related {Works}}
	\label{sec:related_works}
	
	Over the last few decades, many advances have been presented in signature verification literature, which have been covered by some key survey papers \cite{Impedovo2008,Plamondon1989}. More recently, Hafemann et al. \cite{Hafemann2017b} and Diaz et al. \cite{Diaz2019} have
	discussed new trends, such as the use of deep learning techniques applied to handwritten signatures. Such methods have achieved superior results in multiple benchmarks. Readers interested in the state of the art of signature verification systems, should please refer to these works and to \cite{Zois2019b, Yilmaz2019, Souza2020, Diaz2020}. In this section, we cover the core of our work, i.e., signature duplication methods and data augmentation methods in the feature space.
	
	\subsection{Signature Duplication Methods}
	\label{subsec:duplication_methods}
	
	Signature duplication methods consist of algorithms used to generate new artificial signatures, with one or more signatures used as seeds \cite{Diaz2017}. Duplication methods can be divided into: i) the creation of online (dynamic) signatures using real online samples  \cite{Diaz2018,Diaz2015,Ferrer2018,Ferrer2017,Galbally2012b,Galbally2012a,Galbally2009,Munich2003,Song2014}; ii) the creation of offline (static) signatures using real online samples \cite{Diaz2014a,Diaz2014b,Ferrer2013b,Galbally2015,Melo2019,Rabasse2008}; iii) the creation of offline signatures using real offline samples \cite{Diaz2016a,Diaz2017,Diaz2016b,Fang2002,Ferrer2013a,Ferrer2015,Frias2006,Huang1997,Ruiz2019}; and iv) the creation of online signatures by using real offline samples. Despite recent advances in the recovery of dynamic signatures from static signatures \cite{Diaz2017b,Lau2005,Nel2005,Nguyen2010}, the last type of method still an open issue \cite{Diaz2019}. 
	
\begin{figure}[!t]
		\centering
		\includegraphics[width=1\linewidth]{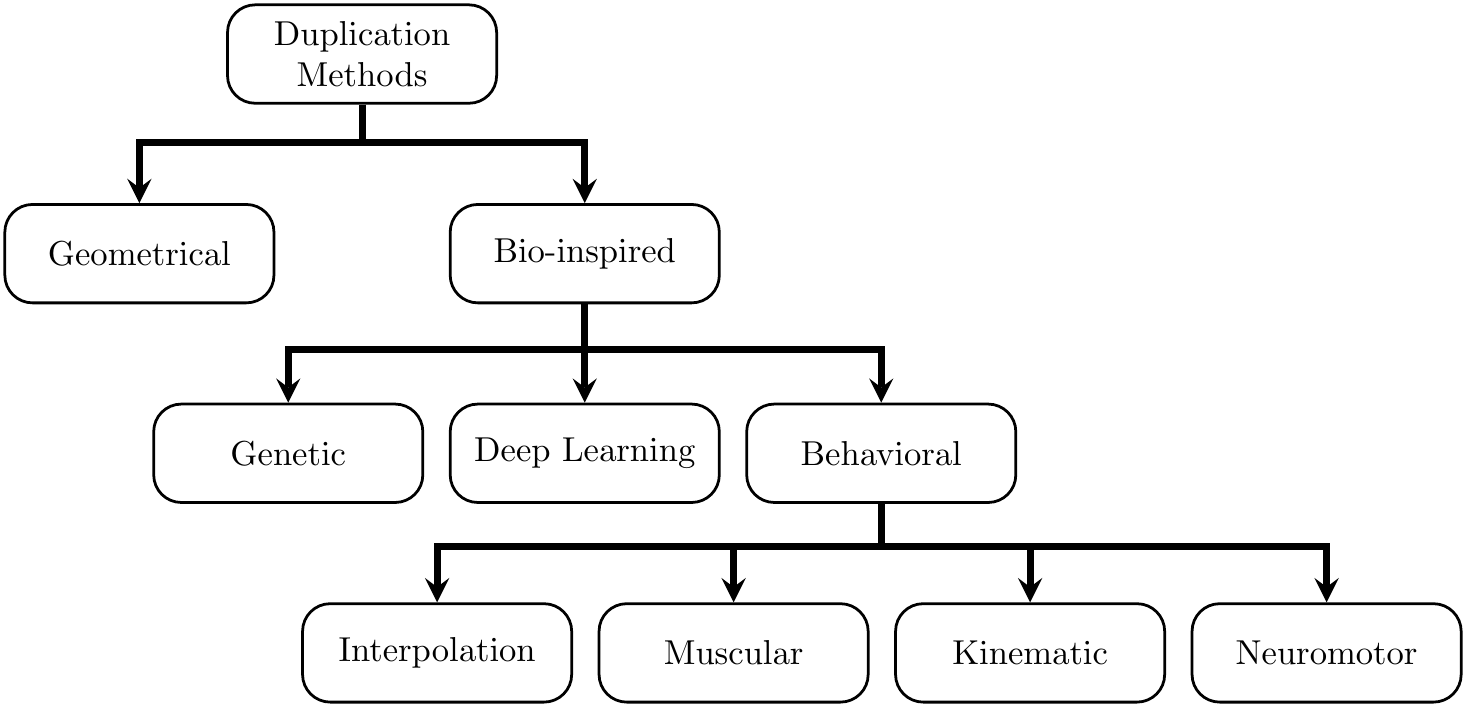}
		\caption{Taxonomy of the duplication methods.}
		\label{fig:taxonomy}
	\end{figure}

	Duplication methods can still be classified according to the approach used to create new signatures, which in turn can be based on geometrical transformations or be bio-inspired. Figure \ref{fig:taxonomy} depicts the taxonomy of duplication methods.
		
	Methods based on geometrical transformations usually employ rotation \cite{Frias2006,Huang1997,Ruiz2019}, scaling \cite{Frias2006,Huang1997,Munich2003}, a perspective view \cite{Huang1997}, displacement \cite{Fang2002,Frias2006,Munich2003}, and warping \cite{Rabasse2008} to increase the number of signatures. They can add some natural and unnatural distortions, which may be used to create artificial genuine signatures \cite{Frias2006,Galbally2009,Huang1997,Munich2003,Rabasse2008} and artificial signature forgeries \cite{Huang1997,Ruiz2019}, respectively. As pointed out by Ruiz et al. \cite{Ruiz2019}, signatures with unnatural distortions can be used as genuine or random forgeries to improve the performance of verification systems. One drawback of geometrical transformations is that they can create duplicates that {are not necessarily visually} similar to genuine signatures \cite{Diaz2017}. Even though duplication methods based on geometrical transformations can improve the performance of signature verification systems, they neglect an important aspect of handwriting, namely, the writer's behavior. To bridge this gap, several bio-inspired methods have been proposed in the literature, and can be divided into three main categories, namely, Genetic Algorithms, Deep Learning and Behavioral approaches.
	
	Song and Sun \cite{Song2014} proposed a method to duplicate dynamic signatures based on genetic algorithms. Firstly, a set of duplicates with successive stroke sections are generated. These duplicates must be similar to some sections of the input signature. Then, the input signatures are used to generate a set of duplicates with some different strokes. Both sets are then combined to generate new duplicates. The resulting population is iteratively updated by a cloning-mutation-selection operation. Although it does consider the variability of input signatures, this bio-inspired method does not provide the signature variability that needs to be maintained during the generation of duplicates.
	
	With regard to deep learning, Melo et al. \cite{Melo2019} proposed a model to generate offline signatures based on online signatures that learn to map online signatures to offline signatures. The method requires that both types of signatures be trained. However, most datasets do not have both types of signatures available to perform this task. The acquisition of an online signature and the corresponding offline signatures requires several precautions and resources. Despite promising results, deep learning models require a great amount of data to be trained. 
	As this requirement could not possibly be met for their specific case, the authors used a dataset of 23,000 mapped online and offline words.
	However, all the words had been written by the same writer. Therefore, the trained model did not consider the individual variabilities of different writers.

	The third category of the bio-inspired methods tries to mimic the behavior of a person when he/she is writing. In this context, different approaches, such as interpolation, muscular, kinematic, and neuromotor approaches have been used to underpin the proposed methods.
	
	Since dynamic signature data is related to writer behavior, some works use this type of data to generate static signatures \cite{Diaz2014b, Galbally2015}. When dynamic information about the signature trajectory is available, it can be interpolated to generate new static signature duplicates. In order to enhance the reliability of this kind of duplicate, information relating to pressure and speed is also used. The reliability of this kind of approach depends strongly on the choice of the interpolation algorithm \cite{Diaz2014b}.   
	
	Although the handwriting process is still not fully understood, some duplication methods try to model human behavior during a signing act. 
	As mentioned above, some of them are based on muscular models, and others on kinematic theory. The methods based on muscular models try to reproduce the trajectory of signature strokes when the writer is moving the muscles \cite{Morasso1982}. For their part, those based on kinematic theory try to reproduce the writer's muscular speed effects on the handwritten signature \cite{Diaz2015, Diaz2018, Ferrer2013b,Galbally2012a,Galbally2012b}, which thus necessitates the presence of online signatures with velocity and pressure information. Methods based on muscular models and kinematic theory are mainly designed to duplicate flourished signatures. Therefore, for legible signatures, the methods can present signature duplicates that are not human-like \cite{Ferrer2013a,Galbally2012b}.
	
	The fourth behavioral class category is neuromotor theory \cite{Diaz2014a, Ferrer2015, Diaz2016a, Diaz2016b, Diaz2017, Ferrer2017, Ferrer2018}, which considers a set of muscles, skeletal parts, eyes, and the central nervous system in creating signatures. In this case, the brain stores the signature generation plan and sends electric impulses to the eyes and to specific muscles, allowing the signature plan to be executed by a series of specific muscular contractions and articulatory movements \cite{Ferrer2015}.
	
	Most neuromotor methods are not concerned with stroke sequences. However, some of them are mainly focused on the signature trajectory plan \cite{Diaz2016a,Ferrer2018}. This plan determines the distribution and sequence of strokes which are used to create a new signature. Generally, the sequence information is only present in dynamic signatures. However, Djioua and Plamondon (2009) \cite{Djioua2009}, introduce an algorithm to extract the speed information from an offline signature.  Such an algorithm was successfully used by Ferrer et al. \cite{Ferrer2015} to create a trajectory plan. Furthermore, Ferrer et al. \cite{Ferrer2018} showed that the trajectory plan and complexity of a method depend on the alphabet used to sign.
	
	To simulate the desired motor effects, most neuromotor methods use a grid to map the distribution of strokes or characters on the written surface. Different architectures (hexagonal \cite{Diaz2016a,Ferrer2018}, quadrangular \cite{Ferrer2015}, and sinusoidal \cite{Diaz2017}) have been investigated for generating the desired deformations. In \cite{Ferrer2018}, the authors showed that the grid density is directly related to the alphabet used to sign. While dense grids are indicated for eastern alphabets such as Bengali and Devanagari, sparse grids are more suitable for western alphabets such as Latin. Diaz et al. \cite{Diaz2017}, on the other hand, showed that the sinusoidal grid produces good duplicates for signatures in Devanagari, Bengali, and Latin. Two other aspects that play an important role when duplicating a handwritten signature are the paper and ink used. In that regard, some methods try to reproduce successive ink depositions on the paper and the roughness of the paper used \cite[p.~195-197]{Koppenhaver2008}\cite{Diaz2016b,Ferrer2013b,Galbally2015}.
	
	Similar to methods based on geometrical transformations, behavior-based methods also use several parameters to control writer variability, which is generally defined empirically. Defining such parameters is time-consuming, and very often produces values that do not describe the real writer variability. Table \ref{tab:dup_methods} summarizes the duplication methods showing the reference, type of input and output of the method, the strategy used by each method, whether or not the method considers the writer variability (V), and the signature alphabet.
	
	\begin{table*} [!t]
		\setlength{\tabcolsep}{5pt}
		\renewcommand\arraystretch{1.05}
		\caption {Signature duplication methods (On-2-Off: from real online to duplicated offline signature; On-2-On: from real online to duplicated online signature; Off-2-Off: from real offline to duplicated offline signature; Off-2-On: from real offline to duplicated offline signature; If the method considers the writer variability, it is marked with a X.)}
		\label{tab:dup_methods}
		\centering
		\begin{tabular}{lllll} \hline 
			\multirow{2}{*}{\textbf{Reference}} &
			\multirow{2}{*}{\textbf{Conversion}} &
			\multirow{2}{*}{\textbf{Method}} & 
			{\textbf{Writer}} & 
			\multirow{2}{*}{\textbf{Alphabet}}\\
			{\textbf{}} &
			{\textbf{}} &
			{\textbf{}} & 
			{\textbf{Variability}} & 
			{\textbf{}}\\
			\hline
			{Rabasse et al., 2008 \cite{Rabasse2008}} & {On-2-Off} & {Affine\textbackslash Geometrical Transformations} & {X} & {Latin\textbackslash {Chinese}}\\
			{Ferrer et al., 2013b \cite{Ferrer2013b}} & {On-2-Off} & {Spectral Analysis\textbackslash}{Kinematic Theory\textbackslash}{Paper and Ink Model}  & {-} & {Latin}\\
			{Diaz et al., 2014a \cite{Diaz2014a}} & {On-2-Off} & {Neuromotor Inspired Model\textbackslash}{Ink Model} & {-} & {Latin}\\
			{Diaz et al., 2014b \cite{Diaz2014b}} & {On-2-Off} & {Interpolation\textbackslash}{Ink Model} & {-} & {Latin}\\
			{Galbally et al., 2015 \cite{Galbally2015}} & {On-2-Off} & {Interpolation\textbackslash}{Ink Model} & {X} & {Latin}\\
			%
			{Melo et al., 2019 \cite{Melo2019}} & {On-2-Off} & {Deep Learning Model} & {-} & {Latin}\\
			{Munich and Perona, 2003 \cite{Munich2003}} & {On-2-On} & {Time Origin Translation\textbackslash}{Affine-scale} & {-} & {Latin}\\
			{Galbally et al., 2009 \cite{Galbally2009}} & {On-2-On} & {Affine\textbackslash Geometrical Transformations}& {X} & {Latin}\\
			{Galbally et al., 2012 \cite{Galbally2012a}\cite{Galbally2012b}} & {On-2-On} & {Spectral Analysis\textbackslash Kinematic Theory} & {X} & {Latin}\\
			{Song and Sun, 2014 \cite{Song2014}} & {On-2-On} & {Clonal Selection Algorithm} & {X} & {Latin\textbackslash Chinese}\\
			{Diaz et al., 2015 \cite{Diaz2015}} & {On-2-On} & {Kinematic Theory} & {X} & {Latin}\\
			{Diaz et al., 2018 \cite{Diaz2018}} & {On-2-On} & {Kinematic Theory} & {X} & {Latin}\\
			{Ferrer et al., 2017 \cite{Ferrer2017}} & {On-2-On\textbackslash}{Off-2-Off} & {Neuromotor Inspired Model} & {X} & {Latin\textbackslash Chinese}\\
			{Ferrer et al., 2018 \cite{Ferrer2018}} & {On-2-On\textbackslash {Off-2-Off}} & {Neuromotor Inspired Model\textbackslash Ink Model} & {X} & {Bengali\textbackslash}{Devanagari}\\
			{Huang and Yan, 1997 \cite{Huang1997}} & {Off-2-Off} & {Affine\textbackslash Geometrical Transformations} & {-} & {Latin\textbackslash Chinese}\\
			{Fang et al, 2002 \cite{Fang2002}} & {Off-2-Off} & {Elastic Matching} & {-} & {Latin}\\
			{Frias-Martinez et al., 2006 \cite{Frias2006}} & {Off-2-Off} & {Affine\textbackslash Geometrical Transformations} & {X} & {Latin}\\
			{Ferrer et al., 2013a \cite{Ferrer2013a}} & {Off-2-Off} & {Active Shape Model\textbackslash}{Muscular Model\textbackslash}{Ink Model} & {X} & {Latin}\\
			{Ferrer et al., 2015 \cite{Ferrer2015}} & {Off-2-Off} & {Neuromotor Inspired Model\textbackslash}{Ink Model} & {X} & {Latin}\\
			{Diaz et al., 2016a \cite{Diaz2016a}} & {Off-2-Off} & {Neuromotor Inspired Model\textbackslash Ink Model} & {X} & Bengali\\
			{Diaz et al., 2016b \cite{Diaz2016b}} & {Off-2-Off} & {Neuromotor Inspired Model\textbackslash Ink Model} & {X} & {Bengali\textbackslash Devanagari}\\
			{Diaz et al., 2017 \cite{Diaz2017}} & {Off-2-Off} & {Neuromotor Inspired Model\textbackslash Ink Model} & {X} & {Latin}\\
			{Ruiz et al., 2019 \cite{Ruiz2019}} & {Off-2-Off} & {Geometrical\textbackslash Morphological Transformations\textbackslash} {Noise Addition} & {-} & {Latin}\\
			{Open Issue} & {Off-2-On} & {-} & {-} & {-}\\ 
			\hline	
			
		\end{tabular}
	\end{table*}
	The literature shows that neuromotor-based methods produce better duplicates than models based on geometrical transformations \cite{Diaz2014a, Diaz2017,Ferrer2017,Ferrer2018}. Among the neuromotor-based methods, we highlight the one proposed by Diaz et al. \cite{Diaz2017}, which uses an ink model to produce high quality duplicates. The duplicates created with this method were used to train a signature verification system and outperformed the system trained with the duplicates created by the geometrical transformation proposed in \cite{Frias2006} by a fair margin. Furthermore, it proved to be robust using signatures of different alphabet systems such as Latin, Bengali, and Devanagari. Notwithstanding the good results presented by the authors in \cite{Diaz2017}, the problem of properly defining the parameters that control the generation of duplicates remains unsolved. In this paper, we aim to bridge this gap by proposing a real parameter black-box optimization strategy.
	
	\subsection{{Data Augmentation Methods in the Feature Space}}
		
	Instead of increasing the number of image samples, some techniques such as linear delta \cite{Kumar2019}, interpolation \cite{Chawla2002} \cite{Bunkhumpornpat2012} \cite{DeVries2017}, extrapolation \cite{DeVries2017}, delta-encoder \cite{Kumar2019}, and application of random noise \cite{Schluter2015} \cite{Kurata2016} \cite{DeVries2017} \cite{Kumar2019} \cite{Teng2020} increase the number of samples in the feature space. Unlike image space augmentation, it is very difficult to interpret the synthetic feature vectors \cite{Shorten2019}. Despite this, most feature space augmentation techniques have low computational complexity and are easy to implement \cite{Kumar2019}. 
	
	The linear delta technique finds the difference between the same class samples and uses it to generate a new feature vector. First, the difference between two feature vectors is computed, and is added to a third feature vector of the same class. The delta-encoder extends the linear delta concept using an autoencoder-based model to learn the differences between pairs of samples. First, it computes the intra-class deformations between pairs of training examples. Subsequently, it generates synthetic feature vectors applying the learned transformations to the original feature vectors \cite{Kumar2019}. 
	
	The interpolation strategy uses two feature vectors with the same label to generate a new one. For each feature vector, the K nearest neighbors with the same label in feature space are computed. For each pair of neighboring vectors $V_k$ and $V_j$, a new vector $V'$ can be generated using Equation \ref{eq:interpolation}. The degree of interpolation $\lambda$ is controlled using an interval from 0 to 1 \cite{DeVries2017}.
	
	\begin{equation}
	{V'} = \left({V}_{k} - {V}_{j}\right)\lambda + V_{j}
	\label{eq:interpolation}
	\end{equation}
	
	Similar to interpolation, extrapolation can be applied to a vector $V_{j}$ using Equation \ref{eq:extrapolation}. In this case, the degree of extrapolation $\lambda$ is controlled using an interval from 0 to $\infty$ \cite{DeVries2017} \cite{Kumar2019}.    
	
	\begin{equation}
	{{V'}_{j}} = \left({V}_{j} - {V}_{k}\right)\lambda + V_{j}
	\label{eq:extrapolation}
	\end{equation}
	
	Despite the simplicity of interpolation, extrapolation, and linear delta techniques, these strategies need a least two or three feature vectors to generate a new one \cite{Chawla2002} \cite{DeVries2017} \cite{Kumar2019}. In some cases, only one sample is available \cite{Diaz2017}. Therefore, in this specific case, these strategies cannot be used.
	
	Another simple way the number of training feature vectors are increased is by applying random noise to them \cite{Schluter2015}. The one-dimensional Gaussian filter is widely used to apply random noise in the feature space \cite {Schluter2015} \cite{DeVries2017} \cite{Kumar2019}. Unlike the extrapolation, interpolation, and linear delta strategies, random noise can be used with only one sample \cite{DeVries2017}. Furthermore, it has low computational complexity, and is easier to implement than the delta-encoders \cite{Kumar2019}.
	
	To the best of our knowledge, none of these techniques have previously been used to increase the number of offline signatures in the feature space. Therefore, we also propose a method to increase the number of offline signatures in the feature space based on writer variability.
		
	
	\section{The Proposed Method}
	\label{sec:method}
	
\begin{figure*}[!t]
		\centering
		\includegraphics[width=0.825\linewidth]{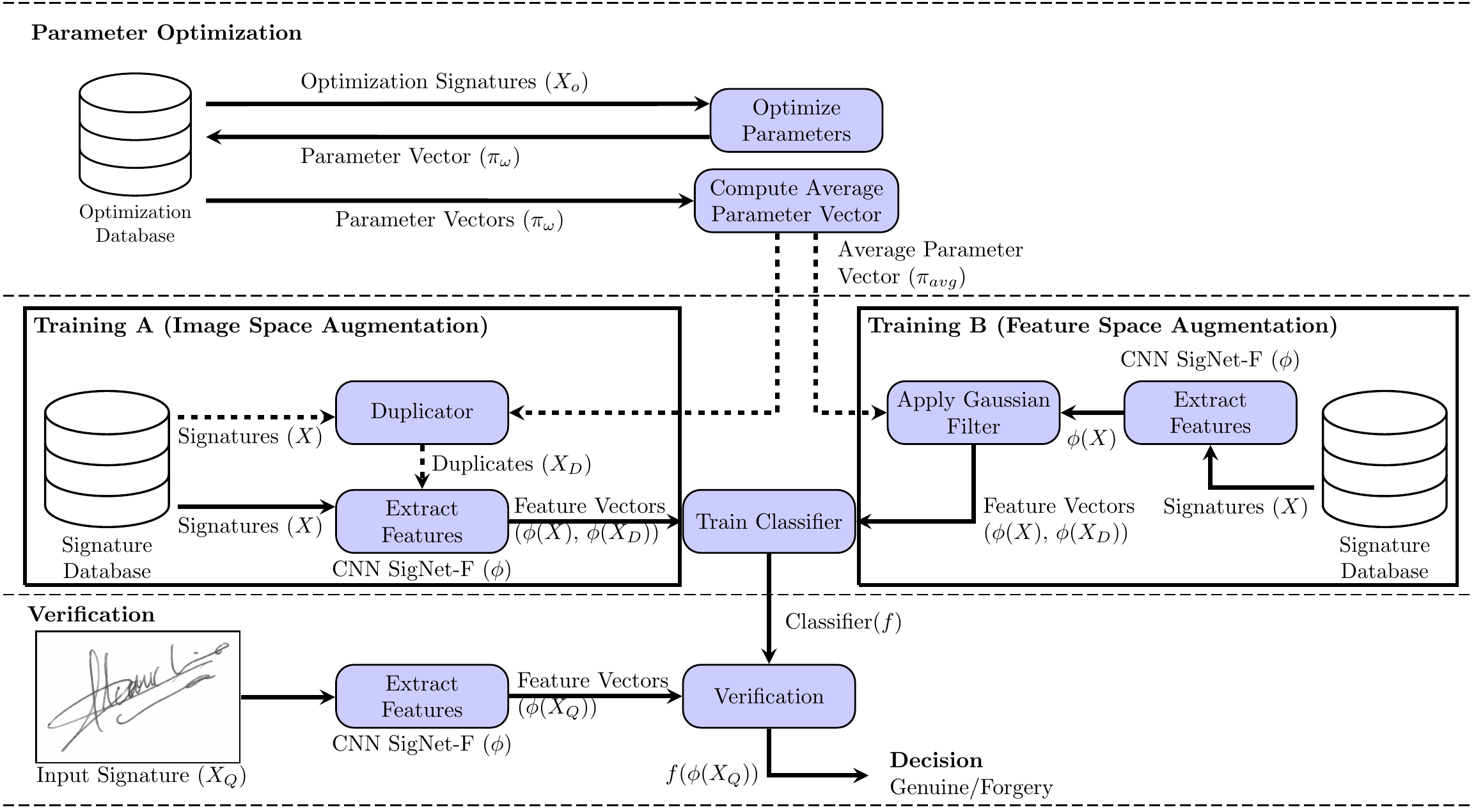}
		\caption{Offline signature verification system using the proposed method. {Training A is used only for data augmentation in the image space, while Training B is used only for data augmentation in the feature space.}}
		\label{fig:method}
	\end{figure*}

	To develop our method, we considered two different data augmentation techniques, one in the image space using Duplicator, and the other in the feature space using a Gaussian filter.	
	Figure \ref{fig:method} depicts the overall framework of the proposed method, from parameter optimization through to the final decision, i.e., assign genuine or forgery to a given query signature. The Convolutional Neural Network SigNet-F ($\phi$) is used to extract a representation $\phi(X)$ from each signature image $X$. For a given set of writers in the optimization database, $\phi(X_{o})$ is then used to optimize the parameters which describe the signature variability of each writer $\omega$. The result of this optimization is a parameter vector ${\pi}_{\omega}$ for each writer that describes his/her variability. Then,  the average parameter vector ${\pi}_{avg}$ for all writers available in the optimization database is computed.
	
	If we consider only the image space augmentation, the average parameter vector ${\pi}_{avg}$ and the signatures $X$ of the training set are used by Duplicator to generate duplicates respecting the writer variability.
	The model $\phi$ is used to extract the feature vectors $\phi(X)$ from signatures $X$, and the feature vectors $\phi(X_D)$ from duplicates $X_D$.
	If only the feature space augmentation is considered, the average parameter vector ${\pi}_{avg}$ and the signature feature vectors $\phi(X)$ of the training set are used by the Gaussian filter to generate new feature vectors $\phi(X_D)$ respecting the writer variability.	
	
	The feature vectors $\phi(X)$ and $\phi(X_D)$ are used to train a classifier $f$ for each writer of the verification system.
	For an input signature $X_Q$, the model $\phi$ extracts the feature vector $\phi(X_Q)$ and sends it to the pre-trained classifier $f$. Using the feature vector $\phi(X_Q)$, the classifier $f$ makes a decision $f(\phi(X_Q))$ as to whether the signature $X_Q$ is genuine or a forgery.   
	
	\subsection{Datasets}
	\label{sec:databases}
	
\begin{figure}[!t]
		\centering
		\includegraphics[width=0.95\linewidth]{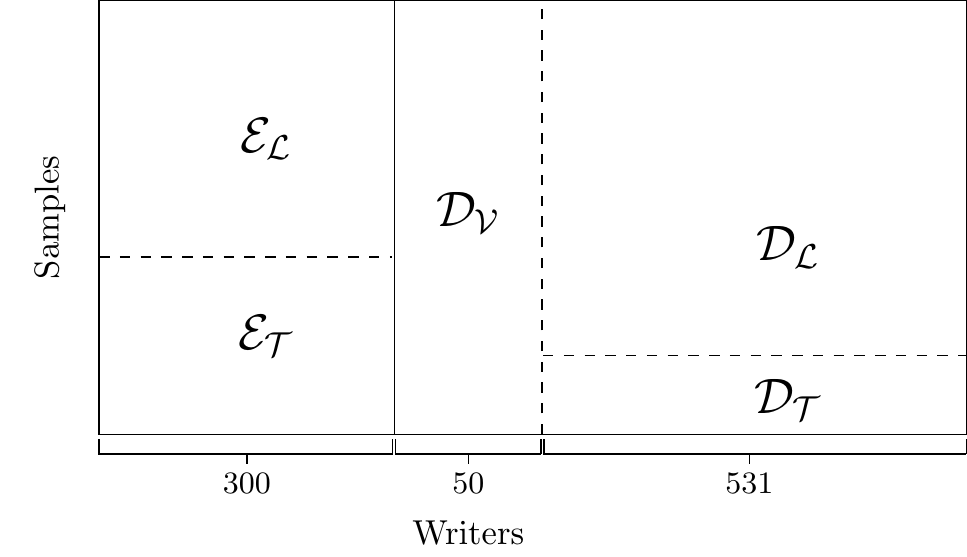}
		\caption{GPDS Dataset partitioning.}
		\label{fig:gpds}
	\end{figure}

	The experimental procedure in the present study was performed using the handwritten signature datasets GPDS-960 \cite{Vargas2007}, CEDAR \cite{Kalera2004}, and MCYT-75 \cite{Fierrez2004}. The datasets are summarized in Table \ref{tab:datasets} considering the number of writers, number of genuine samples, number of skilled forgeries, and the window size (height $\times$ width) used to normalize the signature images in each dataset.
	The GPDS dataset consists of 881 different writers, and has 24 genuine samples and 30 skilled forgeries per writer. To compare the results with Hafemann et al. \cite{Hafemann2017a}, we used the same GPDS partitioning (Figure \ref{fig:gpds}) as them. The last 581 writers compose the development dataset $\mathcal{D}$, which is subdivided into $\mathcal{D_L}$, $\mathcal{D_T}$ and $\mathcal{D_V}$ subsets. Subset $\mathcal{D_L}$ is used to train the Convolutional Neural Network (CNN), while $\mathcal{D_T}$ is used to monitor the evolution of the CNN training. These two subsets contain the same 531 writers, but different signature samples from each writer. Subset $\mathcal{D_L}$ contains 90\% of the signature samples, while $\mathcal{D_T}$ contains 10\% of them. 
	Some writers of subset $\mathcal{D_L}$ are also used to optimize the parameters of our method. Subset $\mathcal{D_V}$ contains 50 writers, and is used to make all the choices regarding the CNN model, hyperparameters of the SVM classifiers, and the initial range used to optimize the parameter vectors.  

	Finally, the first 300 writers (GPDS-300) are used to train and test the SVMs used in this work. They belong to exploitation subset $\mathcal{E}$. The samples that are used to train the classifiers are called $\mathcal{E_{L}}$, while those used to test them are called $\mathcal{E_{T}}$.
	
	The CEDAR consists of 55 different writers, and has 24 genuine samples and 24 skilled forgeries per writer \cite{Kalera2004}. The MCYT-75 consists of 75 different writers, with 15 genuine samples and 15 skilled forgeries \cite{Fierrez2004}. These datasets were used to show the generalization capability of the proposed method.
	
	\begin{table}[!t]		
		\renewcommand{\arraystretch}{1.1}
		\setlength{\tabcolsep}{5pt}
		\caption{The Offline signature datasets used in this work}
		\label{tab:datasets}
		\centering
		\begin{tabular}{lllll}
			\hline
			\multirow{2}{*}{\textbf{Dataset}} &
			\multirow{2}{*}{\textbf{Writers}} &
			{\textbf{Genuine}} &
			\multirow{2}{*}{\textbf{Forgeries}} & \multirow{2}{*}{\textbf{Window Size}}\\
			{~} &
			{~} &
			{\textbf{Signatures}} &
			{~} & {~}\\
			\hline
			{{GPDS-960}}  & {881} & {24} & {30} & {952 $\times$ 1360}\\
			{CEDAR} & {55} & {24} & {24} & {730 $\times$ 1042}\\
			{MCYT-75} & {75} & {15} & {15} & {600 $\times$ 850}\\  
			\hline
		\end{tabular}
	\end{table}
	
	\subsection{Normalization process}
	\label{sec:normalization}
	
	Hafemann et al. \cite{Hafemann2017a} showed that feature extraction can be influenced by the normalization process. Furthermore, the CNN expects signatures with the same size. Therefore, signature images are normalized using the procedure proposed in \cite{Hafemann2017a}. Firstly, the signature images are segmented using the Otsu algorithm \cite{Otsu1979}. The signature pixels remain in grayscale, while the background pixels are converted to white (255). The center of mass of the signature is computed and placed into the center of a window of height $\times$ width pixels (Table \ref{tab:datasets}). Due to the difference between the acquisition protocols of the signature datasets \cite{Vargas2007} \cite{Kalera2004} \cite{Fierrez2004}, their signatures have different sizes. Therefore, each dataset has its window size to normalize them. This process attempts to maintain the proportion of the different signature sizes.
	
	The color of all the pixels is inverted using Equation \ref{eq:inv}. The resulting image is resized to 170 $\times$ 242 pixels. Finally, the central portion of the image with 150  $\times$ 220 pixels is cropped. 
	
	\begin{equation}
	I(x, y) = 255 - I(x, y) 
	\label{eq:inv}
	\end{equation}
	
	\subsection{Convolutional Neural Network SigNet-F}
	\label{sec:cnn_signetf}
	
	We used one of the methods proposed by Hafemann et al. \cite{Hafemann2017a} to extract signature features because of the outstanding results it provided on several benchmarks. The choice was also made thanks to the discriminant nature of the descriptor. It uses a writer-independent feature learning method, in which a development set $\mathcal{D_L}$ is used to learn a feature representation $\phi(X)$. This representation is learned using a Convolutional Neural Network (CNN) to discriminate among writers in $\mathcal{D_L}$. In this context, the CNN (called Signet-F) is trained with both genuine signatures and skilled forgeries, optimizing to jointly discriminate between writers, and between genuine signatures and forgeries. In experiments performed in \cite{Hafemann2017a}, the SigNet-F achieved the best results in the GPDS-300 and CEDAR datasets, and therefore, we use it in our work.
	
	The subset $\mathcal{D_L}$ was used to learn signature features, while the process was monitored using the subset $\mathcal{D_T}$. The training was performed for 60 epochs, with an initial learning rate of 0.001. After every 20 epochs, the learning rate was divided by 10. Considering the need for a high volume of data to train the CNN, random patches of 150 $\times$ 220 pixels were extracted from the normalized 170 $\times$ 242 pixel signatures. During the feature extraction, the CNN layer FC7 was used to extract vectors with 2048 elements.
	
	\subsection{Duplicator}
	\label{sec:duplicator}
	
	Diaz et al. \cite{Diaz2017} proposed a neuromotor method combined with an ink model for signature duplication called {Duplicator}, which uses a set of 30 parameters that control the signature variability. The first 6 parameters (${\alpha}^{min}_{A}$, ${\alpha}^{max}_{A}$, ${\alpha}^{min}_{P}$, ${\alpha}^{max}_{P}$, ${\alpha}^{min}_{S}$, ${\alpha}^{max}_{S}$) are mainly responsible for describing the writer variability. To create the writer variability{, a} sinusoidal transformation is applied. The sine amplitude is determined by ${\alpha}^{min}_{A}$ and ${\alpha}^{max}_{A}$, while the sine period is determined by ${\alpha}^{min}_{P}$ and ${\alpha}^{max}_{P}$. Finally, the sine phase is delimited by ${\alpha}^{min}_{S}$ and ${\alpha}^{max}_{S}$. Considering a flexible surface where the signature is written, these six parameters control how this surface will be deformed. As a consequence, the signature will also be distorted. The next 20 parameters (${\xi}^{1}_{x}$, ${\sigma}^{1}_{x}$, ${\mu}^{1}_{x}$, ${\xi}^{2}_{x}$, ${\sigma}^{2}_{x}$, ${\mu}^{2}_{x}$, ${\xi}^{3}_{x}$, ${\sigma}^{3}_{x}$, ${\mu}^{3}_{x}$, ${\xi}^{1}_{y} $, ${\sigma}^{1}_{y} $, ${\mu}^{1}_{y} $, ${\xi}^{2}_{y} $, ${\sigma}^{2}_{y} $, ${\mu}^{2}_{y} $, ${\xi}^{3}_{y} $, ${\sigma}^{3}_{y} $, ${\mu}^{3}_{y} $, ${k}_{1}$, and ${k}_{2}$) describe the distribution of unconnected strokes in the image. These strokes are displaced taking into account three different kinds of ratio intervals, which for their part are determined by ${k}_{1}$ and ${k}_{2}$. To {choose} an interval, the ratio between the number of stroke pixels and the number of signature pixels is calculated. If a stroke sits within one of these intervals $r$, 6 parameters (${\sigma}^{{r}}_{x} $, ${\mu}^{{r}}_{x} $, ${\xi}^{{r}}_{x}$, ${\sigma}^{{r}}_{y} $, ${\mu}^{{r}}_{y} $, and ${\xi}^{{r}}_{y}$) are used to displace the unconnected stroke. The ink deposition effect is determined by ${\psi}$ {and the} last 3 parameters (${\xi}_{S}$, ${\sigma}_{S}$, ${\mu}_{S}$) are used to control the signature inclination.
	
	In this work, we focused on the optimization of the first six parameters which are mainly responsible for defining the writer variability \cite{Diaz2017}, and the others were kept at their default values. Table \ref{tab:default_parameters} shows the default values of all 31 parameters.

	\begin{table}[!t]
		\renewcommand{\arraystretch}{1.2}
		\caption{Default parameter vector proposed in \cite{Diaz2017} to duplicate offline handwritten signatures.}
		\label{tab:default_parameters}
		\centering
		\begin{tabular}{>{\bfseries}lll}
			\hline
			{\textbf{Description}} & \textbf{Parameter} & {\textbf{Default Values}}\\
			\cline{1-3}
			{\textbf{Intra-class}} & \boldmath ${\alpha}^{min}_{A}$  & {5}\\ 
			{\textbf{Variability}} &\boldmath ${\alpha}^{max}_{A}$  & {30}\\
			{} &\boldmath ${\alpha}^{min}_{P}$  & {0.5}\\
			{} &\boldmath ${\alpha}^{max}_{P}$  & {1}\\
			{} &\boldmath ${\alpha}^{min}_{S}$  & {0}\\
			{} &\boldmath ${\alpha}^{max}_{S}$  & {1}\\
			\hline
			
			{\textbf{Inter-Component}} &\boldmath {$\{{\xi}^{1}_{x}$, ${\sigma}^{1}_{x}$, ${\mu}^{1}_{x}\}$}  & {$\{-0.5$, 20, ${2*\sigma}^{1}_{x}\}$}\\
			
			{\textbf{Variability}} &\boldmath {$\{{\xi}^{2}_{x}$, ${\sigma}^{2}_{x}$, ${\mu}^{2}_{x}\}$}  & {$\{-0.5$, ${1.4*\sigma}^{1}_{x}$, ${2*1.4*\sigma}^{1}_{x}\}$}\\
			
			{} &\boldmath {$\{{\xi}^{3}_{x}$, ${\sigma}^{3}_{x}$, ${\mu}^{3}_{x}\}$}  & {$\{-0.5$, ${1.8*\sigma}^{1}_{x}$, ${2*1.8*\sigma}^{1}_{x}\}$}\\
			
			{} &\boldmath {$\{{\xi}^{1}_{y}$, ${\sigma}^{1}_{y}$, ${\mu}^{1}_{y}\}$}  & {$\{-0.5$, 8, ${\sigma}^{1}_{y}\}$}\\
			
			{} &\boldmath {$\{{\xi}^{2}_{y}$, ${\sigma}^{2}_{y}$, ${\mu}^{2}_{y}\}$}  & {$\{-0.5$, ${1.2*\sigma}^{1}_{y}$, ${1.2*\sigma}^{1}_{y}\}$}\\
			
			{} &\boldmath {$\{{\xi}^{3}_{y}$, ${\sigma}^{3}_{y}$, ${\mu}^{3}_{y}\}$}  & {$\{-0.5$, ${1.5*\sigma}^{1}_{y}$, ${1.5*\sigma}^{1}_{y}\}$}\\
			
			{} &\boldmath {${k}_{1}$}  & {0.33}\\
			{} &\boldmath {${k}_{2}$}  & {0.67}\\
			{} &\boldmath {${\psi}$}  & {0.8}\\	
			\hline
			
			{\textbf{Inclination}} &\boldmath {${\xi}_{S}$}  & {-0.19}\\
			{} &\boldmath {${\sigma}_{S}$}  & {3.28}\\
			{} &\boldmath {${\mu}_{S}$}  & {-1.30}\\
			\hline
		\end{tabular}
	\end{table}
	
	\subsection{Gaussian Filter}
	\label{sec:gaussian_filter}
	
	As previously shown, the one-dimensional low-pass Gaussian filter is widely used to generate synthetic samples in the feature space (Equation \ref{eq:gaussian_filter}). Despite the simplicity of the filter \cite{Kumar2019}, a parameter $\sigma$ is still needed  to control its intensity. The standard deviation $\sigma$ is randomly selected, considering a uniform distribution from $\sigma_{min}$ to $\sigma_{max}$. This interval introduces some variability to the synthetic samples. Based on the same idea as in Duplicator, we used this interval to represent the writer variability, and thus optimized the parameters that determine this interval using the optimization process described in the next section.
	
	\begin{equation}
	{G(x)} = {\frac{1}{\sqrt{2\pi}\sigma}e^{-\frac{x^2}{2{\sigma}^2}}}
	\label{eq:gaussian_filter}
	\end{equation}
	
	\subsection{Parameter Optimization}
	\label{sec:parameter_optimization}
	
	Parameter optimization is performed in a bid to find a set of parameters for to find a set of parameters for data augmentation methods, which allow the methods to generate synthetic samples respecting the distribution of a given writer.
	In a real-world scenario, two or more writers have different intra-class variabilities. This difference notwithstanding, some common behaviors can be shared by these writers along the writing process, and these common writer variability traits can be described by a global set of parameters.
	Since the first six parameters of the duplicator are mainly responsible for the intra-personal variability of writer signatures \cite{Diaz2017}, and optimization is a time-consuming task \cite{Zhang2020}, these parameters are chosen to represent the writer variability traits. Regarding the Gaussian filter, we used the parameters $\sigma_{min}$ and $\sigma_{max}$ to represent the writer variability traits.
	
	In this work, we used a Particle Swarm Optimization (PSO) algorithm \cite{Zhao2016} to find the first six variability parameters used by the Neuromotor-based duplicator{, and the two parameters used by Gaussian filter}. The PSO was originally proposed by Kennedy and Eberhart \cite{Kennedy1995} to optimize continuous nonlinear functions. The algorithm is based on the behavior of a flock of birds or school of fish looking for places with abundant food. Due to its efficiency in solving several {kinds} of optimization problems \cite{Zhao2016} \cite{Huang2011} \cite{Salehi2015} \cite{Fan2020} and its simple implementation \cite{Zhang2020}, the PSO is used in our work.
	
	Considering the search space of $d$ dimensions, each particle {$\pi$} with index $i$ represents the position and a possible solution to an optimization problem.
	During optimization, a velocity $v$ with index $i$ is associated with each particle in the search space. For each iteration $n$ of the algorithm, the velocity ${v_{id}^{n+1}}$ of the next iteration (Equation \ref{eq:pso_speed}) is updated according to the values
	of the local minima particle ${p^{n}_{id}}$ and the global minima particle ${{\pi}^{n}_{{\omega}d}}$ \cite{Zhao2016}.
	
	The ${v_{id}^{n+1}}$ is also computed considering two uniformly distributed variables within [0,1], ${r^{n}_{1id}}$ and ${r^{n}_{2id}}$. These variables introduce some diversity to the particles during the search. This diversification is regulated by the constant $1-\chi_o$. The constant $\chi_o$ itself controls the intensity of the search to find the best solution \cite{Zhao2016}. While the constant ${c_{o1}}$ regulates the tendency of the particles to approach their local minima particle, the constant ${c_{o2}}$ regulates their tendency to approach their global minima particle \cite{Fan2020}. The perturbation constant ${\gamma_o}$ controls the stability of the algorithm regulating the effect of both constants ${c_{o1}}$ and ${c_{o2}}$ concurrently.
	
	\raggedbottom
	\begin{eqnarray}
	{v_{id}^{n+1}} &{}={}& (1-\chi_o){v^{n}_{id}}+{\chi_o}{c_{o1}}{\gamma_o}{r^{n}_{1id}}({p^{n}_{id}}-{{\pi}^{n}_{id}})\nonumber\\ 
	&{}&+{}{{\chi_o}{c_{o2}}{\gamma_o}{r^{n}_{2id}}({{\pi}^{n}_{{\omega}d}}-{{\pi}^{n}_{id}})}
	\label{eq:pso_speed}
	\end{eqnarray}
	
	According to the optimization experiments performed by Zhao \cite{Zhao2016}, when the PSO uses the constants $(1-\chi_o)=(3-\sqrt{5})/2$, ${\chi_o}{c_{o1}}{\gamma_o}=(1+\sqrt{5})/2$, and ${\chi_o}{c_{o2}}{\gamma_o}=1$
	it is more accurate, efficient, and stable than the traditional PSO \cite{Kennedy1995}. Therefore, we used these constants in {Equation \ref{eq:pso_speed}, which resulted} in Equation \ref{eq:pso_speed2}. Considering the computed velocity ${v_{id}^{n+1}}$, the position of the particle ${{\pi}_{id}^{n+1}}$ is updated using {Equation} \ref{eq:pso_position}.
	
	\begin{eqnarray}
	{v_{id}^{n+1}} &{}={}& \frac{(3-\sqrt{5})}{2}{v^{n}_{id}}+{\frac{(1+\sqrt{5})}{2}}{r^{n}_{1id}}({p^{n}_{id}}-{{\pi}^{n}_{id}})\nonumber\\ 
	&{}&+{}{{r^{n}_{2id}}({{\pi}^{n}_{{\omega}d}}-{{\pi}^{n}_{id}})}
	\label{eq:pso_speed2}
	\end{eqnarray}
	
	\begin{equation}
	{{\pi}_{id}^{n+1}} = {{\pi}^{n}_{id}} + {v^{n+1}_{id}}
	\label{eq:pso_position}
	\end{equation}
	
	The parameter vector is encoded into a 6-dimensional particle $\pi$ for image space augmentation, while for feature space augmentation, it is encoded into a 2-dimensional particle $\pi$. The parameters are randomly initialized considering a uniform distribution with low and high limits.
	Table \ref{tab:sigvar_range} shows the low and high limits used during the optimization process. The low limit of the max parameter is defined using the value of the min parameter. For example, if the parameter ${\alpha}^{min}_{A}$ is initialized with the value 20, the low limit of the parameter ${\alpha}^{max}_{A}$ is 20.
	
	For the image space  augmentation, the limits of the parameters were defined considering the extreme values of the default parameters proposed by Diaz et al. \cite{Diaz2017}.
	If we use ${\alpha}^{min}_{A}$ and ${\alpha}^{max}_{A}$ equal to or greater than 10, the duplicator generates duplicates with fewer distortions than when values lower than 10 are used. However, if we use large values of ${\alpha}_{A}$, the duplicates will be equal to the signature used as seed. In order to find the best values of ${\alpha}^{min}_{A}$ and ${\alpha}^{max}_{A}$, we set the limits of ${\alpha}^{min}_{A}$ from 10 to 100. According to the authors in \cite{Diaz2017} the values of ${\alpha}^{min}_{P}$, ${\alpha}^{max}_{P}$, ${\alpha}^{min}_{S}$, and ${\alpha}^{max}_{S}$ should be between 0 and 1. However, when these parameters assume extremely small or extremely large values, the duplicator generates unnatural signature duplicates.
	
	For the feature space augmentation, the limits of the parameters were defined, considering the mathematical constraints of the Gaussian filter. When $\sigma$ is 0, it leads to a mathematical indetermination. We also thought about the degree of perturbation applied by the Gaussian filter in the original feature vectors. If we perturb the original feature vector too much, it can generate a feature vector that does not resemble the same class as the original \cite{DeVries2017}. Therefore, we considered the limits ranging from 0.01 to 1.00.   
	
	\begin{table}[!t]
		\setlength{\tabcolsep}{2pt} 
		\renewcommand{\arraystretch}{1.2} 
		\caption{Parameter Initialization Range used to optimize the parameter vectors.}
		\label{tab:sigvar_range}
		\centering
		\begin{tabular}{>{\bfseries}lclllllll}
			\hline
			\multirow{2}{*}{{\textbf{Limit}}} &
			\multicolumn{8}{c}{{\textbf{Parameter Initialization Range}}}\\
			\cline{2-9}
			\multicolumn{1}{l}{} &
			\multicolumn{1}{c}{{\boldmath ${\alpha}^{min}_{A}$}} &
			\multicolumn{1}{c}{{\boldmath ${\alpha}^{max}_{A}$}} & 
			\multicolumn{1}{c}{{\boldmath ${\alpha}^{min}_{P}$}} &
			\multicolumn{1}{c}{{\boldmath ${\alpha}^{max}_{P}$}} &
			\multicolumn{1}{c}{{\boldmath ${\alpha}^{min}_{S}$}} &
			\multicolumn{1}{c}{{\boldmath ${\alpha}^{max}_{S}$}} &
			\multicolumn{1}{l}{{\boldmath ${\sigma}_{min}$}} &
			\multicolumn{1}{l}{{\boldmath ${\sigma}_{max}$}}\\[1mm]
			\cline{1-9}
			{Low} & {10.0} & {${\alpha}^{min}_{A}$} & {0.0} & {${\alpha}^{min}_{P}$} & {0.0} & {${\alpha}^{min}_{S}$} & {0.01} & {${\sigma}_{min}$}\\
			
			{High} & {100.0} & {100.0} & {1.0} & {1.0} & {1.0} & {1.0} & {1.00} & {1.00}\\
			
			\hline
		\end{tabular}
	\end{table}
	
	To guide the optimization process, we used the silhouette index \cite{Rousseeuw1987}. Usually, this index is used to measure how good two or more clusters are. If the silhouette index is equal to 1, the clusters have a small intraclass and a large interclass variability; if it is equal to -1, it means the data were assigned to the wrong clusters. Finally, if the silhouette index is equal to 0, then the clusters are overlapped.
	This is exactly what we are looking for. Therefore, during the optimization process, our goal is to find a set of parameters such that the silhouette index is close to 0.
	
	The silhouette index evaluates the sparsity and the distance between the clusters concurrently. Several functions can be used to compute the dissimilarity between the elements of the clusters. As recommended by Rousseeuw \cite{Rousseeuw1987}, we used the Euclidean distance as a dissimilarity function $d(.)$. The average dissimilarity between the $i^{th}$ element and the elements of the same cluster $Cs$ can be computed using Equation \ref{eq:dissimilarity_a}. Basically, $a(.)$ measures the sparsity inside each cluster.
	
	\begin{equation}
	{a(\phi(X_i))} = \frac{\sum\limits_{j=1}^{n_{Cs}}{d{(\phi(X_i);\phi(X_j))}}}{{n_{Cs}}-1}
	\label{eq:dissimilarity_a}
	\end{equation}
	
	The average dissimilarity between the $i^{th}$ element and the elements of the other cluster $Cr$ is computed using Equation \ref{eq:dissimilarity_inter}. Subsequently,  Equation \ref{eq:dissimilarity_inter} is used in Equation \ref{eq:dissimilarity_b} to compute the minimum distance between the border of the clusters $Cs$ and $Cr$. This indicates how far the clusters are from each other. It is important to highlight that the clusters $Cs$ and $Cr$ are different.
	
	\begin{equation}
	{d(\phi(X_i); Cr)} = \frac{\sum\limits_{j=1}^{n_{Cr}}{d{(\phi(X_i);\phi(X_j))}}}{{n_{Cr}}}
	\label{eq:dissimilarity_inter}
	\end{equation}
	
	\begin{equation}
	{b(\phi(X_i))} = \min{\left\{{d(\phi(X_i); Cr)}\right\}}
	\label{eq:dissimilarity_b}
	\end{equation}
	
	Equations \ref{eq:dissimilarity_a} and \ref{eq:dissimilarity_b} are combined in Equation \ref{eq:deltai}. For an arbitrary feature vector $\phi(X_i)$ as reference, Equation \ref{eq:deltai} evaluates the intraclass and the interclass variability concurrently. Following that, the value is normalized using the maximum value between the intraclass and the interclass variability.
	
	\begin{equation}
	{\delta(\phi(X_i))} = \frac{b(\phi(X_i))-a(\phi(X_i))}{\max{\{b(\phi(X_i));a(\phi(X_i))\}}}
	\label{eq:deltai}
	\end{equation}
	
	Equation \ref{eq:abs_silhouette} computes the ${\delta(\phi(X_i))}$ for all elements of the clusters. The ${\delta(\phi(X_i))}$s are summed and divided by the number of elements $n_C$ in all clusters. To simplify the optimization process, we used only the absolute value of the silhouette index ($\left|{\Delta}\right|$).	
	
	\begin{equation}
	\left|{\Delta}\right| = \left|\frac{\sum\limits_{i=1}^{n_C}{\delta(\phi(X_i))}}{n_C}\right|, {\left|{\Delta}\right| \in {[0,1]}}
	\label{eq:abs_silhouette}
	\end{equation}
	
	Details of the proposed optimization algorithm are presented in Algorithm \ref{alg:training_algorithm_sigvar}. It receives as input $N$ (the number of samples that will be generated for each genuine signature), $W$ (list of writers with their respective genuine signatures), and $I$ (the number of iterations for which the algorithm will run). The output is the average parameter vector $\pi_{avg}$. While Algorithm \ref{alg:eval_duplicator_params_algorithm} presents the evaluation process of parameters to generate duplicates, Algorithm \ref{alg:eval_filter_params_algorithm} details the evaluation of parameters to generate samples in the feature space.

	Then, the optimization process starts with the PSO generating a set of parameter vectors that are used to create the synthetic samples. 
	In the image space augmentation, for each signature of the writer $X_o$, a duplicate $X_D$ is generated. The duplicates are normalized and the feature vectors $\phi(X_o)$, and $\phi(X_D)$ are extracted.
	In the feature space augmentation, the signatures of the writer are normalized and the feature vectors $\phi(X_o)$ are extracted. For each feature vector $\phi(X_o)$, a synthetic feature vector $\phi(X_D)$ is generated using a Gaussian filter.	
	The feature vectors of the genuine signatures $\phi(X_o)$ and synthetic samples $\phi(X_D)$ are used to compute the absolute value of the silhouette index $\left|{\Delta}\right|$. If the cluster of genuine signatures and the cluster of synthetic samples have an equal or similar variability, $\left|{\Delta}\right| \rightarrow 0$.
	Therefore, the parameter vectors $\pi_{\omega}$ with the lowest absolute silhouette indices are selected and saved for each writer. The parameter vectors are updated for the next iteration using Equations \ref{eq:pso_speed2} and \ref{eq:pso_position}. The process is repeated until the stop condition is satisfied. In the end, the average parameter vector ($\pi_{avg}$) is computed, and describes the common behavioral biometric traits shared by the writers in the optimization database. It is important to highlight that we hypothesized that the writer variability observed on the image space can be reflected in the feature space. Based on this hypothesis, the parameter optimization is performed, considering only the feature vectors. Therefore, we assume that the duplicates will have a human-like appearance if $\left|{\Delta}\right|\rightarrow 0$. In other words, the interaction between the first 6 parameters is taken into account considering the minimization of the absolute value of $\Delta$ (Equation \ref{eq:abs_silhouette}).
	
	\begin{algorithm}[!t]
		\caption{The parameter optimization algorithm (Sigvar)}
		\label{alg:training_algorithm_sigvar}
		\small
		\begin{algorithmic}[1]
			\Require
			\Statex $N$: number of duplicates per signature
			\Statex $W$: list of writers
			\Statex $I$: number of iterations
			\Ensure
			\Statex ${{\pi}_{avg}}$: average parameter vector
			\State ${\pi_{avg}} \gets {\emptyset}$
			\ForEach {$\omega \in W$}
			\State ${X} \gets$ {loadSignatures}$\left({\omega}\right)$
			\LineComment{Initialize the particles}
			\State $\textit{particles} \gets \text{initializeParticles()}$
			\State {${\Delta_{localmin}} \gets \text{9999}$}
			\State ${\Delta_{min}} \gets \text{9999}$
			\State ${{\pi}_{\omega}} \gets \emptyset$
			\ForEach {$iteration \in I$}
			\State {$p \gets \emptyset$}
			\ForEach {$\pi \in particles$}
			\LineComment{Evaluate parameter vector}
			\State {$\Delta \gets $evalParameters($\pi$,$X$,$N$)}
			\If{$|\Delta| < |\Delta_{min}|$}
			\State $|\Delta_{min}| \gets |\Delta|$
			\State {${\pi}_{\omega} \gets \pi$}
			\EndIf
			\If {$|\Delta| < |\Delta_{localmin}|$}
			\State {$|\Delta_{localmin}| \gets |\Delta|$}
			\State {${p} \gets \pi$}
			\EndIf
			\EndFor
			\LineComment{Update the value of each particle using\newline \hspace*{0.95cm} Equations \ref{eq:pso_speed2} and \ref{eq:pso_position}}
			\State {$particles \gets $
				updateParticles(particles, p, $\pi_{\omega}$)}
			\EndFor
			\LineComment {Save the best parameters for each writer}
			\State {saveParameters($\omega$,${\pi}_{\omega}$)}
			\LineComment{Sum all the parameter vectors}
			\State ${{\pi}_{avg}} \gets {{\pi}_{avg}} + {{\pi}_{\omega}}$
			\EndFor
			\LineComment {Count the number of writers in the list $W$}
			\State {$n_W \gets $getNumberOfWriters(W)}
			\LineComment {Compute the average parameter vector}
			\State {${{\pi}_{avg}} \gets {{\pi}_{avg}}/{n_W}$}
		\end{algorithmic}
	\end{algorithm}
	
	\begin{algorithm}[!t]
		\caption{{evalParameters (Duplicator)}}
		\label{alg:eval_duplicator_params_algorithm}
		\small
		\begin{algorithmic}[1]
			\Require
			\Statex {$\pi$: parameter vector}
			\Statex {$X$: signatures}
			\Statex {$N$: number of duplicates per signature}
			\Ensure
			\Statex {${{\Delta}}$: Silhouette Index}
			\State {{${\pi_{avg}} \gets {\emptyset}$}}
			\State {${X_D} \gets \emptyset$}
			\ForEach {$X_i \in X$}
			\State {{{aux}$X_D \gets $ Duplicator($\pi$, N,${X_i}$)}}
			\LineComment{{Concatenate ${X_D}$ and aux$X_D$}}
			\State {{${X_D} \gets {X_D} ^\frown $aux$X_D$}}
			\EndFor
			\LineComment{{Normalize the signatures and duplicates}}
			\State {norm$X \gets $normalize(${X}$)}
			\State {norm$X_D \gets $normalize(${X_D}$)}
			\LineComment{{Extract the features from the normalized\newline \hspace*{0.30cm} signatures and duplicates using the SigNet-F}}
			\State {${\phi(X)} \gets $extractFeatures(norm$X$)}
			\State {${\phi(X_D)} \gets $ extractFeatures(norm$X_D$)}
			\LineComment{{Calculate the value of the silhouette index}}
			\State {$\Delta \gets $silhouetteIndex($\phi(X)$,$ \phi(X_D)$)}
		\end{algorithmic}
	\end{algorithm}
	
	\begin{algorithm}[!t]
		\caption{{evalParameters (Gaussian filter)}}
		\label{alg:eval_filter_params_algorithm}
		\small
		\begin{algorithmic}[1]
			\Require
			\Statex {$\pi$: parameter vector}
			\Statex {$X$: signatures}
			\Statex {$N$: number of new samples per signature}
			\Ensure
			\Statex {${{\Delta}}$: Silhouette Index}
			\LineComment{{Normalize the signatures and duplicates}}
			\State {norm$X \gets $normalize(${X}$)}
			\LineComment{{Extract the features from the normalized\newline \hspace*{0.30cm} signatures using the SigNet-F}}
			\State {${\phi(X)} \gets $extractFeatures(norm$X$)}
			\State {${\phi(X_D)} \gets \emptyset$}
			\ForEach {$\phi(X_i) \in \phi(X)$}
			\State {{{aux}$\phi(X_D) \gets $ applyGaussianfilter($\pi$, N, ${\phi(X_i)}$)}}
			\LineComment{{Concatenate $\phi(X_D)$ and aux$\phi(X_D)$}}
			\State {{${\phi(X_D)} \gets {\phi(X_D)} ^\frown $aux$\phi(X_D)$}}
			\EndFor
			\LineComment{{Calculate the value of the silhouette index}}
			\State {$\Delta \gets $silhouetteIndex($\phi(X)$,$ \phi(X_D)$)}
		\end{algorithmic}
	\end{algorithm}
	
	\subsection{Training}
	\label{sec:training}
	
	Once the search process is complete, we can use the optimized parameters to create synthetic samples to train the Writer-Dependent classifiers used by the signature verification system.
	Four different scenarios were considered to train these classifiers. The first scenario considers that there are no duplicates
	at all to train the SVMs, and is used as a baseline. The second one uses duplicates that were created by using the default parameters reported in Table \ref{tab:default_parameters}. The third creates duplicates using the parameters found by the proposed algorithm described in the previous section. Finally, the last scenario creates synthetic feature vectors using a Gaussian filter and the parameters found by the optimization process.
	
	To validate the impacts of the proposed optimization method and to fairly compare the results with those of Hafemann et al. \cite{Hafemann2017a}, we adopted the same signature verification system proposed by them.
	The SigNet-F is used to extract the feature vectors of the normalized genuine and random forgery signatures. For each writer, an SVM (Support Vector Machine) classifier with an RBF kernel is trained using these feature vectors. The genuine signatures were considered as a positive class and the random forgeries were considered as a negative class. The genuine signatures of other writers are used as random forgeries. The difference in the numbers of positive and negative examples may lead to classifiers that tend to select one class more frequently than others. Therefore, different $C$ weights were adopted to positive and negative classes. For the negative class, $C^-$ is equal to 1. Before determining the $C^+$, it is necessary to compute the skew $\psi$. The skew $\psi$ was computed using the number of positive examples $P$ (genuine signatures) used for training and the number of negative examples $N$ (random forgeries) used for training (Equation \ref{eq:skew}). Then, $C^+$ was computed using $C^-$ and $\psi$ (Equation \ref{eq:positive}).
	
	\begin{equation}
	{\psi} = \frac{N}{P}
	\label{eq:skew}
	\end{equation}
	
	\begin{equation}
	{{C}^{+}} = {\psi}{{C}^{-}}  
	\label{eq:positive}
	\end{equation}  
	
	\subsection{Verification}
	\label{sec:verification}
	
	After training the Writer-Dependent classifiers the system is ready to be deployed. During the verification phase, each query signature image $X_Q$ is normalized and the feature vector $\phi(X_Q)$ is extracted using the SigNet-F. The feature vector $\phi(X_Q)$ is submitted to an SVM classifier $f$, and it makes a decision $f(\phi(X_Q))$. As a result of the decision $f(\phi(X_Q))$, the signature $X_Q$ is classified as a genuine or a forgery sample.
	To assess the performance of the proposed method, the mean Equal Error Rate (EER) of the verification system was computed. The verification system was assessed considering the three previously exposed scenarios.

	\section{Experimental Results}
	\label{sec:results}
	
	To validate the proposed method, two different approaches were considered: feature-level (Section \ref{sec:validation_feature}) and performance-level (Sections \ref{sec:performance_validation_duplicator} and \ref{sec:performance_validation_gaussianfilter}). All experiments were carried out on the three datasets described in Section \ref{sec:databases}.
	
	It is well known that finding a solution for an optimization problem using a meta-heuristic algorithm is a time-consuming task, and it was no different in our case. For this reason, instead of using all writers available in the $\mathcal{D_L}$, we selected a subset of 20 writers representing the entire population to compose our optimization dataset. The 20 writers that covered the distribution of $\mathcal{D_L}$ were randomly selected (writers 431, 490, 503, 525, 588, 611, 631, 641, 643, 654, 673, 676, 701, 716, 797, 825, 897, 912, 935, and 945) and used as input to the optimization algorithm.
	
	As discussed in Section \ref{sec:parameter_optimization}, at the end of the optimization process, for each writer from the optimization dataset, there is one optimized parameter vector, which is used to calculate 
	the average parameter vector ($\pi_{avg}$) that is used by the data augmentation method. While the duplicator uses the average parameter vector $\pi_{dup}$, the Gaussian filter uses the average parameter vector $\pi_{gauss}$.
	Table \ref{tab:common_parametersLR_20DLDT} shows the silhouette index, the average parameter vectors {$\pi_{dup}$ and $\pi_{gauss}$}, and the default parameter vector. As we can see, {$\pi_{dup}$} is quite distant from the default vector. Mainly, the first two parameters ($\alpha_{A}^{min}$ and $\alpha_{A}^{max}$) have greater {average parameter vector values} than those {specified for} the default parameter vector. {Consequently, the} silhouette index ($\left|\Delta\right|$) of the {$\pi_{dup}$} is lower than that of the $\pi_{def}$. Therefore, {$\pi_{dup}$} {better} represents the variability of these writers than $\pi_{def}$ does.
			
\begin{figure}[!t]
		\pgfplotsset{cycle list/Dark2-6}
		\pgfplotsset{cycle list/Spectral}
		\centering
		\begin{tikzpicture}[scale=0.9, auto, inner sep=0, outer sep=0]
		\definecolor{clr3}{RGB}{31,119,180}
		\definecolor{clr18}{RGB}{174,199,232}
		\definecolor{clr15}{RGB}{44,160,44}
		\definecolor{clr6}{RGB}{152,223,138}
		\definecolor{clr2}{RGB}{23,190,207}
		\definecolor{clr7}{RGB}{158,218,229}
		\definecolor{clr20}{RGB}{188,189,34}
		\definecolor{clr5}{RGB}{219,219,141}
		\definecolor{clr8}{RGB}{140,86,75}
		\definecolor{clr11}{RGB}{196,156,148}
		\definecolor{clr16}{RGB}{127,127,127}
		\definecolor{clr9}{RGB}{199,199,199}
		\definecolor{clr19}{RGB}{148,103,189}
		\definecolor{clr10}{RGB}{197,176,213}
		\definecolor{clr12}{RGB}{227,119,194}
		\definecolor{clr13}{RGB}{247,182,210}
		\definecolor{clr4}{RGB}{255,187,120}
		\definecolor{clr14}{RGB}{255,127,14}
		\definecolor{clr17}{RGB}{255,152,150}
		\definecolor{clr1}{RGB}{214,39,40}
		\begin{axis}[ 
		width=9.5cm,
		xlabel=$\sigma$,            
		xmin = 0.0,
		xmax = 4.01,
		xtick={0.0,0.1,...,4.01},
		x tick label style={
			rotate=90,
			anchor=east,
			font=\scriptsize,
			xshift=-3pt
		},
		x label style={
			font=\footnotesize,
			text width=0.5cm,
			align=center,
			yshift=-2pt
		},
		ylabel=$\left|\Delta\right|$,
		ylabel style={
			anchor=west,
			text width=0.5cm,
			align=center
		},
		y tick label style={
			font=\scriptsize,
			xshift=-2pt
		},	
		y label style={
			font=\footnotesize,
			yshift=8pt
		},
		ymin = 0.,
		ymax = 1.01,
		ytick={0.0,0.1,...,1.01},
		legend columns=6,
		legend style={at={(0.5,0.85)},
			draw=none,
			fill=none, 
			anchor=center,
			align=center,
			font=\scriptsize},
		cycle multi list={
			Spectral\nextlist
			[1 of]mark list}]
		
		\addplot [clr1, mark=+] table [x=sigma, y=si] {\tableCorrSiA};
		\addlegendentry{431}
		
		\addplot [clr2] table [x=sigma, y=si] {\tableCorrSiB};
		\addlegendentry{490}
		
		\addplot [clr3, mark=o] table [x=sigma, y=si] {\tableCorrSiC};
		\addlegendentry{503}
		
		\addplot [clr4, mark=asterisk] table [x=sigma, y=si] {\tableCorrSiD};
		\addlegendentry{525}
		
		\addplot [clr5] table [x=sigma, y=si] {\tableCorrSiE};
		\addlegendentry{588}
		
		\addplot [clr6] table [x=sigma, y=si] {\tableCorrSiF};
		\addlegendentry{611}
		
		\addplot [clr7] table [x=sigma, y=si] {\tableCorrSiG};
		\addlegendentry{631}
		
		\addplot [clr8] table [x=sigma, y=si] {\tableCorrSiH};
		\addlegendentry{641}
		
		\addplot [clr9] table [x=sigma, y=si] {\tableCorrSiI};
		\addlegendentry{643}
		
		\addplot [clr10] table [x=sigma, y=si] {\tableCorrSiJ};
		\addlegendentry{654}
		
		\addplot [clr11, mark=triangle] table [x=sigma, y=si] {\tableCorrSiK};
		\addlegendentry{673}
		
		\addplot [clr12] table [x=sigma, y=si] {\tableCorrSiL};
		\addlegendentry{676}
		
		\addplot [clr13] table [x=sigma, y=si] {\tableCorrSiM};
		\addlegendentry{701}
		
		\addplot [clr14, mark=o] table [x=sigma, y=si] {\tableCorrSiN};
		\addlegendentry{716}
		
		\addplot [clr15] table [x=sigma, y=si] {\tableCorrSiO};
		\addlegendentry{797}
		
		\addplot [clr16] table [x=sigma, y=si] {\tableCorrSiP};
		\addlegendentry{825}
		
		\addplot [clr17, mark=x] table [x=sigma, y=si] {\tableCorrSiQ};
		\addlegendentry{897}
		
		\addplot [clr18, mark=star] table [x=sigma, y=si] {\tableCorrSiR};
		\addlegendentry{912}
		
		\addplot [clr19] table [x=sigma, y=si] {\tableCorrSiS};
		\addlegendentry{935}
		
		\addplot [clr20] table [x=sigma, y=si] {\tableCorrSiT};
		\addlegendentry{945}
		\draw [dashed] (100.00,0) -- (100.00,80);
		\draw [dashed] (200.00,0) -- (200.00,80);
		\draw [dashed] (300.00,0) -- (300.00,80);
		\end{axis}
		\end{tikzpicture}
		\caption{Effect of Gaussian filter in the $\left|\Delta\right|$ for each writer of $20\mathcal{D_L}$.}
		\label{fig:sigma_si} 
	\end{figure}
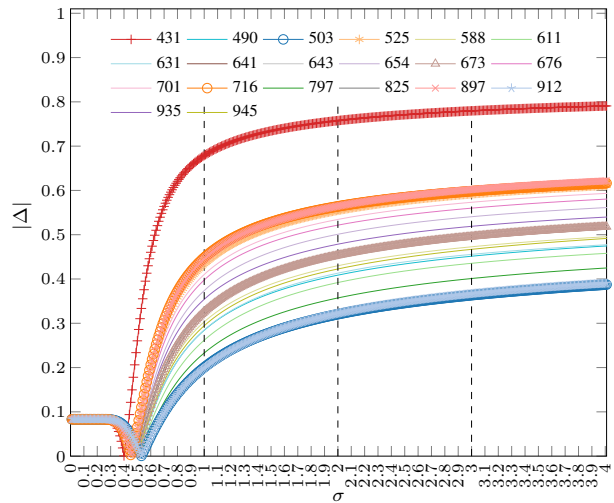

	Figure \ref{fig:sigma_si} shows the effect of the Gaussian filter on the silhouette index for each writer of 20$\mathcal{D_L}$. It can be seen that each writer has their own ideal sigma interval denoting their specific writer variability. Furthermore, even a simple noise addition technique needs some kind of optimization to generate more realistic synthetic samples. 
	
	\begin{table}[!t]
		\renewcommand{\arraystretch}{1.1} 
		\caption{{Average parameter vectors ${\pi}_{dup}$ and ${\pi}_{gauss}$,  default parameters (${\pi}_{def}$) proposed in \cite{Diaz2017}, and Silhouette indices.}}
		\label{tab:common_parametersLR_20DLDT}
		\centering
		\begin{tabular}{>{\bfseries}lrrr}
			\hline
			\multirow{2}{*}{{\textbf{Parameter}}} &
			\multicolumn{3}{c}{{\textbf{Parameter Vectors}}}\\[1mm]
			\cline{2-4}
			\multicolumn{1}{l}{~} &
			\multicolumn{1}{l}{{\boldmath ${\pi}_{def}$}} &
			\multicolumn{1}{c}{{\boldmath ${\pi}_{dup}$}} &
			\multicolumn{1}{c}{{\boldmath ${\pi}_{gauss}$}}\\[1mm]
			\cline{1-4}
			{\boldmath ${\alpha}^{min}_{A}$} & {5.000} & {69.300} &	{-}\\
			{\boldmath ${\alpha}^{max}_{A}$} & {30.000} & {88.70} &	{-}\\
			{\boldmath ${\alpha}^{min}_{P}$} & {0.500} & {0.320} & {-}\\
			{\boldmath ${\alpha}^{max}_{P}$} & {1.000} & {0.530} &	{-}\\
			{\boldmath ${\alpha}^{min}_{S}$} & {0.000} & {0.470} &	{-}\\
			{\boldmath ${\alpha}^{max}_{S}$} & {1.000} & {0.740} &	{-}\\
			{\boldmath ${\sigma}_{min}$} & {-} & {-} &	{0.290}\\
			{\boldmath ${\sigma}_{max}$} & {-} & {-} &	{0.720}\\
			\hline
			{\boldmath{$\left|\Delta\right|$}} & {0.153} & {0.047} &	{0.040}\\
			\hline
		\end{tabular}
	\end{table}
	
	\subsection{Validation at Feature Level}
	\label{sec:validation_feature}
	
	Before discussing the impacts of the proposed optimization method in terms of performance, i.e., reduction of the EER in the signature verification system, we will present an analysis of the quality of the 
	synthetic samples that are created by the duplicator and Gaussian filter using the average parameter vectors.
	
	Since we are only using the feature vectors of handwritten signatures, it was necessary to perform the validation of the method at the feature level. For the same writer, it is expected that genuine signatures and synthetic samples will have similar aspects, and as a result, the signature and synthetic feature vectors should be similar as well. Besides the similarity, the synthetic samples are also expected to keep the original writer variability. It is important to highlight that the feature descriptor must be sufficiently discriminant to measure the dissimilarity between the signatures and synthetic samples \cite{Hafemann2017a}. If the feature descriptor is poor, then the distinction between them will be poor as well \cite{Souza2020}.
	
	In this experiment, for each writer, we used the first 12 genuine signatures to build the genuine cluster in the 2048 dimensional feature space created by the SigNet-F. Therefore, 4 clusters with 12 feature vectors were created: one with genuine signatures, one with duplicates created with the optimized parameter vector ${\pi}_{dup}$, one with the duplicates created with the default parameter vector ${\pi}_{def}$, and one with the feature vectors generated with the optimized parameter vector ${\pi}_{gauss}$.
	
	To measure how close the synthetic samples and genuine signatures are, we used the absolute value of the silhouette index $\left|\Delta\right|$. As earlier explained, if both clusters have a similar variability, then $\left|\Delta\right| \rightarrow 0$.
	For each writer, three $\left|\Delta\right|$ values were computed using Equation \ref{eq:abs_silhouette}: the $\left|\Delta\right|$ between the genuine cluster and the cluster of duplicates using the default parameter vector ${\pi}_{def}$, the $\left|\Delta\right|$ between the genuine cluster and the duplicates using the average parameter vector ${\pi}_{dup}$, and the $\left|\Delta\right|$ between the genuine cluster and the synthetic feature vectors using the average parameter vector ${\pi}_{gauss}$. Then, the average silhouette indices ${\left|\Delta\right|_{avg}}$ and the standard deviations were computed. To measure the sparsity of each dataset in the feature space, we used the average cohesion of all writers in each dataset. The cohesion $co$ is computed using the sum of the squared differences between all elements $\phi(X_i)$ of the cluster and the cluster centroid $\mu$ (Equation \ref{eq:cohesion}) \cite[p.~578]{Tan2018}. For each dataset, we computed the cohesions of all genuine clusters and their average, ${co}_{avg}$. 
	Table \ref{tab:si_param20DL} shows the ${\left|\Delta\right|_{avg}}${, {the} average sparsities ${co}_{avg}$}, and standard deviations of the three datasets described in Section \ref{sec:databases}.
	
	\begin{equation}
	{{{co}} = \sum\limits_{i=1}^{n}{{(\phi(X_i)-\mu)}^{2}}}
	\label{eq:cohesion}
	\end{equation} 
	
	\begin{table}[!t]
		\setlength{\tabcolsep}{2.5pt} 
		\renewcommand{\arraystretch}{1.1}
		\caption{Average absolute values of the silhouette indices ($\left|\Delta\right|_{avg}$), {Average {Sparsity} of genuine clusters (${co}_{avg}$),} and standard deviations for GPDS-300, CEDAR, and MCYT-75.}
		\label{tab:si_param20DL}
		\centering
		\begin{tabular}{>{\bfseries}lllll}
			\hline
			\multirow{2}{*}{\textbf{Dataset}} &  \multicolumn{3}{c}{\boldmath{${\left|\Delta\right|}_{avg}$}} &
			\multicolumn{1}{c}{\multirow{2}{*}{{\boldmath ${{co}}_{avg}$}}}\\
			\cline{2-4}
			\multicolumn{1}{l}{} &
			\multicolumn{1}{c}{\boldmath ${{\pi}}_{def}$} & 
			\multicolumn{1}{c}{{\boldmath ${\pi}_{dup}$}} &
			\multicolumn{1}{c}{{\boldmath ${{\pi}}_{gauss}$}} &
			\multicolumn{1}{c}{}\\
			\cline{1-5}
			{GPDS-300} & {$0.14\pm0.10$} & {$0.04\pm0.05$} & {$0.04\pm0.04$} & {$18860.60\pm1854.13$}\\
			{CEDAR} & {$0.70\pm0.14$} & {$0.56\pm0.18$} & {{$0.28\pm0.13$}} & {$13788.87\pm804.96$}\\
			{MCYT-75} & {$0.37\pm0.12$} & {$0.15\pm0.10$} & {{$0.10\pm0.06$}} & {$15900.48\pm945.49$}\\
			\hline 
		\end{tabular}
	\end{table}
	
\begin{figure}[!t]
		\centering
		\subfloat[]{\includegraphics[width=0.33\linewidth]{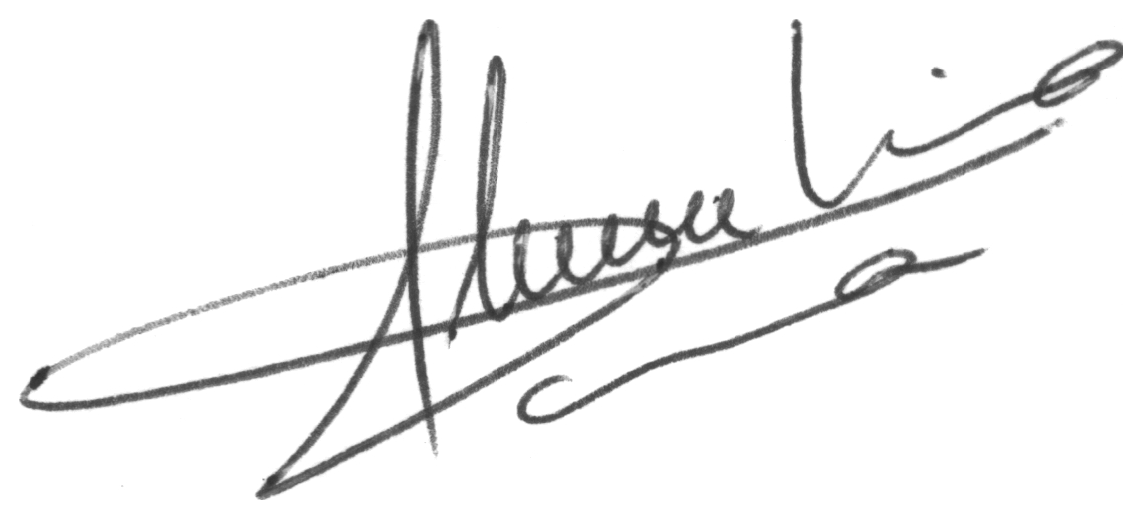}%
			\label{sig_1_default}}
		\subfloat[]{\includegraphics[width=0.33\linewidth]{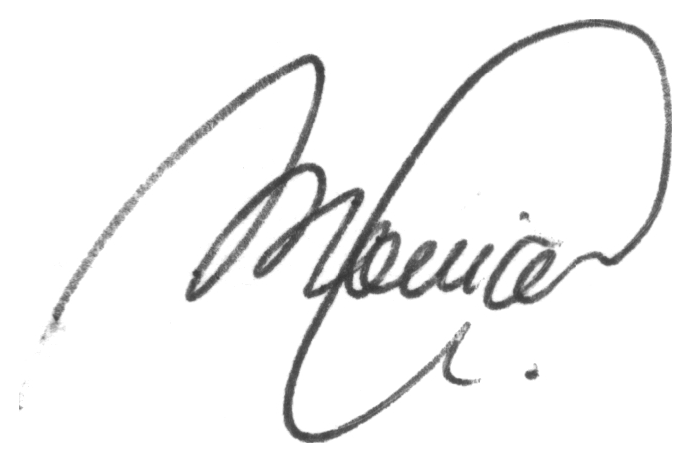}%
			\label{sig_2_default}}
		\subfloat[]{\includegraphics[width=0.33\linewidth]{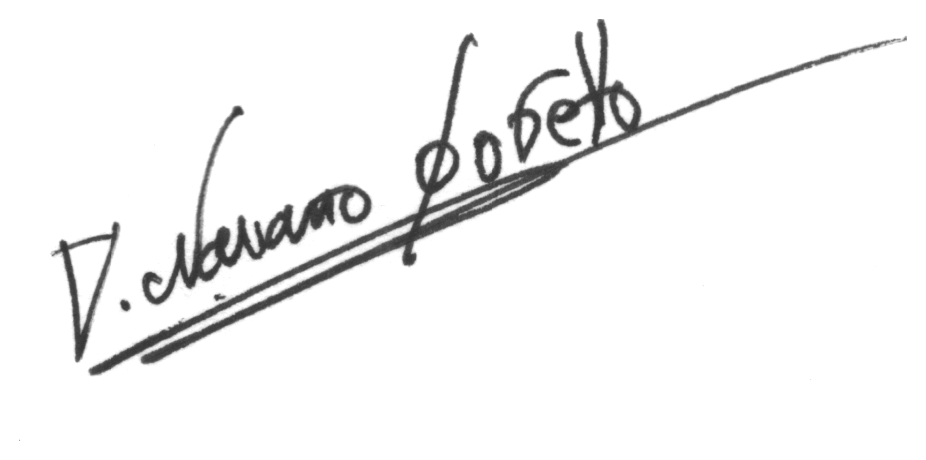}%
			\label{sig_5_default}}\\
		\subfloat[]{\includegraphics[width=0.33\linewidth]{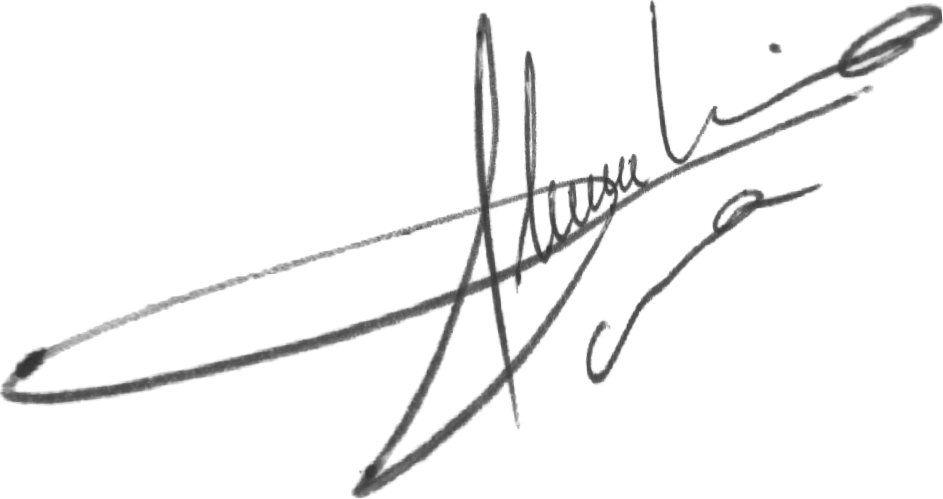}%
			\label{sig_1_dist_default}}
		\subfloat[]{\includegraphics[width=0.33\linewidth]{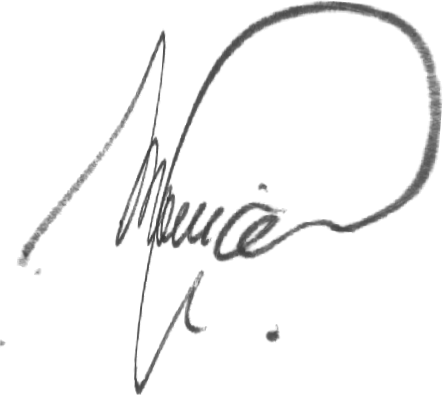}%
			\label{sig_2_dist_default}}
		\subfloat[]{\includegraphics[width=0.33\linewidth]{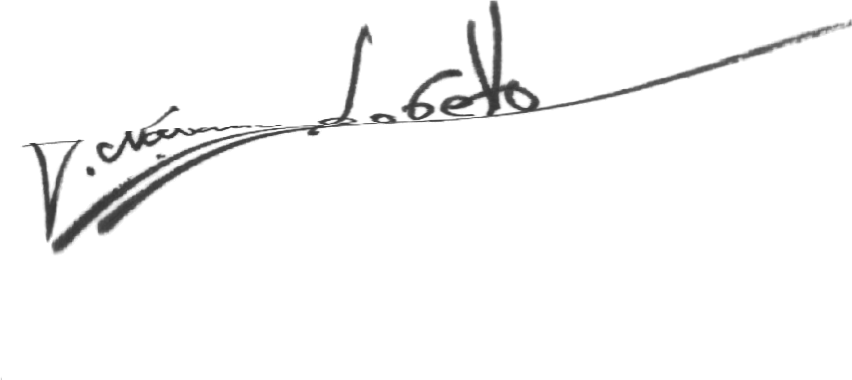}%
			\label{sig_5_dist_default}}\\
		\subfloat[]{\includegraphics[width=0.33\linewidth]{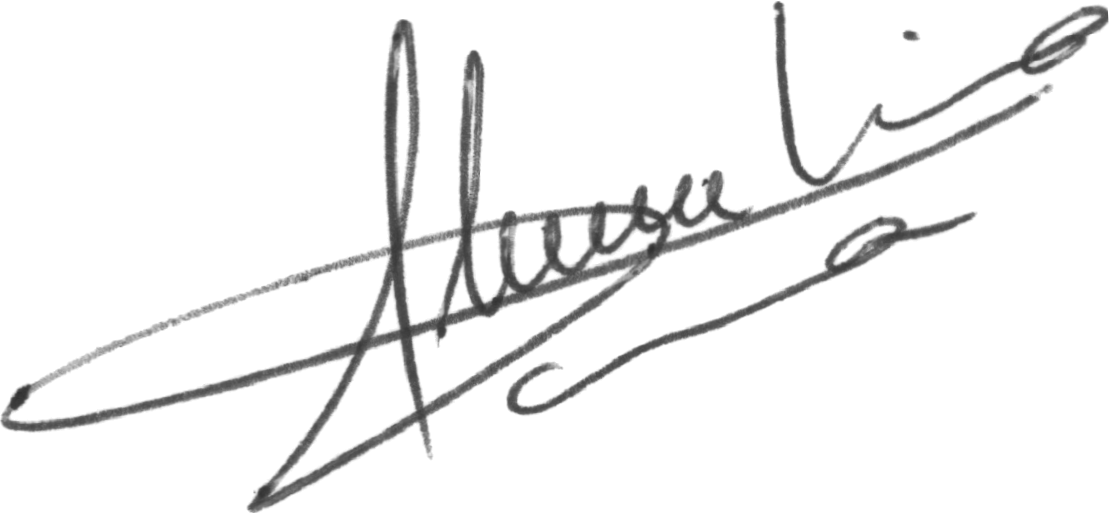}%
			\label{sig_1_sigvarLR_gpds_20dldt_avg}}
		\subfloat[]{\includegraphics[width=0.33\linewidth]{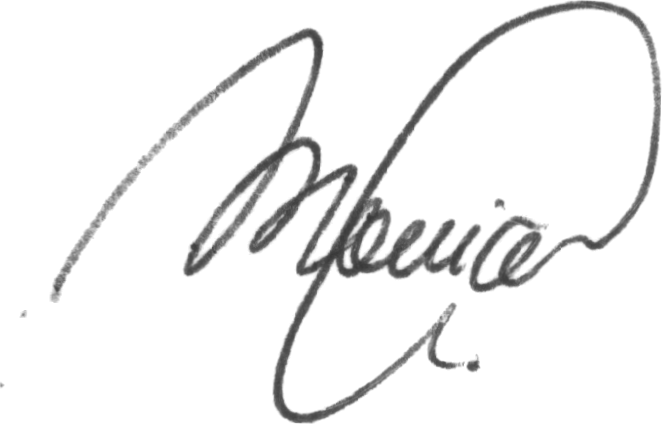}%
			\label{sig_2_sigvarLR_gpds_20dldt_avg}}
		\subfloat[]{\includegraphics[width=0.33\linewidth]{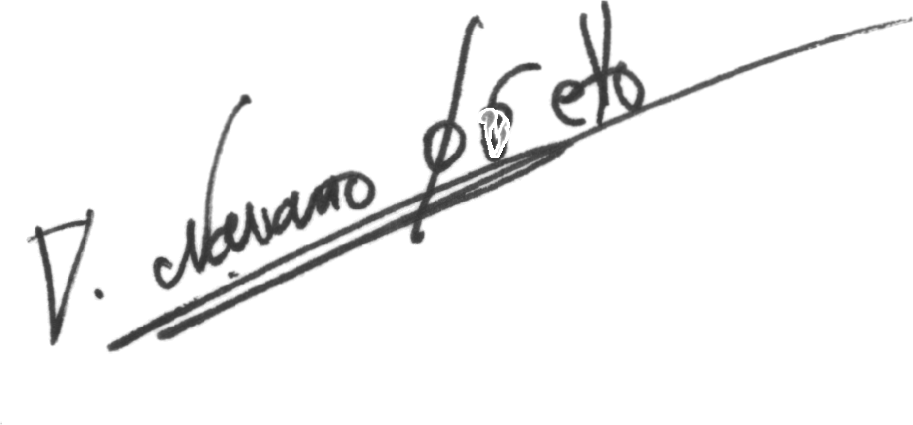}%
			\label{sig_5_sigvarLR_gpds_20dldt_avg}}\\
		\caption{Genuine signatures (a-c), duplicates (d-f) generated by Duplicator with default parameters, and duplicates (g-i) generated by Duplicator with the average parameter vector {${\pi}_{dup}$}.}
		\label{fig:distorted_duplicates_default}
	\end{figure}

	Although the average parameter vector ${\pi}_{dup}$ does not represent the individual variability of each writer, it does represent the variability better than the ${{\pi}}_{def}$. Besides, duplicates created using ${{\pi}}_{def}$ may introduce some distortions that may lead to unnatural signature duplicates (Figure \ref{fig:distorted_duplicates_default}). As can be seen in Figure \ref{fig:distorted_duplicates_default} and Table \ref{tab:si_param20DL}, minimizing the silhouette index helps improve the quality of duplicates. Furthermore, the average parameter vector ${\pi}_{gauss}$ helps generate synthetic samples in the feature space that {resemble} the original signatures.
	
	Since the average parameter vectors ${\pi}_{dup}$ and ${\pi}_{gauss}$ were optimized using a subset of GPDS-960, the lowest ${\left|\Delta\right|_{avg}}$ was achieved using the GPDS-300 dataset. Even using the signatures of just 20 writers, the method was able to represent the writer variability of 300 different writers. As well, using just the parameters optimized in the GPDS dataset, the proposed method proved able to better represent the writer variability in the CEDAR and MCYT-75 datasets.
	
	As can be seen, ${\pi}_{def}$, ${\pi}_{dup}$, and ${\pi}_{gauss}$ have difficulty representing the variability of the writers present in the CEDAR dataset. In the case of ${\pi}_{dup}$ and ${\pi}_{gauss}$, this may be due to the kind of signatures present in the GPDS dataset, where the parameters have been optimized. Since the GPDS dataset has highly sparse signatures in the feature space, it is expected that the parameter vector will be able to reproduce the sparsity of signatures with the same nature. According to Table \ref{tab:si_param20DL}, the writers in the CEDAR dataset present a smaller sparsity (${co}_{avg}$) in the feature space than those writers in the other two datasets. Since we applied two indirect transformations and one direct one in the feature space, it can be seen that the nature of the transformations applied in the feature space also impacts the quality of synthetic samples. The results presented in Table \ref{tab:si_param20DL} suggest that it is more difficult to generate synthetic samples that are close to the low-sparsity clusters than it is to generate them close to high-sparsity ones.
	
	\subsection{Validation at Performance Level Using the Duplicator}
	\label{sec:performance_validation_duplicator}
	
	To respect the constraints imposed by problems in the real world, where few genuine signatures per writer are available, no more than three genuine signatures per writer were used in these experiments to train the Writer-Dependent SVMs of the signature verification system described in Section \ref{sec:verification}. The genuine signatures of other writers of the dataset were used as random forgeries to train the classifiers. Each genuine signature and forgery was randomly selected for training. During the testing, the genuine signatures and forgeries were randomly selected as well.
	
	To better assess the impacts of the number of duplicates in reducing the EER, for each genuine signature, we created up to 22 duplicates. Furthermore, to compare the performance using the duplicates with that achieved by Hafemann et al. \cite{Hafemann2017a}, we used the same experimental protocol. In our case, we used fewer original genuine signatures for training and included the duplicates during the training process. Table \ref{tab:dataset_separation} summarizes the separation of the dataset into training and testing. For each number of duplicates, the experiment was repeated 10 times, and the average EER and standard deviation were reported for the three datasets described in Section \ref{sec:databases}. As in the previous experiments, duplicates were created using ${\pi}_{def}$ and ${\pi}_{dup}$.
		
	\begin{table*}[!t]
		\renewcommand{\arraystretch}{1.1}
		\caption{Dataset separation into training and testing. The number of genuine signatures G, random forgeries R, and skilled forgeries S are specified. The number of duplicates used for training also is specified.}
		\label{tab:dataset_separation}
		\centering
		\begin{tabular}{lllllll}
			\hline
			\multirow{2}{*}{\textbf{Dataset}} & \multicolumn{2}{c}{\textbf{Training set}} &
			{} &
			\multicolumn{3}{c}{\textbf{Testing set}} \\
			\cline{2-3}
			\cline{5-7}
			\multicolumn{1}{l}{} & 
			\multicolumn{1}{l}{G} & 
			\multicolumn{1}{l}{R} & {} &
			\multicolumn{1}{l}{G} & 
			\multicolumn{1}{l}{R} &
			\multicolumn{1}{l}{S} \\
			\cline{5-7}
			\hline
			{GPDS-300}  & r $\in\{1,...,3\}$ + r $\times\left(\text{d} \in\{0,...,22\}\right)$  & {$\left(14\times581\right)  + \left(14 \times581\times\text{d}\right)$} & {} & {10} &{10} & {10}\\
			{MCYT-75}  & r $\in\{1,...,3\}$ + r $\times\left(\text{d} \in\{0,...,22\}\right)$ & {$\left(10\times74\right)  + \left(10 \times74\times\text{d}\right)$} & {} & {5} &{-} & {15}\\
			{CEDAR}  & r $\in\{1,...,3\}$ + r $\times\left(\text{d} \in\{0,...,22\}\right)$ & {$\left(12\times54\right)  + \left(12 \times54\times\text{d}\right)$} & {} & {10} &{-} & {10}\\
			\hline
		\end{tabular}
	\end{table*}
	
	For each random forgery used to train the SVMs, from 0 to 22 duplicates were also used, meaning that the random forgeries and the corresponding  duplicates were used for training.  
	In GPDS, 14 genuine signatures of 581 other writers and the corresponding duplicates are used as random forgeries. For example, if we use 14 genuine signatures with 22 duplicates each for 581 writers, we have the original random forgeries, plus the corresponding duplicates (($14\times581) +(14\times581\times22) =187,082$).
	The classifiers were tested using 10 genuine signatures, 10 random forgeries and 10 skilled forgeries per writer from subset $\mathcal{E_T}$.
	
	For the MCYT-75 dataset, 10 genuine signatures of 74 other writers and the corresponding number of duplicates were used as random forgeries. The classifiers were tested using 5 genuine signatures and 15 skilled forgeries. For the CEDAR dataset, 12 genuine signatures of 54 other writers were used as random forgeries. The classifiers were tested using 10 genuine signatures and 10 skilled forgeries. 
	
\begin{figure}[!t]
		\pgfplotsset{cycle list/Dark2-6}
		\centering
		\begin{tikzpicture}[scale=0.475] 
		\begin{axis}[ 
		xlabel=Number of duplicates per sample,            
		xmin=0,
		xmax = 23,
		minor y tick num={1},
		minor x tick num={1},
		ylabel=Average EER (\%),
		ymin = 0,
		ymax = 6.5,
		ystep = 1,
		label style={font=\Large},
		tick label style={font=\Large},	
		legend columns=1,
		legend style={at={(0.7,0.85)},draw=none,anchor=center,align=center,font=\Large},
		cycle multi list={
			Dark2-6\nextlist
			[1 of]mark list
		},
		width=\textwidth]
		
		\addplot
		plot [error bars/.cd, y dir = both, y explicit]
		table[row sep=crcr, x index=0, y index=1, x error index=0, y error index=2,]{
			0  5.71 	0.38\\
			1  5.35 	0.36\\
			2  4.47		0.25\\
			3  4.34		0.32\\
			4  3.77		0.16\\
			5  3.42		0.22\\
			6  3.40		0.23\\
			7  3.09 	0.20\\
			8  2.82		0.21\\
			9  2.70		0.15\\
			10 2.66		0.17\\
			11 2.49		0.19\\
			12 2.48		0.14\\
			13 2.31		0.20\\
			14 2.25 	0.10\\
			15 2.12		0.18\\
			16 2.22		0.11\\
			17 2.17		0.24\\
			18 2.09		0.20\\
			19 1.85		0.17\\
			20 1.99		0.22\\
			21 1.92		0.18\\
			22 1.79		0.18\\
		};
		\addlegendentry{1 Genuine Samples, RBF SVM, {$\pi_{def}$}}
		
		\addplot
		plot [error bars/.cd, y dir = both, y explicit]
		table[row sep=crcr, x index=0, y index=1, x error index=0, y error index=2,]{
			0  4.01 	0.24\\
			1  3.42 	0.24\\
			2  3.12		0.20\\
			3  2.87		0.15\\
			4  2.67		0.15\\
			5  2.53		0.13\\
			6  2.27		0.19\\
			7  2.08 	0.17\\
			8  2.04		0.17\\
			9  1.89		0.18\\
			10 1.79		0.14\\
			11 1.70		0.15\\
			12 1.65		0.13\\
			13 1.69		0.15\\
			14 1.54 	0.20\\
			15 1.47		0.08\\
			16 1.41		0.14\\
			17 1.35		0.10\\
			18 1.37		0.15\\
			19 1.30		0.14\\
			20 1.22		0.16\\
			21 1.16		0.14\\
			22 1.14		0.16\\
		};
		\addlegendentry{2 Genuine Samples, RBF SVM, {$\pi_{def}$}}
		
		\addplot
		plot [error bars/.cd, y dir = both, y explicit]
		table[row sep=crcr, x index=0, y index=1, x error index=0, y error index=2,]{
			0  3.38 	0.15\\
			1  2.84 	0.24\\
			2  2.58		0.21\\
			3  2.50		0.16\\
			4  2.24		0.12\\
			5  2.03		0.21\\
			6  1.83		0.18\\
			7  1.77 	0.14\\
			8  1.59		0.10\\
			9  1.46		0.09\\
			10 1.40		0.14\\
			11 1.35		0.14\\
			12 1.25		0.13\\
			13 1.23		0.16\\
			14 1.14 	0.13\\
			15 1.14		0.14\\
			16 1.02		0.15\\
			17 1.03		0.11\\
			18 1.04		0.11\\
			19 0.91		0.13\\
			20 0.83		0.09\\
			21 0.91		0.14\\
			22 0.92		0.10\\
		};
		\addlegendentry{3 Genuine Samples, RBF SVM, {$\pi_{def}$}}
		
		\addplot
		plot [dotted, mark=square*, error bars/.cd, y dir = both, y explicit]
		table[row sep=crcr, x index=0, y index=1, x error index=0, y error index=2,]{
			0  5.90		0.22\\
			1  4.08		0.16\\
			2  3.42		0.21\\
			3  2.94		0.13\\
			4  2.67		0.12\\
			5  2.38		0.21\\
			6  2.22		0.18\\
			7  2.08		0.14\\
			8  1.88		0.14\\
			9  1.87		0.15\\
			10 1.73		0.08\\
			11 1.56		0.06\\
			12 1.61		0.16\\
			13 1.54		0.13\\
			14 1.43		0.06\\
			15 1.42		0.09\\
			16 1.27		0.11\\
			17 1.22		0.12\\
			18 1.26		0.17\\
			19 1.18		0.17\\
			20 1.19		0.12\\
			21 1.10		0.14\\
			22 1.08		0.17\\
		};
		\addlegendentry{1 Genuine Samples, RBF SVM, {$\pi_{dup}$}}

		\addplot
		plot [dotted, mark=square*, error bars/.cd, y dir = both, y explicit]
		table[row sep=crcr, x index=0, y index=1, x error index=0, y error index=2,]{
			0  4.01		0.24\\
			1  2.95		0.26\\
			2  2.49		0.22\\
			3  2.23		0.15\\
			4  1.87		0.10\\
			5  1.66		0.16\\
			6  1.55		0.14\\
			7  1.46		0.10\\
			8  1.24		0.09\\
			9  1.19		0.08\\
			10 1.02		0.10\\
			11 0.99		0.11\\
			12 0.97		0.08\\
			13 0.89		0.07\\
			14 0.84		0.09\\
			15 0.79		0.11\\
			16 0.74		0.10\\
			17 0.70		0.11\\
			18 0.60		0.09\\
			19 0.60		0.12\\
			20 0.50		0.10\\
			21 0.50		0.07\\
			22 0.47		0.07\\
		};
		\addlegendentry{2 Genuine Samples, RBF SVM, {$\pi_{dup}$}}

		\addplot
		plot [dotted, mark=square*, error bars/.cd, y dir = both, y explicit]
		table[row sep=crcr, x index=0, y index=1, x error index=0, y error index=2,]{
			0  3.28		0.24\\
			1  2.47		0.15\\
			2  1.99		0.11\\
			3  1.73		0.13\\
			4  1.54		0.17\\
			5  1.44		0.16\\
			6  1.19		0.10\\
			7  1.09		0.13\\
			8  0.89		0.11\\
			9  0.86		0.08\\
			10 0.79		0.12\\
			11 0.68		0.09\\
			12 0.56		0.10\\
			13 0.51		0.08\\
			14 0.49		0.08\\
			15 0.38		0.10\\
			16 0.40		0.07\\
			17 0.40		0.07\\
			18 0.33		0.06\\
			19 0.28		0.06\\
			20 0.25		0.04\\
			21 0.27		0.06\\
			22 0.24		0.07\\
		};
		\addlegendentry{3 Genuine Samples, RBF SVM, {$\pi_{dup}$}}

		\end{axis}
		\end{tikzpicture}
		
		\caption{Average EER achieved using GPDS-300 dataset, Signet-F and the Proposed Method with a Large Range of parameters.}
		\label{fig:eer_gpds300_duplicator_sigvarLR_0999_20dldt} 
	\end{figure}
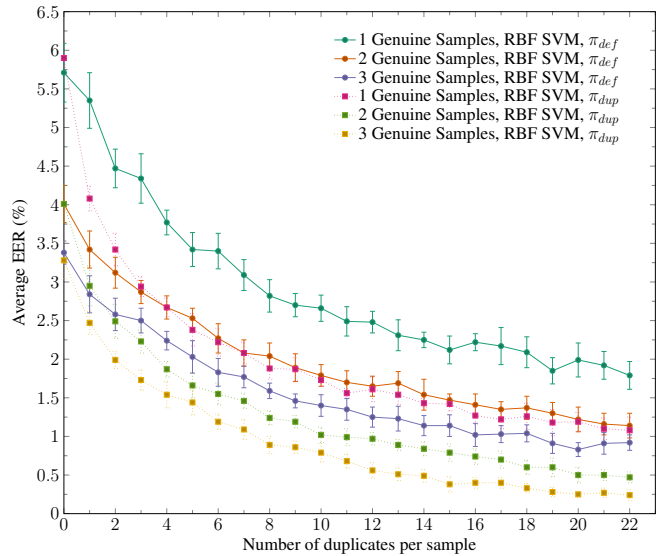

	Figure \ref{fig:eer_gpds300_duplicator_sigvarLR_0999_20dldt} shows the average EER of each number of duplicates in the GPDS dataset. As expected, the greater the number of duplicates, the smaller the EER. It should, however, be noted that the system trained with the duplicates created with ${\pi}_{dup}$ outperforms the one trained with ${\pi}_{def}$ in all scenarios. This is somewhat attributable to the quality of the duplicates created using the optimized parameters. As shown in the previous section, the default parameters may sometimes lead to unnatural duplicates that negatively affect the performance of the verification system. Like the real signatures, the duplicates can also provide complementary information about the signatures of a writer. 
	
\begin{figure}[!t]
		\pgfplotsset{cycle list/Dark2-6}
		\centering
		\begin{tikzpicture}[scale=0.475] 
		\begin{axis}[ 
		xlabel=Number of duplicates per sample,            
		xmin=0,
		xmax = 23,
		minor y tick num={1},
		minor x tick num={1},
		ylabel=Average EER (\%),
		ymin = 0,
		ymax = 10,
		ystep = 1,
		label style={font=\Large},
		tick label style={font=\Large},  	
		legend columns=1,
		legend style={at={(0.7,0.85)},draw=none,anchor=center,align=center,font=\Large},
		cycle multi list={
			Dark2-6\nextlist
			[1 of]mark list
		},
		width=\textwidth]
		
		\addplot
		plot [error bars/.cd, y dir = both, y explicit]
		table[row sep=crcr, x index=0, y index=1, x error index=0, y error index=2,]{
			0  9.63 	1.54\\
			1  7.27 	0.86\\
			2  5.65		0.93\\
			3  5.19		0.53\\
			4  4.40		0.95\\
			5  3.76		0.57\\
			6  3.42		0.78\\
			7  3.30 	0.59\\
			8  2.70		0.52\\
			9  2.55		0.71\\
			10 2.77		0.69\\
			11 2.33		0.38\\
			12 2.31		0.44\\
			13 2.32		0.63\\
			14 1.94 	0.44\\
			15 1.93		0.25\\
			16 1.75		0.41\\
			17 1.92		0.68\\
			18 1.69		0.38\\
			19 1.67		0.42\\
			20 1.64		0.35\\
			21 1.58		0.51\\
			22 1.55		0.50\\
		};
		\addlegendentry{1 Genuine Samples, RBF SVM, {${\pi}_{def}$}}

		\addplot
		plot [error bars/.cd, y dir = both, y explicit]
		table[row sep=crcr, x index=0, y index=1, x error index=0, y error index=2,]{
			0  6.96 	0.95\\
			1  4.28 	0.85\\
			2  3.49		0.72\\
			3  3.34		0.71\\
			4  2.63		0.65\\
			5  2.57		0.56\\
			6  2.00		0.49\\
			7  2.02 	0.46\\
			8  2.04		0.72\\
			9  1.71		0.35\\
			10 1.63		0.48\\
			11 1.46		0.31\\
			12 1.66		0.45\\
			13 1.47		0.33\\
			14 1.28 	0.42\\
			15 1.28		0.29\\		
			16 1.18		0.30\\
			17 1.27		0.42\\
			18 1.29		0.36\\		
			19 1.27		0.52\\
			20 1.11		0.45\\
			21 0.96		0.29\\
			22 0.91		0.28\\
		};
		\addlegendentry{2 Genuine Samples, RBF SVM, {${\pi}_{def}$}}

		\addplot
		plot [error bars/.cd, y dir = both, y explicit]
		table[row sep=crcr, x index=0, y index=1, x error index=0, y error index=2,]{
			0  5.66 	0.90\\
			1  3.25 	0.86\\
			2  2.69		0.69\\
			3  2.29		0.56\\
			4  2.19		0.51\\
			5  1.95		0.47\\
			6  1.79		0.61\\
			7  1.64 	0.58\\
			8  1.74		0.53\\
			9  1.66		0.46\\
			10 1.29		0.37\\
			11 1.39		0.40\\
			12 1.25		0.39\\
			13 1.00		0.43\\
			14 1.37 	0.36\\
			15 1.07		0.36\\
			16 0.92		0.46\\
			17 0.74		0.22\\
			18 0.73		0.20\\
			19 1.03		0.37\\
			20 0.99		0.35\\
			21 0.70		0.32\\
			22 1.04		0.47\\
		};
		\addlegendentry{3 Genuine Samples, RBF SVM, {${\pi}_{def}$}}		
		
		\addplot
		plot [dotted, mark=square*, error bars/.cd, y dir = both, y explicit]
		table[row sep=crcr, x index=0, y index=1, x error index=0, y error index=2,]{
			0  8.70		0.91\\
			1  5.26		0.81\\
			2  4.52		1.05\\
			3  3.58		0.52\\
			4  2.92		0.81\\
			5  2.50		0.36\\
			6  2.41		0.61\\
			7  2.32		0.57\\
			8  2.11		0.62\\
			9  1.47		0.38\\
			10 1.47		0.46\\
			11 1.47		0.46\\
			12 1.58		0.31\\
			13 1.15		0.37\\
			14 1.03		0.26\\
			15 0.92		0.29\\
			16 0.98		0.33\\
			17 0.76		0.32\\
			18 0.97		0.19\\
			19 0.69		0.36\\
			20 0.74		0.41\\
			21 0.82		0.43\\
			22 0.90		0.49\\
		};
		\addlegendentry{1 Genuine Samples, RBF SVM, {${\pi}_{dup}$}}
		
		\addplot
		plot [dotted, mark=square*, error bars/.cd, y dir = both, y explicit]
		table[row sep=crcr, x index=0, y index=1, x error index=0, y error index=2,]{
			0  5.42		0.81\\
			1  4.05		0.90\\
			2  2.74		0.59\\
			3  2.28		0.59\\
			4  1.84		0.36\\
			5  1.62		0.52\\
			6  1.47		0.32\\
			7  1.06		0.43\\
			8  1.01		0.41\\
			9  0.71		0.37\\
			10 0.57		0.27\\
			11 0.66		0.41\\
			12 0.63		0.25\\
			13 0.52		0.32\\
			14 0.48		0.29\\
			15 0.39		0.26\\
			16 0.32		0.20\\
			17 0.44		0.27\\
			18 0.19		0.22\\
			19 0.30		0.22\\
			20 0.18		0.18\\
			21 0.20		0.19\\
			22 0.12		0.12\\
		};
		\addlegendentry{2 Genuine Samples, RBF SVM, {${\pi}_{dup}$}}
		
		\addplot
		plot [dotted, mark=square*, error bars/.cd, y dir = both, y explicit]
		table[row sep=crcr, x index=0, y index=1, x error index=0, y error index=2,]{
			0  4.04		0.95\\
			1  2.81		0.61\\
			2  1.88		0.62\\
			3  1.48		0.61\\
			4  1.33		0.37\\
			5  1.08		0.38\\
			6  0.82		0.47\\
			7  0.81		0.21\\
			8  0.51		0.24\\
			9  0.45		0.27\\
			10 0.46		0.30\\
			11 0.53		0.32\\
			12 0.30		0.25\\
			13 0.20		0.25\\
			14 0.27		0.20\\
			15 0.14		0.11\\
			16 0.17		0.15\\
			17 0.11		0.14\\
			18 0.13		0.15\\
			19 0.14		0.14\\
			20 0.10		0.11\\
			21 0.12		0.09\\
			22 0.07		0.07\\
		};
		\addlegendentry{3 Genuine Samples, RBF SVM, {${\pi}_{dup}$}}
		
		\end{axis}
		\end{tikzpicture}
		\caption{Average EER achieved using MCYT-75 dataset, Signet-F and the Proposed Method with a Large Range of parameters.}
		\label{fig:eer_mcyt75_duplicator_sigvarLR_0999_20dldt} 
	\end{figure}
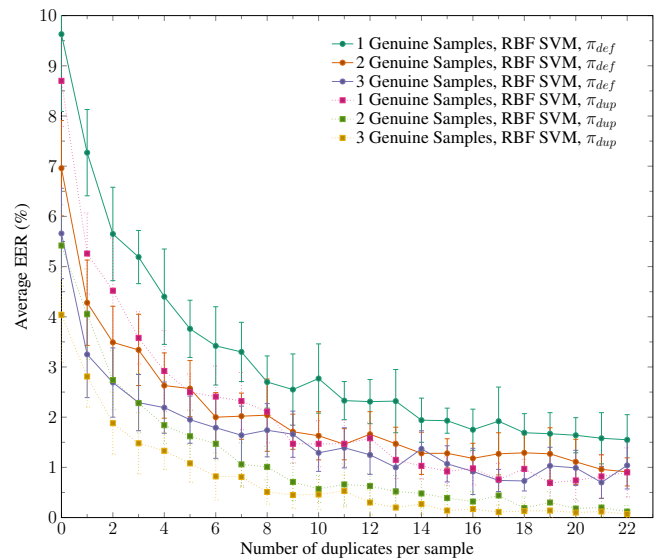

	Figure \ref{fig:eer_mcyt75_duplicator_sigvarLR_0999_20dldt} shows the average EER of each number of duplicates in the MCYT-75 dataset. As observed earlier, when the number of duplicates increases, the EER drops {simultaneously}. This experiment also {showed} the generalization capability of the proposed method. Even though {${\pi}_{dup}$} was optimized on a subset of GPDS-960, it was able to produce high quality duplicates for MCYT-75 as well, since this dataset contains {highly variable} signatures, similar to those found in the  GPDS dataset. 
	
\begin{figure}[!t]
		\pgfplotsset{cycle list/Dark2-6}
		\centering
		\begin{tikzpicture}[scale=0.475] 
		\begin{axis}[ 
		xlabel=Number of duplicates per sample,            
		xmin=0,
		xmax = 23,
		minor y tick num={1},
		minor x tick num={1},
		ylabel=Average EER (\%),
		ymin = 0,
		ymax = 12,
		ystep = 1,
		label style={font=\Large},
		tick label style={font=\Large},
		legend columns=1,
		legend style={at={(0.7,0.85)},draw=none,anchor=center,align=center,font=\Large},
		cycle multi list={
			Dark2-6\nextlist
			[1 of]mark list
		},
		width=\textwidth]
		
		\addplot
		plot [error bars/.cd, y dir = both, y explicit]
		table[row sep=crcr, x index=0, y index=1, x error index=0, y error index=2,]{
			0  11.33 	1.84\\
			1  7.91 	1.12\\
			2  7.03		1.01\\
			3  6.61		0.62\\
			4  5.88		1.15\\
			5  5.97		0.91\\
			6  5.68		0.61\\
			7  5.30 	0.83\\
			8  4.85		0.78\\
			9  4.69		0.73\\
			10 4.59		0.53\\
			11 5.02		0.78\\
			12 4.64		0.55\\
			13 4.07		0.60\\
			14 4.29 	0.43\\
			15 4.52		0.73\\
			16 4.29		0.54\\
			17 4.31		0.57\\
			18 4.17		0.46\\
			19 4.12		0.40\\
			20 3.91		0.36\\
			21 4.03		0.76\\
			22 4.29		0.65\\
		};
		\addlegendentry{1 Genuine Sample, RBF SVM, {${\pi}_{def}$}}
		
		\addplot
		plot [error bars/.cd, y dir = both, y explicit]
		table[row sep=crcr, x index=0, y index=1, x error index=0, y error index=2,]{
			0  8.63 	1.04\\
			1  5.42 	0.69\\
			2  4.55		0.67\\
			3  4.93		0.58\\
			4  5.01		0.75\\
			5  4.43		0.48\\
			6  4.49		0.37\\
			7  4.42 	0.66\\
			8  4.00		0.40\\
			9  4.21		0.71\\
			10 3.97		0.54\\
			11 3.83		0.48\\
			12 3.56		0.52\\
			13 3.73		0.71\\
			14 3.50 	0.29\\
			15 3.81		0.41\\
			16 3.39		0.43\\
			17 3.73		0.45\\
			18 3.51		0.33\\
			19 3.35		0.32\\
			20 3.14		0.44\\
			21 3.26		0.42\\
			22 3.72		0.43\\
		};
		\addlegendentry{2 Genuine Samples, RBF SVM, {${\pi}_{def}$}}
		
		\addplot
		plot [error bars/.cd, y dir = both, y explicit]
		table[row sep=crcr, x index=0, y index=1, x error index=0, y error index=2,]{
			0  7.39		0.65\\
			1  5.29 	0.64\\
			2  5.16		0.75\\
			3  4.19		0.47\\
			4  4.39		0.54\\
			5  4.12		0.74\\
			6  3.56		0.60\\
			7  3.93 	0.62\\
			8  3.59		0.56\\
			9  3.85		0.52\\
			10 3.70		0.57\\
			11 3.58		0.47\\
			12 3.27		0.49\\
			13 3.63		0.59\\
			14 3.46 	0.60\\
			15 3.51		0.63\\
			16 3.17		0.26\\
			17 3.21		0.53\\
			18 3.12		0.44\\
			19 3.30		0.26\\
			20 3.36		0.39\\
			21 3.17		0.50\\
			22 3.04		0.33\\
		};
		\addlegendentry{3 Genuine Samples, RBF SVM, {${\pi}_{def}$}}
		
		\addplot
		plot [dotted, mark=square*, error bars/.cd, y dir = both, y explicit]
		table[row sep=crcr, x index=0, y index=1, x error index=0, y error index=2,]{
			0 11.09		1.43\\
			1  6.99		1.18\\
			2  5.73		1.12\\
			3  5.21		0.89\\
			4  5.20		0.72\\
			5  4.56		0.62\\
			6  4.41		0.74\\
			7  4.07		0.59\\
			8  4.61		0.67\\
			9  4.42		0.28\\
			10 4.38		0.41\\
			11 4.15		0.76\\
			12 4.06		0.63\\
			13 3.95		0.36\\
			14 3.76		0.65\\
			15 3.66		0.70\\
			16 3.98		0.33\\
			17 3.70		0.30\\
			18 3.80		0.72\\
			19 3.49		0.42\\
			20 3.25		0.36\\
			21 3.37		0.51\\
			22 3.39		0.40\\
		};
		\addlegendentry{1 Genuine Samples, RBF SVM, {${\pi}_{dup}$}}
		
		\addplot
		plot [dotted, mark=square*, error bars/.cd, y dir = both, y explicit]
		table[row sep=crcr, x index=0, y index=1, x error index=0, y error index=2,]{
			0  8.48		1.12\\
			1  5.67		0.78\\
			2  4.86		0.88\\
			3  4.62		0.76\\
			4  4.24		0.47\\
			5  4.02		0.67\\
			6  3.59		0.44\\
			7  3.55		0.64\\
			8  3.49		0.69\\
			9  3.32		0.58\\
			10 3.26		0.55\\
			11 3.44		0.49\\
			12 3.12		0.45\\
			13 2.82		0.69\\
			14 2.85		0.40\\
			15 3.08		0.47\\
			16 2.94		0.39\\
			17 3.01		0.43\\
			18 2.85		0.34\\
			19 2.86		0.39\\
			20 2.85		0.35\\
			21 2.37		0.43\\
			22 2.93		0.37\\
		};
		\addlegendentry{2 Genuine Samples, RBF SVM, {${\pi}_{dup}$}}
		
		\addplot
		plot [dotted, mark=square*, error bars/.cd, y dir = both, y explicit]
		table[row sep=crcr, x index=0, y index=1, x error index=0, y error index=2,]{
			0  6.83		0.95\\
			1  4.43		0.66\\
			2  4.40		0.63\\
			3  3.96		0.69\\
			4  3.70		0.48\\
			5  3.58		0.65\\
			6  3.35		0.66\\
			7  3.02		0.48\\
			8  3.05		0.32\\
			9  2.96		0.42\\
			10 2.69 	0.47\\
			11 2.58		0.41\\
			12 2.60		0.57\\
			13 2.41		0.54\\
			14 2.60		0.37\\
			15 2.35		0.40\\
			16 2.40		0.31\\
			17 2.20		0.42\\
			18 2.16		0.48\\
			19 2.20		0.28\\
			20 2.22		0.30\\
			21 2.18		0.35\\
			22 2.39		0.40\\
		};
		\addlegendentry{3 Genuine Samples, RBF SVM, {${\pi}_{dup}$}}

		\end{axis}
		\end{tikzpicture}
		\caption{Average EER achieved using CEDAR dataset, Signet-F and the Proposed Method with a Large Range of parameters.}
		\label{fig:eer_cedar_duplicator_sigvarLR_0999_20dldt} 
	\end{figure}
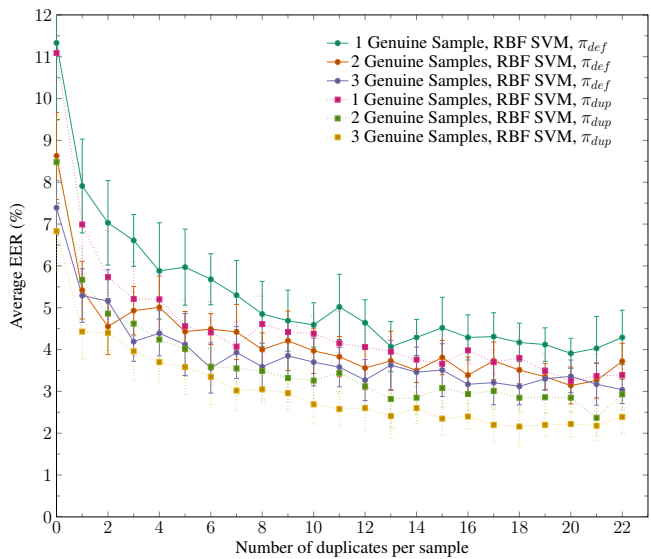

	Figure \ref{fig:eer_cedar_duplicator_sigvarLR_0999_20dldt} shows the performance of each number of duplicates in the CEDAR dataset. As observed in the previous datasets, while the EER follows the same trend, the drop in the EER is, however, more subtle. This corroborates the hypothesis that the distribution of the CEDAR signatures is hard to represent due to the difference between the writer variability of the optimization dataset and the CEDAR dataset.
	Nevertheless, the performance achieved using duplicates generated with {${\pi}_{dup}$} is better than that obtained using ${\pi}_{def}$.

	Table \ref{tab:cmp_results} summarizes the best results achieved in each experiment and compares them with the signature verification system  proposed in \cite{Hafemann2017a}{,} which uses up to 12 genuine signatures{{; the one proposed} in \cite{Zois2019b}{,} which uses 10 and 12 genuine signatures,} and the original duplicator proposed in \cite{Diaz2017}. It should be noted that the results published in \cite{Diaz2017} {and \cite{Zois2019b}} use {other representations}, and therefore, a direct comparison is not possible. As can be seen, the proposed method achieves outstanding results. Using the optimized parameter vector, it achieves state-of-the art results using no more than three genuine signatures. 
	
	Even duplicates generated using an {${\pi}_{dup}$} provide additional data that enable the classifiers to learn to distinguish between genuine signatures and skilled forgeries. Similar to Frias-Martinez et al. \cite{Frias2006} and Diaz et al. \cite{Diaz2017}, the best results were achieved using between 15 and 22 duplicates (Table \ref{tab:cmp_results}).
	This may be an indication of the ideal number of duplicates that can be used to improve the performance of a verification system. Notwithstanding the fact that parameters were optimized using the GPDS dataset, the proposed method was able to generate real duplicates when it was tested using CEDAR and MCYT-75.	
	
	\begin{table}[!t]
		\caption{Summary of the experimental results where \#W, \#S, and \#D stand for the number of writers used for training, the number of genuine samples used for training, and the number of duplicates per sample used for training, respectively.}
		\label{tab:cmp_results}
		\renewcommand{\arraystretch}{1.01}
		\setlength{\tabcolsep}{3pt}
		\begin{tabular}{>{\bfseries}llllllll}
			\hline
			\multicolumn{1}{l}{\bfseries Reference} & \multicolumn{1}{l}{\bfseries Feature} & 
			\multicolumn{1}{l}{\bfseries Classifier} & \multicolumn{1}{l}{\bfseries Dataset} & 
			\multicolumn{1}{l}{\bfseries \#W} & 
			\multicolumn{1}{l}{\bfseries \#S} & 
			\multicolumn{1}{l}{\bfseries \#D} &
			\multicolumn{1}{l}{\bfseries EER (\%)}\\
			\hline
			
			{Diaz et al.,} & {LDerivP} & {SVM} &
			{GPDS} & {300} & {2} & {20} & {21.63}\\
			{2017 \cite{Diaz2017}} & {~} & {~} &
			{~} & {~} & {5} & {20} & {17.19}\\
			{~} & {~} & {~} &
			{~} & {~} & {8} & {20} & {14.58}\\
			
			{~} & {~} & {~} &
			{MCYT} & {75} & {2} & {20} & {16.06}\\
			{~} & {~} & {~} &
			{~} & {~} & {5} & {20} & {11.90}\\
			{~} & {~} & {~} &
			{~} & {~} & {8} & {20} & {9.12}\\ \hline
			
			{Hafemann} & {SigNet-F} & {SVM} &
			{GPDS} & {300} & {5} & {0} & {2.42}\\
			{et al., 2017} & {~} & {~} &
			{~} & {~} & {12} & {0} & {1.69}\\
			
			{\cite{Hafemann2017a}} & {~} & {~} &
			{CEDAR} & {55} & {4} & {0} & {5.92}\\
			{~} & {~} & {~} &
			{~} & {~} & {8} & {0} & {4.77}\\
			{~} & {~} & {~} &
			{~} & {~} & {12} & {0} & {4.63}\\
			
			{~} & {~} & {~} &
			{MCYT} & {75} & {5} & {0} & {3.70}\\ 
			{~} & {~} & {~} &
			{~} & {~} & {10} & {0} & {3.00}\\ \hline
			
			{Zois et al.} & {KSVD/OMP} & {SVM} &
			{GPDS} & {300} & {12} & {0} & {0.70}\\
			{2019 \cite{Zois2019b}} & {($F_3$)} & {~} &
			{CEDAR} & {55} & {10} & {0} & {0.79}\\
			{~} & {~} & {~} &
			{MCYT} & {75} & {10} & {0} & {1.37}\\ \hline
			
			{Baseline} & {SigNet-F} & {SVM} &
			{GPDS} & {300} & {1} & {0} & {5.71}\\
			{(Without} & {~} & {~} &
			{~} & {~} & {2} & {0} &{4.01}\\
			{Duplicates)} & {~} & {~} &
			{~} & {~} & {3} & {0} & {3.38}\\
			
			{~} & {~} & {~} &
			{CEDAR} & {55} & {1} & {0} & {11.33}\\
			{~} & {~} & {~} &
			{~} & {~} & {2} & {0} &{8.63}\\
			{} & {~} & {~} &
			{~} & {~} & {3} & {0} & {7.39}\\
			
			{~} & {~} & {~} &
			{MCYT} & {75} & {1} & {0} & {9.63}\\
			{~} & {~} & {~} &
			{~} & {~} & {2} & {0} & {6.96}\\
			{~} & {~} & {~} &
			{~} & {~} & {3} & {0} & {5.66}\\ \hline
			
			{Duplicator} & {SigNet-F} & {SVM} &
			{GPDS} & {300} & {1} & {22} & {1.79}\\
			{\boldmath{${\pi}_{def}$}} & {~} & {~} &
			{~} & {~} & {2} & {22} & {1.14}\\
			{~} & {~} & {~} &
			{~} & {~} & {3} & {22} & {0.92}\\
			
			{~} & {~} & {~} &
			{CEDAR} & {55} & {1} & {22} & {4.29}\\
			{~} & {~} & {~} &
			{~} & {~} & {2} & {22} & {3.72}\\
			{~} & {~} & {~} &
			{~} & {~} & {3} & {22} & {3.04}\\  

			{~} & {~} & {~} &
			{MCYT} & {75} & {1} & {22} & {1.55}\\
			{~} & {~} & {~} &
			{~} & {~} & {2} & {22} & {0.91}\\
			{~} & {~} & {~} &
			{~} & {~} & {3} & {22} & {1.04}\\ \hline
			
			{Proposed} & {SigNet-F} & {SVM} &
			{GPDS} & {300} & {1} & {22} & {1.08}\\
			{Method} & {~} & {~} &
			{~} & {~} & {2} & {22} & {0.47}\\
			{\boldmath{$20\mathcal{D_L}$}} & {~} & {~} &
			{~} & {~} & {3} & {22} & \textbf{0.24}\\
			
			{\boldmath{${\pi}_{dup}$}} & {~} & {~} &
			{CEDAR} & {55} & {1} & {22} & {3.39}\\
			{~} & {~} & {~} &
			{~} & {~} & {2} & {22} & {2.93}\\
			{~} & {~} & {~} &
			{~} & {~} & {3} & {22} & {\textbf{2.39}}\\
			
			{~} & {~} & {~} &
			{MCYT} & {75} & {1} & {22} & {0.90}\\
			{~} & {~} & {~} &
			{~} & {~} & {2} & {22} & {0.12}\\
			{~} & {~} & {~} &
			{~} & {~} & {3} & {22} & \textbf{0.07}\\ \hline
			
			{Proposed} & {SigNet-F} & {SVM} &
			{GPDS} & {300} & {1} & {22} & {1.04}\\
			{Method} & {~} & {~} &
			{~} & {~} & {2} & {22} & {0.48}\\
			{\boldmath{$20\mathcal{D_L}$}} & {~} & {~} &
			{~} & {~} & {3} & {22} & {\textbf{0.20}}\\
			
			{\boldmath{${\pi}_{gauss}$}} & {~} & {~} &
			{CEDAR} & {55} & {1} & {22} & {2.47}\\
			{~} & {~} & {~} &
			{~} & {~} & {2} & {22} & {1.31}\\
			{~} & {~} & {~} &
			{~} & {~} & {3} & {22} & {\textbf{0.82}}\\
			
			{~} & {~} & {~} &
			{MCYT} & {75} & {1} & {22} & {0.72}\\
			{~} & {~} & {~} &
			{~} & {~} & {2} & {22} & {0.12}\\
			{~} & {~} & {~} &
			{~} & {~} & {3} & {22} & {\textbf{0.01}}\\ 
			
			\hline
		\end{tabular}
	\end{table}
	
	\subsection{{Validation at Performance Level Using the Gaussian Filter}}
	\label{sec:performance_validation_gaussianfilter}
	
	As mentioned before, we hypothesized that adding some intraclass variability in the image space would induce some intraclass variability in the feature space as well, which allows measuring the fitness function in the feature space. To prove that this a valid strategy, we also
	present a method in which the feature data points increase when the duplicate feature points are placed on the feature domain.
	This was implemented by perturbing the genuine feature vector with correlated noise, which was added using a low-pass Gaussian filter (Equation \ref{eq:gaussian_filter}). 
	
	We used the same protocol adopted in Section \ref{sec:performance_validation_duplicator} to evaluate the performance in the GPDS-300, MCYT-75, and CEDAR datasets. Instead of using the duplicates, we used the feature vectors with noise to train the SVM classifiers. To generate the feature vectors with noise, the standard deviation $\sigma$ was defined considering the average parameter vector ${\pi}_{gauss}$.
	
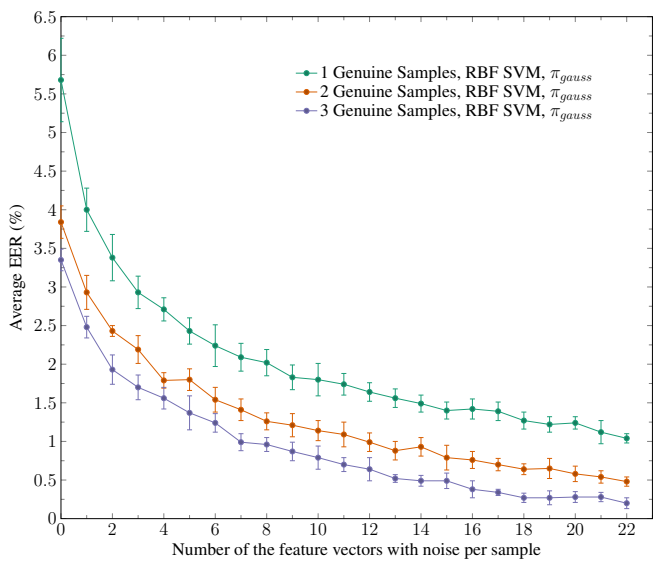
\begin{figure}[!t]
		\pgfplotsset{cycle list/Dark2-6}
		\centering
		\begin{tikzpicture}[scale=0.475] 
		\begin{axis}[ 
		xlabel=Number of the feature vectors with noise per sample,            
		xmin=0,
		xmax = 23,
		minor y tick num={1},
		minor x tick num={1},
		ylabel=Average EER (\%),
		ymin = 0,
		ymax = 6.5,
		ystep = 1,
		label style={font=\Large},
		tick label style={font=\Large},
		legend columns=1,
		legend style={at={(0.65,0.85)},draw=none,anchor=center,align=center,font=\Large},
		cycle multi list={
			Dark2-6\nextlist
			[1 of]mark list
		},
		width=\textwidth]
		
		\addplot
		plot [error bars/.cd, y dir = both, y explicit]
		table[row sep=crcr, x index=0, y index=1, x error index=0, y error index=2,]{
			0  5.68 	0.54\\
			1  4.00 	0.28\\
			2  3.38 	0.30\\
			3  2.93 	0.21\\
			4  2.71 	0.15\\
			5  2.43 	0.17\\
			6  2.24 	0.27\\
			7  2.09 	0.18\\
			8  2.02 	0.17\\
			9  1.83 	0.16\\
			10 1.80 	0.21\\
			11 1.74 	0.14\\
			12 1.64 	0.12\\			
			13 1.56 	0.12\\			
			14 1.49 	0.11\\
			15 1.40 	0.11\\			
			16 1.42 	0.13\\
			17 1.39 	0.12\\
			18 1.27 	0.11\\
			19 1.22 	0.10\\
			20 1.24 	0.08\\
			21 1.12 	0.15\\
			22 1.04 	0.06\\
		};
		\addlegendentry{{1 Genuine Samples, RBF SVM, ${\pi}_{gauss}$}}
		
		\addplot
		plot [error bars/.cd, y dir = both, y explicit]
		table[row sep=crcr, x index=0, y index=1, x error index=0, y error index=2,]{
			0  3.84 	0.21\\
			1  2.93 	0.22\\
			2  2.43 	0.07\\
			3  2.19 	0.18\\
			4  1.79 	0.10\\
			5  1.80 	0.14\\
			6  1.54 	0.16\\
			7  1.41 	0.14\\
			8  1.26 	0.11\\
			9  1.21 	0.15\\
			10 1.14 	0.13\\
			11 1.09 	0.16\\
			12 0.99 	0.12\\			
			13 0.88 	0.12\\			
			14 0.93 	0.12\\
			15 0.79 	0.16\\			
			16 0.76 	0.11\\
			17 0.70 	0.08\\
			18 0.64 	0.07\\
			19 0.65 	0.13\\
			20 0.58 	0.10\\
			21 0.54 	0.08\\
			22 0.48 	0.06\\
		};
		\addlegendentry{{2 Genuine Samples, RBF SVM, ${\pi}_{gauss}$}}
		
		\addplot
		plot [error bars/.cd, y dir = both, y explicit]
		table[row sep=crcr, x index=0, y index=1, x error index=0, y error index=2,]{
			0  3.35 	0.14\\
			1  2.48 	0.14\\
			2  1.93 	0.19\\
			3  1.70 	0.16\\
			4  1.56 	0.14\\
			5  1.37 	0.22\\
			6  1.24 	0.12\\
			7  0.99 	0.11\\
			8  0.96 	0.09\\
			9  0.87 	0.12\\
			10 0.79 	0.15\\
			11 0.70 	0.09\\
			12 0.64 	0.15\\			
			13 0.52 	0.05\\			
			14 0.49 	0.07\\
			15 0.49 	0.10\\			
			16 0.38 	0.11\\
			17 0.34 	0.04\\
			18 0.27 	0.06\\
			19 0.27 	0.09\\
			20 0.28 	0.07\\
			21 0.28 	0.06\\
			22 0.20 	0.07\\
		};
		\addlegendentry{{3 Genuine Samples, RBF SVM, ${\pi}_{gauss}$}}
		
		\end{axis}
		\end{tikzpicture}
		\caption{Average EER achieved using GPDS-300 dataset, Signet-F and the feature vectors with noise.}
		\label{fig:eer_gpds_featurevector_with_noise_gaussian_filter_sigmamax_1} 
	\end{figure}

\begin{figure}[!t]
		\pgfplotsset{cycle list/Dark2-6}
		\centering
		\begin{tikzpicture}[scale=0.475] 
		\begin{axis}[ 
		xlabel=Number of the feature vectors with noise per sample,            
		xmin=0,
		xmax = 23,
		minor y tick num={1},
		minor x tick num={1},
		ylabel=Average EER (\%),
		ymin = 0,
		ymax = 10,
		ystep = 1,
		label style={font=\Large},
		tick label style={font=\Large},
		legend columns=1,
		legend style={at={(0.65,0.85)},draw=none,anchor=center,align=center,font=\Large},
		cycle multi list={
			Dark2-6\nextlist
			[1 of]mark list
		},
		width=\textwidth]
		
		\addplot
		plot [error bars/.cd, y dir = both, y explicit]
		table[row sep=crcr, x index=0, y index=1, x error index=0, y error index=2,]{
			0  9.20 	1.33\\
			1  5.03 	0.86\\
			2  4.40 	0.99\\
			3  3.17 	0.68\\
			4  3.20 	0.93\\
			5  2.67 	0.73\\
			6  2.34 	0.48\\
			7  2.36 	0.64\\
			8  1.75 	0.51\\
			9  1.50 	0.45\\
			10 1.20 	0.51\\
			11 1.32 	0.44\\
			12 1.56 	0.41\\			
			13 1.25 	0.53\\			
			14 1.08 	0.31\\
			15 1.06 	0.25\\			
			16 0.87 	0.32\\
			17 0.87 	0.37\\
			18 0.75 	0.22\\
			19 0.68 	0.23\\
			20 0.90 	0.33\\
			21 0.70 	0.36\\
			22 0.72 	0.24\\
		};
		\addlegendentry{{1 Genuine Samples, RBF SVM, ${\pi}_{gauss}$}}
		
		\addplot
		plot [error bars/.cd, y dir = both, y explicit]
		table[row sep=crcr, x index=0, y index=1, x error index=0, y error index=2,]{
			0  5.57 	1.06\\
			1  3.75 	0.90\\
			2  2.60 	0.52\\
			3  2.09 	0.68\\
			4  2.05 	0.47\\
			5  1.43 	0.69\\
			6  1.08 	0.53\\
			7  1.01 	0.58\\
			8  0.88 	0.35\\
			9  0.80 	0.25\\
			10 0.67 	0.18\\
			11 0.55 	0.26\\
			12 0.48 	0.22\\			
			13 0.54 	0.32\\			
			14 0.41 	0.22\\
			15 0.34 	0.29\\			
			16 0.13 	0.08\\
			17 0.27 	0.24\\
			18 0.21 	0.21\\
			19 0.18 	0.24\\
			20 0.26 	0.22\\
			21 0.17 	0.17\\
			22 0.12 	0.17\\
		};
		\addlegendentry{{2 Genuine Samples, RBF SVM, ${\pi}_{gauss}$}}		
		
		\addplot
		plot [error bars/.cd, y dir = both, y explicit]
		table[row sep=crcr, x index=0, y index=1, x error index=0, y error index=2,]{
			0  4.47		0.57\\
			1  2.46 	0.61\\
			2  2.20 	0.47\\
			3  1.70 	0.35\\
			4  1.37 	0.54\\
			5  1.02 	0.37\\
			6  0.77 	0.50\\
			7  0.60 	0.13\\
			8  0.32 	0.28\\
			9  0.22 	0.19\\
			10 0.29 	0.25\\
			11 0.33 	0.27\\
			12 0.14 	0.10\\			
			13 0.15 	0.18\\			
			14 0.19 	0.23\\
			15 0.09 	0.13\\			
			16 0.08 	0.10\\
			17 0.04 	0.06\\
			18 0.04 	0.07\\
			19 0.06 	0.11\\
			20 0.00 	0.01\\
			21 0.01 	0.04\\
			22 0.01 	0.01\\
		};
		\addlegendentry{{3 Genuine Samples, RBF SVM, ${\pi}_{gauss}$}}		
		
		\end{axis}
		\end{tikzpicture}
		\caption{Average EER achieved using MCYT-75 dataset, Signet-F and the feature vectors with noise.}
		\label{fig:eer_mcyt_featurevector_with_noise_gaussian_filter_sigmamax_1} 
	\end{figure}
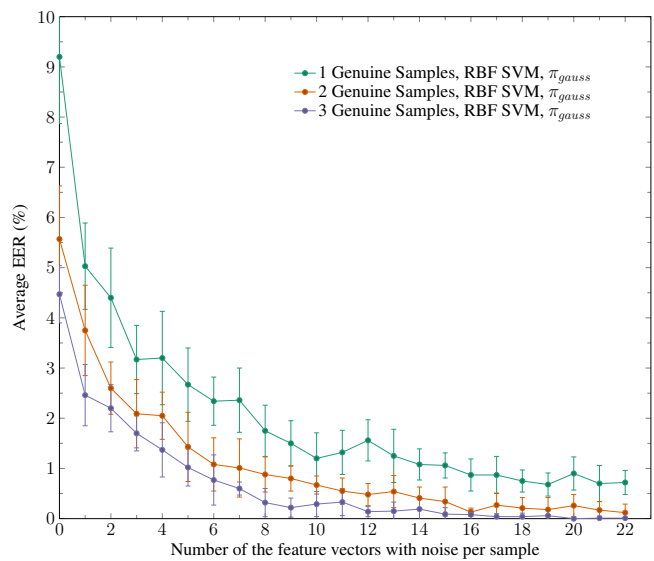

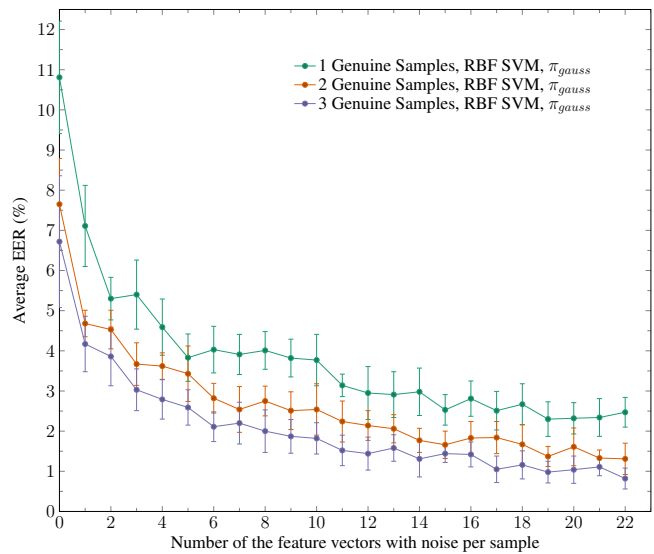
\begin{figure}[!t]
		\pgfplotsset{cycle list/Dark2-6}
		\centering
		\begin{tikzpicture}[scale=0.475] 
		\begin{axis}[ 
		xlabel=Number of the feature vectors with noise per sample,            
		xmin=0,
		xmax = 23,
		minor y tick num={1},
		minor x tick num={1},
		ylabel=Average EER (\%),
		ymin = 0,
		ymax = 12.5,
		ystep = 1,
		label style={font=\Large},
		tick label style={font=\Large},
		legend columns=1,
		legend style={at={(0.65,0.85)},draw=none,anchor=center,align=center,font=\Large},
		cycle multi list={
			Dark2-6\nextlist
			[1 of]mark list
		},
		width=\textwidth]
		
		\addplot
		plot [error bars/.cd, y dir = both, y explicit]
		table[row sep=crcr, x index=0, y index=1, x error index=0, y error index=2,]{
			0  10.81 	1.40\\
			1  7.11 	1.01\\
			2  5.30 	0.53\\
			3  5.40 	0.86\\
			4  4.59 	0.70\\
			5  3.83 	0.59\\
			6  4.03 	0.58\\
			7  3.91 	0.50\\
			8  4.01 	0.47\\
			9  3.82 	0.47\\
			10 3.77 	0.64\\
			11 3.14 	0.28\\
			12 2.95 	0.66\\			
			13 2.91 	0.57\\			
			14 2.98 	0.59\\
			15 2.53 	0.38\\			
			16 2.81 	0.44\\
			17 2.51 	0.48\\
			18 2.67 	0.51\\
			19 2.30 	0.43\\
			20 2.32 	0.39\\
			21 2.34 	0.47\\
			22 2.47 	0.37\\
		};
		\addlegendentry{{1 Genuine Samples, RBF SVM, ${\pi}_{gauss}$}}
		
		\addplot
		plot [error bars/.cd, y dir = both, y explicit]
		table[row sep=crcr, x index=0, y index=1, x error index=0, y error index=2,]{
			0  7.65 	1.14\\
			1  4.68 	0.33\\
			2  4.53 	0.48\\
			3  3.67 	0.53\\
			4  3.62 	0.33\\
			5  3.43 	0.69\\
			6  2.82 	0.37\\
			7  2.54 	0.57\\
			8  2.75 	0.37\\
			9  2.51 	0.47\\
			10 2.54 	0.64\\
			11 2.24 	0.51\\
			12 2.14 	0.37\\			
			13 2.06 	0.35\\			
			14 1.77 	0.30\\
			15 1.66 	0.34\\			
			16 1.83 	0.41\\
			17 1.84 	0.40\\
			18 1.67 	0.49\\
			19 1.37 	0.25\\
			20 1.61 	0.47\\
			21 1.33 	0.20\\
			22 1.31 	0.39\\
		};
		\addlegendentry{{2 Genuine Samples, RBF SVM, ${\pi}_{gauss}$}}
		
		\addplot
		plot [error bars/.cd, y dir = both, y explicit]
		table[row sep=crcr, x index=0, y index=1, x error index=0, y error index=2,]{
			0  6.72 	1.64\\
			1  4.17 	0.69\\
			2  3.86 	0.73\\
			3  3.03 	0.52\\
			4  2.79 	0.49\\
			5  2.59 	0.44\\
			6  2.11 	0.37\\
			7  2.20 	0.52\\
			8  2.00 	0.53\\
			9  1.87 	0.42\\
			10 1.82 	0.39\\
			11 1.52 	0.38\\
			12 1.44 	0.41\\			
			13 1.58 	0.33\\			
			14 1.31 	0.45\\
			15 1.44 	0.22\\			
			16 1.42 	0.31\\
			17 1.05 	0.33\\
			18 1.16 	0.35\\
			19 0.98 	0.27\\
			20 1.04 	0.34\\
			21 1.11 	0.22\\
			22 0.82 	0.26\\
		};
		\addlegendentry{{3 Genuine Samples, RBF SVM, ${\pi}_{gauss}$}}
		
		\end{axis}
		\end{tikzpicture}
		\caption{Average EER achieved using CEDAR dataset, Signet-F and the feature vectors with noise.}
		\label{fig:eer_cedar_featurevector_with_noise_gaussian_filter_sigmamax_1} 
	\end{figure}

	Figures \ref{fig:eer_gpds_featurevector_with_noise_gaussian_filter_sigmamax_1}, \ref{fig:eer_mcyt_featurevector_with_noise_gaussian_filter_sigmamax_1}, and \ref{fig:eer_cedar_featurevector_with_noise_gaussian_filter_sigmamax_1} respectively show the performance on the GPDS-300, MCYT-75, and CEDAR datasets, respectively. The performance is summarized in Table \ref{tab:cmp_results}. As can be observed, most of the results achieved by the classifiers trained with the synthetic feature vectors achieved results similar to those of the proposed method reported in Section \ref{sec:performance_validation_duplicator}. For the CEDAR dataset, the synthetic feature vectors achieved better results than did the duplicates.
	According to {Figure \ref{fig:sigma_si}}, 
	the sigma interval regulates the noise intensity applied in the feature vectors. If the sigma interval is high, it will generate synthetic feature vectors with more noise. Therefore, these synthetic feature vectors will be far from the original feature vectors. Consequently, the EER will be higher than when a small interval is used.
	
	The feature space augmentation does not provide the signature image that can be used by different feature descriptors. It is worth to mention that generating duplicated signatures with realistic appearances helps provide a better understanding of signature execution from several neuroscientific perspectives. Moreover, such synthetic specimens can be used with any feature extraction method \cite{Diaz2017}.
	
	\section{Conclusion}
	\label{sec:conclusion}
	
	In this work, we proposed a method to automatically optimize and select the parameters used to generate synthetic samples of offline handwritten signatures in the image and feature spaces. We showed that the proposed method can be used to increase the number of signatures used to train an automatic handwritten signature verification system. In addition to improving the performance of the system, the proposed method was able to represent writers' variability better than the parameters proposed by Diaz et al. (2017) \cite{Diaz2017}. Furthermore, a new approach to validate the writer variability of synthetic signature samples using their features was proposed. The experimental results support the hypothesis that the writer variability observed on the image space is reflected in the feature space as well.
	
	The proposed method achieved an EER almost equal to zero in MCYT-75. This dataset is characterized by writers with a great variability. Since the writers from the GPDS dataset used to optimize the parameters also showed great variability, this behavior was expected. The proposed method achieved low EERs in the CEDAR dataset. However, the optimization of the six parameters here may not be enough to generate more compact clusters in a feature space such as in the CEDAR dataset. Therefore, more parameters can be optimized to solve this issue. The different distributions of the three datasets suggest that other transformations can be investigated to improve the performance in the CEDAR dataset. The proposed method showed interesting results for three different signature datasets based on the Latin alphabet. To verify its generalization capability for different writing systems, the method can also be evaluated using signatures based on other alphabetical systems.
	
	Notwithstanding the fact that the feature space augmentation in our proposed method has a low computational complexity \cite{Kumar2019} and presented promising results, it nonetheless needs to have optimized parameters in order to generate more realistic samples. As well, it does not provide a signature image that can be used by different feature descriptors.
	Since the method can represent the variability of signatures during data augmentation in the image space, it can therefore be used to create more robust offline handwritten signature datasets. In addition, the generated signatures can be used to train more robust CNN models.

	\ifCLASSOPTIONcaptionsoff
	\newpage
	\fi

	
	
	\bibliographystyle{IEEEtran}
	\bibliography{refer2}
	
	%
	
	
	
	
	%
	
	\vspace*{-1.9cm}
	\begin{IEEEbiography}[{\includegraphics[width=1in,height=1.25in,clip,keepaspectratio]{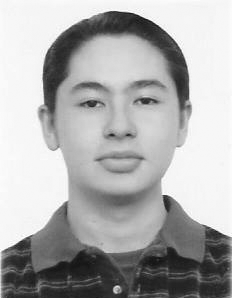}}]{Teruo M. Maruyama}
		received the B.S. degree in computer engineering and the M.Sc. degree in applied computing from the State University of Ponta Grossa, Ponta Grossa, Brazil, in 2013 and 2017, respectively. He is currently a Doctoral candidate in informatics in the Federal University of Paraná, Curitiba, Brazil. His current interests include pattern recognition, machine learning, and image processing.
	\end{IEEEbiography}
	
	\vspace*{-1.9cm}
	\begin{IEEEbiography}[{\includegraphics[width=1in,height=1.25in,clip,keepaspectratio]{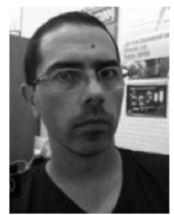}}]{Luiz S. Oliveira}
		received the B.S. degree in computer science from Unicenp, Curitiba, PR, Brazil, the M.Sc. degree in electrical engineering and industrial informatics from the Centro Federal de Educação Tecnológica do Paraná (CEFET-PR), Curitiba, and the Ph.D. degree in computer science from the École de Technologie Supérieure, Université du Québec, in 1995, 1998, and 2003, respectively. From 2004 to 2009, he was a Professor with the Computer Science Department, Pontifical Catholic University of Paraná, Curitiba. In 2009, he joined the Federal University of Paraná, Curitiba, where he is a Professor with the Department of Informatics and the Head of the Graduate Program in computer science. His current interests include Pattern Recognition, Machine Learning, Image Analysis, and Evolutionary Computation.
	\end{IEEEbiography}
	
	\vspace*{-1.8cm}
	\begin{IEEEbiography}[{\includegraphics[width=1in,height=1.25in,clip,keepaspectratio]{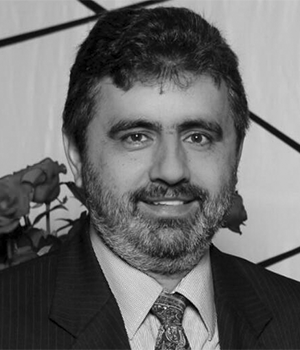}}]{Alceu S. Britto Jr.}
		A. S. Britto Jr. received M.Sc. degree in Industrial Informatics from the Centro Federal de Educação Tecnológica do Paraná (CEFET-PR, Brazil) in 1996, and Ph.D. degree in Computer Science from the Pontifícia Universidade Católica do Paraná (PUCPR, Brazil) in 2001. In 1989, he joined the Informatics Department of the Universidade Estadual de Ponta Grossa (UEPG, Brazil). In 1995, he also joined the Computer Science Department of the Pontifícia Universidade Católica do Paraná (PUCPR) and, in 2001, the Post-graduate Program in Informatics (PPGIa). His current interests include Pattern Recognition, Machine Learning, Image Analysis, and Evolutionary Computation.
	\end{IEEEbiography}
	
	\vspace*{-1.9cm}
	\begin{IEEEbiography}[{\includegraphics[width=1in,height=1.25in,clip,keepaspectratio]{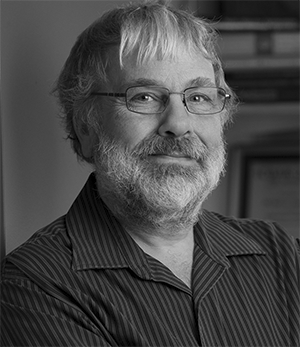}}]{Robert Sabourin}
		joined the Physics Department, Montreal University, in 1977, where his main contribution was the design and implementation of a microprocessor-based fine tracking system combined with a low-light level CCD detector. In 1983, he joined the staff of the École de Technologie Supérieure, Université du Québec, Montreal, where he co-founded the Department of Automated Manufacturing Engineering, where he is currently a Full Professor and teaches pattern recognition, evolutionary algorithms, neural networks, and fuzzy systems. In 1992, he joined the Computer Science Department, Pontificia Universidade Católica do Paraná, Curitiba, Brazil. Since 1996, he has been a Senior Member of the Centre for Pattern Recognition and Machine Intelligence (CENPARMI), Concordia University. Since 2012, he has been the Research Chair specializing in adaptive surveillance systems in dynamic environments. He is the author or coauthor of more than 450 scientific publications, including journals and conference proceedings. His research interests are in the areas of adaptive biometric systems, adaptive classification systems in dynamic environments, dynamic classifier selection, and evolutionary computation.
	\end{IEEEbiography}
	

	
	
	
\end{document}